\documentclass[runningheads]{llncs}

\usepackage{eccv}

\usepackage{eccvabbrv}

\usepackage{graphicx}
\usepackage{booktabs}

\usepackage[accsupp]{axessibility}  %

\usepackage[colorlinks,citecolor=eccvblue]{hyperref}

\usepackage{orcidlink}

\usepackage{graphicx}

\usepackage{tikz}
\usepackage{comment}
\usepackage{amsmath,amssymb} %
\usepackage{color}
\usepackage[numbers,sort&compress]{natbib}

\usepackage[export]{adjustbox} %
\usepackage{booktabs} %

\usepackage{enumitem}%

\usepackage{amssymb}
\usepackage{mathtools}
\usepackage{stmaryrd}
\usepackage{tabularx}
\usepackage{makecell}

\usepackage{url}            %
\usepackage{booktabs}       %
\usepackage{amsfonts}       %
\usepackage{nicefrac}       %
\usepackage{microtype}      %
\usepackage{multirow}       %
\usepackage{algorithm}
\usepackage{algorithmic}

\usepackage{wrapfig}
\usepackage{comment}
\usepackage{amsmath,amssymb} %
\usepackage{symbols}

\usepackage{amsmath,amsfonts,bm}

\def\1{\bm{1}}

\newcommand{\test}{\mathcal{D_{\mathrm{test}}}}

\def\eps{{\epsilon}}
\def\sp{space}

\def\rvc{{\mathbf{c}}}

\def\rvx{{\mathbf{x}}}

\def\vc{{\bm{c}}}

\def\vx{{\bm{x}}}

\def\mA{{\bm{A}}}
\def\mB{{\bm{B}}}

\DeclareMathAlphabet{\mathsfit}{\encodingdefault}{\sfdefault}{m}{sl}
\SetMathAlphabet{\mathsfit}{bold}{\encodingdefault}{\sfdefault}{bx}{n}

\def\gA{{\mathcal{A}}}

\def\gD{{\mathcal{D}}}

\def\gI{{\mathcal{I}}}

\def\gM{{\mathcal{M}}}
\def\gN{{\mathcal{N}}}

\def\gS{{\mathcal{S}}}

\def\gU{{\mathcal{U}}}

\def\sR{{\mathbb{R}}}

\DeclareMathOperator*{\argmin}{arg\,min}

\newcommand*{\ShowNotes}{} %
\usepackage{color}
\usepackage{soul}
\usepackage{multirow}
\usepackage{xcolor}
\usepackage{wrapfig}

\definecolor{darkred}{rgb}{0.7,0.1,0.1}
\definecolor{darkgreen}{rgb}{0.1,0.7,0.1}
\definecolor{cyan}{rgb}{0.7,0.0,0.7}
\definecolor{dblue}{rgb}{0.2,0.2,0.8}
\definecolor{maroon}{rgb}{0.76,.13,.28}
\definecolor{burntorange}{rgb}{0.81,.33,0}
\definecolor{tealblue}{rgb}{0.212,0.459, 0.533}
\definecolor{myyellow}{rgb}{0.8627451 , 0.75294118, 0.20784314]}

\definecolor{mypink}{rgb}{0.93359375, 0.62109375, 0.83984375}

\definecolor{pp}{rgb}{0.43921569, 0.18823529, 0.62745098}
\definecolor{rr}{rgb}{0.5254902 , 0.00784314, 0.12941176}
\definecolor{bb}{rgb}{0.09019608, 0.23529412, 0.37647059}
\definecolor{yy}{rgb}{0.49803922, 0.3372549 , 0.0}
\definecolor{gg}{rgb}{0.02352941, 0.3372549 , 0.17647059}

\ifdefined\ShowNotes
  \newcommand{\colornote}[3]{{\color{#1}\bf{#2: #3}\normalfont}}
\else
  \newcommand{\colornote}[3]{}
\fi

\definecolor{imma}{rgb}{0.23, 0.45, 0.69} %
\definecolor{noimma}{rgb}{0.95 , 0.53, 0.17} %
\definecolor{immao}{rgb}{0.17254902, 0.62745098, 0.17254902} %
\definecolor{mist}{rgb}{0.5, 0.3, 0.3} %

\definecolor{lightred}{rgb}{0.9,0.4,0.4}

\newlength\savewidth

\definecolor{turquoise}{cmyk}{0.65,0,0.1,0.1}
\definecolor{purple}{rgb}{0.65,0,0.65}
\definecolor{darkgreen}{rgb}{0.0, 0.5, 0.0}
\definecolor{darkred}{rgb}{0.5, 0.0, 0.0}
\definecolor{darkblue}{rgb}{0.0, 0.0, 0.5}
\definecolor{blue}{rgb}{0.0, 0.0, 1.0}
\definecolor{orange}{rgb}{1.0,0.5,0.0}

\renewcommand{\vec}[1]{\mathbf{#1}}

\makeatletter
\renewcommand{\paragraph}{%
  \@startsection{paragraph}{4}%
  {\z@}{0.3ex \@plus 1ex \@minus .1ex}{-1em}%
  {\normalfont\normalsize\bfseries}%
}
\makeatother

\newif\ifproofread

\usepackage[multiple]{footmisc}

\newcommand{\myparagraph}[1]{\vspace*{1pt}{\bf\noindent #1}}

\definecolor{mybrown}{rgb}{0.87058824, 0.56078431, 0.01960784}
\definecolor{myblue}{rgb}{0.3372549 , 0.70588235, 0.91372549}
\definecolor{mypurple}{rgb}{0.8, 0.47058824, 0.7372549 }
\definecolor{myorange}{rgb}{0.835, 0.368, 0}
\definecolor{mygreen}{rgb}{0.00784314, 0.61960784, 0.45098039}
\definecolor{mygt}{rgb}{0.0078125 , 0.57421875, 0.40625}
\definecolor{mysp}{rgb}{0.84765625, 0.515625  , 0.0234375}
\definecolor{mycitecolor}{rgb}{0,0.08,0.45}
\definecolor{mygr}{rgb}{0.9607,0.9607,0.9607}
\definecolor{myoo}{rgb}{0.992,0.9176,0.9019}

\newcommand{\SGR}{\text{Similarity Gap Ratio}\xspace}
\newcommand{\sgr}{\texttt{SGR}\xspace}

\newcommand{\rsgr}{\texttt{RSGR}\xspace}

\usepackage{arydshln}

\begin{document}
\pagestyle{headings}
\mainmatter
\def\ECCVSubNumber{5625}  %

\title{IMMA: Immunizing text-to-image Models against Malicious Adaptation} %

\author{Amber Yijia Zheng \quad\quad
Raymond A. Yeh}
\authorrunning{A.Y.~Zheng and R.A.~Yeh}
\institute{Department of Computer Science, Purdue University \\
\email{\tt\small \{zheng709,rayyeh\}@purdue.edu}
}
\maketitle

\begin{abstract}
Advancements in open-sourced text-to-image models and fine-tuning methods have led to the increasing risk of malicious adaptation, \ie, fine-tuning to generate harmful/unauthorized content. Recent works, \eg, Glaze or MIST, have developed data-poisoning techniques which protect the data against adaptation methods. In this work, we consider an alternative paradigm for protection. We propose to ``immunize'' the model by learning model parameters that are difficult for the adaptation methods when fine-tuning malicious content; in short IMMA. Specifically, IMMA should be applied before the release of the model weights to mitigate these risks.
Empirical results show IMMA's effectiveness against malicious adaptations, including mimicking the artistic style and learning of inappropriate/unauthorized content, over three adaptation methods: LoRA, Textual-Inversion, and DreamBooth. The code is available at \url{https://github.com/amberyzheng/IMMA}.
\end{abstract}

\section{Introduction}
With the open-source of large-scale text-to-image models~\cite{rombach2022high,deep-floyd-if} %
the entry barrier to generating images has been drastically lowered. %
Building on top of these models, methods such as Textual Inversion~\cite{gal2022textual}, DreamBooth~\cite{ruiz2022dreambooth}, and LoRA~\cite{hu2022lora} allow quick adaptation to generate personalized content. These newly introduced capabilities come with great responsibility and trust in individuals to do good instead of harm to society. 
Unfortunately, the capabilities of adapting text-to-image generative models have already had negative impacts, \eg, the generation of sexual content~\cite{harwell2023wash}, copying artists' work without consent~\cite{heikkila2022artist,noveck2023}, duplicating celebrity images~\cite{Moore_2023}, \etc. We broadly encapsulate all these harmful fine-tuning of models under the term {\it malicious adaptation}.

To countermeasure these adaptations, open-source models have users agree that they will not use the software for ``the purpose of harming others'' in their licenses~\cite{sd_license} or implement safety checks that would censor inappropriate generated content~\cite{rando2022red}. Nonetheless, these approaches do not have real enforcing power. Users can trivially disregard the license and remove the safety filters~\cite{smith2022}. 

To address these loopholes, recent data poisoning techniques have shown a promising path towards preventing malicious adaptations~\cite{shan2023glaze,liang2023adversarial,pmlr-v202-salman23a}. The main idea is to \textit{protect images} by modifying them with imperceivable changes, \eg, adversarial noise, such that adaptation methods confuse the style and content of the images and fail to generalize. A key shortcoming of these methods is that the burden of enforcement is on \textit{content creators.} The artists need to apply these techniques before releasing their artwork. Contrarily, we present an alternative paradigm that places this burden on the \textit{releaser of open-source models.}

\begin{wrapfigure}[14]{r}{0.5\linewidth}
\vspace{-.76cm}
\input{figs/teaser}
\end{wrapfigure}
In this paper, we propose to ``{\bf I}mmunnize'' {\bf M}odels, \ie, make them more resilient, against {\bf M}alicious {\bf A}daptation; in short \textbf{IMMA}. The goal of IMMA is to make adaptation more difficult for concepts that are deemed malicious while maintaining the models' adaptability for other concepts. At a high level, we propose an algorithm to learn model parameters that perform \textit{poorly} when being adapted to malicious concepts, \eg, mimicry artistic style. In~\figref{fig:teaser}, we illustrate the effect of IMMA when mimicking Picasso's style; the model with IMMA  fails to mimic the style.

To demonstrate the effectiveness of IMMA, we consider three adaptation methods, namely, Textual Inversion~\cite{gal2022textual}, DreamBooth~\cite{ruiz2022dreambooth}, and LORA~\cite{hu2022lora}. We experimented with immunizing against several malicious settings, including, mimicking artistic styles, restoring erased concepts, and learning personalized concepts. 
Overall, IMMA successfully makes text-to-image models more resilient against adaptation to malicious concepts while maintaining the usability of the model. 
{\bf \noindent Our contributions are as follows:}
\begin{itemize}[topsep=1pt]
\item We propose a novel paradigm for preventing malicious adaptations. In contrast to the data poisoning for protection paradigm, we aim to protect the model instead of the data.  See~\figref{fig:intro_comp}.
\item We present an algorithm (IMMA) that learns difficult model initialization for the adaptation methods.
\item We conduct extensive experiments evaluating IMMA against various malicious adaptations.
\end{itemize}

\section{Related Work}
{\noindent\bf Diffusion models.}
By learning to reverse a process of transforming data into noise, diffusion models achieve impressive generative capabilities~\cite{sohl2015deep, ho2020denoising, dbeatgan}. With the aid of Internet-scale datasets~\cite{laion}, these models are capable of generating diverse images with high realism~\cite{imagen,dalle2,glide,rombach2022high,deep-floyd-if,podell2023sdxl,dai2023emu}. This progress led to new excitements in artificial intelligence generated content (AIGC) and interest in how to mitigate the associated risks, which we discuss next.

\begin{figure}[t]
\centering
\small
\setlength{\tabcolsep}{8pt}
\begin{tabular}{cc}
\textbf{\color{mist} Data Poisoning Paradigm} & \textbf{\color{imma} Model Immunization Paradigm (Ours)}\\
\includegraphics[height=2.7cm]{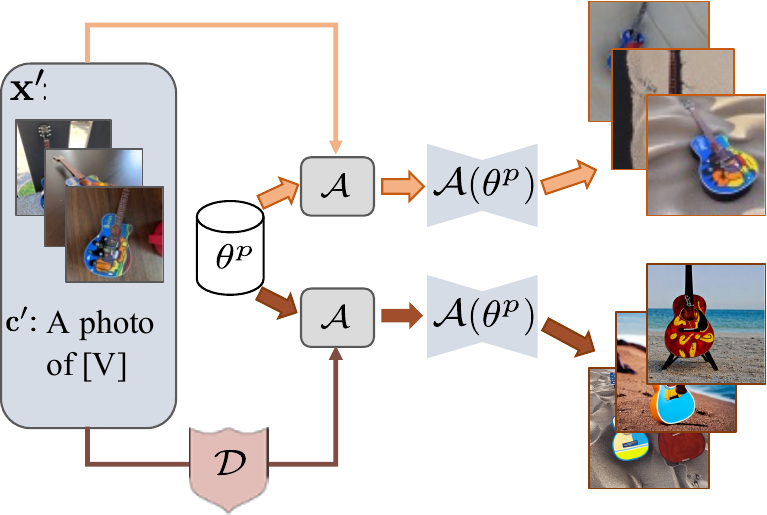} &
\includegraphics[height=2.7cm]{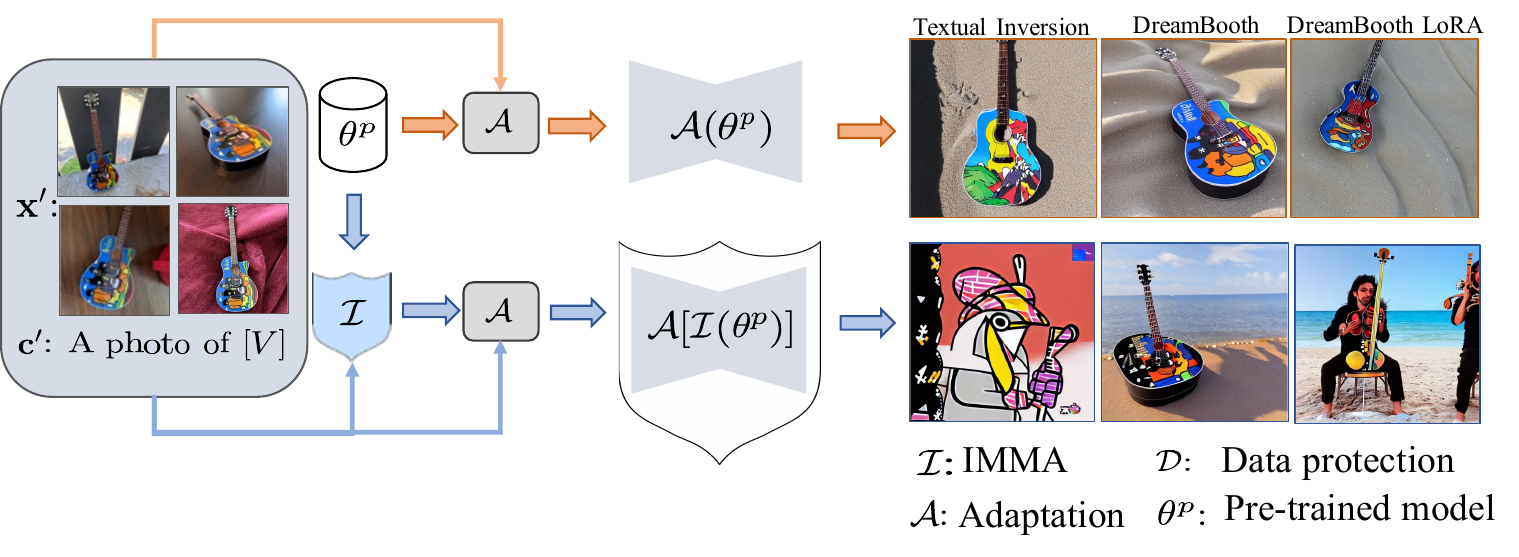}
\end{tabular}
\vspace{-0.4cm}
\caption{{\bf Paradigms for preventing malicious adaptation.
} \textit{\color{mist}Data poisoning:} modify training images $\rvx'$ with imperceivable changes, such that $\gA$ fails to capture $\rvc'$ by training with modified images. \textit{\color{imma}Model immunization (ours):} modify pre-trained model weights $\theta^p$ with immunization methods $\gI$ before adaptation $\gA$, such that $\gA$ fails to capture $\rvc'$ by training on immunized model weights $\gI(\theta^p)$.
\vspace{-0.4cm}
}
\label{fig:intro_comp}
\end{figure}

\myparagraph{Preventing generative AI misuse.} There are many potential risks associated with the advancement in generative capabilities~\cite{bird2023typology, qu2023unsafe}. Recent works have started to address these risks. For example, \citet{wang2023detect} study how to detect unauthorized images that are used in the training set of these models.~\citet{schramowski2023safe} study how to suppress  the generation of inappropriate content, \eg, nudity, self-harm, \etc, during diffusion's generation process. Other works~\cite{gandikota2023erasing,gandikota2023unified,zhang2023forgetmenot,kumari2023conceptablation,heng2023selective}, take a further step in trying to remove these inappropriate content from the diffusion models. 

More closely related to this work, GLAZE~\cite{shan2023glaze}, MIST~\cite{liang2023adversarial, liang2023mist}, and EUDP~\cite{zhao2023unlearnable} studied the misuse of artistic mimicry, \ie, preventing diffusion models from being used to copy artistic styles. \citet{pmlr-v202-salman23a} have also studied how to raise the cost of image manipulation using diffusion models. Notably, these works aim to protect the image, \ie, a form of data-poisioning~\cite{biggio2011support,mei2015using} which modifies the data, \eg, adding adversarial noise~\cite{goodfellow2015explaining}, such that the adaptation techniques fail. Different from these works, we \textit{protect the model}, against misuse, \ie, model immunization. We illustrate the difference between our proposed model immunization \vs data poisoning in~\figref{fig:intro_comp}.

\myparagraph{Meta-learning.} As our approach to model immunization is based on learning against an adaptation method, we briefly review meta-learning; also known as learning to learn. The area covers learning many aspects of a learning algorithm, \eg, learning good initialization suitable to adapation~\cite{finn2017model}, or other hyperparameters, \eg, learning rate, or weight decay, via hypergradient~\cite{Larsen,bengio2000gradient,lorraine2019opt,ren2020not,yeh2022equivariance, zheng2024graph, zheng2024learning}. Various hypergradients approximations have been proposed, \eg, MAML~\cite{finn2017model} uses a single step gradient unrolling, with a summary provided by~\citet{lorraine2019opt}. Different from these works, we aim to learn \textit{poor} initializations for the adaptation methods to prevent the misuse of generative models.

\section{Preliminaries}
\label{sec:prelim}
We briefly review the concepts necessary to understand our approach and introduce a common notation.

\myparagraph{Text-to-image diffusion models.} 
The goal of a text-to-image diffusion model~\cite{rombach2022high} is to learn a conditional distribution of images $\rvx$ given concept embedding $\rvc$, \ie, modeling $p(\rvx| \rvc; \theta)$, where $\theta$ denotes the model parameters. The learning objective is formulated via variational lower bound~\cite{ho2020denoising,kingma2021variational} or from a denoising score-matching perspective~\cite{vincent2011connection,song2019generative} which boils down to minimizing 
\begin{equation} \label{eq:sd}
    L(\rvx, \rvc; \theta) = \sE_{t, \eps \sim \gN(0, I)} \left[ w_t \|\eps_\theta(\rvx_t, \rvc, t) - \eps\|_2^2 \right],
\end{equation}
where $\eps_\theta$ denotes the denoising network,  $\rvx_t$ and $w_t$ denote the noised images and loss weights for a given time-step $t\sim\gU{\{0,T\}}$ sampled from a discrete uniform distribution. 

\myparagraph{Concept erasing methods. }%
To erase a target concept $\vc'$ from a model, erasing algorithms~\cite{gandikota2023erasing, kumari2023conceptablation} fine-tunes a pre-trained model's parameters $\theta^p$ such that the model no longer generate images corresponding to that concept, \ie, 
\bea\label{eq:erased}
    p(\tilde \vx| \rvc'; \theta^p_{-\rvc'}) \approx 0, ~\forall \tilde \vx \sim p(\rvx| \rvc'; \theta^p),
\eea
where $\theta^p_{-\rvc'}$ denotes the parameters of the erased model after fine-tuning. The main idea is to train the target concept to generate images of some other concept. A common choice is an empty token, \ie, the model performs unconditional generation when prompted with the target concept $\vc'$.

\myparagraph{Personalization of text-to-image diffusion models.} 
In contrast to erasing a concept, personalization algorithms aim to \textit{add} a novel concept to the pre-trained model. Given a set of images $\{\rvx'\}$ representative of the concept $\rvc'$. Personalization methods 
introduce a novel token $[V]$ in the word space and train the model to associate $[V]$ to the new concept. To learn this new concept, the model is trained using the data pair $(\rvx', \rvc')$, where $\rvc' \triangleq \Gamma([V])$ is extracted with the text encoder $\Gamma$ using the novel word token $[V]$. We broadly use the notation
$L_{\gA}(\rvx',\rvc'; \theta, \phi)$
to denote the loss function of each adaptation method $\gA$, where $\phi$ denotes the parameters that are being fine-tuned. For example, DreamBooth~\cite{ruiz2022dreambooth,kumari2022customdiffusion} fine-tunes on a subset of parameters, \eg, cross-attention layers, whereas Textual Inversion~\cite{gal2022textual} only optimizes the new word token. %

Another common approach to make the fine-tuning more data efficient is to use Low-Rank Adaptation (LoRA)~\cite{hu2022lora}. Given a pre-trained weight $\theta^p \in \sR^{n \times d}$, LoRA aims to learn an adaptor $\Delta$, such that the final weights become $\hat{\theta} = \theta^p + \Delta $. LoRA specifically restricts the $\Delta$ to be low-rank, \ie, $\Delta = \mA\mB$ where $\mA\in\sR^{n \times r}$ and $\mB\in\sR^{r \times d}$ with $r \ll \min(n,d)$. 
In this work, we show that LoRA can easily learn back the concepts that were previously erased which highlights the need for our proposed research direction of model immunization.

\section{Approach}
\label{sec:app}

Given a pre-trained text-to-image model with parameters $\theta^p$,~\eg, StableDiffusion~\cite{rombach2022high}, existing adaptation methods $\gA$~\cite{gal2022textual, ruiz2022dreambooth, kumari2022customdiffusion, hu2022lora} can fine-tune $\theta^p$ such that the model generates images of a concept $\rvc'$. 
Our goal is to \textit{prevent} the adaptation methods from successfully doing so on harmful concepts, \eg, unauthorized artistic style. To accomplish this, we present an algorithm IMMA $\gI$ that takes pre-trained parameters $\theta^p$ as input, and outputs the immunized parameters $\theta^I$. When applying adaptation $\gA$ with $\theta^I$, it should fail to learn the concept $\rvc'$, \ie, the model is ``immunized'' against adaptation. At a high level, IMMA aims to learn a \textit{poor model initialization} when being fine-tuned by the adaptation methods. We now describe the algorithmic details.

\subsection{Model immunization}
To achieve this goal of learning a poor initialization for adaptation, we propose the following bi-level program:
\bea
\hspace{-0.05cm}
\label{eq:bilevel}
\overbracket{\max_{\theta_{\in\gS}} L_{\gA}(\rvx'_{\gI}, \rvc'; \theta, \phi^\star)}^{\text{upper-level task}}
\text{~s.t.~}\phi^\star =
\overbracket{
\argmin_\phi L_{\gA}(\rvx'_{\gA}, \rvc'; \theta, \phi)}^{\text{lower-level task}}.
\hspace{-0.2cm}
\eea
Here, the set $\gS$ denotes a subset of $\theta$ that is trained by IMMA. This set is a hyperparameter that we choose empirically. Given a dataset of images $\gD = \{\rvx'\}$ representative of the concept $\rvc'$, we sample the training data $\rvx'_{\gA}$ and $\rvx'_{\gI}$ independently. As reviewed in~\secref{sec:prelim}, $\phi$ denotes the parameters modified by the adaptation method $\gA$. Note, $\phi$ may include newly introduced parameters by $\gA$ or the parameters in the pre-trained model. We use the notation $\gU$ to be the set of overlapping parameters between $\phi$ and $\theta$.

Intuitively, the lower-level task simply performs the adaptation $\gA$ given the model initialization $\theta$ by minimizing the loss function $L_\gA$ with respect to (\wrt) $\phi$.
On the other hand, the upper-level task is \textit{maximizing} the loss function of $\gA$ \wrt $\theta$. That is, the upper-level task aims for $\theta$ that results in \textit{poor performance} when the adaptation method $\gA$ is applied to the target concept $\rvc'$. The worse the performance is when adapted by $\gA$, the more immunized a model is.

To solve the program in~\equref{eq:bilevel}, we use gradient-based methods %
to solve the upper-level optimization. This leads to the following update steps:
\bea
        \phi^\star& \leftarrow& \arg\min_\phi L_{\gA}(\rvx'_{\gA}, \rvc'; \theta^{i-1}, \phi) \\ \label{eq:min}
        \theta^i& \leftarrow& \theta^{i-1} + \alpha \nabla_\theta L_{\gA}(\rvx'_{\gI}, \rvc'; \theta^{i-1}, \phi^\star) \label{eq:max},
\eea
\begin{wrapfigure}[18]{l}{0.5\linewidth}%
\vspace{-1.2cm}
\begin{minipage}{0.5\textwidth}
    \input{figs/alg_imma}
\end{minipage}
\end{wrapfigure}%
where $\alpha$  denotes learning rate. We summarized the procedure in~\algref{alg:imma}.
Given the pre-trained model parameters $\theta^p$, the algorithm returns an immunized parameter $\theta^I$. 
We will next describe the subtle choices that were made regarding re-initialization and overlapping parameters during the optimization of $\theta$ and $\phi$.\\
\myparagraph{Details on updating $\phi$ and $\theta$.} 
In~\algref{alg:imma} line 5, in theory, we should reinitialize $\phi$ following the initialize scheme of $\gA$, as the lower-level task is performing the adaptation. In practice, computing the lower-level task until convergence for each outer-loop iteration is prohibitively expensive. To reduce computation, we only solve the lower-level task with a fixed number of update steps. However, this leads to lower-level tasks being not very well trained due to the small number of updates on $\phi$.
To address this, we initialize $\phi$ from $\phi^{i-1}$, \ie, the result from the previous outer-loop iteration. Empirically, this leads to faster convergence of the lower-level tasks; we suspect that this is because $\theta^{i}$ and $\theta^{i-1}$ remain quite similar after one outer-loop update.

The next subtle detail is when updating $\theta$ in~\algref{alg:imma} line 7. Recall, that depending on the adaptation method, the pre-trained parameters $\theta$ and adapted parameters $\phi$ may overlap. In such a scenario, there are two ways to compute the upper-level task's gradient: (a) we assume that $\theta^{i-1}$ and $\phi^i$ are separate parameters, \ie, the result of $\phi^i$ will not change $\theta^{i-1}$; (b) we update the overlapping parameters in $\gU$ with the value from $\phi^i$, leading to line 7 in~\algref{alg:imma}. Empirically, we found (b) to perform better. We conduct an ablation study for~\algref{alg:imma} lines 5 and 7 in~\secref{sec:ablation}, where we found both to improve immunization quality.

\myparagraph{Implementation details.} %
To apply IMMA in practice, we approximate the solution of the lower-level task by taking a single gradient step for each of the adaptation methods. For the upper-level task, we use the Adam optimizer~\cite{kingma2015adam}. We choose $\gS$ in~\equref{eq:bilevel} to contain only the cross-attention layer, \ie, only these layers are being optimized. This choice follows the intuition that cross-attention layers are important as they mix the features of the target concept and image representation. Additional hyperparameters and experiment details are documented in the appendix.

\subsection{Applications of model immunization}
\myparagraph{Immunizing concept erased models.} Recent works~\cite{gandikota2023erasing, zhang2023forgetmenot, kumari2023conceptablation}, reviewed
\begin{wrapfigure}[14]{r}{0.55\textwidth}%
\vspace{-.85cm}
    \centering
    \small
    \setlength{\tabcolsep}{1.4pt}
    \renewcommand{\arraystretch}{1.2}
    \resizebox{\linewidth}{!}{%
    \begin{tabular}{lcc@{\hskip 6pt}cc}
    & \multicolumn{2}{c}{Stable Diffusion} & \multicolumn{2}{c}{Re-learning}\vspace{-3pt}\\
    & Reference  & Erased (ESD) & \color{noimma} w/o IMMA &  \color{imma} w/ IMMA\\
    \multirow{2}{*}[2cm]{\rotatebox[origin=c]{90}{Thomas Kinkade}} & 
    \includegraphics[height=1.92cm]{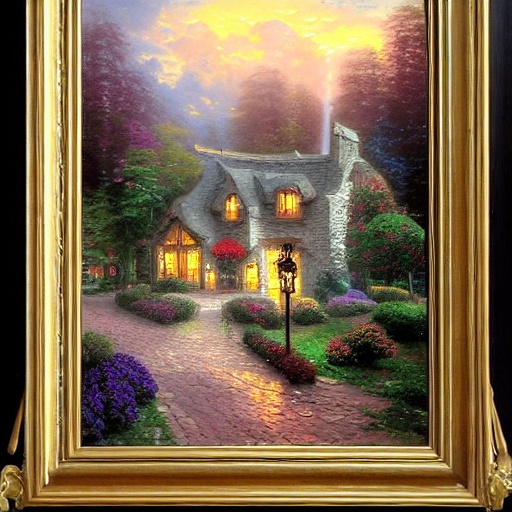} & 
    \includegraphics[height=1.92cm]{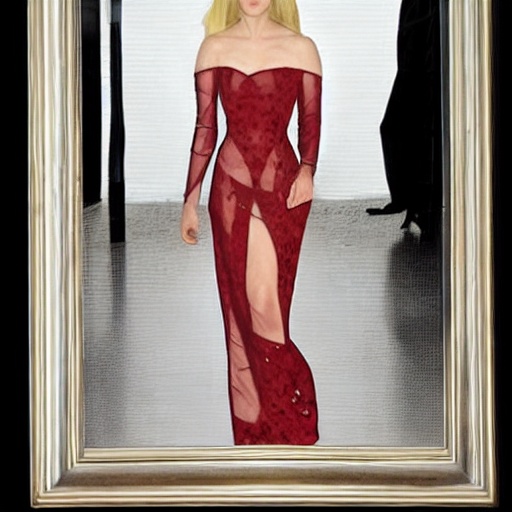}  &
    \includegraphics[height=1.92cm]{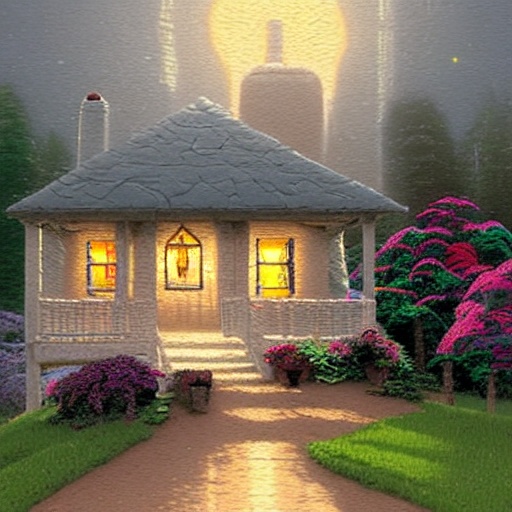}  &
    \includegraphics[height=1.92cm]{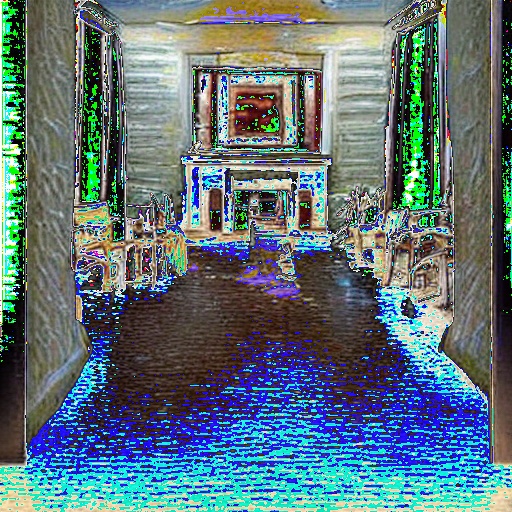}  \\
    \multirow{2}{*}[1.5cm]{\rotatebox[origin=c]{90}{Kilian Eng}} & 
    \includegraphics[height=1.92cm]{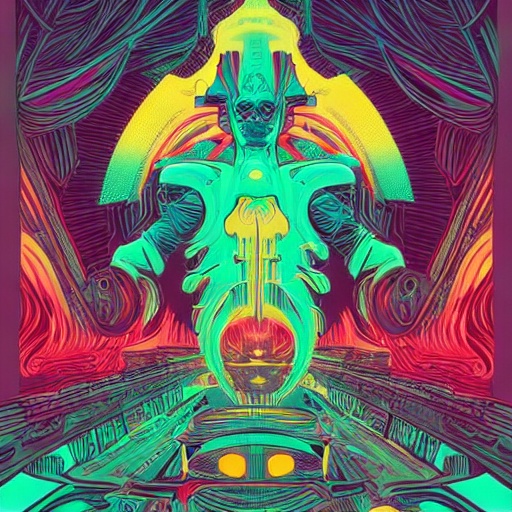} & 
    \includegraphics[height=1.92cm]{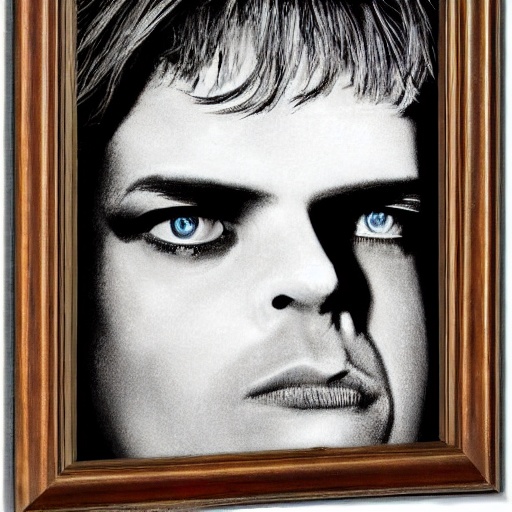}  &
    \includegraphics[height=1.92cm]{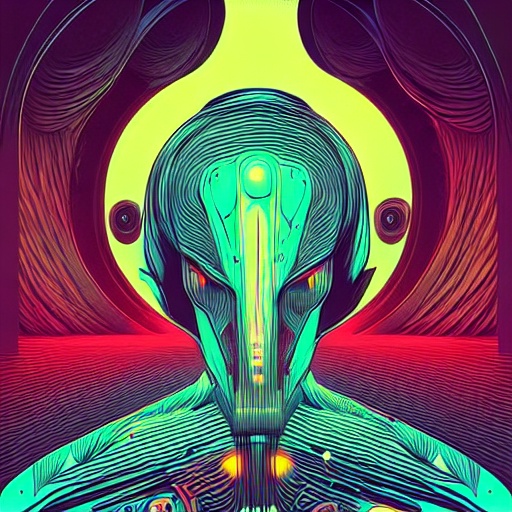}  &
    \includegraphics[height=1.92cm]{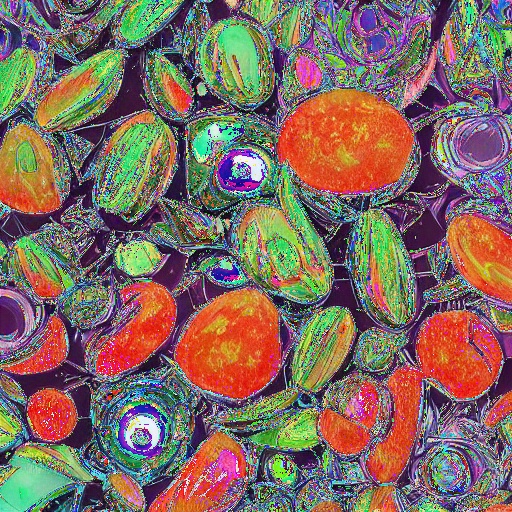}  \\
    \multirow{2}{*}[2.cm]{\rotatebox[origin=c]{90}{Kelly McKernan }} & 
    \includegraphics[height=1.92cm]{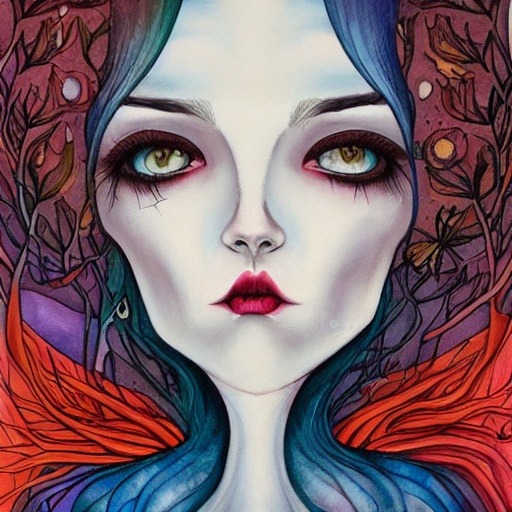} & 
    \includegraphics[height=1.92cm]{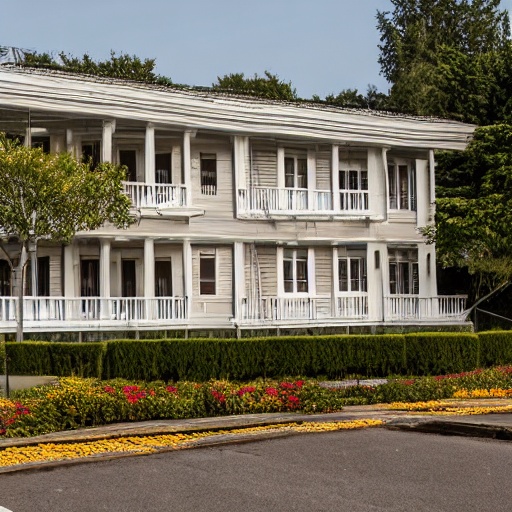}  &
    \includegraphics[height=1.92cm]{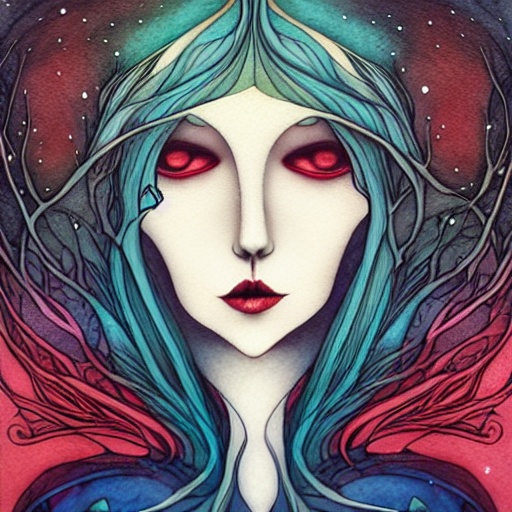}  &
    \includegraphics[height=1.92cm]{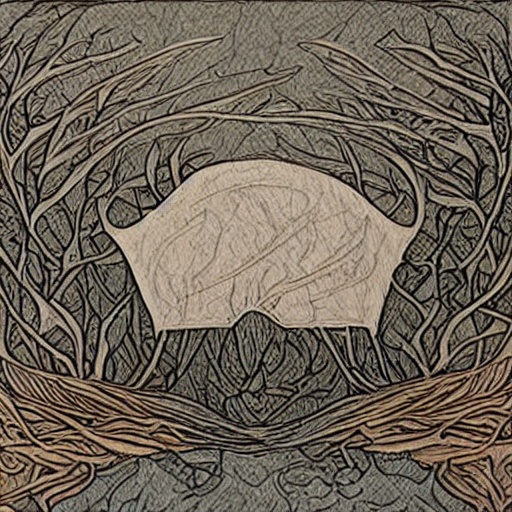}  \\
    \end{tabular}
    }
    \vspace{-0.3cm}
    \caption{\textbf{IMMA's result against re-learning.} %
    }
    \label{fig:lora_qual_art}
\end{wrapfigure}%
in~\secref{sec:prelim}, 
have shown that they can erase concepts from diffusion models without retraining the model from scratch. 
After a target concept $\rvc'$ is erased, the erased model
$\theta^p_{-\rvc'}$ can no longer generate that object or style given the concept's text prompt. 
However, in our experiments, we show that the model can easily \textit{re-learn} the target concept again in just a few training epochs by using LoRA~\cite{hu2022lora}. In~\figref{fig:lora_qual_art}, we illustrate that the erased stable diffusion (ESD~\cite{gandikota2023erasing}) successfully removed a target concept of artists' style. 

This motivated us to immunize the concept erased model $\theta^p_{-\rvc'}$ to make the adaptation of re-learning $\rvc'$ more difficult.
Ideally, the immunized model is no longer able to generate images of the target concept, \ie, 
\bea
    p(\tilde \rvx| \rvc'; \gA[\gI(\theta^p_{-\rvc'})]) \approx 0, ~\forall \tilde \rvx \sim p(\rvx| \rvc'; \theta^p).
\eea

\myparagraph{Immunizing against personalization adaptation.} 
Another potential for misuse is with personalization adaptation. Methods such as DreamBooth~\cite{ruiz2022dreambooth} or Textual Inversion~\cite{gal2022textual} allow for a stable diffusion model to quickly learn to generate a personalized/unique concept given a few images of the unique concept. This motivated us to study immunization against these personalization adaptations. In practice, IMMA will be applied before a pre-trained model's release such that it fails to generate the unique concept even after adaptation.

\section{Experiments}

To evaluate IMMA, we conduct comprehensive experiments over multiple applications and adaptation methods. %

\subsection{Immunizing erased models against re-learning}
\label{sec:erase}

{\bf \noindent Experiment setup.}
We employ LoRA as the adaptation method, and the pre-trained erased models are from ESD~\cite{gandikota2023erasing} using their publicly released code base.
Following ESD, we consider experiments on eight artistic styles including both well-recognized and modern artists, and ten object classes from a subset of ImageNet~\cite{deng2009imagenet}. For immunization, we use 20 images generated by Stable Diffusion (SD V1-4), prior to erasing, with the prompt of the target artistic style or object class. For adaptation, we generate \textit{different} 20 images for each dataset and use LoRA to fine-tune the erased model for 20 epochs. 
For style, the prompt used in the evaluation is \textit{``An artwork by \{artist name\}''}. 
For object, %
the prompt used in evaluation is \textit{``a \{concept\}''}, for example, \textit{``a parachute''}.
We also evaluate on re-learning Not-Safe-For-Work (NSFW) content where we followed I2P prompts proposed by~\citet{schramowski2023safe}.

\myparagraph{Evaluation metrics.} To measure the effect of immunization against re-learning, we aim for a metric that measures the \textit{performance gap} with and without IMMA, where the larger value indicates a stronger effect of IMMA, and vice versa. For this, we propose \textit{\SGR} (\sgr). Let $\rvx_\gI$ and $\rvx_\gA$ denote the generated images with and without IMMA, and $\rvx_r$ to be the reference images of the target concept then~\sgr is defined as follows: 
\bea
\label{eq:sgr}
\sgr(\rvx_\gI, \rvx_\gA, \rvx_r) = \frac{\gM(\rvx_r, \rvx_\gA) - \gM(\rvx_r, \rvx_\gI)}{\gM(\rvx_r, \rvx_\gA)},
\eea
where $\gM$ is an image similarity metric. Common choices, following prior works~\cite{gandikota2023erasing,gal2022textual,ruiz2022dreambooth}, include Learned Perceptual Image Patch Similarity~\cite{zhang2018unreasonable} (LPIPS), and similarity measured in the feature space of CLIP~\cite{gal2022textual} or DINO~\cite{caron2021emerging} each denoted as \sgr(\texttt{L}), \sgr(\texttt{C}), and \sgr(\texttt{D}).
For consistency, we report \textit{one minus} LPIPS, such that larger values for all three metrics mean higher image similarity. %

{\bf\noindent User study.} To check the quantitative metrics against human perception, we prepare four reference images and one pair of generated images (w/ and w/o IMMA) for the participants to select the generated image that is more similar to the reference images in terms of content and quality. \\
\begin{wrapfigure}[21]{r}{0.5\textwidth}
\vspace{-1.2cm}
\centering
\captionof{table}{\textbf{\sgr$\uparrow$(\%) on artistic styles for ESD with LoRA adaptation.}
}
\label{tab:lora_style}
\setlength{\tabcolsep}{2pt}
\resizebox{\linewidth}{!}{%
\begin{tabular}{lccccccccc}
\specialrule{.15em}{.05em}{.05em}
           \multirow{2}{*}{\sgr} & Van  & Pablo & Tyler & Kelly  & Kilian & Claude  & Thomas& Kirbi  & %
          \\
          & Gogh & Picasso &Edlin & Mckernan &  Eng & Monet & Kinkade & Fagan & \\
\hline
\texttt{(L)} & 17.61    & 28.08         & 28.67       & 23.34          & 26.78      & 31.14        & 18.47          & 16.18       \\ %
\texttt{(C)}     & 4.77     & 4.56          & 5.87        & 9.34           & 4.59       & 7.27               & 9.44           & 1.18  \\
\texttt{(D)}     & 18.4     & 21.81         & 26.05       & 14.99          & 25.73      & 31.48           & 39.59          & 15.07  \\ %

\specialrule{.15em}{.05em}{.05em}
\end{tabular}
}

\vspace{0.3cm}
    \centering
    \setlength{\tabcolsep}{0.5pt}
    \resizebox{\linewidth}{!}{%
    \begin{tabular}{lcccc}
     \includegraphics[height=2.3cm, trim={0 0.35cm 0 0.35cm},clip]{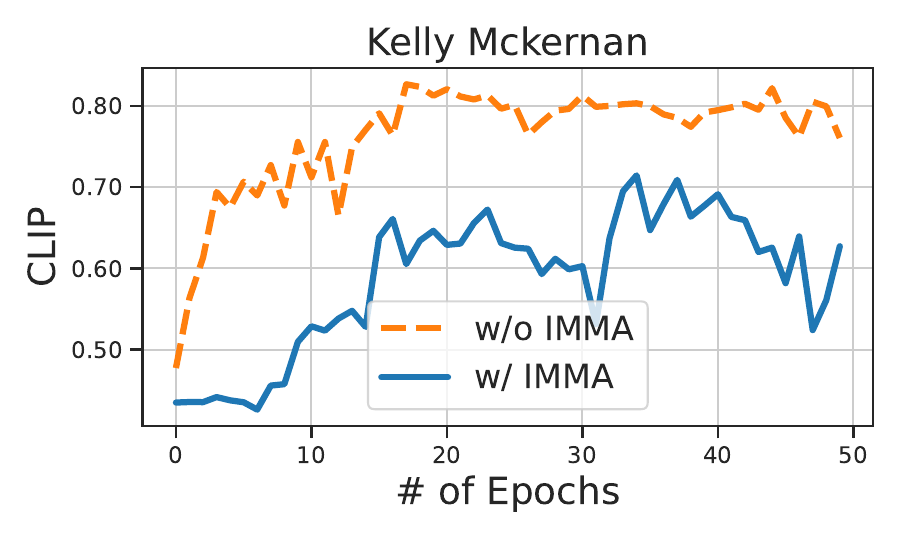} & \includegraphics[height=2.3cm, trim={0 0.35cm 0 0.35cm},clip]{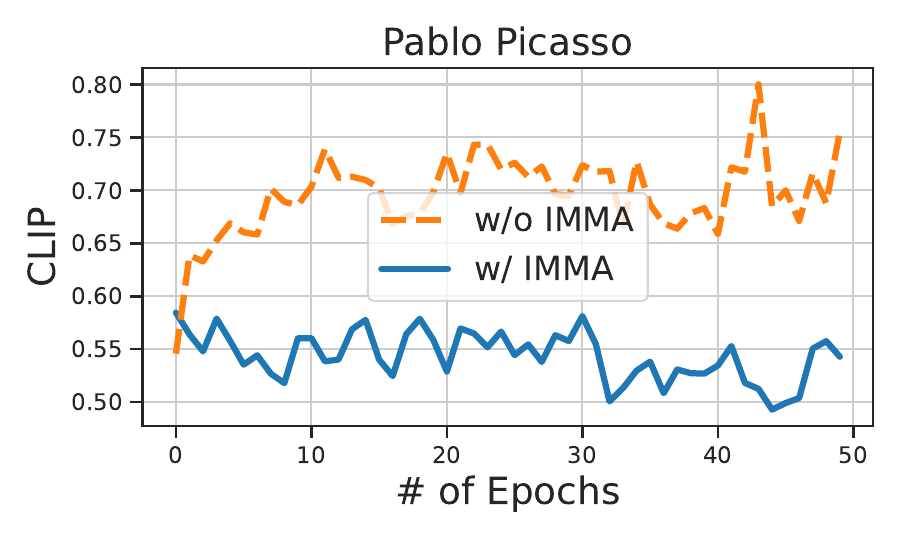}\\
     \includegraphics[height=2.3cm, trim={0 0.35cm 0 0.35cm},clip]{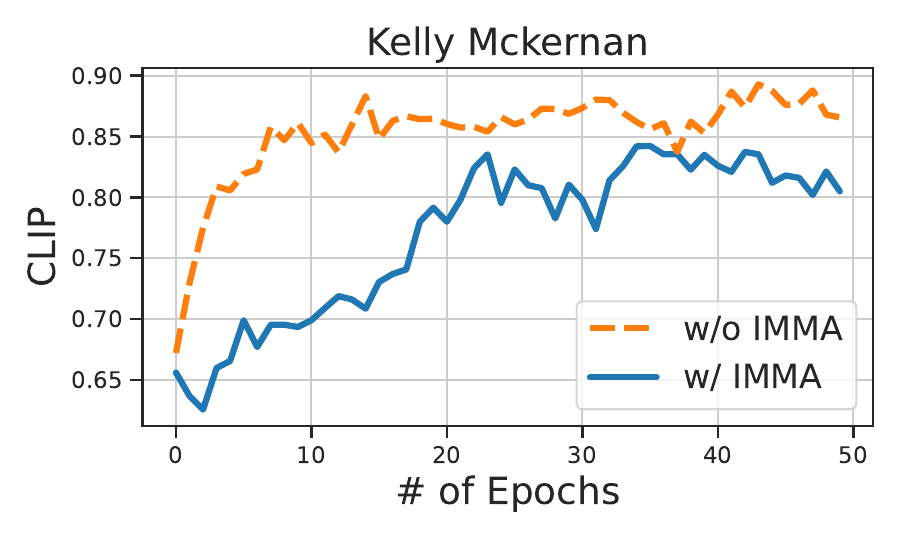} & \includegraphics[height=2.3cm, trim={0 0.35cm 0 0.35cm},clip]{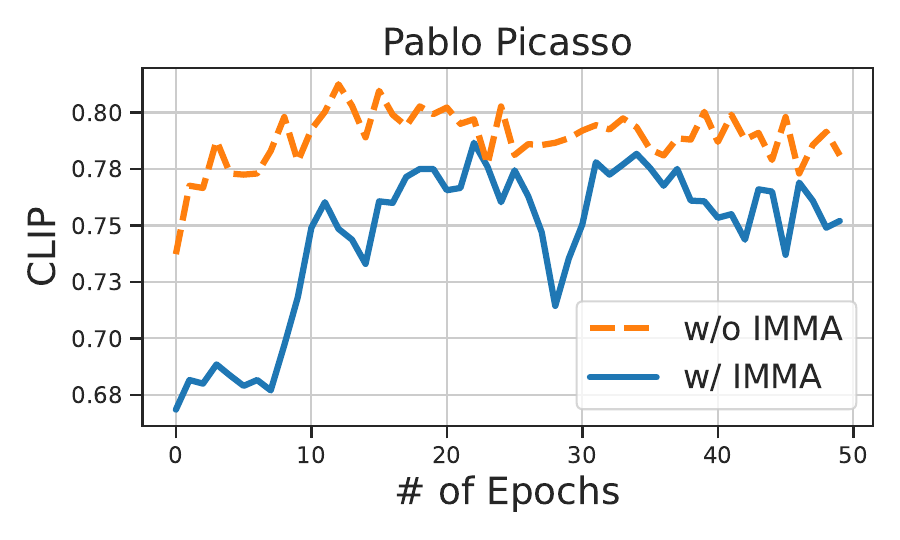}\\
     \includegraphics[height=2.3cm, trim={0 0.35cm 0 0.35cm},clip]{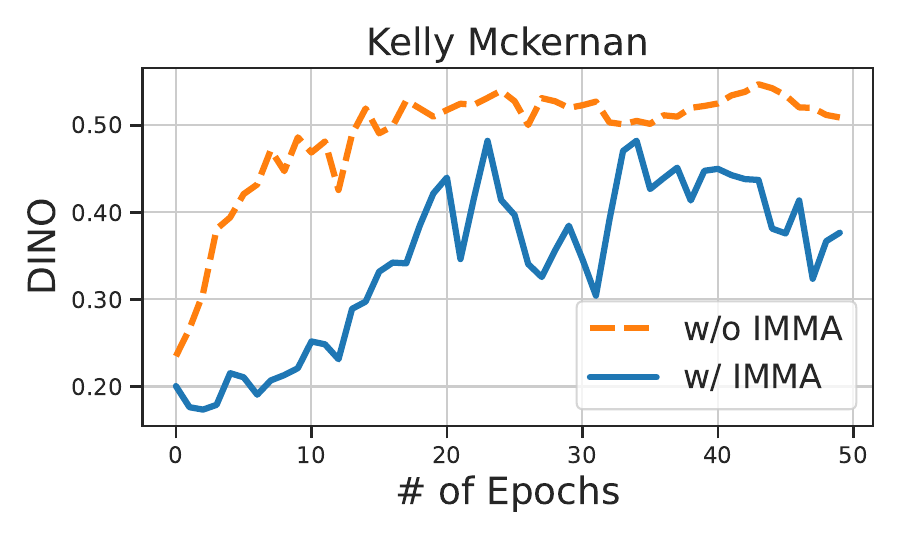} & \includegraphics[height=2.3cm, trim={0 0.35cm 0 0.35cm},clip]{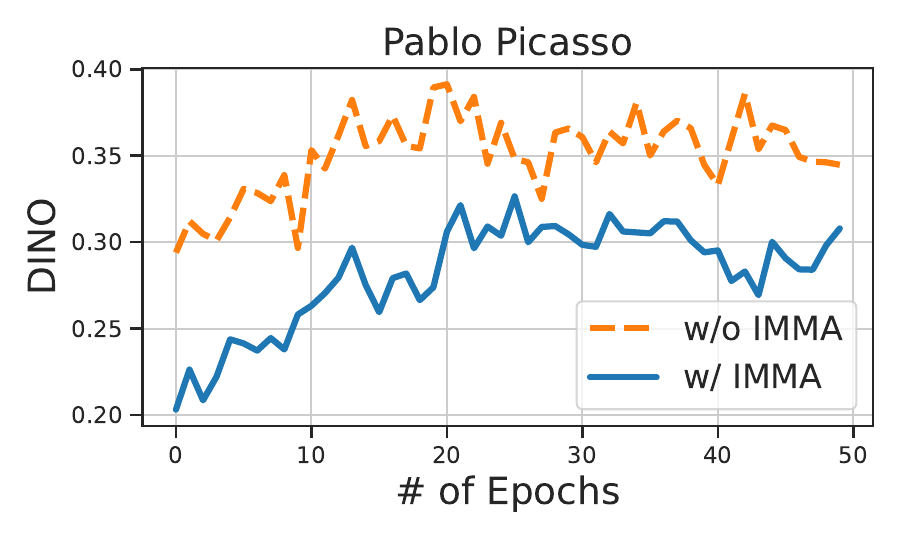}\\
    \end{tabular}
    }
    \vspace{-0.4cm}
    \captionof{figure}{{\bf Similarity \vs epochs for LoRA on styles.} Models with IMMA achieve lower similarity throughout LoRA's epochs. 
    }
    \label{fig:clip_lpips_lora_art}

\end{wrapfigure}
\myparagraph{Results on style.}
\label{sec:style}
In~\tabref{tab:lora_style}, we report \sgr for re-learning artist styles for erased models. %
Over eight artists, we consistently observe a positive gap among all three \sgr based on LPIPS, CLIP, and DINO with averages of 21.84\%, 5.5\%, and 23.35\%. %

In~\figref{fig:clip_lpips_lora_art}, we directly show the LPIPS, CLIP, and DINO metrics at each epoch during the LoRA fine-tuning to visualize the gap. A lower value represents that the generated images are \textit{less similar} to the reference images. Across all of the plots, we observe models without IMMA $\uparrow$ ({\color{noimma}orange}) are more similar to the reference images than models with IMMA $\downarrow$ ({\color{imma}blue}). For artistic styles, the gap remains steady throughout the epochs. Overall, models with IMMA struggle to generate images containing the target concepts. 

\begin{wrapfigure}[21]{r}{0.5\textwidth}%
\vspace{-1.15cm}
\captionof{table}{\textbf{\sgr$\uparrow$(\%)  on objects for ESD with LoRA adaptation.} 
}
\label{tab:lora_object}
\centering
\setlength{\tabcolsep}{1.5pt}
\renewcommand{\arraystretch}{1.185}
\resizebox{\linewidth}{!}{%
\begin{tabular}{lccccccccccc}
\specialrule{.15em}{.05em}{.05em}
 \multirow{2}{*}{\sgr} & Cass. & Garbage & Gas   & Chain & En.  & Golf  & \multirow{2}{*}{church} & French & \multirow{2}{*}{Parachute} & \multirow{2}{*}{Tench} \\ %
          & player   & truck   & pump  & saw   & springer & ball  &        & horn   &           &       &         \\ %
\hline
\texttt{(L)} & 11.66           & 7.48          & 16.64    & 27.33     & 22.42            & 41.96     & 10.09  & 6.88        & 50.97     & 39.14 \\ %
\texttt{(C)}      & 3.66            & 5.18          & 19.72    & 8.72      & 15.08            & 10.67     & 12.51  & 8.11        & 19.34     & 11.07 \\ %
\texttt{(D)}     & 12.16           & 20.0          & 31.31    & 50.86     & 68.85            & 58.09     & 14.33  & 30.25       & 68.56     & 61.84     \\ %
\specialrule{.15em}{.05em}{.05em}
\end{tabular}
}

\vspace{0.3cm}%
    \centering
    \setlength{\tabcolsep}{1.1pt}
        \resizebox{\linewidth}{!}{%
    \begin{tabular}{cc}
     \includegraphics[height=2.3cm, trim={0 0.35cm 0 0.35cm},clip]{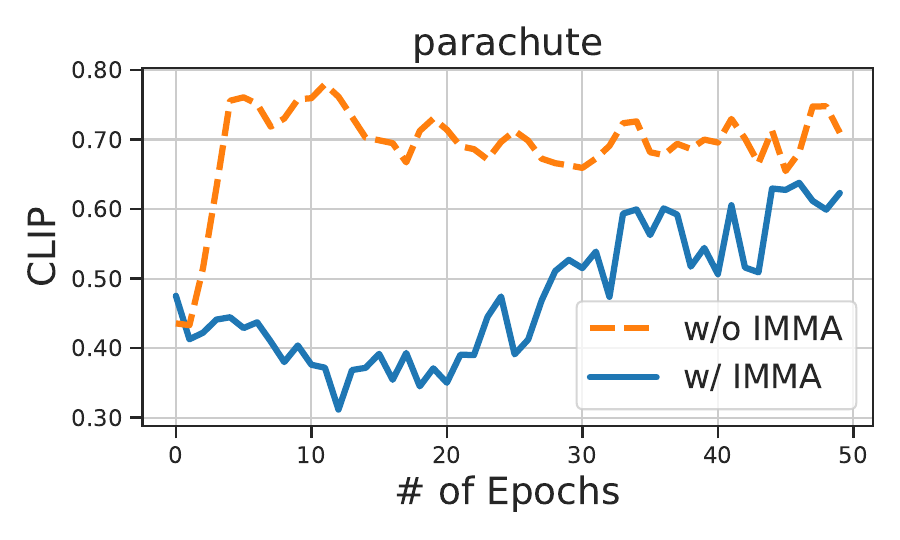} & \includegraphics[height=2.3cm, trim={0 0.35cm 0 0.35cm},clip]{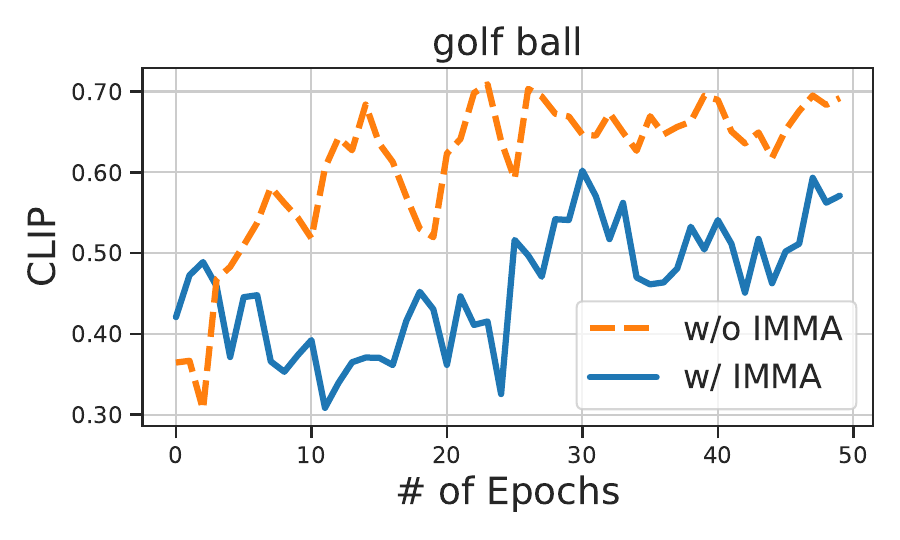}\\
     \includegraphics[height=2.3cm, trim={0 0.35cm 0 0.35cm},clip]{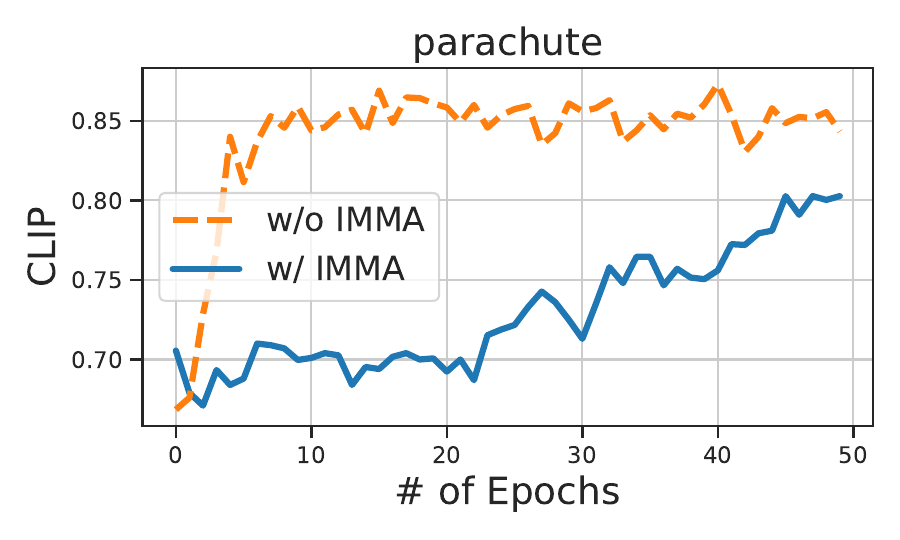} & \includegraphics[height=2.3cm, trim={0 0.35cm 0 0.35cm},clip]{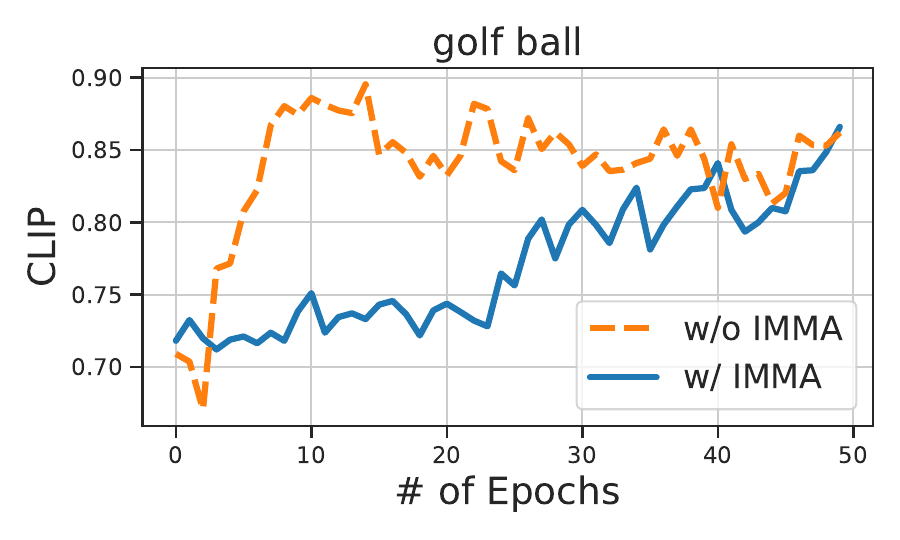}\\
     \includegraphics[height=2.3cm, trim={0 0.35cm 0 0.35cm},clip]{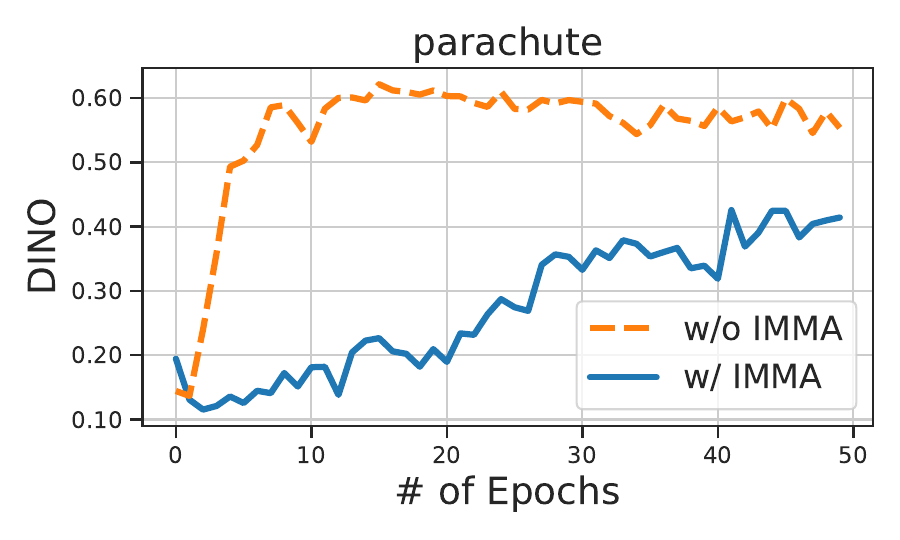} & \includegraphics[height=2.3cm, trim={0 0.35cm 0 0.35cm},clip]{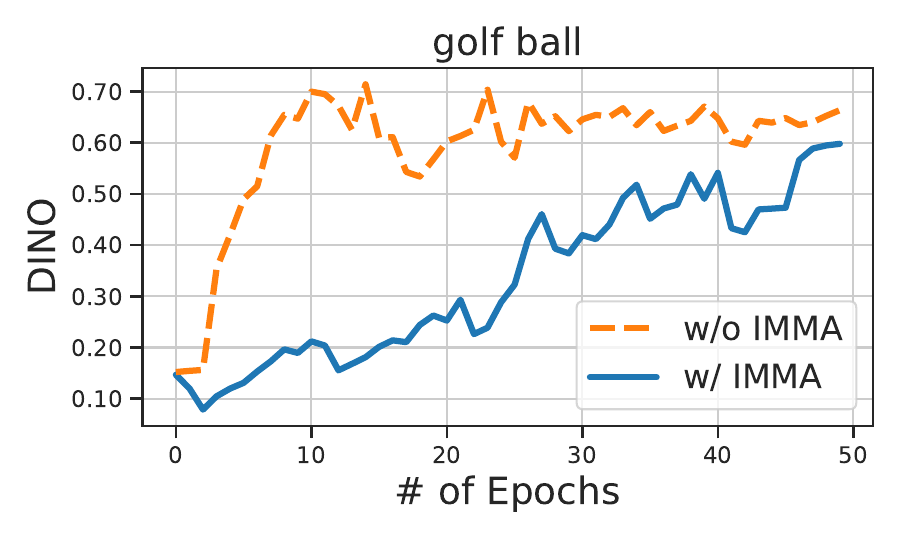}\\
    \end{tabular}
    }
    \vspace{-0.45cm}
    \captionof{figure}{{\bf Similarity \vs epochs for LoRA on objects.} Models with IMMA achieve lower similarity throughout LoRA's epochs.
    }
    \label{fig:clip_lpips_lora_obj}

\end{wrapfigure}%
In~\figref{fig:lora_qual_art}, we provide qualitative comparisons. 
We observe that LoRA successfully re-learned the target concept (third column).  On the other hand, the model with IMMA (last column) either generates an image with lower quality or an unrelated image. Our user study also validates this observation.
 All of the 30 respondents selected generations without IMMA as the one with high similarity and quality across all compared samples. This shows that models with IMMA generate worse images of the target styles.\\
\myparagraph{Results on objects.} %
As in artistic styles, we report~\sgr of re-learning target objects for the erased models in~\tabref{tab:lora_object}.
The average \sgr(\texttt{L}), \sgr(\texttt{C}), and \sgr(\texttt{D}) across ten classes are 21.84\%, 11.41\%, and 41.62\%. 
We also visualize LPIPS, CLIP, and DINO in~\figref{fig:clip_lpips_lora_obj}. Overall, we observe the same trend as in the result for artistic styles. Similarity metrics across all the plots drop for models with IMMA. That is, with the same number of fine-tuning epochs, generations from models with IMMA exhibit lower quality or are less related to the object.

\begin{table}[t]
\centering
\captionof{table}{\textbf{Acc. (\%) of object erased models on 500 images.} 
{\it Col.~1:} Original ESD model with the target concept erased. {\it Col.~2:} ESD without IMMA after LoRA. We observe that the target concept is sucessfully relearned. {\it Col.~3:} ESD with IMMA after LoRA. {\it Col.~4 \& 5:} Acc. of other objects of ESD before and after IMMA. 
}
\label{tab:lora_cls}
\small %
\setlength{\tabcolsep}{2.3pt}
{%
\begin{tabular}{lccc|cc}
\specialrule{.15em}{.05em}{.05em}
Class name / &\multicolumn{3}{c|}{Acc. of LoRA's  target ($\downarrow$)} & \multicolumn{2}{c}{Acc. of others ($\uparrow$)} \\ \cline{2-6}
Methods & ESD    &  w/o IMMA        & w/ IMMA     & ESD                  & w/ IMMA                 \\\hline
Cassette player  & 0.2       & 2.0          & 0.2             & 72.0             & 46.4     \\
Garbage truck    & 3.0       & 40.8         & 7.6            & 63.8             & 25.5    \\
Gas pump         & 0.0       & 63.4         & 0.0            & 62.7             & 39.8      \\
Chain saw        & 0.0       & 15.2         & 0.8             & 79.3             & 52.5     \\
EN springer & 0.4       & 15.6         & 0.8            & 67.2             & 49.6     \\
Golf ball        & 0.4       & 22.4         & 0.0             & 56.4             & 36.7      \\
Church           & 5.6       & 73.4         & 11.8            & 82.3             & 70.0      \\
French horn      & 0.2       & 80.2         & 0.4             & 64.9             & 57.6       \\
Parachute        & 2.0       & 91.0         & 0.0             & 78.8             & 58.9      \\
Tench            & 1.0       & 50.8         & 0.4             & 78.0             & 55.5      \\ \hline
Average          & 1.3       & 45.5         & 2.2            & 70.6             & 49.3      \\
\specialrule{.15em}{.05em}{.05em}
\end{tabular}
}

\end{table}
As the target concept contains objects, we consider classification accuracy (ResNet50 pre-trained on ImageNet) for evaluation reported in~\tabref{tab:lora_cls}. First, without IMMA, ESD can relearn to generate the object. With a mere three epochs of LoRA, the average accuracy of the target concept increased from 1.3\% to 45.5\%. On the other hand, ESD with IMMA, the average accuracy remains low at 2.2\%, demonstrating the effectiveness of IMMA at preventing relearning. 

Thus far, the evaluation has focused on the \textit{target concept} for IMMA models. We are also interested in how well IMMA preserves the \textit{other concepts}. We define ``other concepts'' to be the remaining nine object categories beside the target object that is being adapted. In~\tabref{tab:lora_cls} (rightmost two columns), we observe that the original ESD has an average of 70.6\% and after IMMA the accuracy
\begin{wrapfigure}[12]{r}{0.5\linewidth}%
\vspace{-0.83cm}
    \hspace{-0.55cm}
    \centering
    \small
    \setlength{\tabcolsep}{1.5pt}
    \renewcommand{\arraystretch}{1.2}
    \resizebox{\linewidth}{!}{%
    \begin{tabular}{lcc@{\hskip 8pt}cc}
    & \multicolumn{2}{c}{Stable Diffusion} & \multicolumn{2}{c}{Re-learning}\vspace{-3pt}\\
    & Reference  & Erased (ESD) & \color{noimma} w/o IMMA &  \color{imma} w/ IMMA\\
    \multirow{2}{*}[1.2cm]{\rotatebox[origin=c]{90}{Church}} & 
    \includegraphics[height=1.92cm]{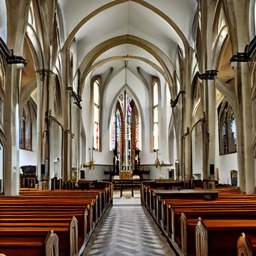} & 
    \includegraphics[height=1.92cm]{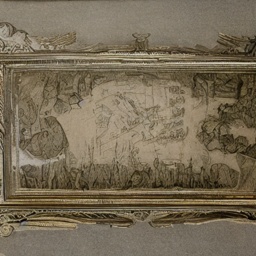}  &
    \includegraphics[height=1.92cm]{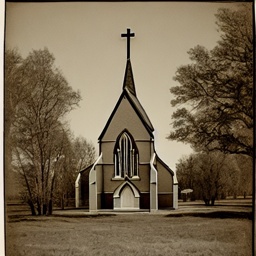}  &
    \includegraphics[height=1.92cm]{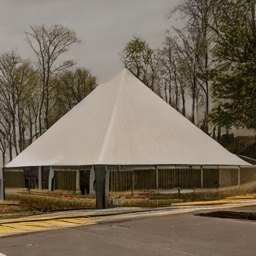}  \\
    \multirow{2}{*}[1.7cm]{\rotatebox[origin=c]{90}{Cassette Player}} & 
    \includegraphics[height=1.92cm]{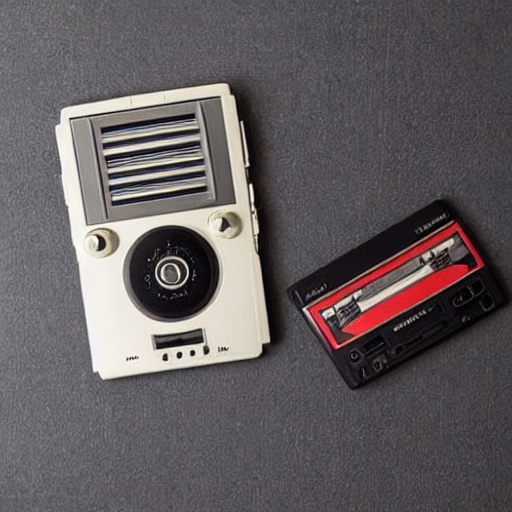} & 
    \includegraphics[height=1.92cm]{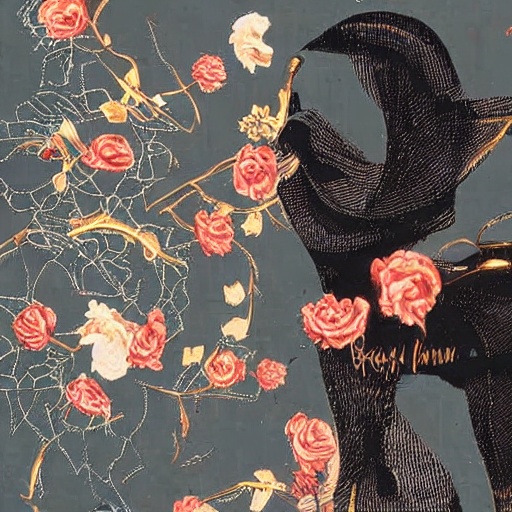}  &
    \includegraphics[height=1.92cm]{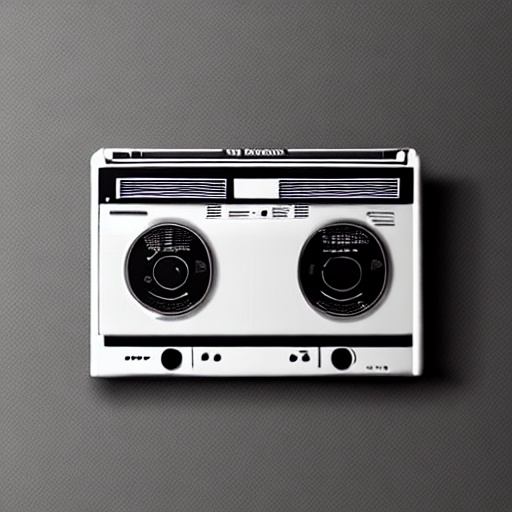}  &
    \includegraphics[height=1.92cm]{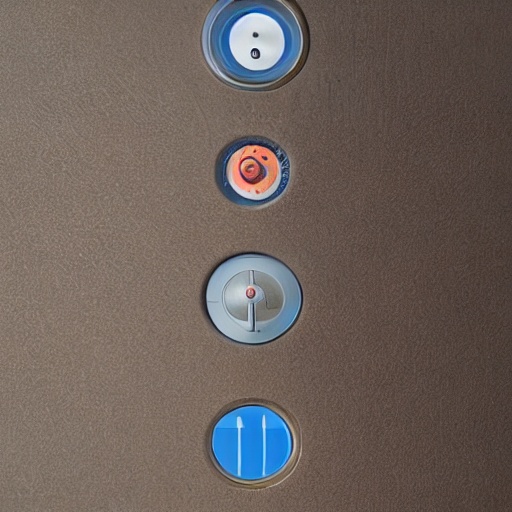}  \\
    \end{tabular}
    }
    \vspace{-0.3cm}
    \captionof{figure}{\textbf{Qualitative result of IMMA against re-learning.} %
    }
    \label{fig:lora_qual_obj}
\end{wrapfigure}%
dropped to 49.3\%. 
We fully acknowledge that this is \textit{a limitation} of IMMA. Prevention against certain target concepts may degrade other concepts. IMMA roughly trades off 43\% in the target concept with 20\% in other concepts.
Finally, qualitative comparisons are shown in~\figref{fig:lora_qual_obj}, where we observe the same conclusion that IMMA is effective against re-learning.

\begin{figure}[t]
    \centering
    \setlength{\tabcolsep}{-1pt}
    \renewcommand{\arraystretch}{1}
    \fontsize{7.5}{9}\selectfont
        \resizebox{0.75\linewidth}{!}{%
    \begin{tabular}{rcc}
     & \color{noimma} \quad~~ w/o IMMA &  \color{imma} \quad~~ w/ IMMA     \vspace{-0.09cm} \\
    Male Genitalia (19)  & \multirow{9}{*}{
    \includegraphics[height=3.2cm, max width=0.9\linewidth]{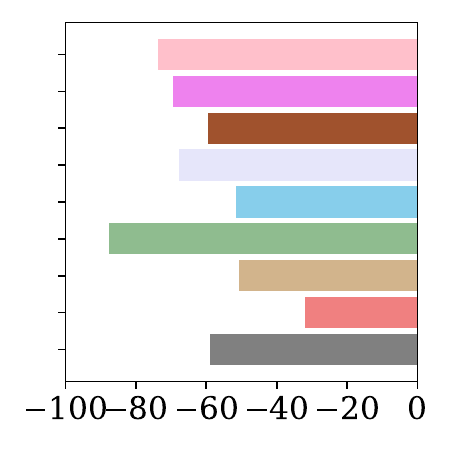}}  & \multirow{9}{*}{\includegraphics[height=3.2cm]{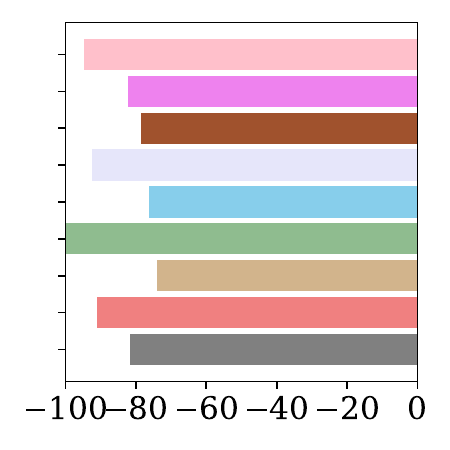} } \\
    Belly (85) && \\ 
    Armpits (89) && \\ 
    Feet (65) && \\ 
    Male Breast (173) && \\ 
    Female Genitalia (8) && \\ 
    Female Breast (134)  && \\ 
    Buttocks (22)  && \\ 
    Total (454) && \\
    && \\
    &  ~~\% of Change & ~~\% of Change 
    
    \end{tabular}
    }
    \vspace{-0.2cm}
    \caption{\textbf{IMMA on NSFW content.} We report the \% of change, relative to SD V1-4, in the number of detected nudity content after LoRA on the nudity-erased model.
    }
    \label{fig:nsfw}
    \vspace{-1.cm}
\end{figure}

\myparagraph{Results on NSFW.} %
For experiments %
on NSFW concepts, we use SD V1-4 to generate 4,703 unsafe images, among which 351 images contain 454 nudity counts based on NudeNet~\cite{bedapudi2019nudenet} with a threshold of 0.05. We randomly select two sets of 50 images containing nudity and their corresponding prompts for re-learning adaptation and the other for IMMA. The evaluation is conducted on the remaining 251 unsafe prompts.

In~\figref{fig:nsfw}, we compare the percentage of change before and after IMMA adaptation with respect to the base SD V1-4 model. We observe that IMMA successfully generated less nudity content after LoRA, \ie, w/ IMMA achieved a negative 80\% of change in the detected nudity content compared to the negative 60\% change for the model without IMMA. This shows that IMMA successfully immunized the model making it more difficult to re-learn nudity from unsafe images/prompts.

\vspace{-.3cm}

\subsection{Immunizing against personalized content}
\label{sec:person}
{\bf \noindent Experiment setup.}
We consider three adaptation methods for learning new unique/personalized concepts: Textual Inversion 
 (TI)~\cite{gal2022textual}, DreamBooth (DB)~\cite{ruiz2022dreambooth}, and DreamBooth LoRA. For DreamBooth LoRA (DB LoRA), instead of modifying all of the parameters during fine-tuning, LoRA is applied on top of the cross-attention layers. %
We follow the exact adaptation procedures following prior works~\cite{ruiz2022dreambooth, gal2022textual},~\ie, adding a special token for the new unique concept. Note, we use \textit{different} novel tokens during adaptation and IMMA. This is because, for a realistic evaluation, we would not know the novel token that would be used during the adaptation.

We perform experiments on ten different datasets released by~\citet{kumari2022customdiffusion} which include a variety of unique objects. Each of them contains four to six images taken in the real world. 
The evaluation prompt for all concepts in this section is \textit{``A $[V]$ on the beach''} following DreamBooth.

\myparagraph{Evaluation metrics.}
In this task, we report~\sgr in~\equref{eq:sgr} with CLIP and DINO following DreamBooth~\cite{ruiz2022dreambooth}. Next, to show that the model maintains its capability to be personalized for \textit{other concepts}, we propose \textit{Relative Similarity Gap Ratio} (\texttt{\rsgr}), which is given by
\bea
\label{eq:rslr}
\texttt{RSGR}(\rvx_\gI, \rvx_\gA, \rvx_\gI^o, \rvx_\gA^o) = \frac{\gM(\rvx_\gA^o, \rvx_\gI^o) - \gM(\rvx_\gA, \rvx_\gI)}{\gM(\rvx_\gA^o, \rvx_\gI^o)},
\eea
where $\rvx_\gA^o$ and $\rvx_\gI^o$ are generated images without and with IMMA on \textit{other unique concepts}. 
\begin{wrapfigure}[15]{r}{0.5\linewidth}%
\vspace{-0.8cm}
        \centering
    \setlength{\tabcolsep}{1.1pt}
    \resizebox{\linewidth}{!}{%
    \begin{tabular}{lcccc}
     \multirow{2}{*}[1.3cm]{\rotatebox{90}{\small TI}} & \includegraphics[height=2.3cm, trim={0 0.35cm 0 0.35cm},clip]{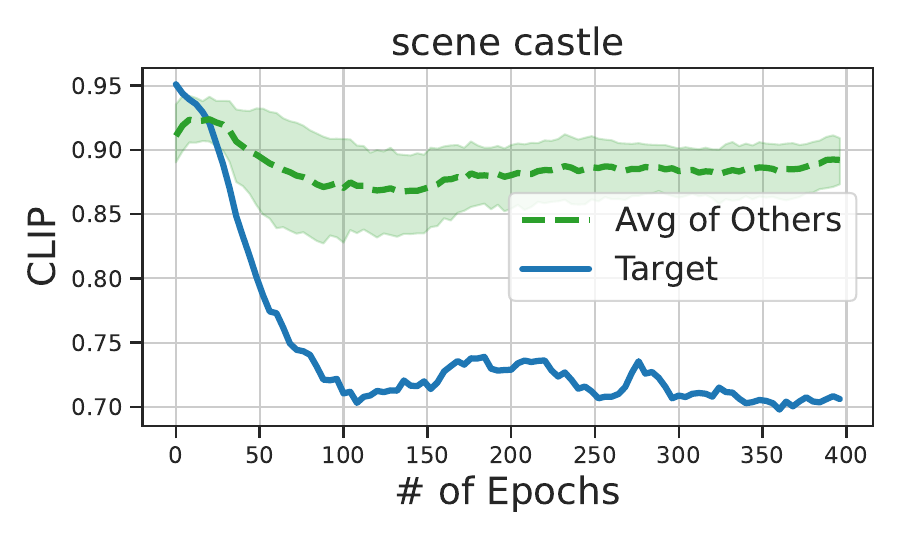} & \includegraphics[height=2.3cm, trim={0 0.35cm 0 0.35cm},clip]{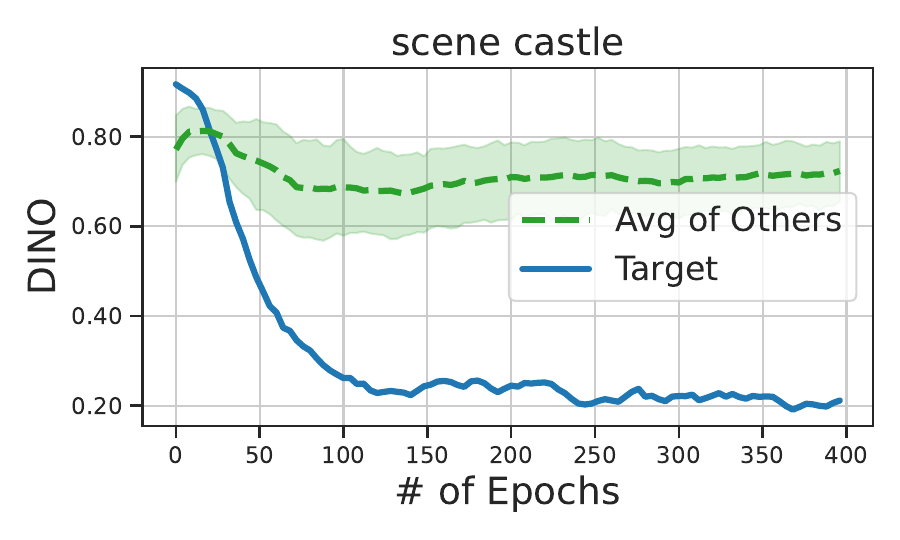} \\ %
     \multirow{2}{*}[1.3cm]{\rotatebox{90}{\small DB}} & \includegraphics[height=2.3cm, trim={0 0.35cm 0 0.35cm},clip]{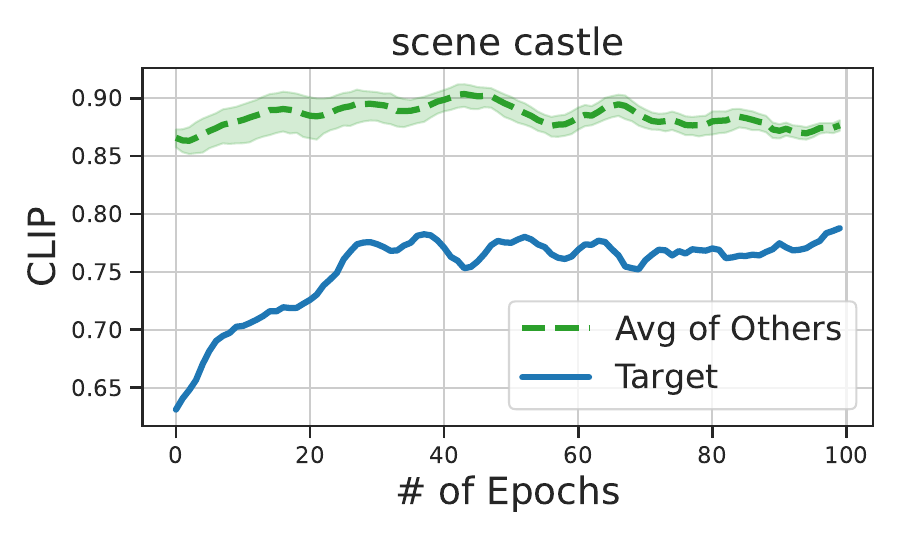} & \includegraphics[height=2.3cm, trim={0 0.35cm 0 0.35cm},clip]{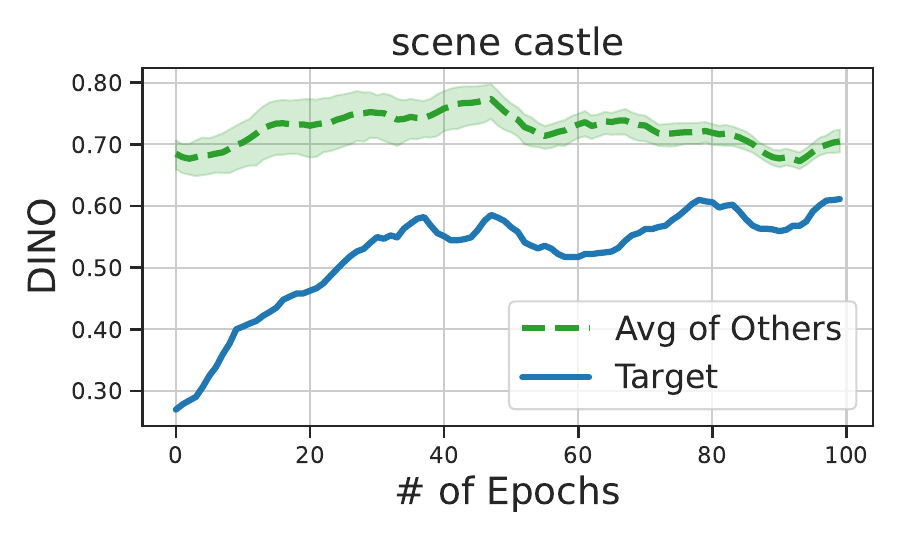} \\ %
     \multirow{2}{*}[1.7cm]{\rotatebox{90}{\small DB Lora}} & \includegraphics[height=2.3cm, trim={0 0.35cm 0 0.35cm},clip]{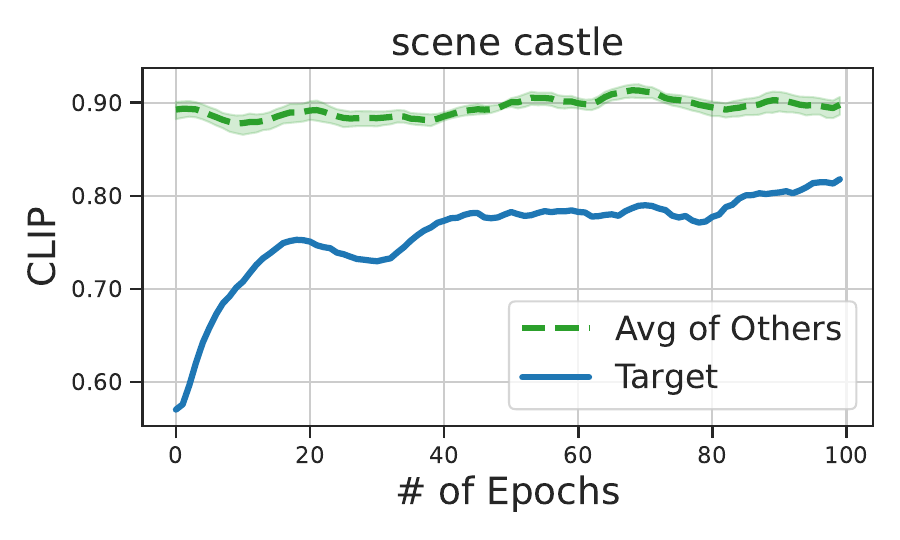} & \includegraphics[height=2.3cm, trim={0 0.35cm 0 0.35cm},clip]{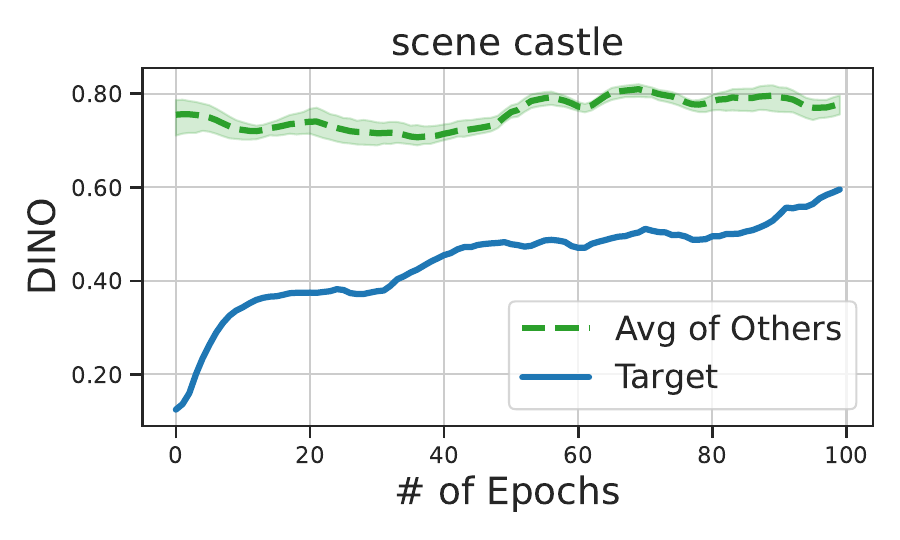} \\ %
    \end{tabular}
    }
    \vspace{-0.4cm}
    \caption{{\bf CLIP and DINO similarity on other concept \vs target concept. } The gap between the two lines shows \rsgr.
    }
    \label{fig:clip_lpips_others}
\end{wrapfigure}%

The term $\gM(\rvx_\gA^o, \rvx_\gI^o)$ captures image similarity for other concepts with and without IMMA. Ideally, this term should be high as IMMA should not affect other concepts. The term $\gM(\rvx_\gA, \rvx_\gI)$ captures the image similarity for target concepts with and without IMMA. In this case, the similarity should be low. \rsgr reports the difference between these two terms as a ratio. Intuitively, larger \rsgr indicates IMMA is better at preserving the other concepts while removing the target concept. 

We also conducted a user study for personalized adaptation using the setting as in~\secref{sec:erase}.
\begin{table}[t]
\caption{ {\bf \sgr(\%) on personalized content adaptation.}
Higher positive values indicate better immunization quality.
}
\vspace{-0.3cm}
\label{tab:personalization_sgr}
\small
\centering
\setlength{\tabcolsep}{1.6pt}
\resizebox{1.0\columnwidth}{!}{
\begin{tabular}{clccccccccccc}
\specialrule{.15em}{.05em}{.05em}
\multicolumn{2}{l}{}                           & castle & car       & chair     & glasses  & instrument      & woodenpot  & lighthouse & motorbike & houseplant & purse   &      \textbf{Average}                    \\ \hline %

\multirow{2}{*}{TI} 
& \texttt{SGR(C)}       & 9.35         & 8.32          & 8.55            & 13.78            & 8.61             & 0.75                 & 7.82             & 12.16               & 15.35                 & 5.80           & 9.05       \\ 
& \texttt{SGR(D)} & 53.49        & 37.08         & 40.19           & 28.50             & 48.85            & 7.04                 & 36.33            & 48.36               & 41.23                 & 24.43         & 36.55               \\
                                   \hline
\multirow{2}{*}{DB} 
&  \texttt{SGR(C)}       & 2.03         & 14.44         & {\color{red}-8.60}            & 0.91             & 2.18             & 5.80                  & {\color{red} -1.19}            & 2.37                & 1.21                  & 3.59          & 2.27                \\
& \texttt{SGR(D)} & 12.56        & 27.01         & {\color{red}-14.67}          & 13.37            & 24.64            & 29.31                & {\color{red} -7.41}            & 0.75                & 9.21                  & 24.97         & 11.97               \\

                                   \hline
\multirow{2}{*}{\makecell{DB\\Lora}}   
& \texttt{SGR(C)}      & 8.60          & 17.49         & 0.25            & 11.73            & 13.66            & 2.16                 & 7.67             & 0.98                & 0.05                  & 3.58          & 6.62   \\ 
& \texttt{SGR(D)} & 43.21        & 34.37         & 8.94            & 37.63            & 40.38            & 32.70                 & 48.36            & 6.40                 & 24.88                 & 0.52          & 27.74               \\
\specialrule{.15em}{.05em}{.05em}           
\end{tabular}
}
\vspace{-0.3cm}
\end{table}

\begin{table}[t]
\caption{
{\bf \rsgr(\%) on personalized content adaptation.} Higher positive values indicate better performance at maintaining other concepts.
}
\vspace{-0.3cm}
\label{tab:personalization_rsgr}
\small
\centering
\setlength{\tabcolsep}{1.4pt}
\resizebox{1.0\columnwidth}{!}{
\begin{tabular}{clccccccccccc}
\specialrule{.15em}{.05em}{.05em}
\multicolumn{2}{l}{}                           & castle & car       & chair     & glasses  & instrument      & woodenpot  & lighthouse & motorbike & houseplant & purse   &      \textbf{Average}           \\ %
\hline
\multirow{2}{*}{TI}  %
& \rsgr\texttt{(C)}      & 15.24        & 20.19         & 8.49            & 9.78             & 19.14            & 8.89                 & 20.57            & 18.59               & 16.32                 & 0.99          & 13.82 \\
& \rsgr\texttt{(D)}      & 70.69        & 63.57         & 22.5            & 40.1             & 57.66            & 34.15                & 51.85            & 67.18               & 55.22                 & 8.88          & 47.18 \\ 
\hline
\multirow{2}{*}{DB} %
& \rsgr\texttt{(C)}      & 19.66        & 13.63         & 6.62            & 4.97             & 4.44             & 4.61                 & 16.53            & 6.08                & 5.12                  & 10.36         & 9.20   \\
& \rsgr\texttt{(D)}      & 36.99        & 35.98         & 17.72           & 32.04            & 24.59            & 23.28                & 35.06            & 2.60                 & 25.05                 & 50.1          & 28.34 \\ %
\hline
\multirow{2}{*}{\makecell{DB\\LoRA}}    %
& \rsgr\texttt{(C)}       & 17.55        & 21.00          & 4.49            & 3.89             & 12.24            & 8.42                 & 24.28            & 8.67                & 11.77                 & 4.83          & 11.71 \\
& \rsgr\texttt{(D)}       & 49.66        & 41.57         & 18.9            & 25.41            & 29.92            & 35.92                & 53.12            & 28.63               & 41.02                 & 24.74         & 34.89 \\   \specialrule{.15em}{.05em}{.05em}           
\end{tabular}
}
\vspace{-0.5cm}

\end{table}

{\bf\noindent Results on personalization.}
In~\tabref{tab:personalization_sgr} we observe a positive ratio among most of the \texttt{SGR} metrics, except for ``furniture chair'' and ``lighthouse'' with Dreambooth highlighted in {\color{red} red}. We show the generations with negative \sgr in 
the appendix. Overall, IMMA effectively prevents the pre-trained model from
learning personalization concepts across the three adaptation methods. 

Next,~\tabref{tab:personalization_rsgr} reports \rsgr to evaluate how well IMMA preserves the ability to personalize other concepts. As we can see, the \rsgr values are consistently positive across all the datasets. This indicates IMMA immunizes against the target concept without hurting the adaptability for personalizing for other concepts. We directly visualize this relative gap in~\figref{fig:clip_lpips_others}. As shown,  the lines of the nine concepts {\color{immao} ($\gM(\rvx_\gA^o, \rvx_\gI^o)$ $\uparrow$ in green) 
} and the adaptation of the target concept {\color{imma} ($\gM(\rvx_\gA, \rvx_\gI)$ $\downarrow$ in  blue)} 
can be easily distinguished; consistent with results in~\tabref{tab:personalization_rsgr}.

\begin{figure}[t]
    \centering
    \small
    \setlength{\tabcolsep}{1.3pt}
    \renewcommand{\arraystretch}{1.2}
    \resizebox{0.71\linewidth}{!}{%
    \begin{tabular}{c@{\hskip 6pt}cccc}
      Reference & \quad TI & DB & DB LoRA\\
    
    \includegraphics[height=1.9cm]{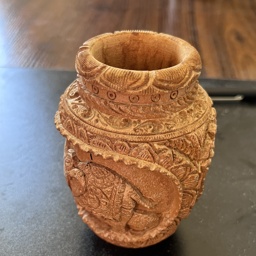} &
    \multirow{1}{*}[1.45cm]{\rotatebox[origin=c]{90}{\color{noimma} w/o IMMA}}
    \includegraphics[height=1.9cm]
    {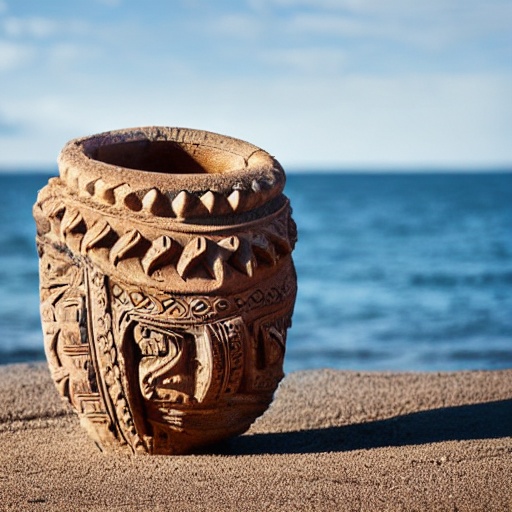} & \includegraphics[height=1.9cm]{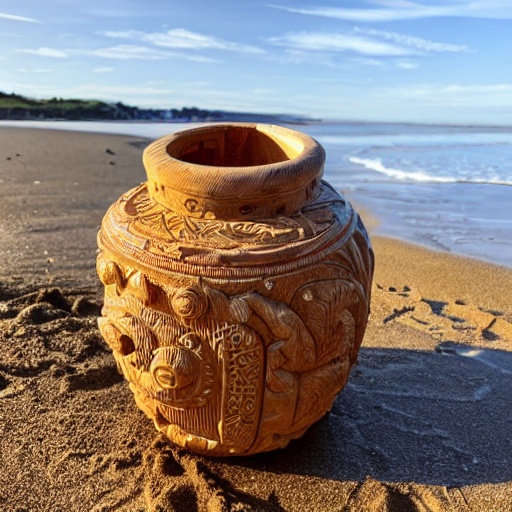} & \includegraphics[height=1.9cm]{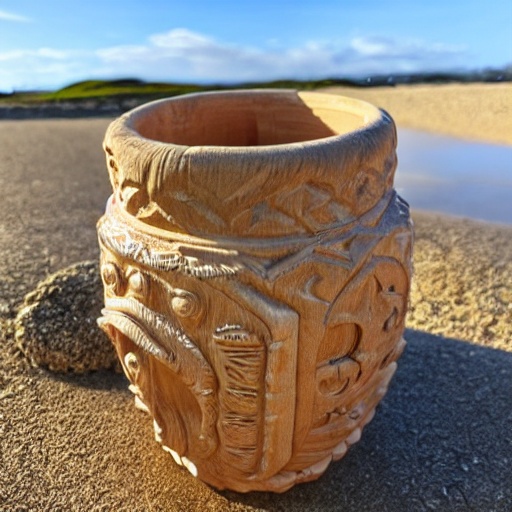} \\
    \includegraphics[height=1.9cm]{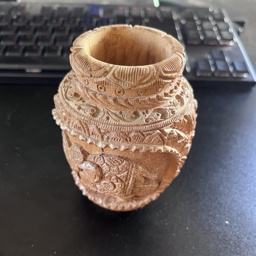} &
     \multirow{2}{*}[1.35cm]{\rotatebox[origin=c]{90}{\color{imma} w/ IMMA}}
     \includegraphics[height=1.9cm]{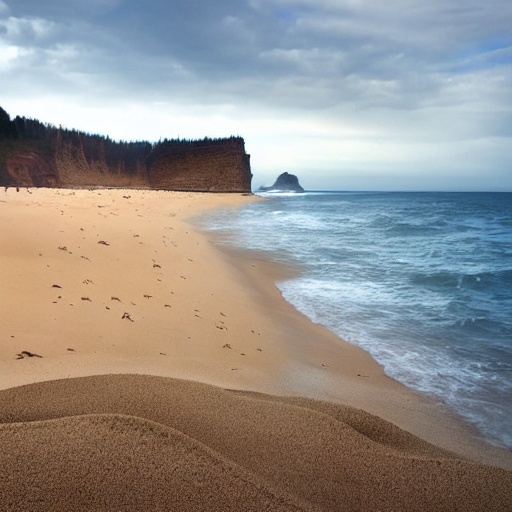} & \includegraphics[height=1.9cm]{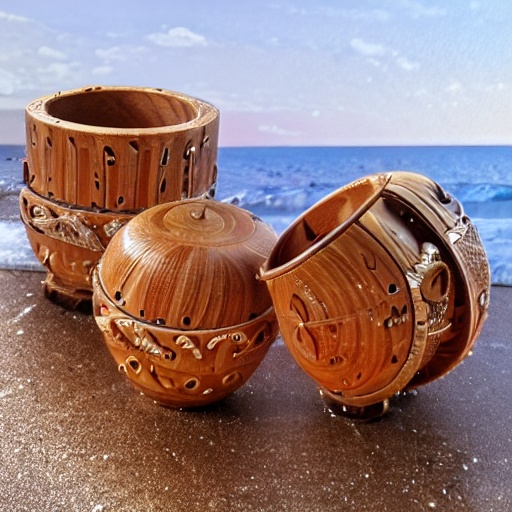} & \includegraphics[height=1.9cm]{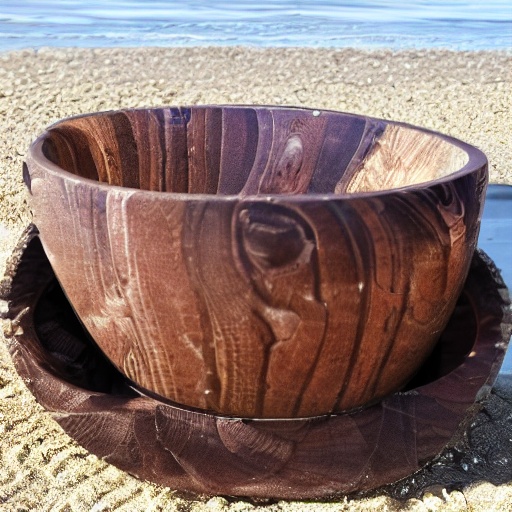} \\
    \hdashline\noalign{\vskip 0.8ex}
    
    \includegraphics[height=1.9cm]{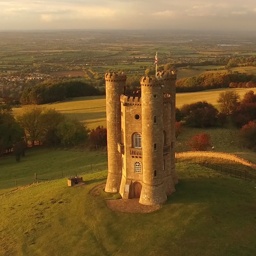} &
    \multirow{1}{*}[1.45cm]{\rotatebox[origin=c]{90}{\color{noimma}w/o IMMA}}
    \includegraphics[height=1.9cm]{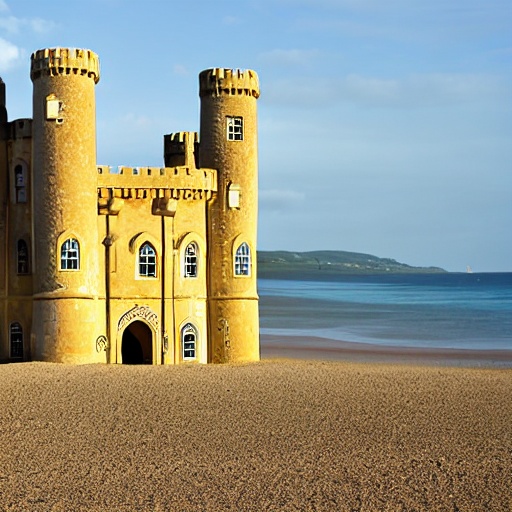} & \includegraphics[height=1.9cm]{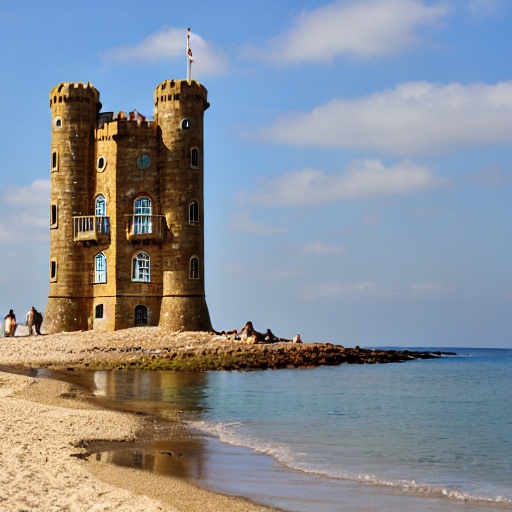} & \includegraphics[height=1.9cm]{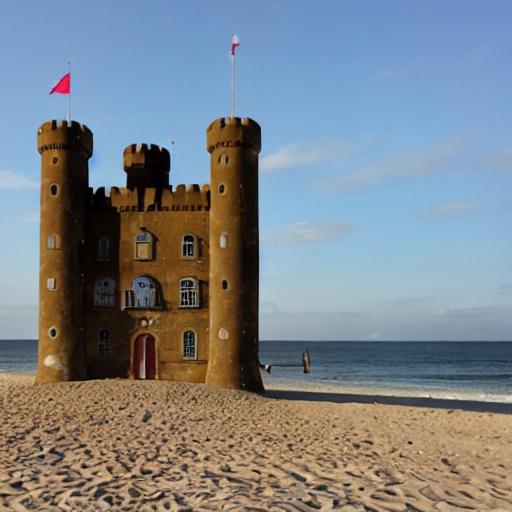} \\
    \includegraphics[height=1.9cm]{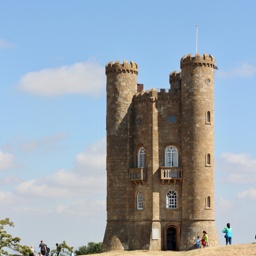} &
     \multirow{2}{*}[1.35cm]{\rotatebox[origin=c]{90}{\color{imma}w/ IMMA}}
     \includegraphics[height=1.9cm]{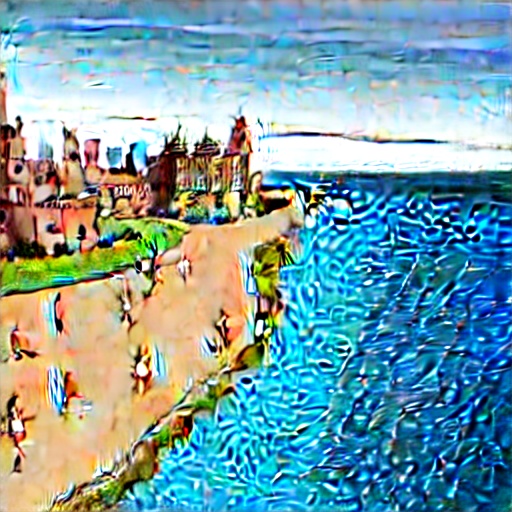} & \includegraphics[height=1.9cm]{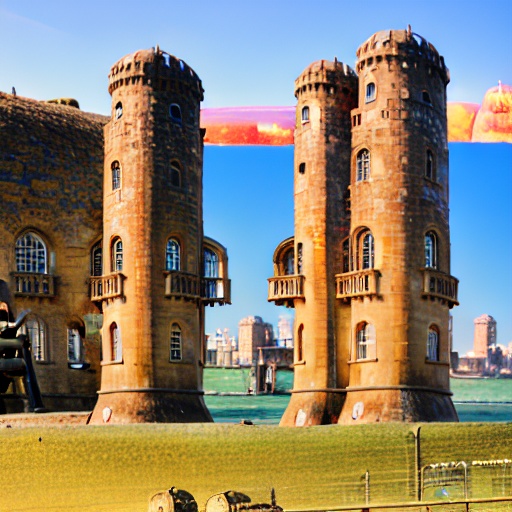} & \includegraphics[height=1.9cm]{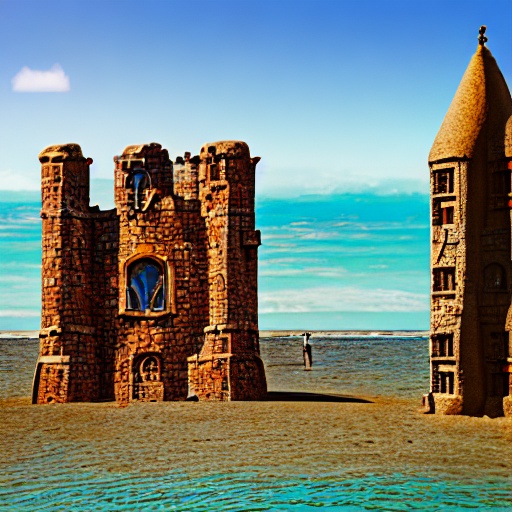}
    \end{tabular}
    }
    \vspace{-0.3cm}
    \caption{{\bf Personalization adaptation w/ and w/o IMMA.
    }
    }
    \label{fig:finetune_qual}
        \vspace{-0.6cm}
\end{figure}

Lastly, we show the generated images in~\figref{fig:finetune_qual}. Comparing the generated images with and without IMMA, we observe models with IMMA are either unable to learn the target concept or generate unrealistic images. 
We also conducted a user study to validate this observation. All of the 30 participants selected generation without IMMA to be more similar to reference images, \ie, models after IMMA fail to generate the target concepts. %

\vspace{-.1cm}

\begin{figure}[t]
\setlength{\tabcolsep}{2pt}
\centering
\small
\begin{tabular}{c@{\hskip 4pt}ccc}
Reference & & {\color{mist} w/ MIST} & {\color{imma} w/ IMMA} \\
\includegraphics[width=0.19\linewidth]{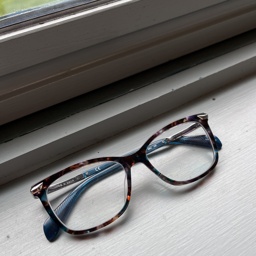} & 
\multirow{1}{*}[1.45cm]{\rotatebox[origin=c]{90}{Original}} &
\includegraphics[width=0.19\linewidth]{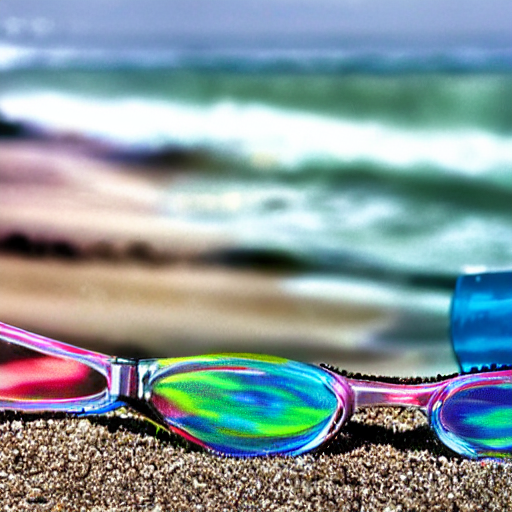} &
\includegraphics[width=0.19\linewidth]{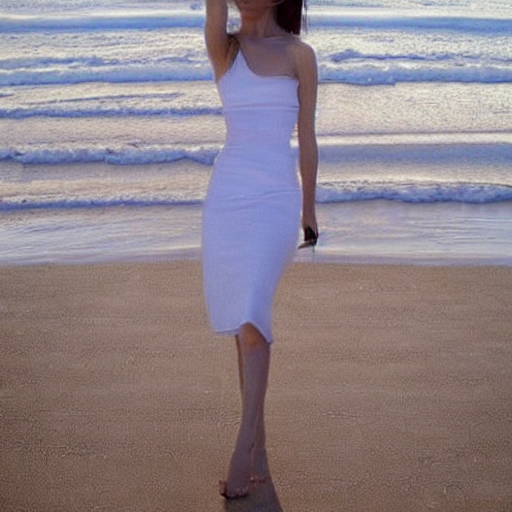} \\
\includegraphics[width=0.19\linewidth]{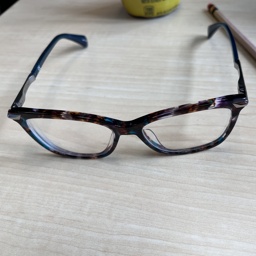} & 
\multirow{1}{*}[1.45cm]{\rotatebox[origin=c]{90}{JPEG}} &
\includegraphics[width=0.19\linewidth]{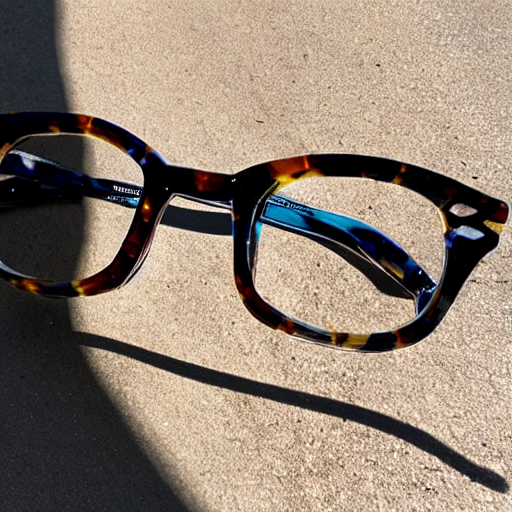} &
\includegraphics[width=0.19\linewidth]{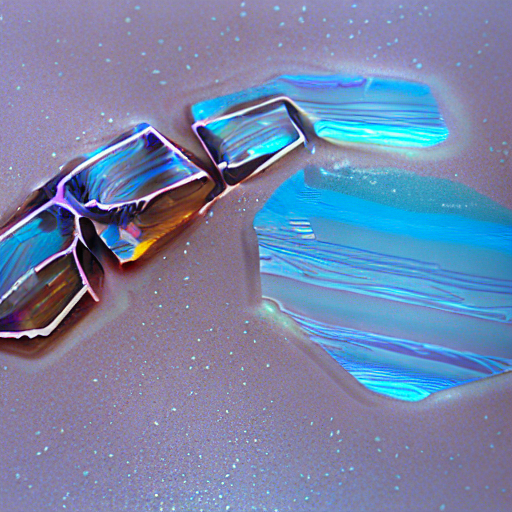} \\
\end{tabular}
\vspace{-0.38cm}
\caption{\textbf{MIST vs. IMMA.} Generation with Textual Inversion on training images w/ and w/o JPEG compression.
}
\vspace{-0.38cm}
\label{fig:mist}
\end{figure}

\begin{figure}[t]
\setlength{\tabcolsep}{2pt}
\centering
\small
\begin{tabular}{ccc}
Reference & Maximization & {\color{imma} IMMA} \\
\includegraphics[width=0.19\linewidth]{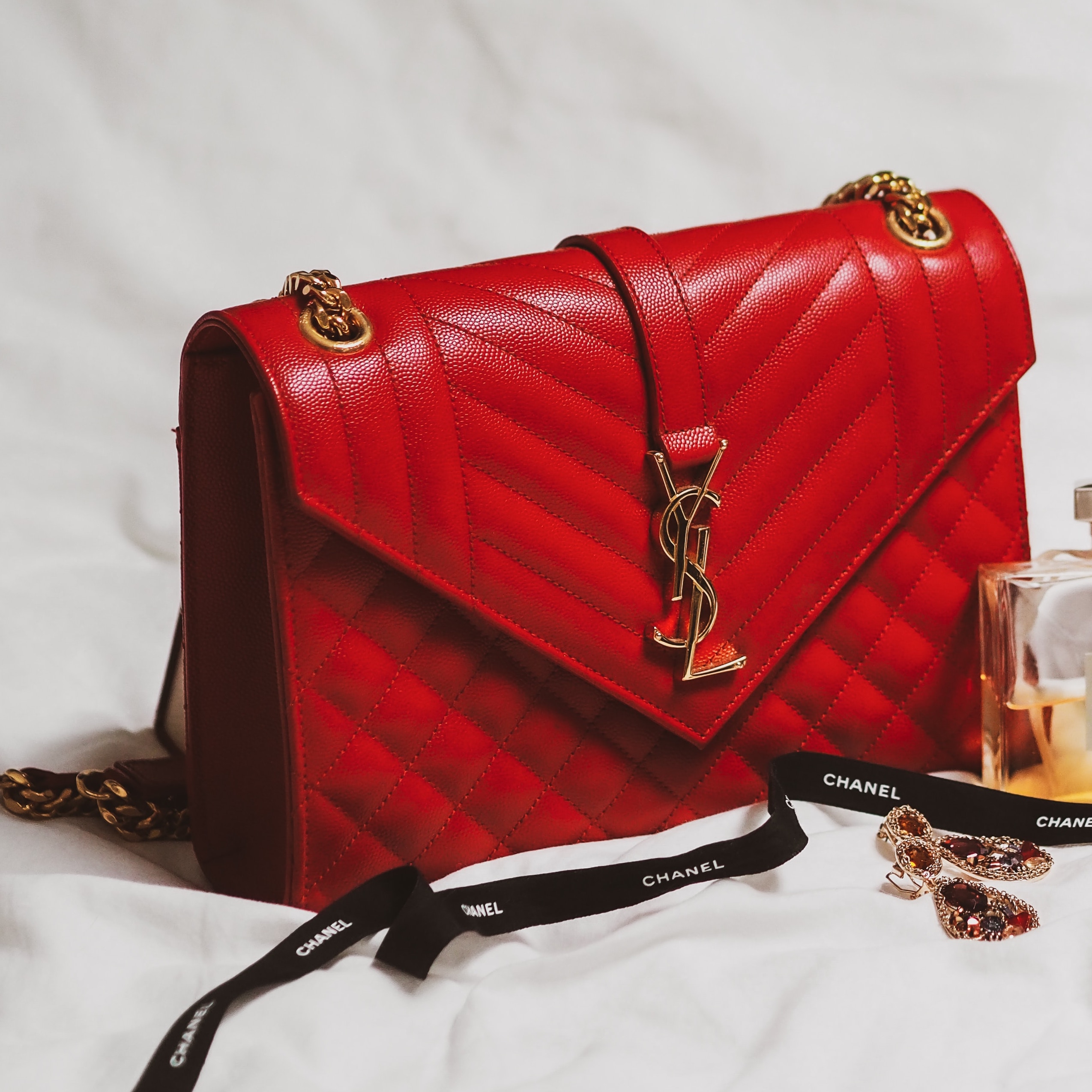} &
\includegraphics[width=0.19\linewidth]{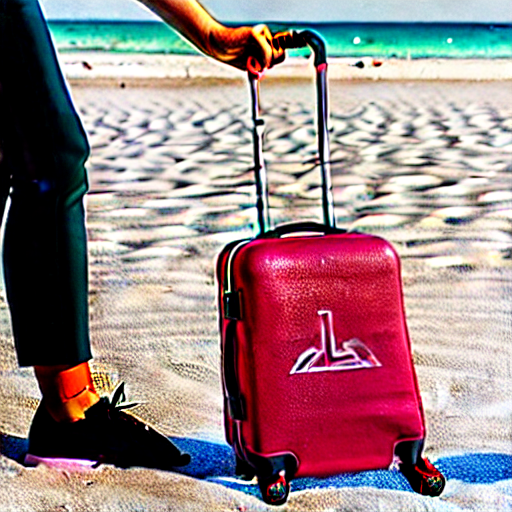} &
\includegraphics[width=0.19\linewidth]{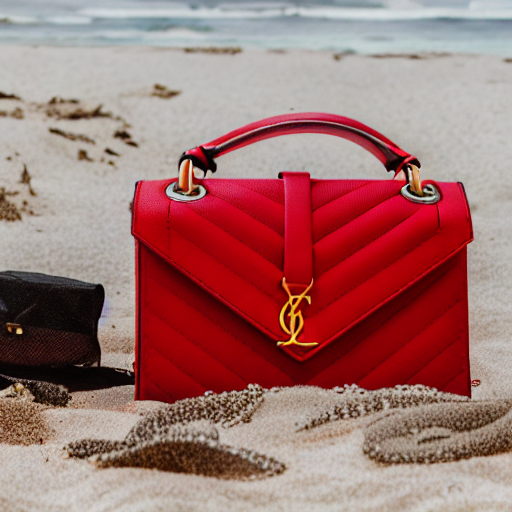}
\end{tabular}
\vspace{-0.38cm}
\caption{\textbf{Ablation on direct maximization.} DreamBooth adaptation on \textit{``luggage purse''} after immunization on \textit{``woodenpot''}.
}
\vspace{-0.7cm}
\label{fig:maximization}
\end{figure}

\subsection{Additional discussion}
\label{sec:ablation}

{\bf\noindent Comparison with data poisoning.} We compare IMMA with MIST~\cite{liang2023adversarial}, one of the data poisoning (DP) methods in the personalized content setup. In~\figref{fig:mist} (top-row) we show both MIST and IMMA successfully prevent the model from learning the target concept after Textual Inversion.

This observation is also supported by a user study. On seven out of ten evaluated datasets, the majority of the 30 users found the generation without MIST to be of higher quality and more similar to the reference images. 

\myparagraph{Adaption on images with JPEG compression.}
As reported in MIST~\cite{liang2023adversarial}, one way to weaken the effect of poisoned data is by compressing the images with JPEG after adaptation. On the other hand, IMMA can defend against JPEG compression by  
including JPEG images in the training data. 
In~\figref{fig:mist} (bottom row), we observe that after compression, MIST fails to prevent generating the target concept, while IMMA remains robust against JPEG.

\begin{figure}[t]
\centering
\includegraphics[width=0.48\linewidth, trim={0.4cm 0.5cm 0.4cm 0.4cm},clip]{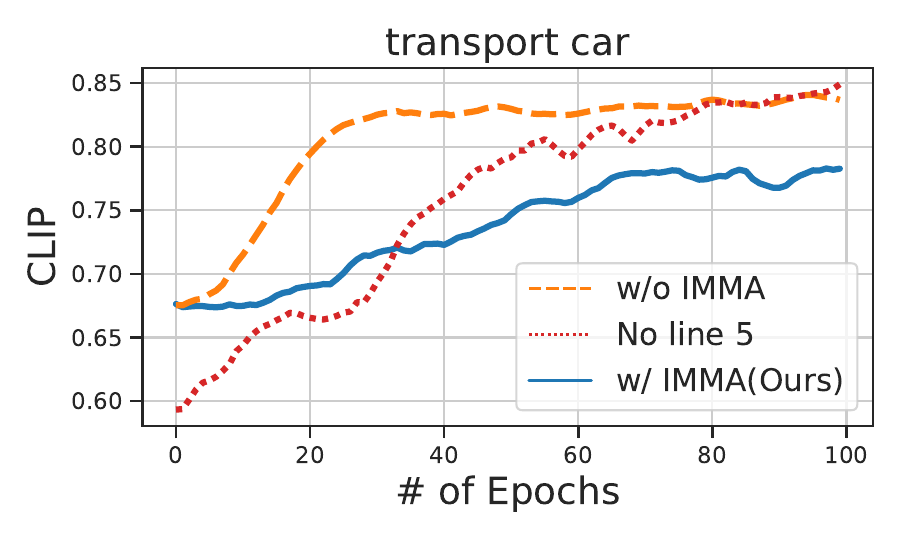}
\includegraphics[width=0.48\linewidth, trim={0.4cm 0.5cm 0.4cm 0.4cm},clip]{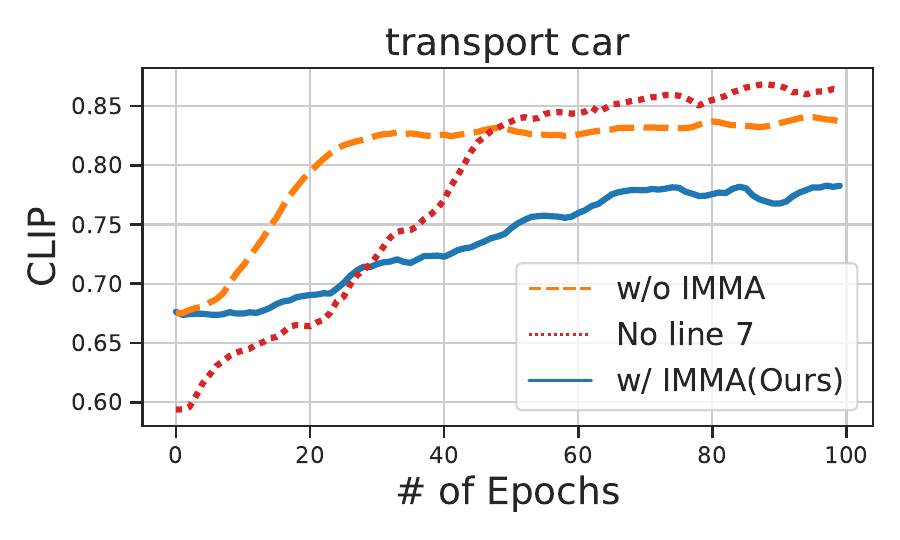}
\vspace{-0.2cm}
\caption{\textbf{Ablation on line 5 and line 7 of~\algref{alg:imma}.} 
}
\vspace{-0.2cm}
\label{fig:ablation}
\end{figure}

\begin{figure}[t]
    \centering
    \setlength{\tabcolsep}{1.3pt}
    \renewcommand{\arraystretch}{1.4}
    \hspace{-0.184cm}
        \resizebox{0.67\linewidth}{!}{%
    \begin{tabular}{lcccc}
     & & \color{noimma} w/o IMMA & \color{imma} w/ IMMA ({\color{black}DB}) & \color{imma} w/ IMMA ({\color{black}TI})\\
    & \multirow{2}{*}[1.4cm]{\rotatebox[origin=c]{90}{\bf DB}}  & \includegraphics[width=0.22\linewidth]{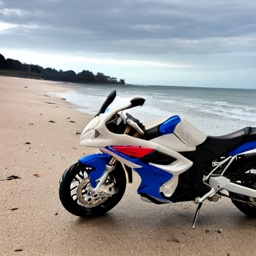} & \includegraphics[width=0.22\linewidth]{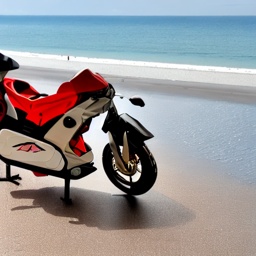}
    & \includegraphics[width=0.22\linewidth]{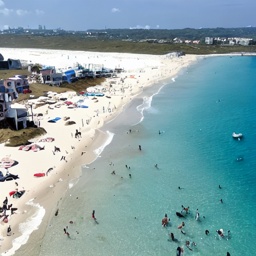} \\
     &\multirow{2}{*}[1.4cm]{\rotatebox[origin=c]{90}{\bf TI}}  & \includegraphics[width=0.22\linewidth]{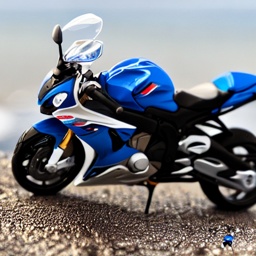} & \includegraphics[width=0.22\linewidth]{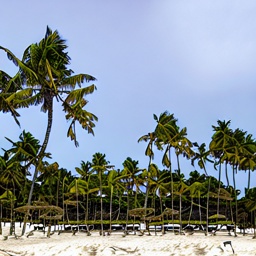}
    & \includegraphics[width=0.22\linewidth]{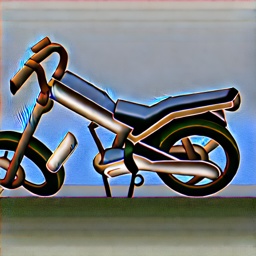}
    \end{tabular}
    }
    \vspace{-0.18cm}
    \caption{\textbf{Results on crossed adaptation immunization.} \textit{First row:} DB after IMMA with either DB or TI. \textit{Second row:} TI after IMMA the model with either DB or TI.}
    \vspace{-0.4cm}
    \label{fig:finetune_cross}
\end{figure}

\myparagraph{Ablation studies.}
We conduct ablations using the DreamBooth personalization setup in~\secref{sec:person}. First, we experiment with a direct maximization baseline, \ie, only the upper-level task in~\equref{eq:bilevel}. Results are shown in~\figref{fig:maximization}. We observe that direct maximization ruins the immunized model, \ie, low image quality when being adapted for another concept. %

Next, we ablate line 5 and line 7 in~\algref{alg:imma} and report CLIP similarity. 
Shown in~\figref{fig:ablation} (left), without line 5, the adaptation can learn the target concept, \ie, high CLIP similarity. Next, we ablate whether to update the overlapping parameters in $\phi$ by removing line 7. The result is shown in~\figref{fig:ablation} (right). We observe that without line 7, the adaptation successfully learns the target concept. These results show the necessity of the proposed steps in~\algref{alg:imma}.

\myparagraph{Crossed adaptation with IMMA.} Thus far, we report results for IMMA by immunizing the model against \textit{the same} adaptation method $\gA$. We now investigate whether IMMA remains effective under a \textit{different} adaptation method during IMMA and adaptation. In other words, we consider IMMA with {\it crossed adaptation methods}, where we immunize the pre-trained model using $\gA_1$ and perform malicious adaptation on $\gA_2$. \figref{fig:finetune_cross} shows the qualitative results across DB and TI. We observe that the model with IMMA against DreamBooth is also effective when being adapted with Textual Inversion, and vice versa.

\myparagraph{Limitations.} Our proposed IMMA and experiments focus on the immunization of a single concept. In this work, we methodically study a variety of adaptation settings and leave the support for multiple concepts for future work. We believe that results in a single concept demonstrate a convincing step towards model immunization. If the reader is interested, we present preliminary results in the Appendix showing that IMMA can generalize to multiple concepts, albeit, not as comprehensive as the single concept study.

\section{Conclusion}
We propose to Immunize Models against Malicious Adaptation (IMMA).
Unlike the data-poisoning paradigm, which protects images, our method focuses on protecting pre-trained models from being used by adaptation methods. We formulate "Immunization" as a bi-level optimization program to learn a poor model initialization that would make adaptation more difficult.
To validate the efficacy of IMMA, we conduct extensive experiments on relearning concepts for erased models and immunizing against the adaptation of personalized content. 
We believe that model immunization is a promising paradigm for combatting the risk of malicious adaptation, and that IMMA is an encouraging first step. We are hopeful that the advancement of IMMA will result in safer open-source text-to-image models, benefiting both the research community and society.

\section*{Acknowledgements}

This project is supported in part by an NSF Award \#2420724. We thank Renan A. Rojas-Gomez for proofreading and helpful discussions.

\bibliographystyle{splncs04nat}
\bibliography{ref}

\clearpage
\onecolumn

\appendix
\newcommand{\beginsupplementary}{%
    \setcounter{section}{0}
	\renewcommand{\thesection}{A\arabic{section}}
	\renewcommand{\thesubsection}{\thesection.\arabic{subsection}}

	\renewcommand{\thetable}{A\arabic{table}}%
	\setcounter{table}{0}

	\renewcommand{\thefigure}{A\arabic{figure}}%
	\setcounter{figure}{0}
	
	\setcounter{algorithm}{0}
}
\beginsupplementary

{\noindent\bf \LARGE Appendix
}
\vspace{0.25cm}

\noindent The appendix is organized as follows:
\begin{itemize}[topsep=2pt]
    \item In~\secref{sec:add_results}, we provide additional quantitative results. Additionally, we include results of IMMA against the adaptation on multiple concepts in~\figref{fig:multi_quan_sgr} and~\figref{fig:multi_quan_rsgr}.
    \item In~\secref{sec:supp_qual}, we provide additional qualitative results. We have also included interactive results (in HTML) along with the supplemental materials.
    \item In~\secref{sec:supp_user_study}, we document the details of our conducted user study.
    \item In~\secref{sec:exp_details}, we provide additional experimental details, \eg, model architecture, hyperparameters, and description of baseline. We have also attached the code in the supplementary materials and will release the code.
\end{itemize}

\section{Additional quantitative results}
\label{sec:add_results}

\subsection{Comparison with data poisoning method}

\begin{figure*}[h]
    \centering
    \setlength{\tabcolsep}{1pt}
    \begin{tabular}{cccc}
     \includegraphics[height=1.7cm, trim={0.3cm 0.35cm 0.3cm 0.35cm},clip]{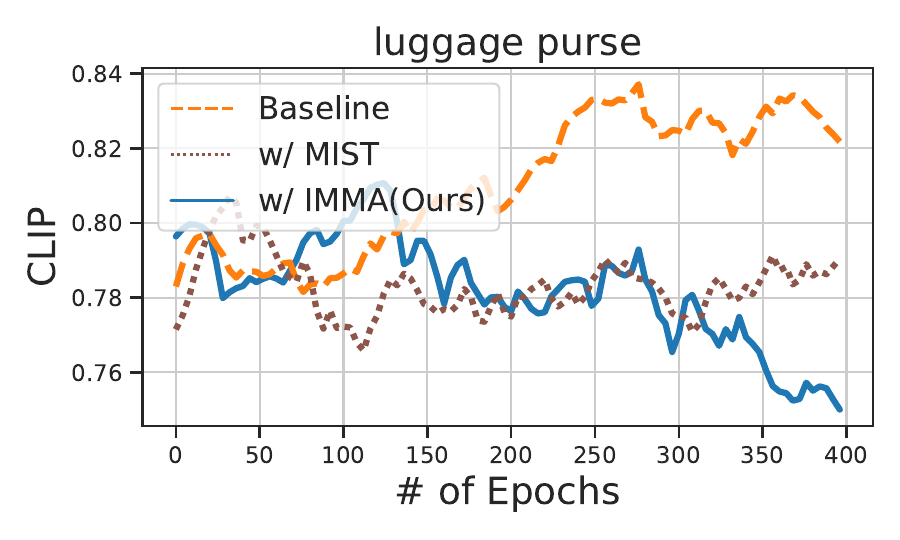} & 
     \includegraphics[height=1.7cm, trim={0.3cm 0.35cm 0.3cm 0.35cm},clip]{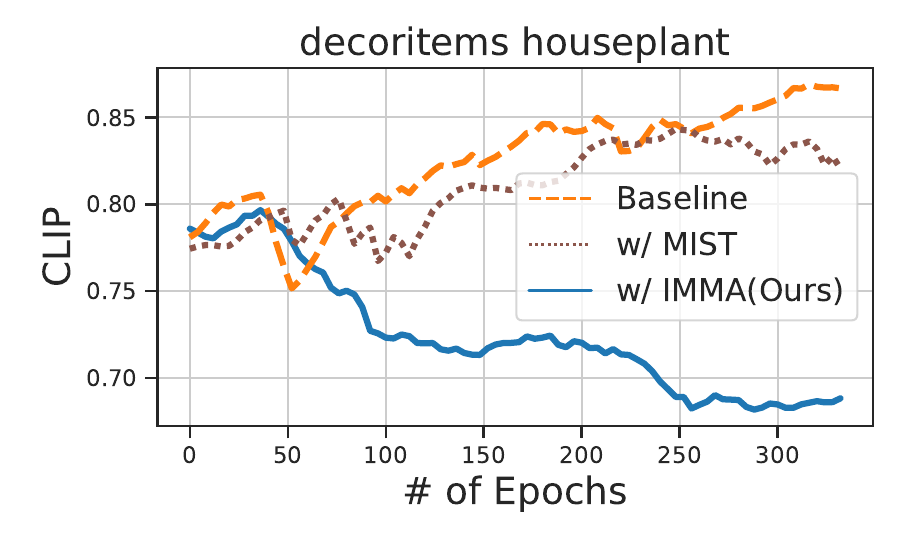} & 
     \includegraphics[height=1.7cm, trim={0.3cm 0.35cm 0.3cm 0.35cm},clip]{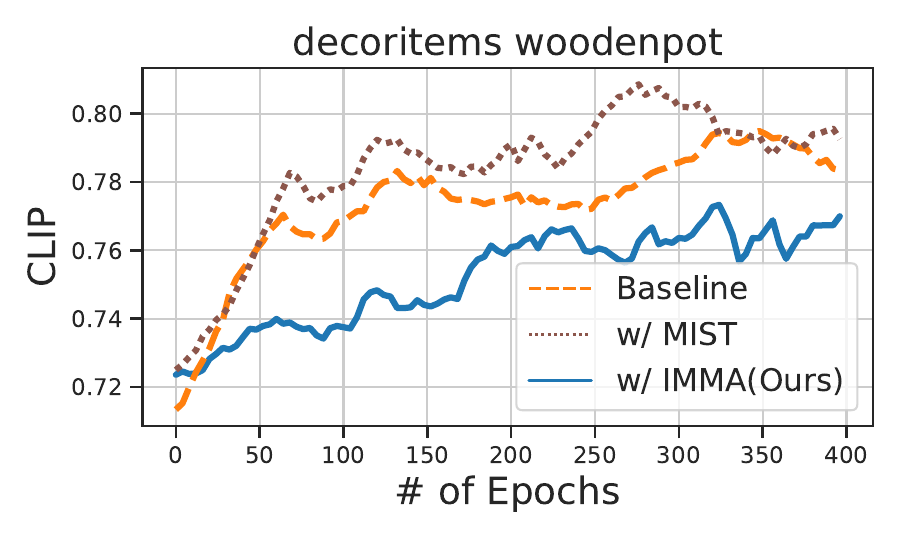} & 
     \includegraphics[height=1.7cm, trim={0.3cm 0.35cm 0.3cm 0.35cm},clip]{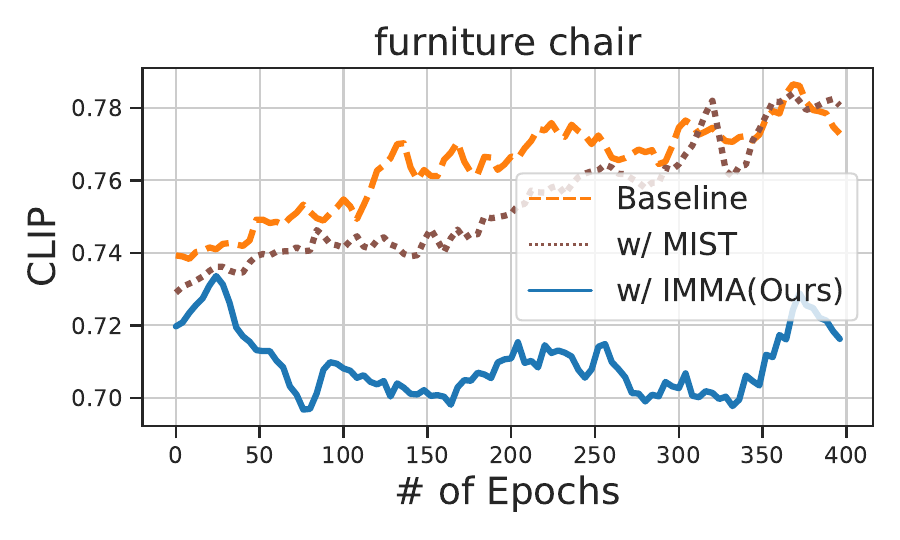} \\
     \includegraphics[height=1.7cm, trim={0.3cm 0.35cm 0.3cm 0.35cm},clip]{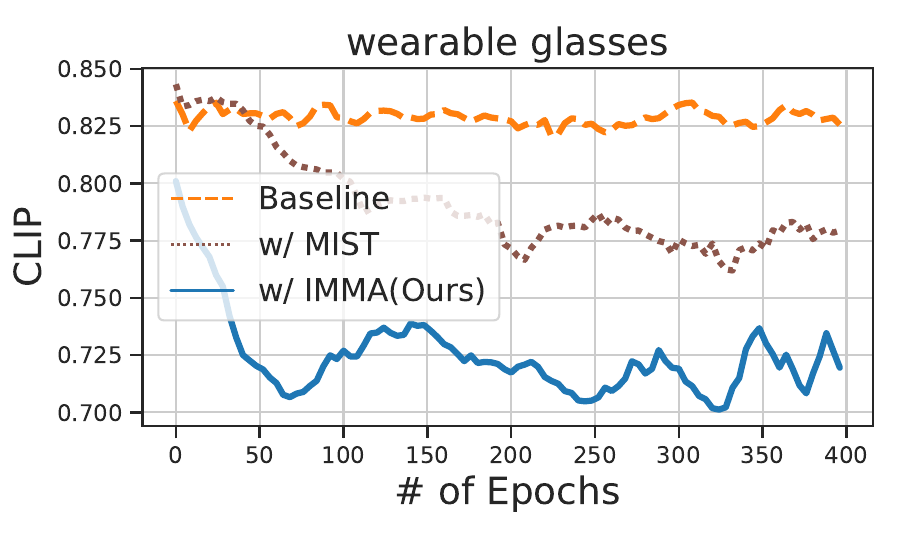} &
     
     \includegraphics[height=1.7cm, trim={0.3cm 0.35cm 0.3cm 0.35cm},clip]{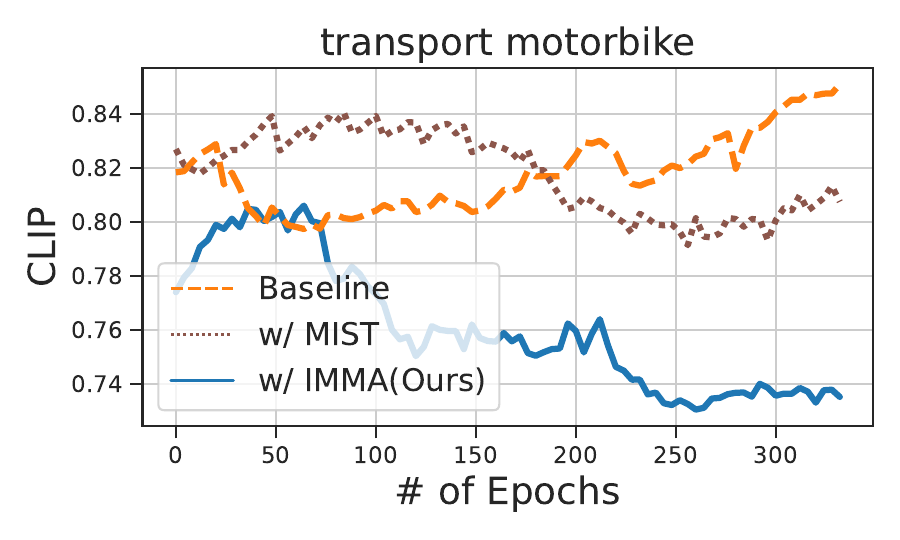} & 
     \includegraphics[height=1.7cm, trim={0.3cm 0.35cm 0.3cm 0.35cm},clip]{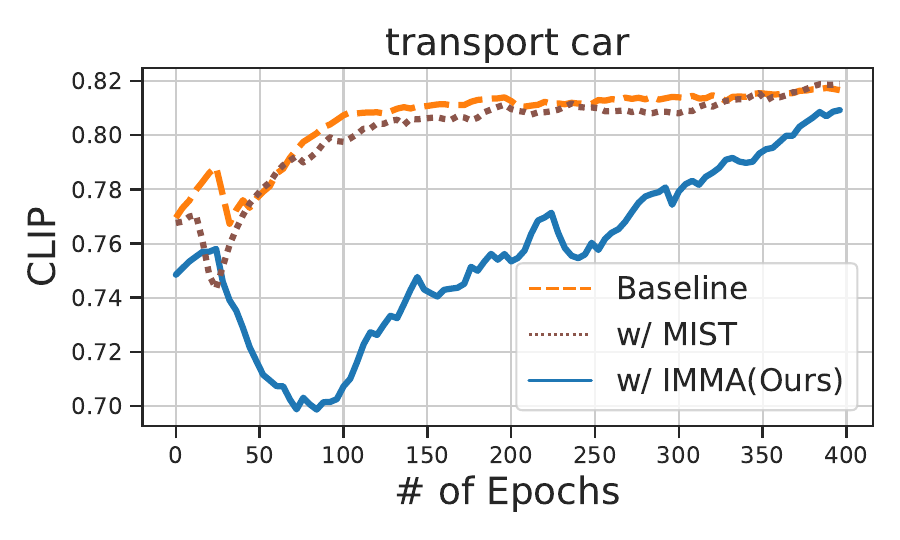} & 
     \includegraphics[height=1.7cm, trim={0.3cm 0.35cm 0.3cm 0.35cm},clip]{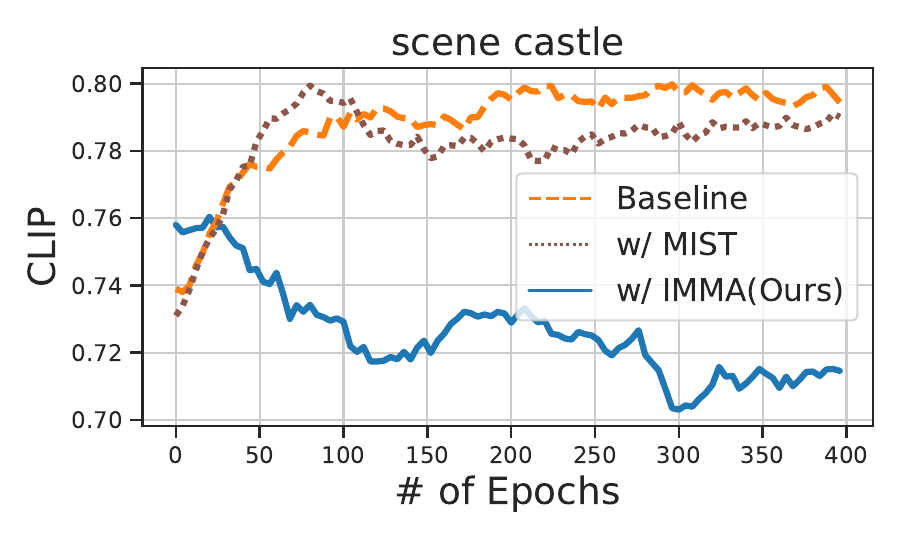} \\
     \includegraphics[height=1.7cm, trim={0.3cm 0.35cm 0.3cm 0.35cm},clip]{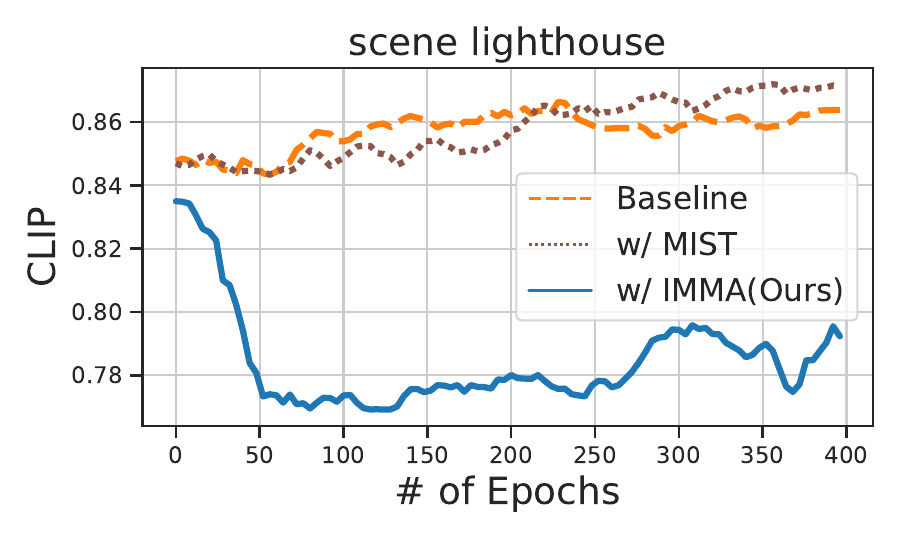}& 
     \includegraphics[height=1.7cm, trim={0.3cm 0.35cm 0.3cm 0.35cm},clip]{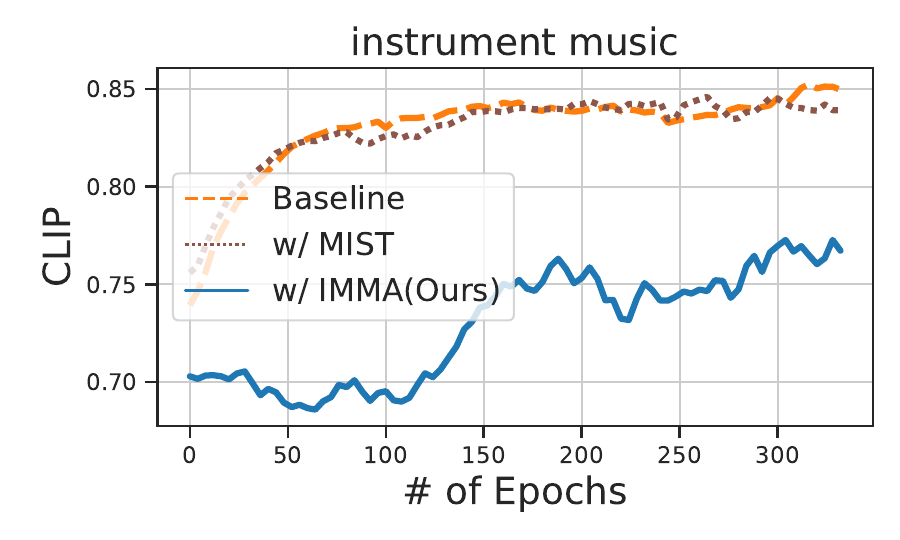} \\
    \end{tabular}
    \caption{\textbf{CLIP versus reference images after Textual Inversion Adaptation.}}
    \label{fig:mist_ti}
\end{figure*}

We provide quantitative results of MIST~\cite{liang2023adversarial}, a data poisoning method for defending against adaptation. We show the CLIP values after Textual Inversion adaptation in~\figref{fig:mist_ti}. We observe a gap between the two lines in \textit{purse} and \textit{glasses} which indicates MIST can prevent the model from learning personalized concepts in some datasets. However, compared with IMMA, MIST is less robust and fails in other datasets,~\eg, \textit{car}.

\subsection{Additional quantitative results of IMMA}

\begin{figure*}[t]
    \centering
    \setlength{\tabcolsep}{0pt}
    \renewcommand{\arraystretch}{1}
    \begin{tabular}{cccc}
     \includegraphics[height=1.7cm, trim={0.3cm 0.35cm 0.3cm 0.35cm},clip]{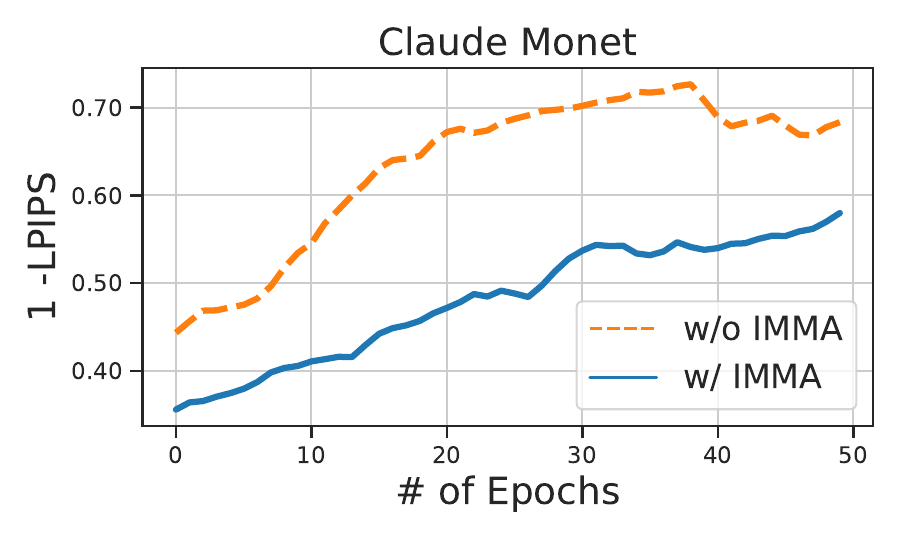} &  \includegraphics[height=1.7cm, trim={0.3cm 0.35cm 0.3cm 0.35cm},clip]{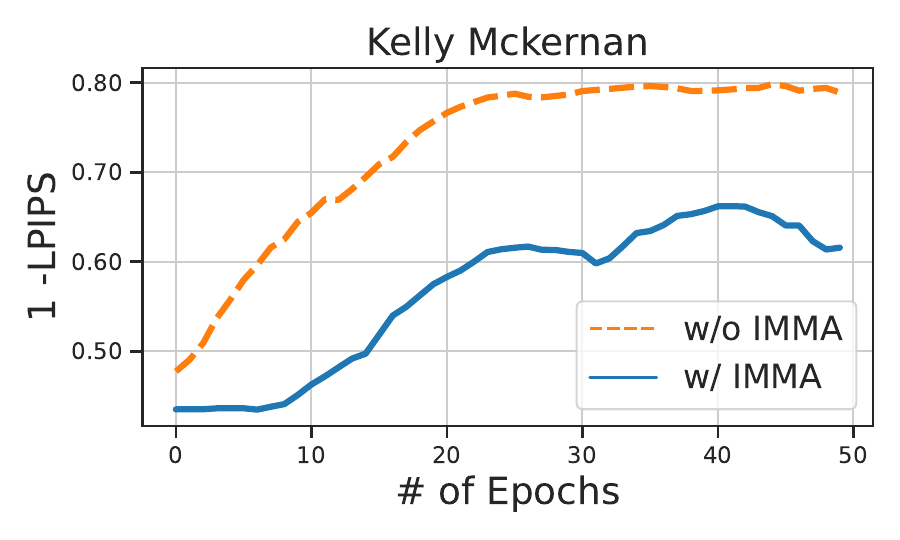} & \includegraphics[height=1.7cm, trim={0.3cm 0.35cm 0.3cm 0.35cm},clip]{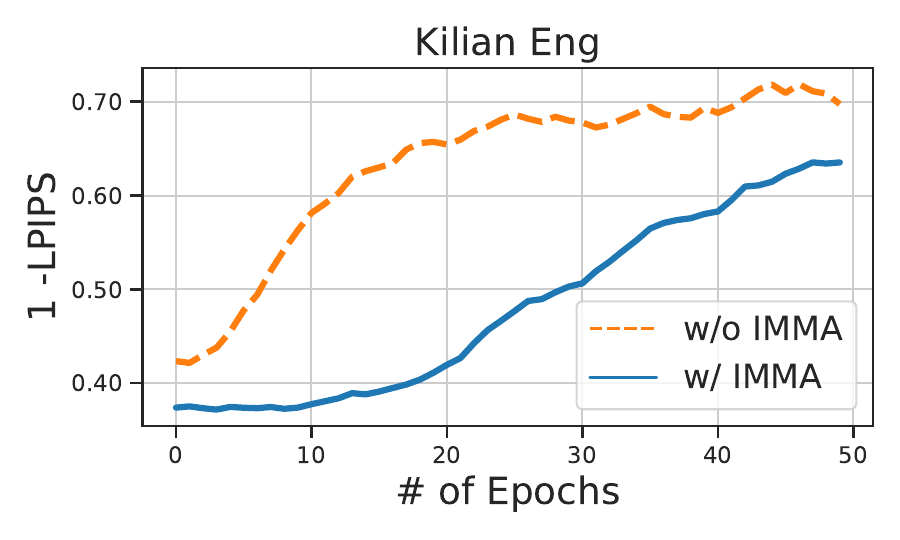}& \includegraphics[height=1.7cm, trim={0.3cm 0.35cm 0.3cm 0.35cm},clip]{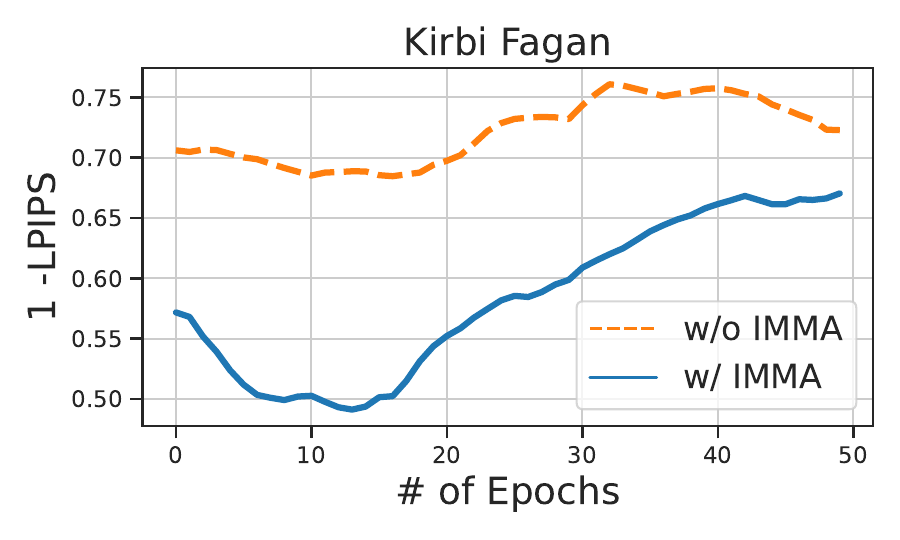} \\
     \includegraphics[height=1.7cm, trim={0.3cm 0.35cm 0.3cm 0.35cm},clip]{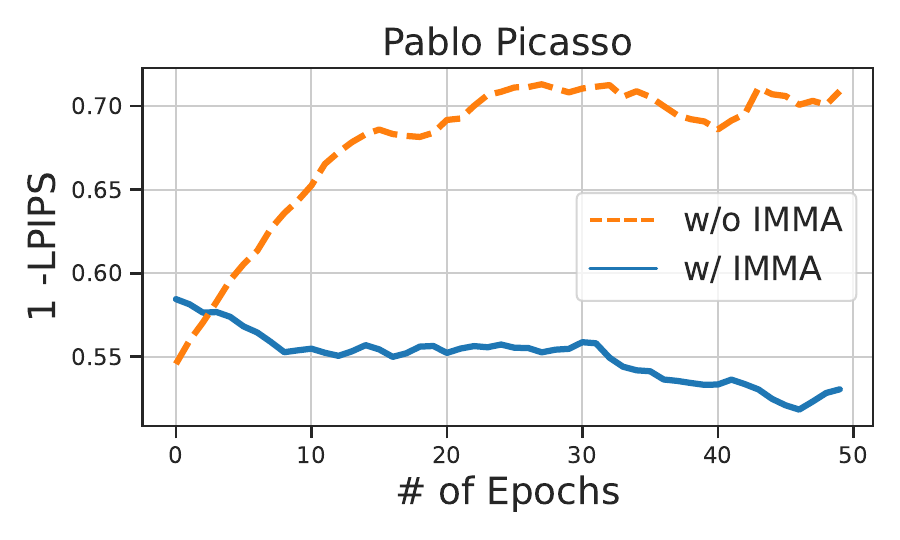} & \includegraphics[height=1.7cm, trim={0.3cm 0.35cm 0.3cm 0.35cm},clip]{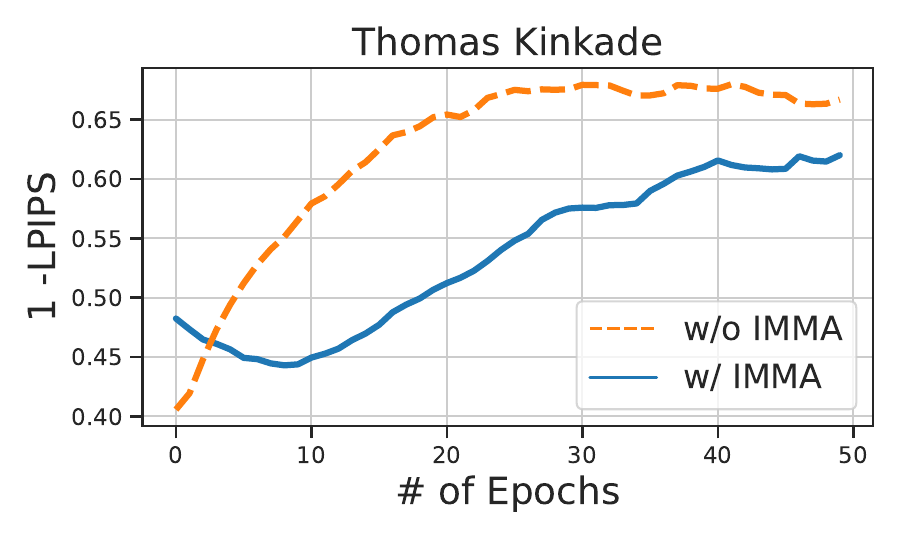} & \includegraphics[height=1.7cm, trim={0.3cm 0.35cm 0.3cm 0.35cm},clip]{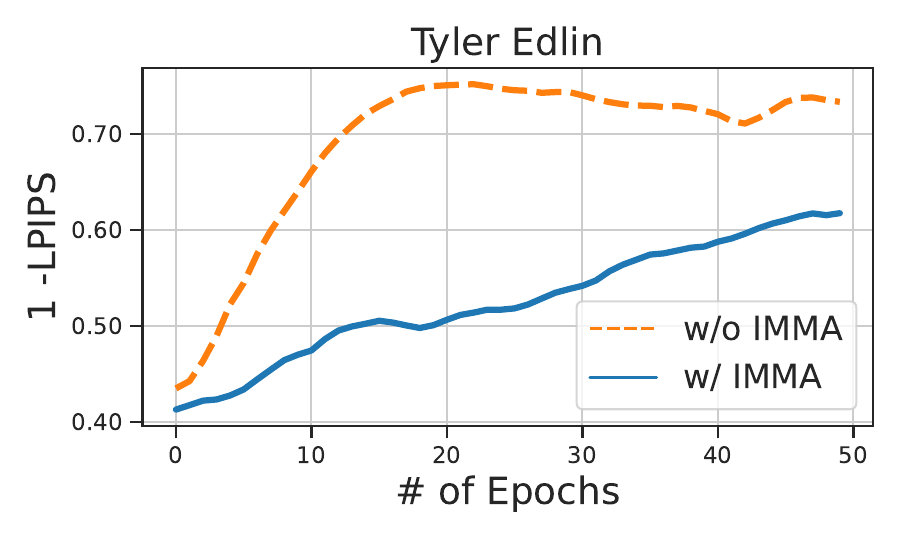} & \includegraphics[height=1.7cm, trim={0.3cm 0.35cm 0.3cm 0.35cm},clip]{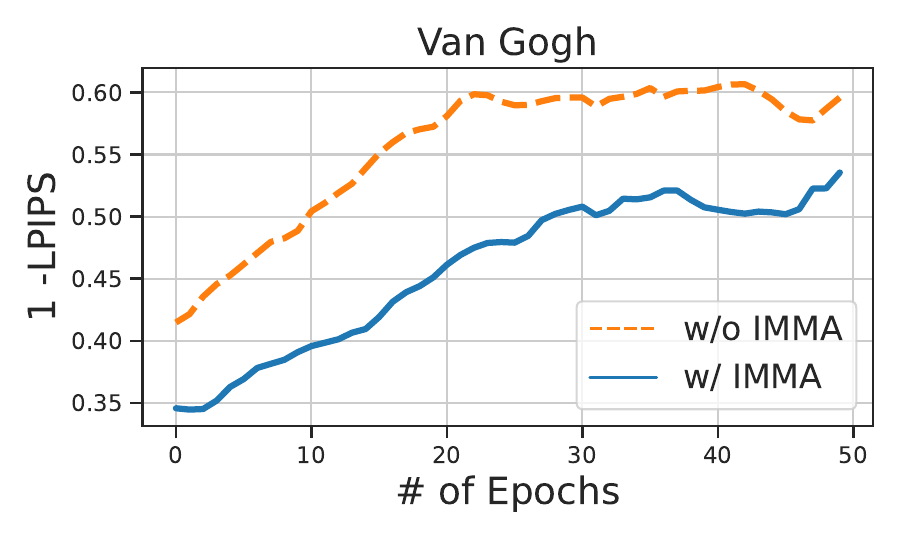}  \\
     \includegraphics[height=1.7cm, trim={0.3cm 0.35cm 0.3cm 0.35cm},clip]{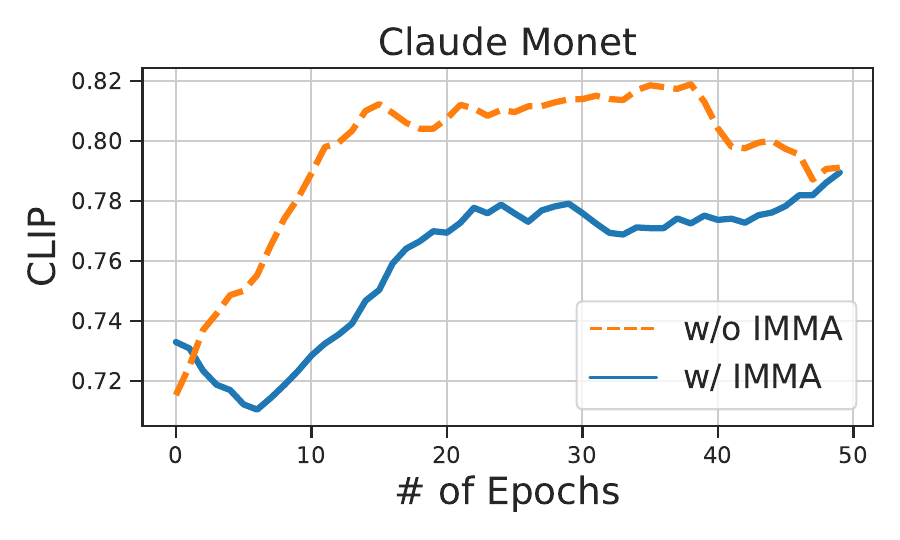} & \includegraphics[height=1.7cm, trim={0.3cm 0.35cm 0.3cm 0.35cm},clip]{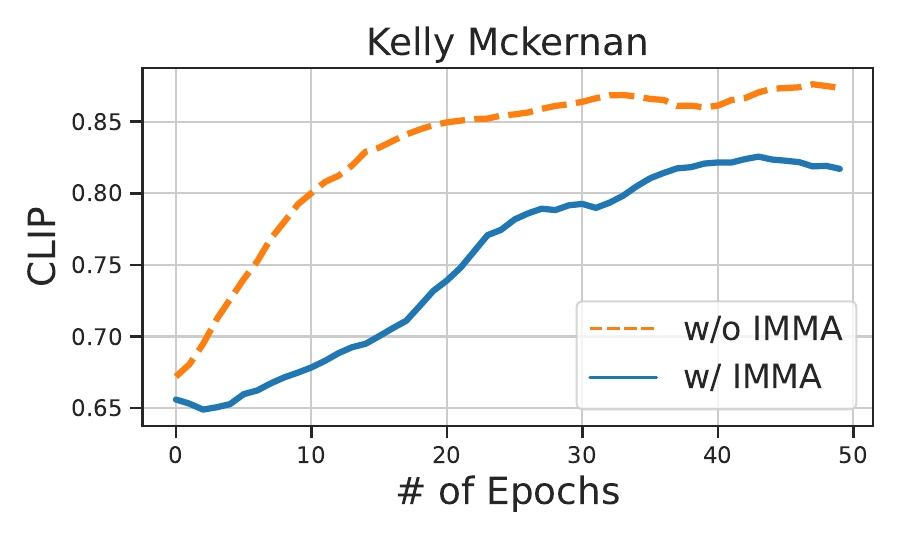} & \includegraphics[height=1.7cm, trim={0.3cm 0.35cm 0.3cm 0.35cm},clip]{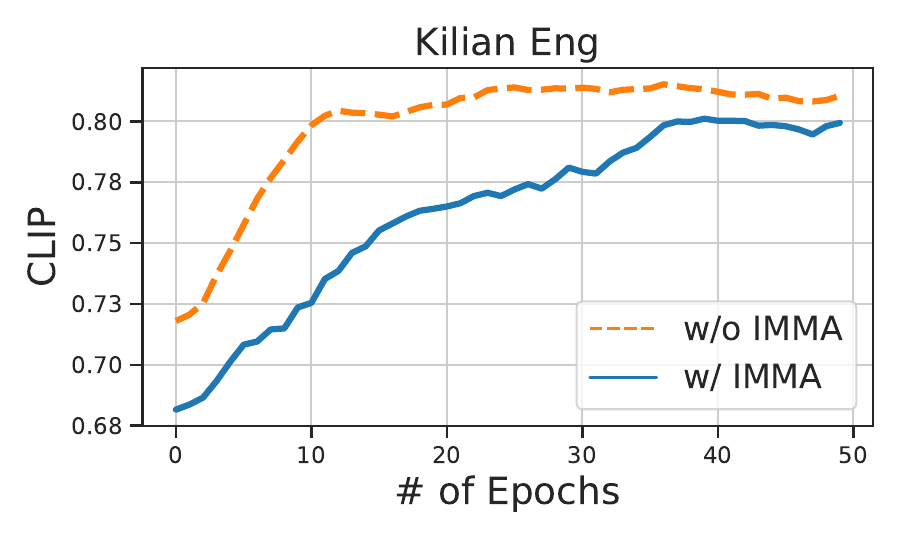}& \includegraphics[height=1.7cm, trim={0.3cm 0.35cm 0.3cm 0.35cm},clip]{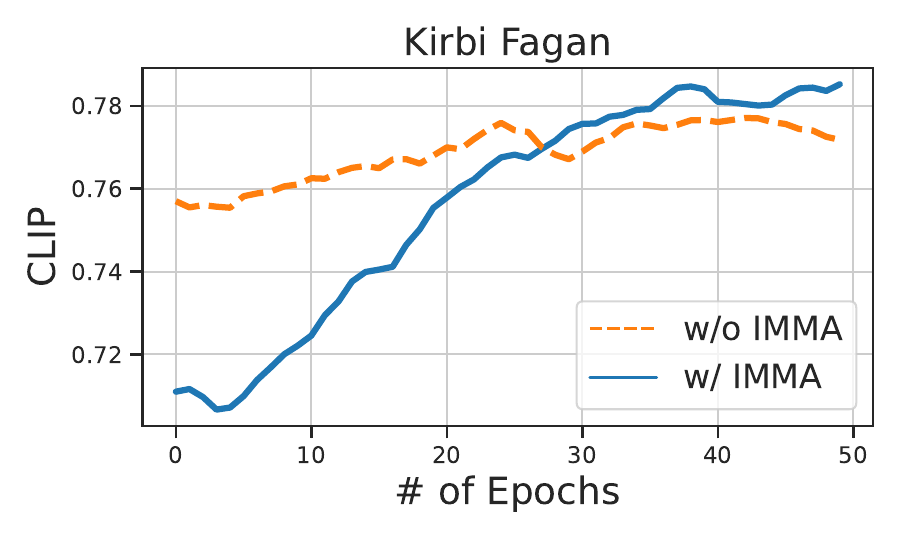} \\
     \includegraphics[height=1.7cm, trim={0.3cm 0.35cm 0.3cm 0.35cm},clip]{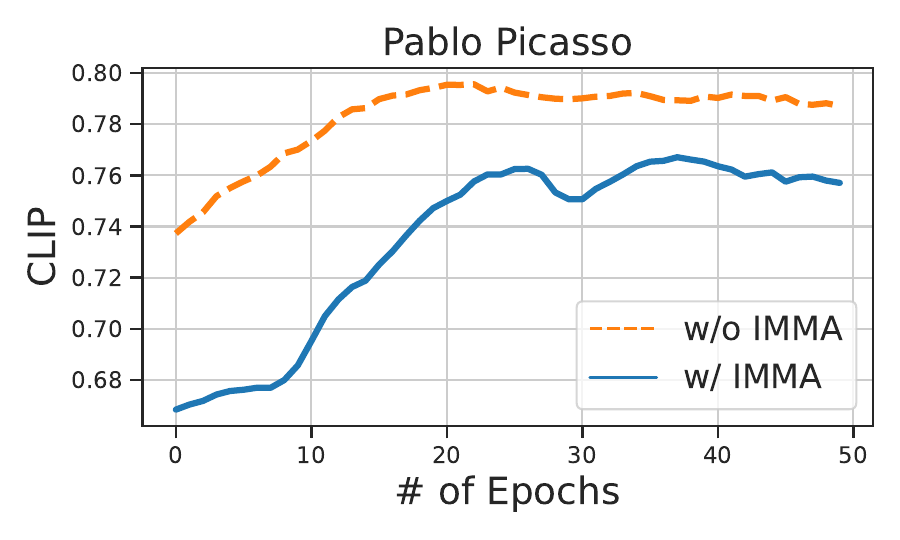} & \includegraphics[height=1.7cm, trim={0.3cm 0.35cm 0.3cm 0.35cm},clip]{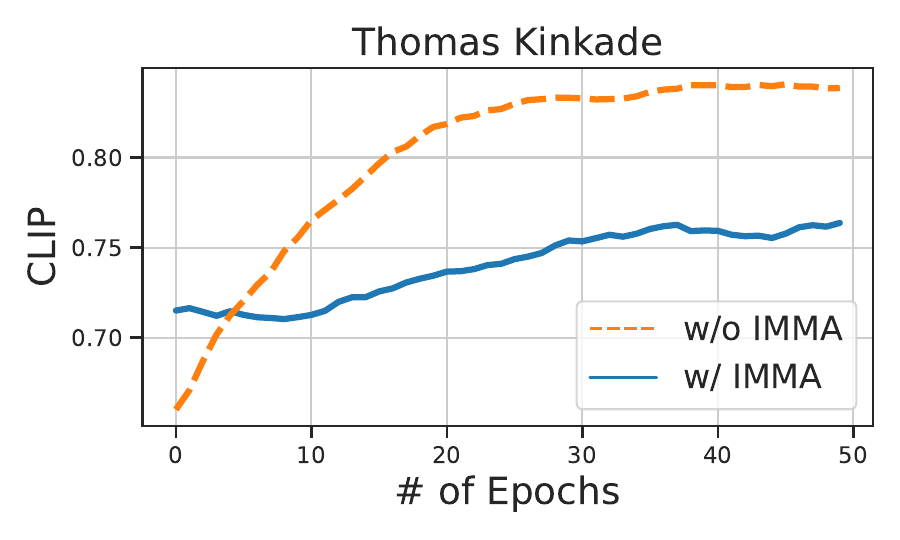} & \includegraphics[height=1.7cm, trim={0.3cm 0.35cm 0.3cm 0.35cm},clip]{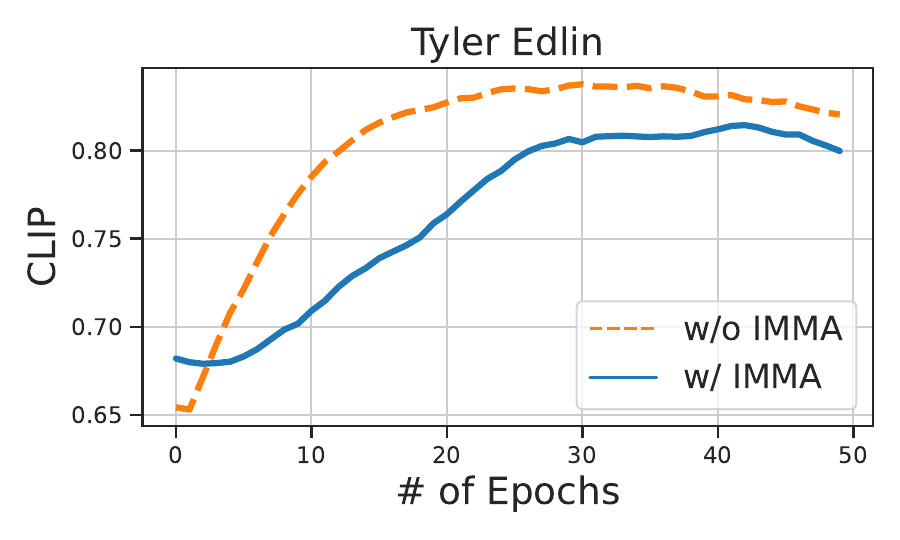} & \includegraphics[height=1.7cm, trim={0.3cm 0.35cm 0.3cm 0.35cm},clip]{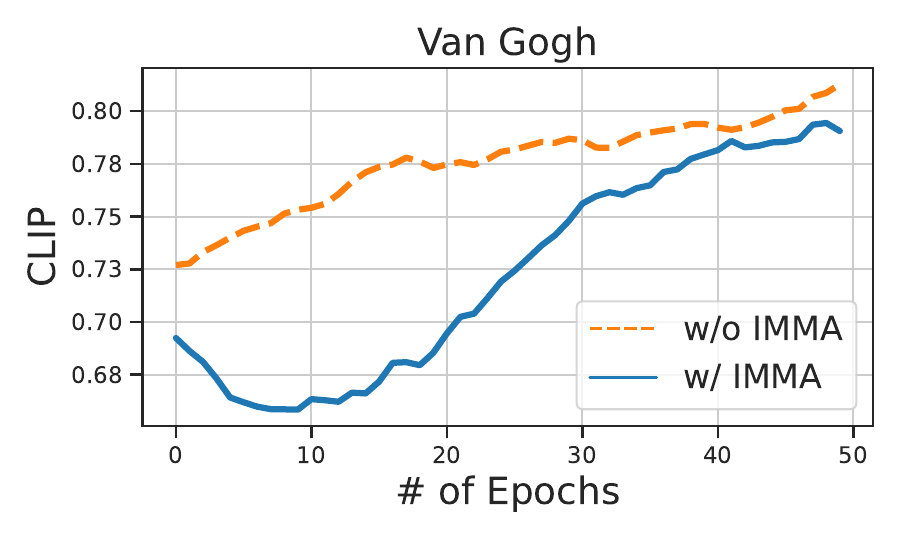}  \\
     \includegraphics[height=1.7cm, trim={0.3cm 0.35cm 0.3cm 0.35cm},clip]{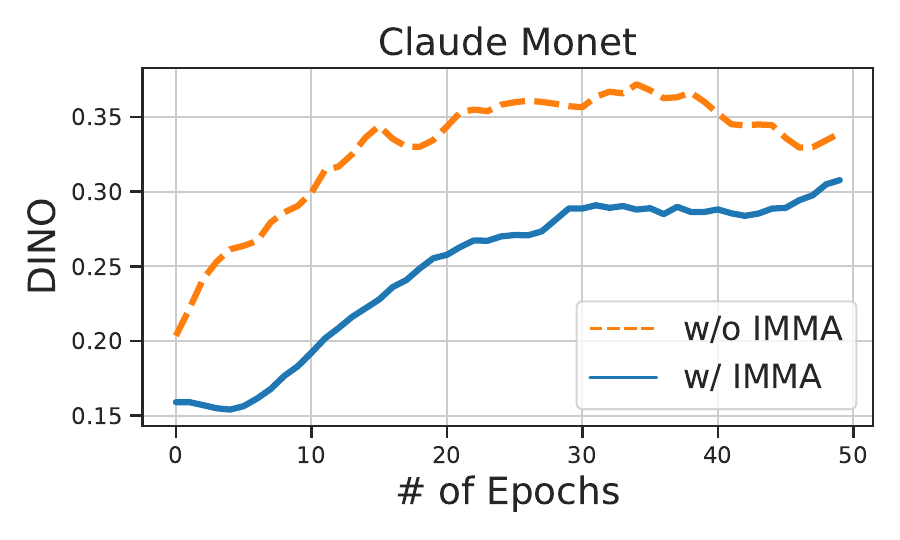} & \includegraphics[height=1.7cm, trim={0.3cm 0.35cm 0.3cm 0.35cm},clip]{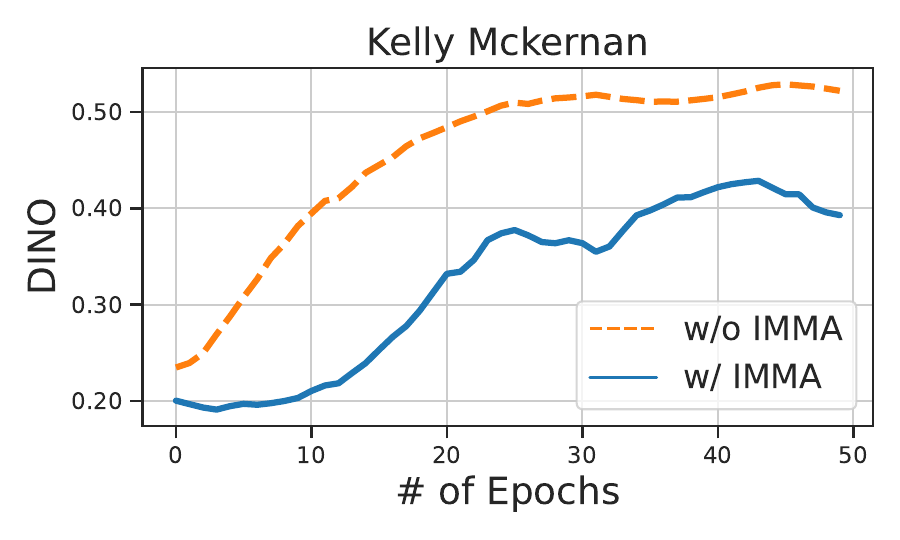} & \includegraphics[height=1.7cm, trim={0.3cm 0.35cm 0.3cm 0.35cm},clip]{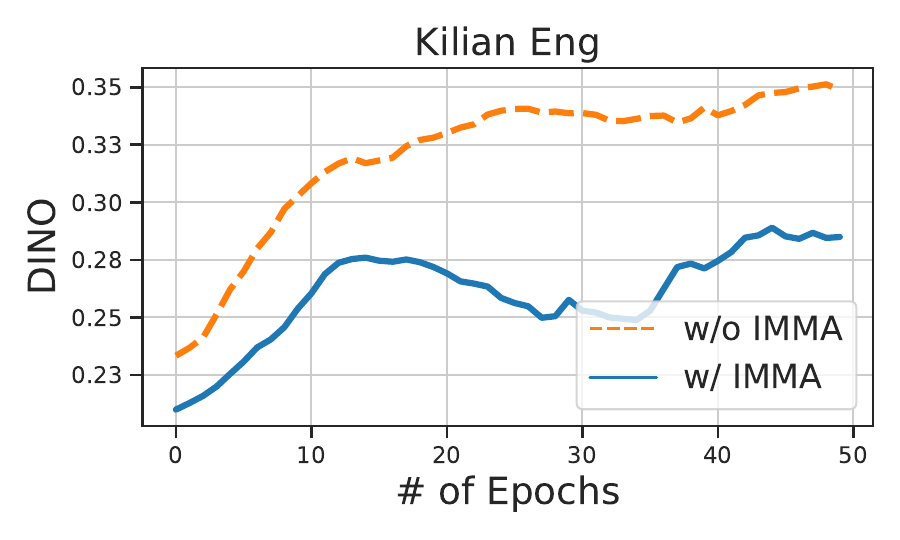}& \includegraphics[height=1.7cm, trim={0.3cm 0.35cm 0.3cm 0.35cm},clip]{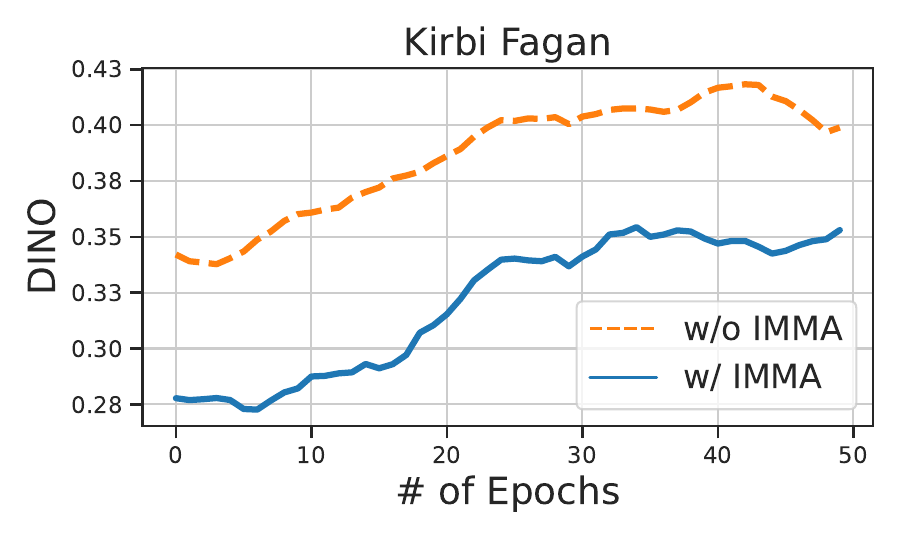} \\
     \includegraphics[height=1.7cm, trim={0.3cm 0.35cm 0.3cm 0.35cm},clip]{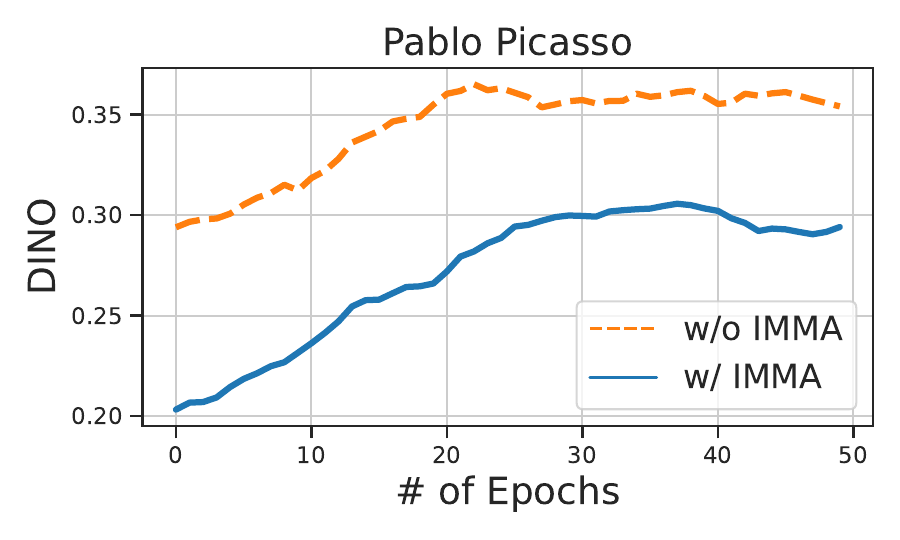} & \includegraphics[height=1.7cm, trim={0.3cm 0.35cm 0.3cm 0.35cm},clip]{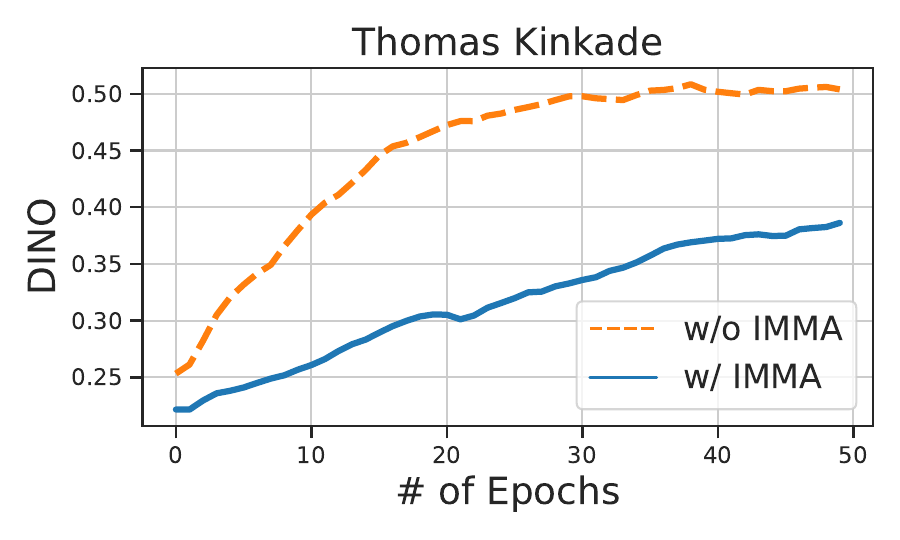} & \includegraphics[height=1.7cm, trim={0.3cm 0.35cm 0.3cm 0.35cm},clip]{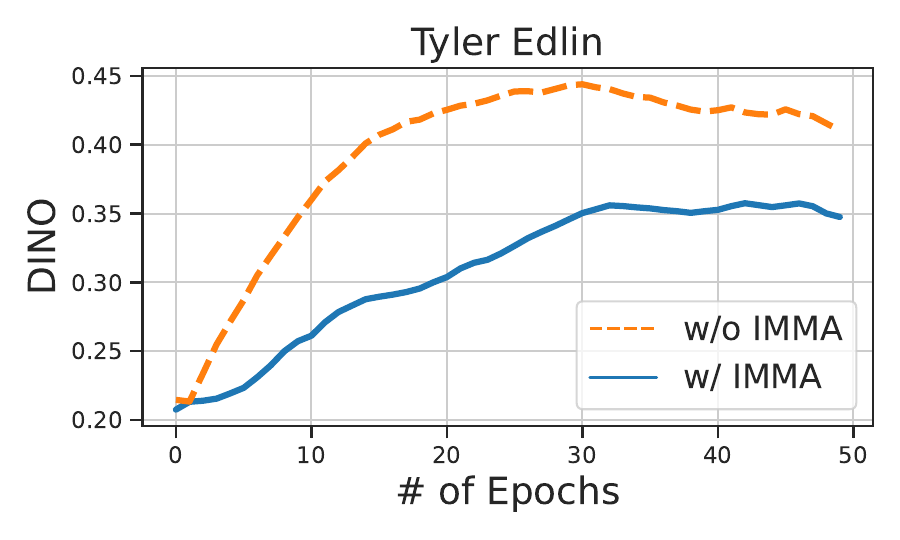} & \includegraphics[height=1.7cm, trim={0.3cm 0.35cm 0.3cm 0.35cm},clip]{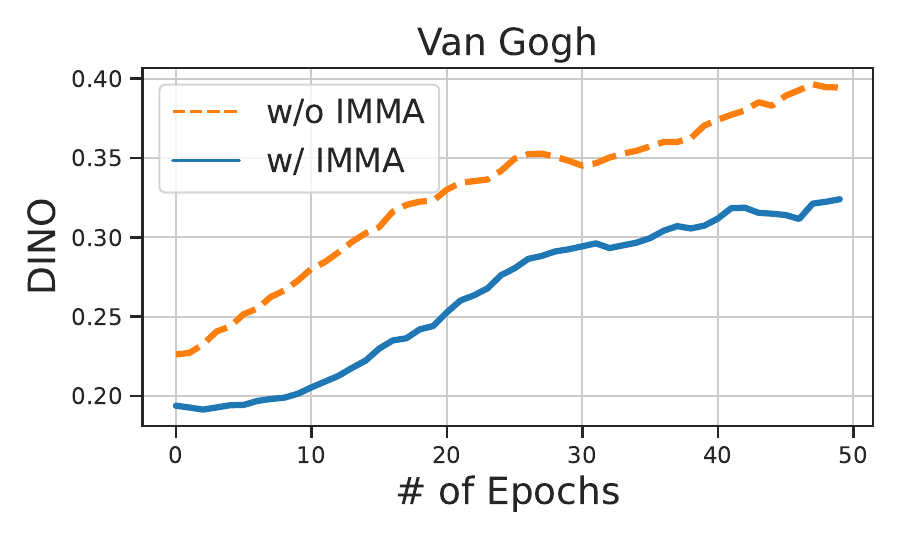}  \\
    \end{tabular}
    \caption{\textbf{LPIPS, CLIP, and DINO of LoRA on artistic style erased model Adaptation {\color{noimma}w/o} and {\color{imma} w/} IMMA.}}
    \label{fig:lora_style_quan}
\end{figure*}

\begin{figure*}[t]
    \centering
    \setlength{\tabcolsep}{1pt}
    \begin{tabular}{cccc}
     \includegraphics[height=1.7cm, trim={0.3cm 0.35cm 0.3cm 0.35cm},clip]{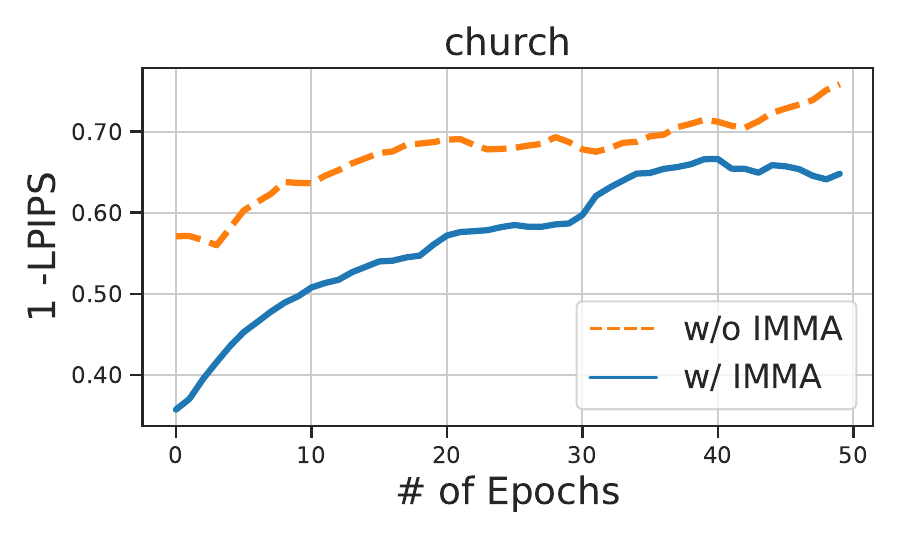} & \includegraphics[height=1.7cm, trim={0.3cm 0.35cm 0.3cm 0.35cm},clip]{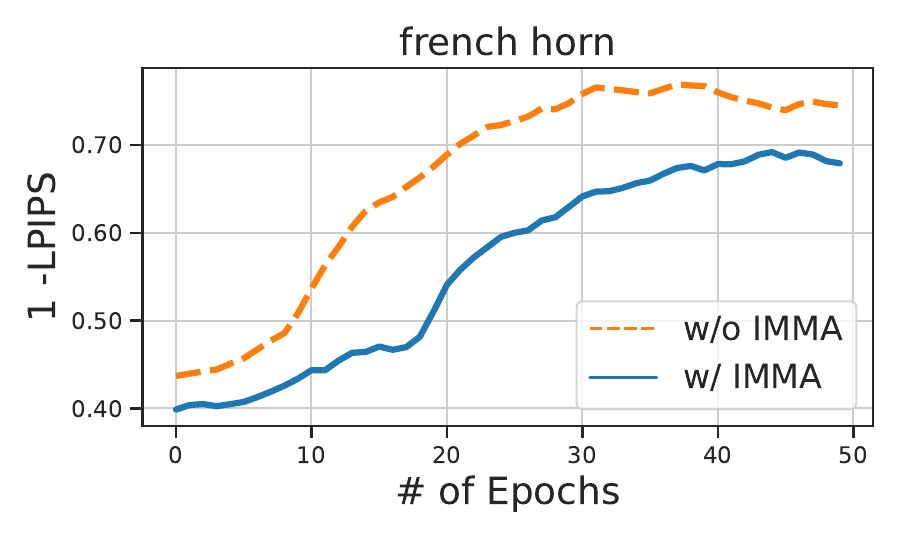} & \includegraphics[height=1.7cm, trim={0.3cm 0.35cm 0.3cm 0.35cm},clip]{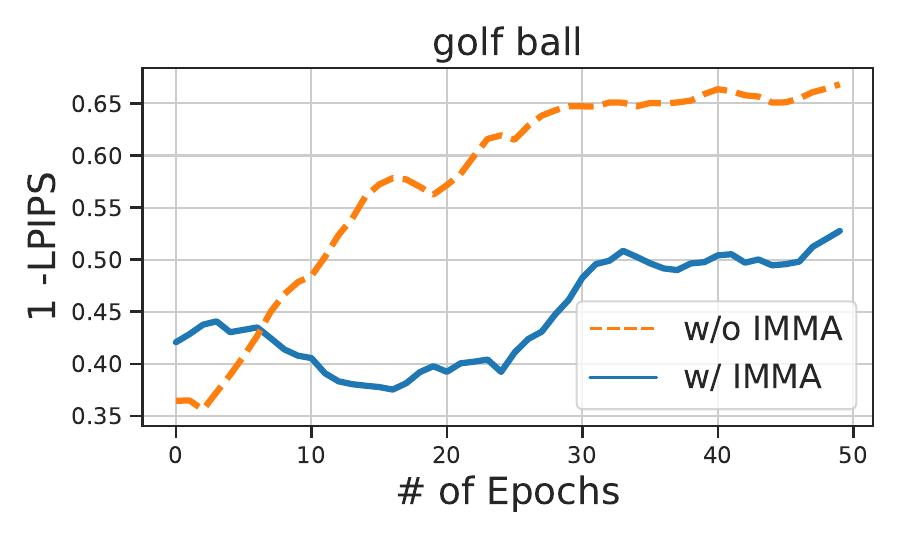} & \includegraphics[height=1.7cm, trim={0.3cm 0.35cm 0.3cm 0.35cm},clip]{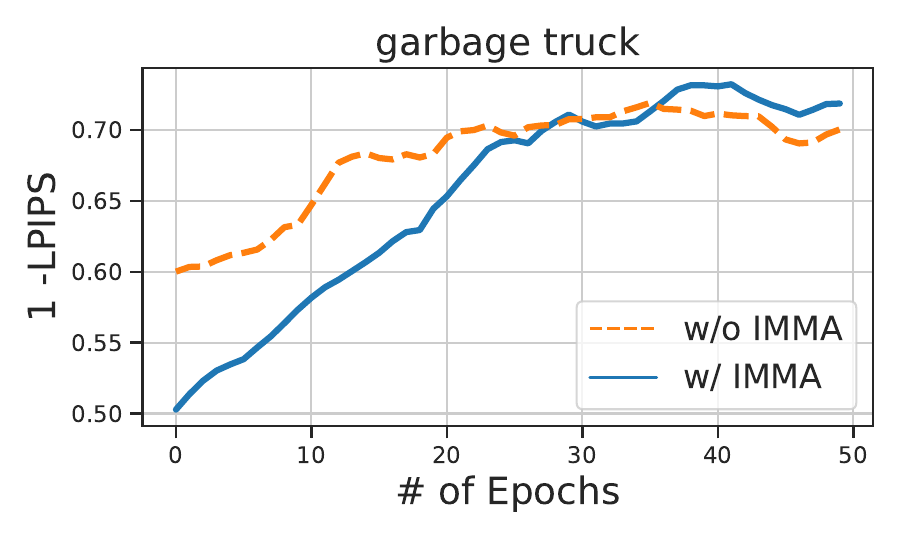} \\ 
     \includegraphics[height=1.7cm, trim={0.3cm 0.35cm 0.3cm 0.35cm},clip]{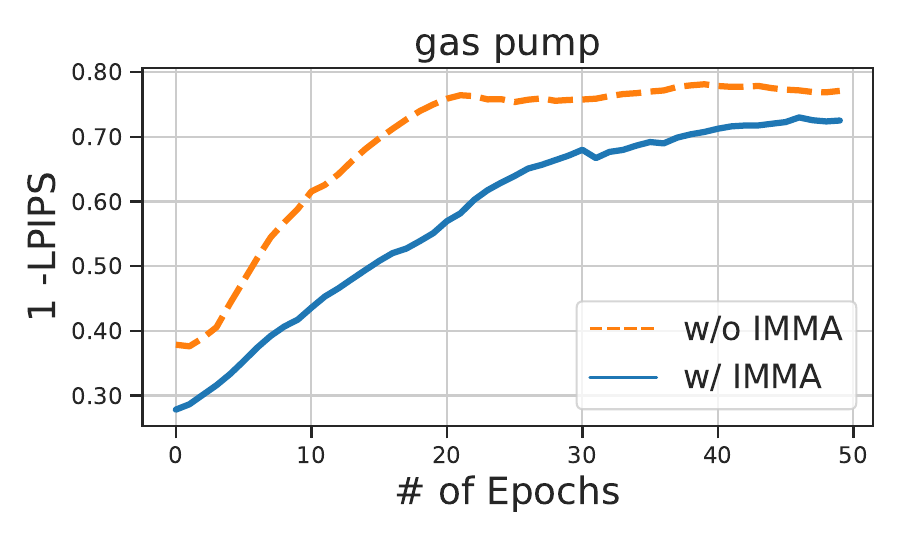} & \includegraphics[height=1.7cm, trim={0.3cm 0.35cm 0.3cm 0.35cm},clip]{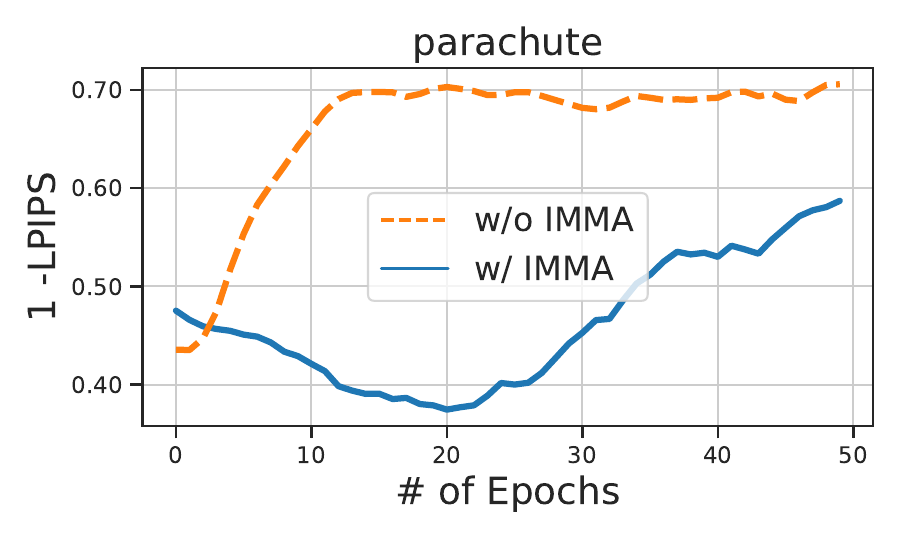} & \includegraphics[height=1.7cm, trim={0.3cm 0.35cm 0.3cm 0.35cm},clip]{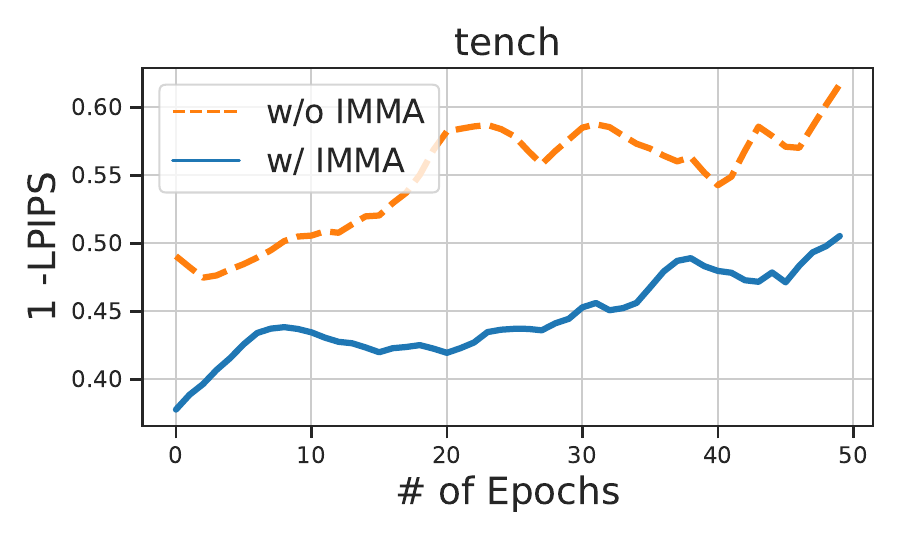} & \includegraphics[height=1.7cm, trim={0.3cm 0.35cm 0.3cm 0.35cm},clip]{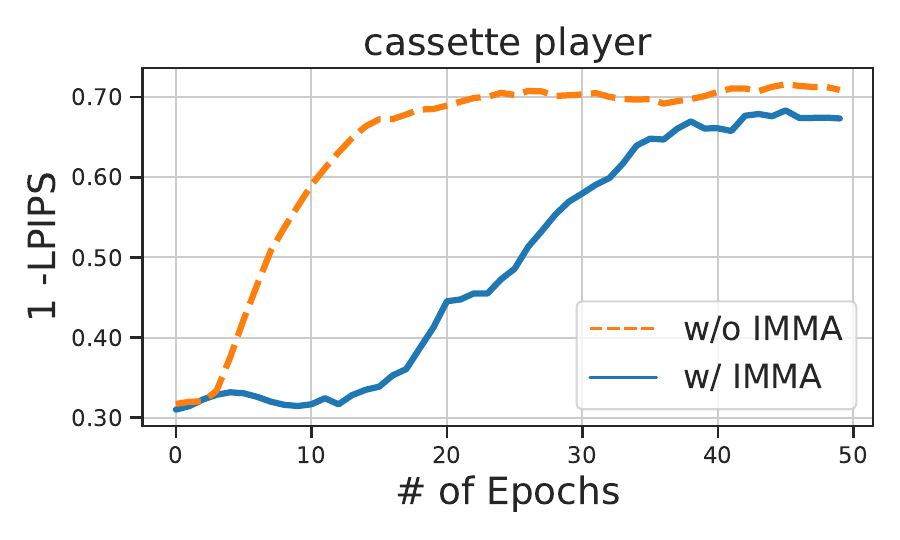} \\
     \includegraphics[height=1.7cm, trim={0.3cm 0.35cm 0.3cm 0.35cm},clip]{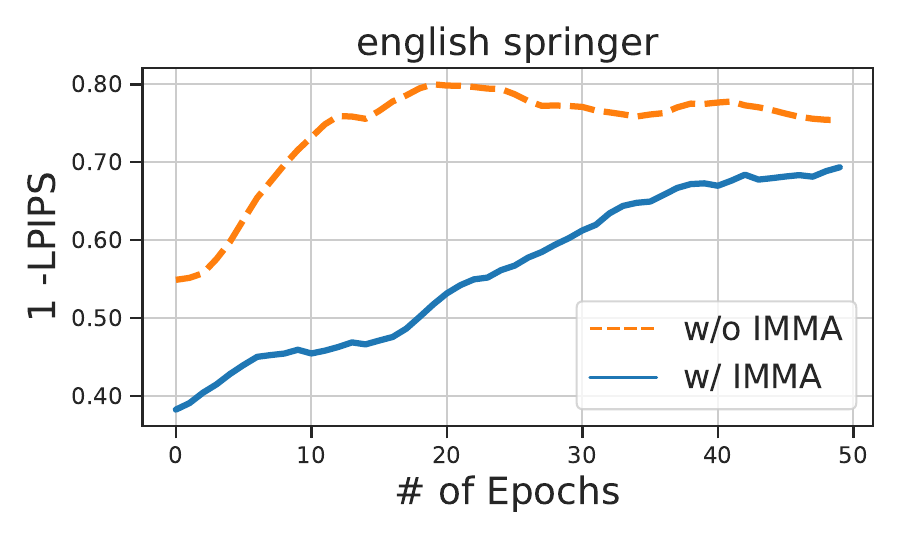}& \includegraphics[height=1.7cm, trim={0.3cm 0.35cm 0.3cm 0.35cm},clip]{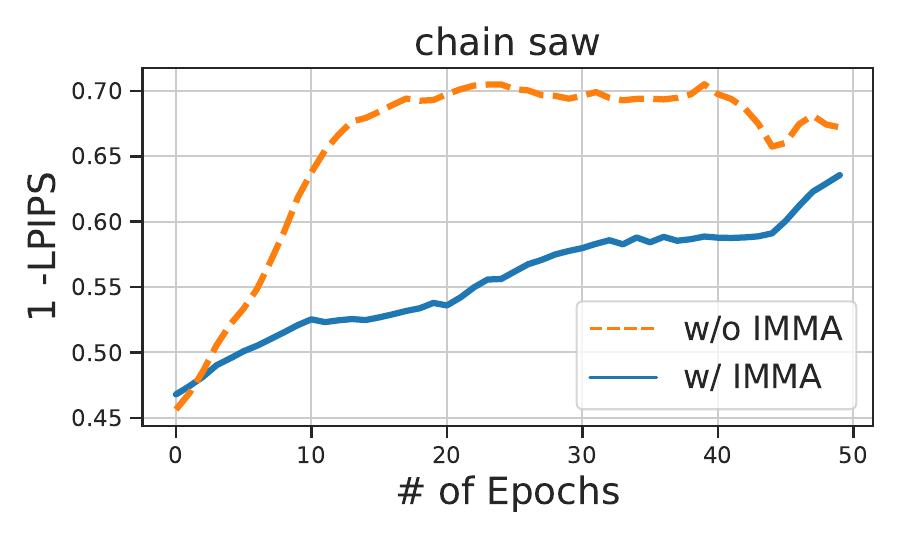} \\
     
     \includegraphics[height=1.7cm, trim={0.3cm 0.35cm 0.3cm 0.35cm},clip]{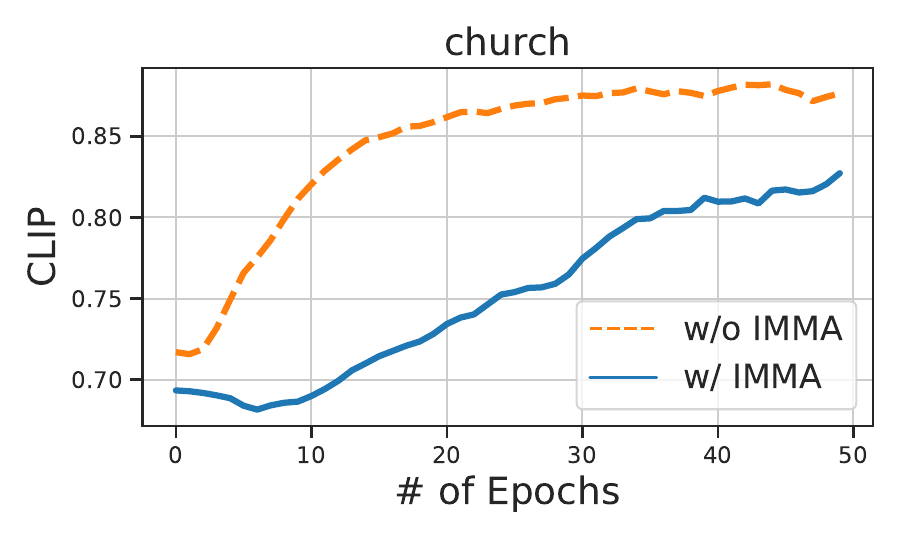} & \includegraphics[height=1.7cm, trim={0.3cm 0.35cm 0.3cm 0.35cm},clip]{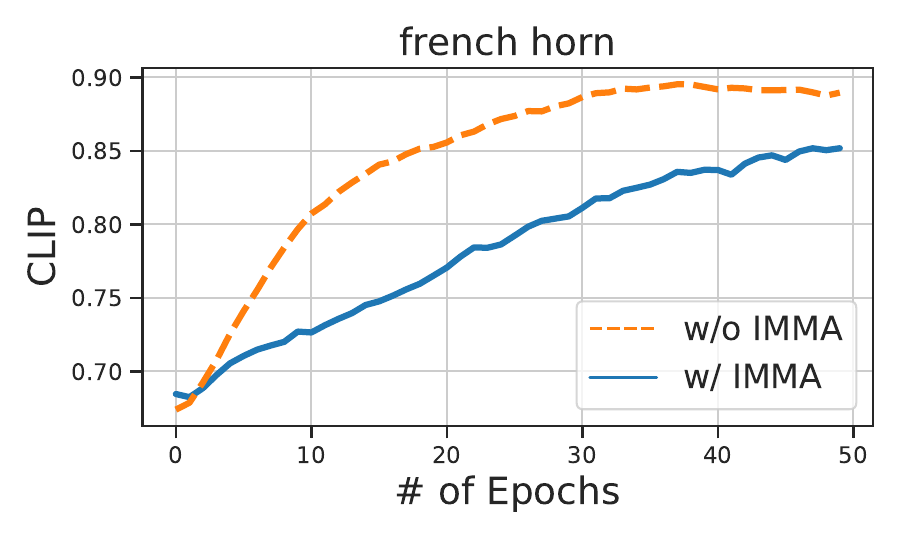} & \includegraphics[height=1.7cm, trim={0.3cm 0.35cm 0.3cm 0.35cm},clip]{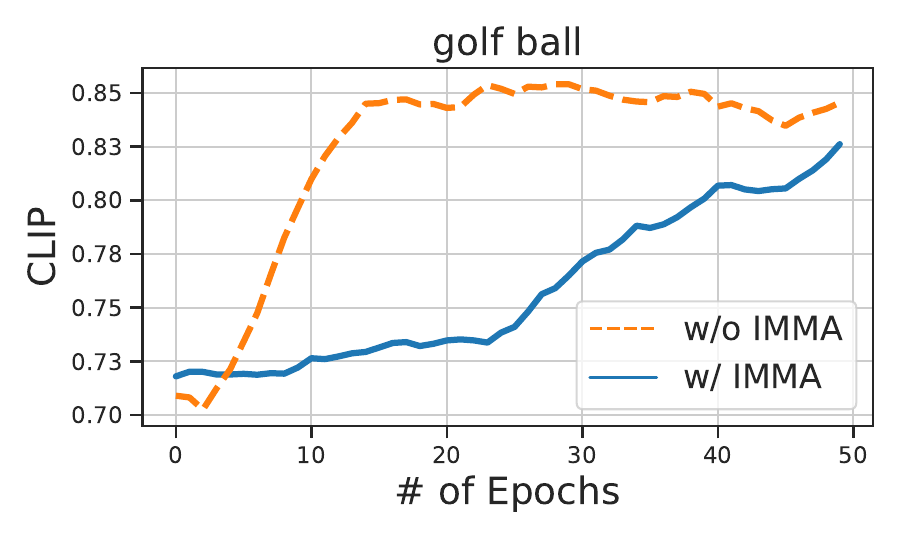} & \includegraphics[height=1.7cm, trim={0.3cm 0.35cm 0.3cm 0.35cm},clip]{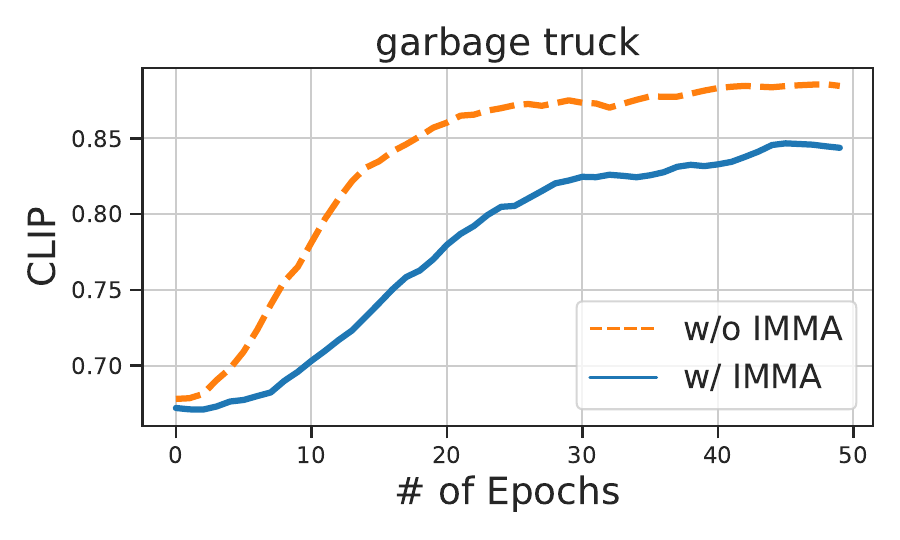} \\ 
     \includegraphics[height=1.7cm, trim={0.3cm 0.35cm 0.3cm 0.35cm},clip]{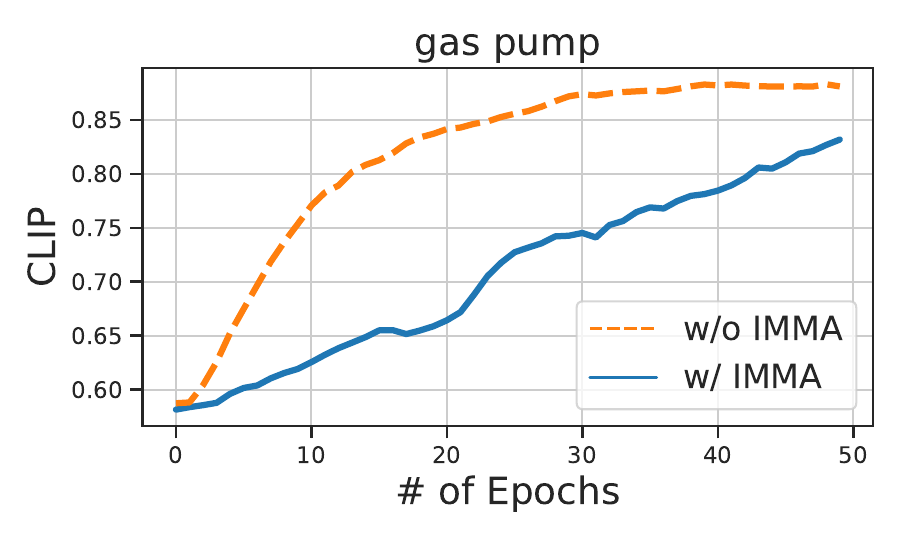} & \includegraphics[height=1.7cm, trim={0.3cm 0.35cm 0.3cm 0.35cm},clip]{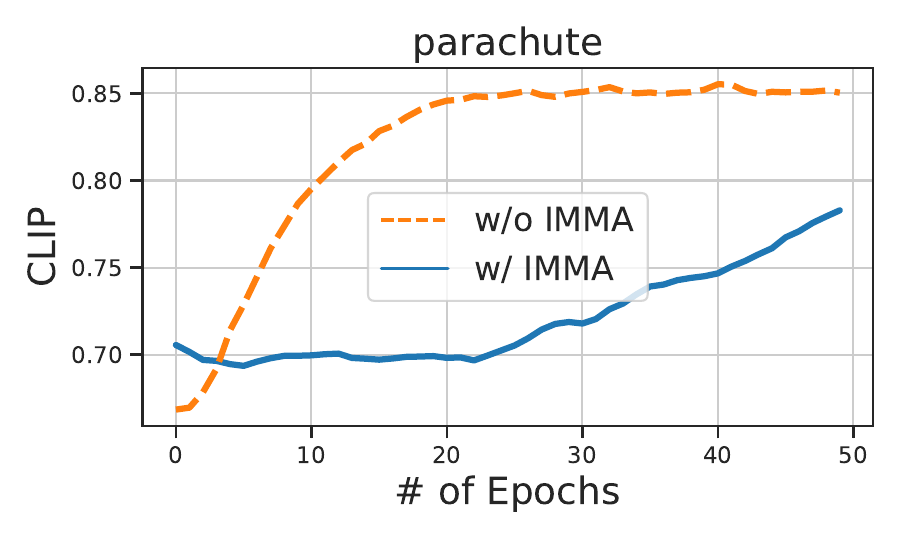} & \includegraphics[height=1.7cm, trim={0.3cm 0.35cm 0.3cm 0.35cm},clip]{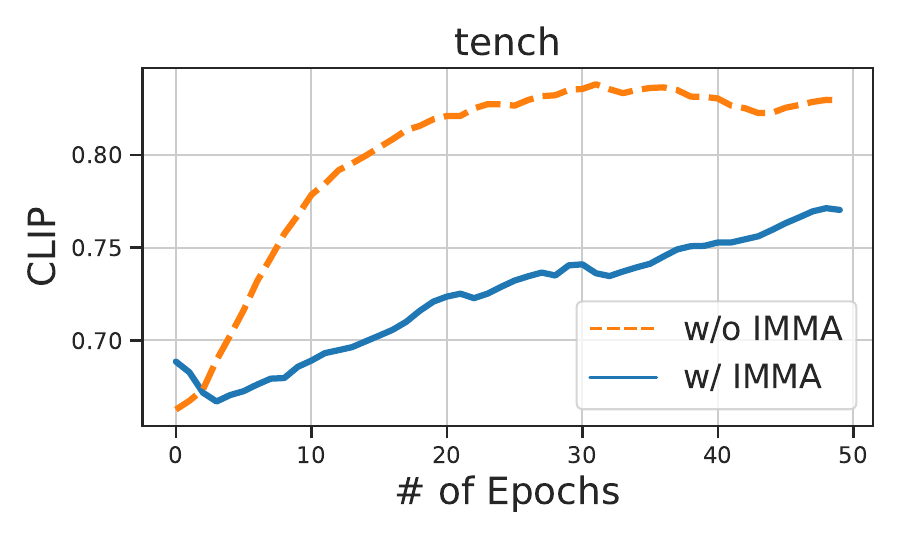} & \includegraphics[height=1.7cm, trim={0.3cm 0.35cm 0.3cm 0.35cm},clip]{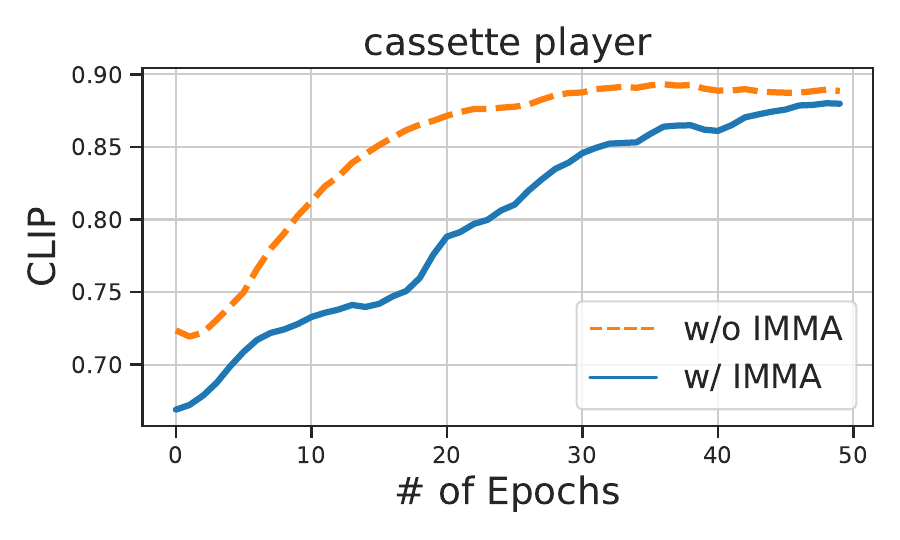} \\ 
     \includegraphics[height=1.7cm, trim={0.3cm 0.35cm 0.3cm 0.35cm},clip]{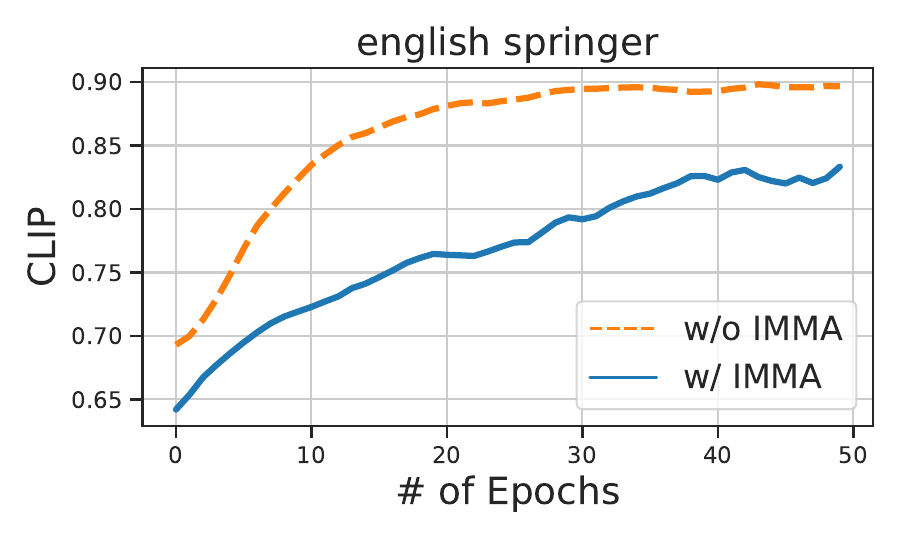}& \includegraphics[height=1.7cm, trim={0.3cm 0.35cm 0.3cm 0.35cm},clip]{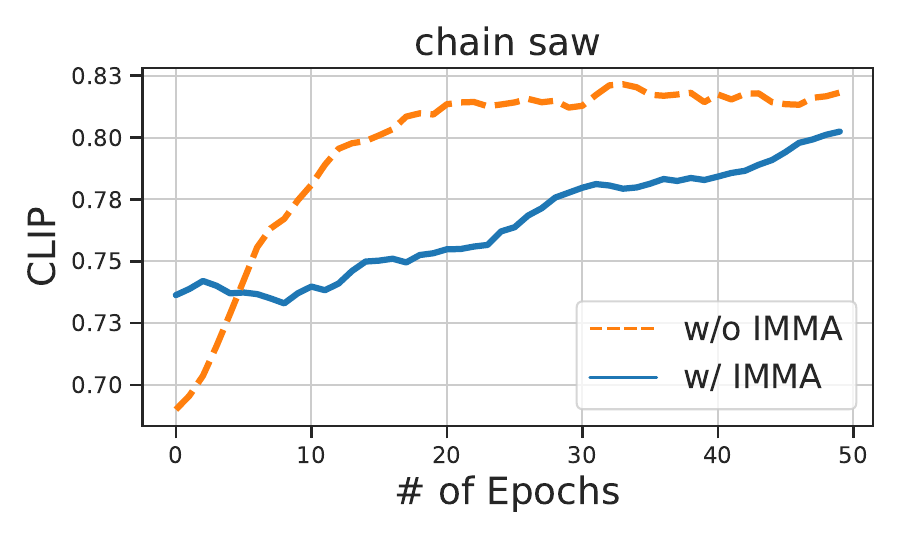} \\
     
      \includegraphics[height=1.7cm, trim={0.3cm 0.35cm 0.3cm 0.35cm},clip]{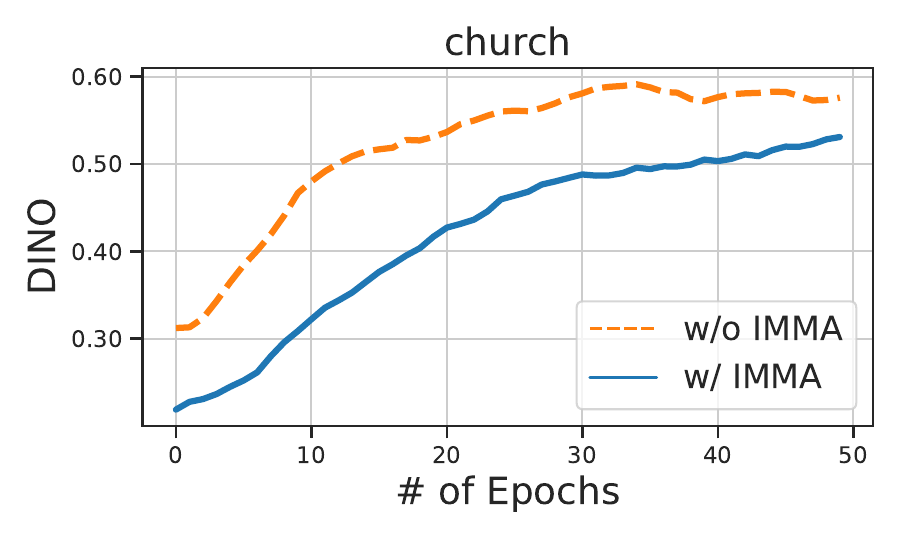} & \includegraphics[height=1.7cm, trim={0.3cm 0.35cm 0.3cm 0.35cm},clip]{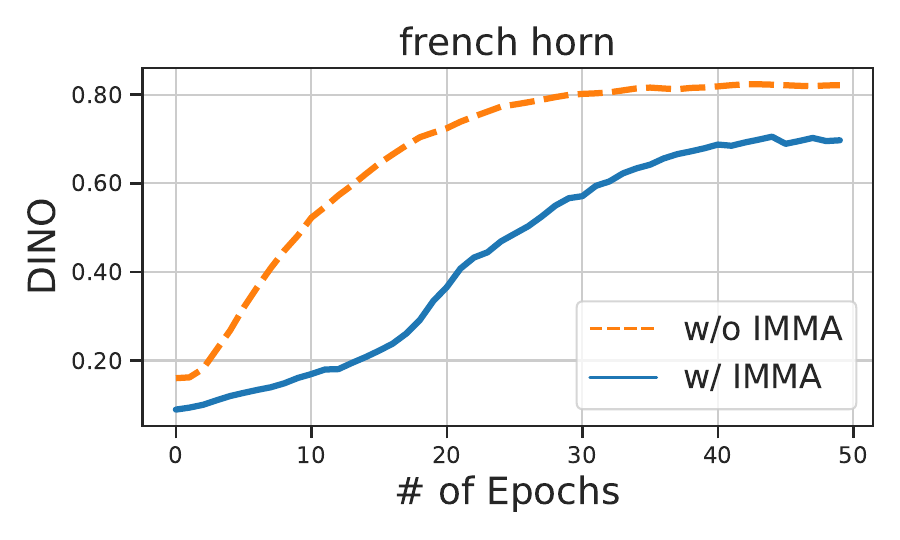} & \includegraphics[height=1.7cm, trim={0.3cm 0.35cm 0.3cm 0.35cm},clip]{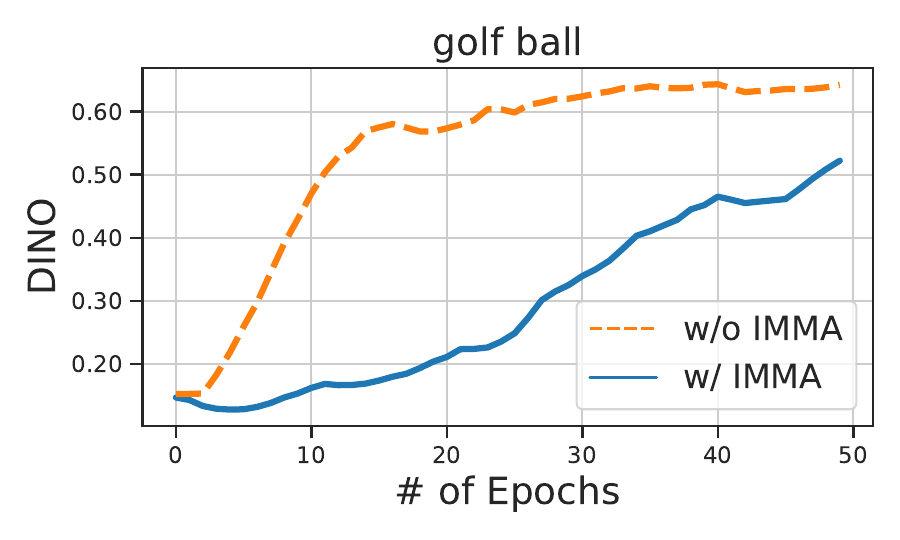} & \includegraphics[height=1.7cm, trim={0.3cm 0.35cm 0.3cm 0.35cm},clip]{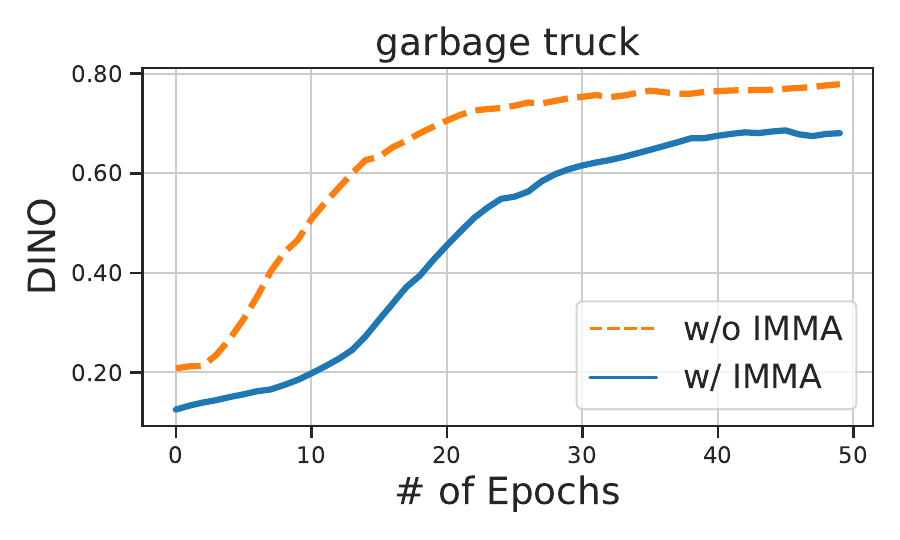}\\ 
      \includegraphics[height=1.7cm, trim={0.3cm 0.35cm 0.3cm 0.35cm},clip]{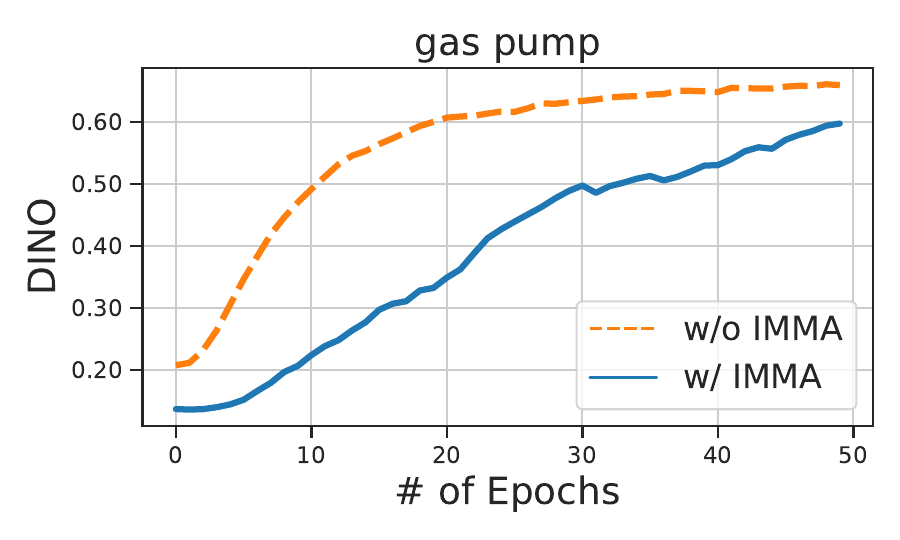} & \includegraphics[height=1.7cm, trim={0.3cm 0.35cm 0.3cm 0.35cm},clip]{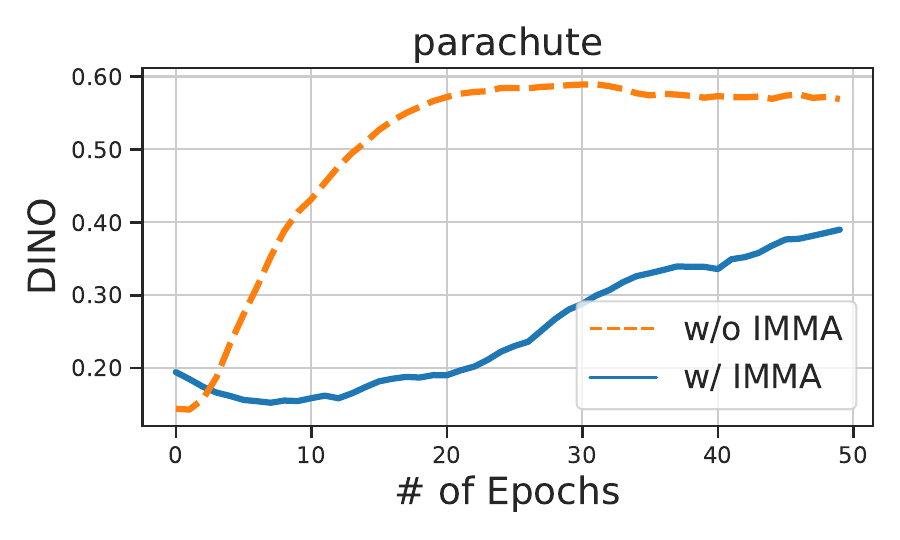} & \includegraphics[height=1.7cm, trim={0.3cm 0.35cm 0.3cm 0.35cm},clip]{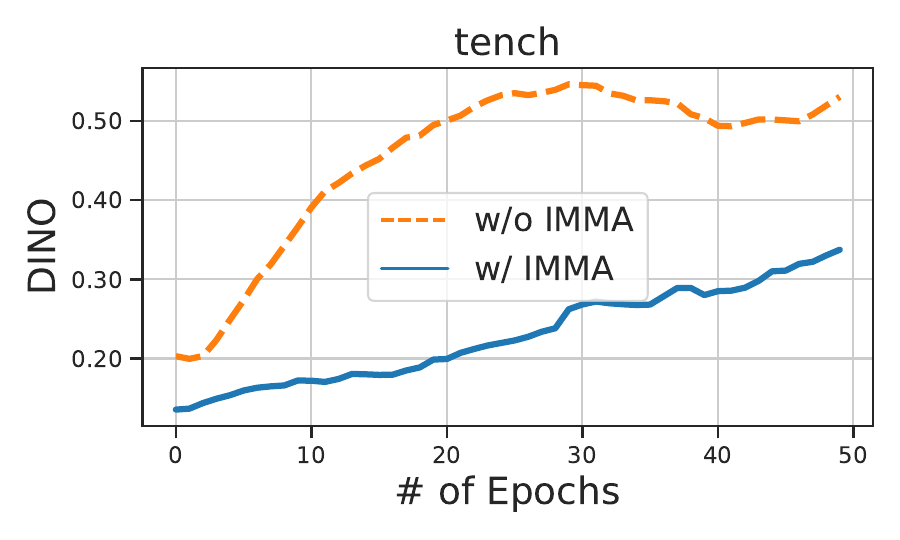} & \includegraphics[height=1.7cm, trim={0.3cm 0.35cm 0.3cm 0.35cm},clip]{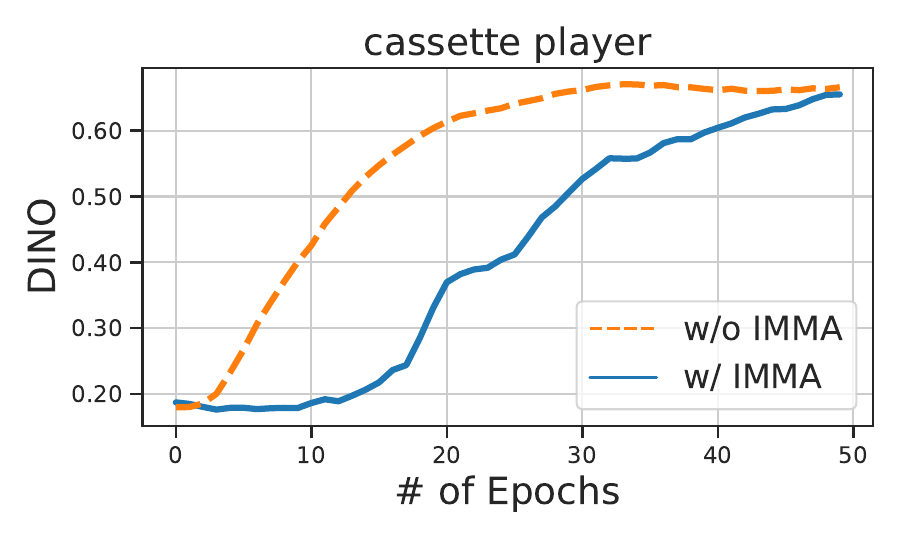} \\
      \includegraphics[height=1.7cm, trim={0.3cm 0.35cm 0.3cm 0.35cm},clip]{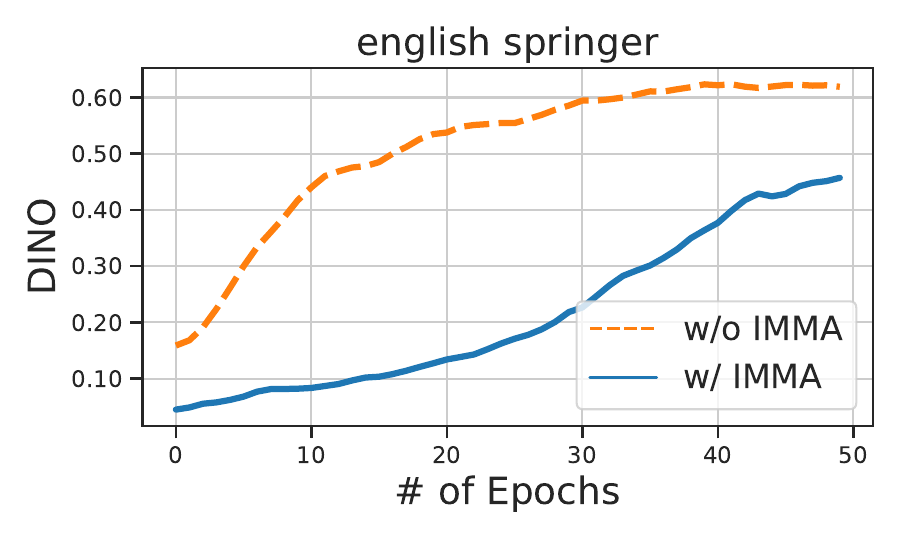}& \includegraphics[height=1.7cm, trim={0.3cm 0.35cm 0.3cm 0.35cm},clip]{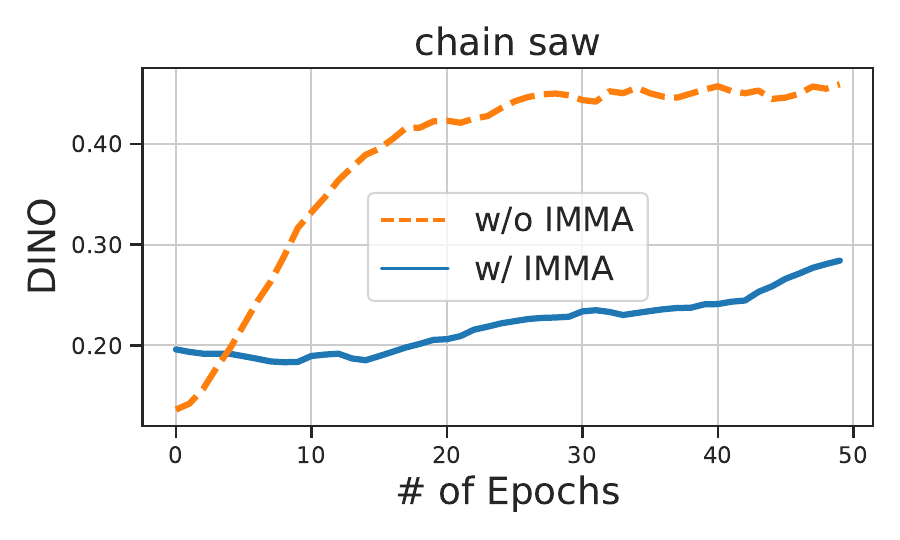} \\
    \end{tabular}
    \caption{\textbf{LPIPS, CLIP, and DINO of LoRA on object erased model Adaptation {\color{noimma}w/o} and {\color{imma} w/} IMMA.} Our method can prevent models from generating images with target concepts and good quality, as indicated by the high LPIPS, and low CLIP scores.}
    \label{fig:lora_object_quan}
\end{figure*}

\myparagraph{Results on immunizing erased model against re-learning.} In~\figref{fig:lora_style_quan}, we show the metric values \vs the number of LoRA adaptation epochs on eight artistic style datasets. We can observe a gap between the two lines, which indicates that implementing IMMA immunizes the model from re-learning the target artistic style. In~\figref{fig:lora_object_quan}, we show the metric values \vs the number of LoRA adaptation epochs on ten objects from ImageNet. We can also see a consistent gap between the two lines across all datasets.

\begin{figure*}[t]
    \centering
    \setlength{\tabcolsep}{1pt}
    \begin{tabular}{cccc}
     \includegraphics[height=1.7cm, trim={0.3cm 0.35cm 0.3cm 0.35cm},clip]{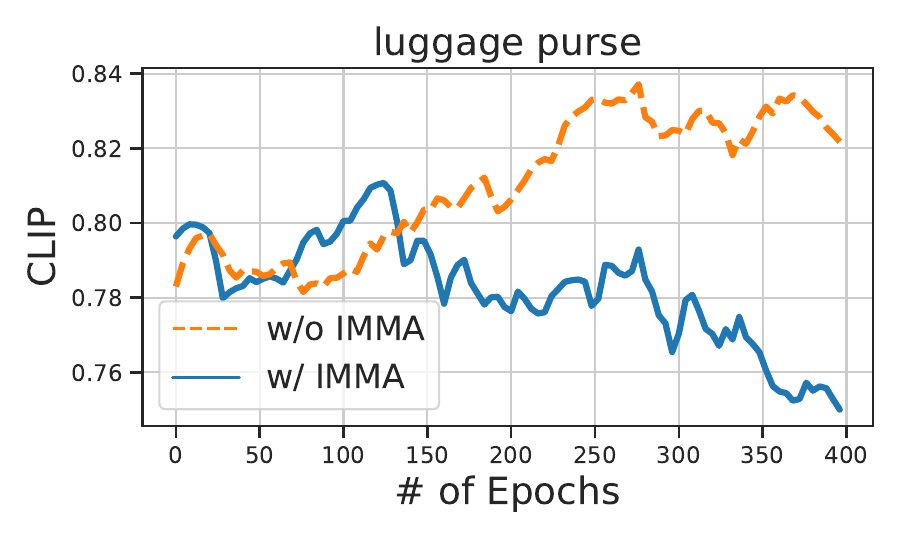} & \includegraphics[height=1.7cm, trim={0.3cm 0.35cm 0.3cm 0.35cm},clip]{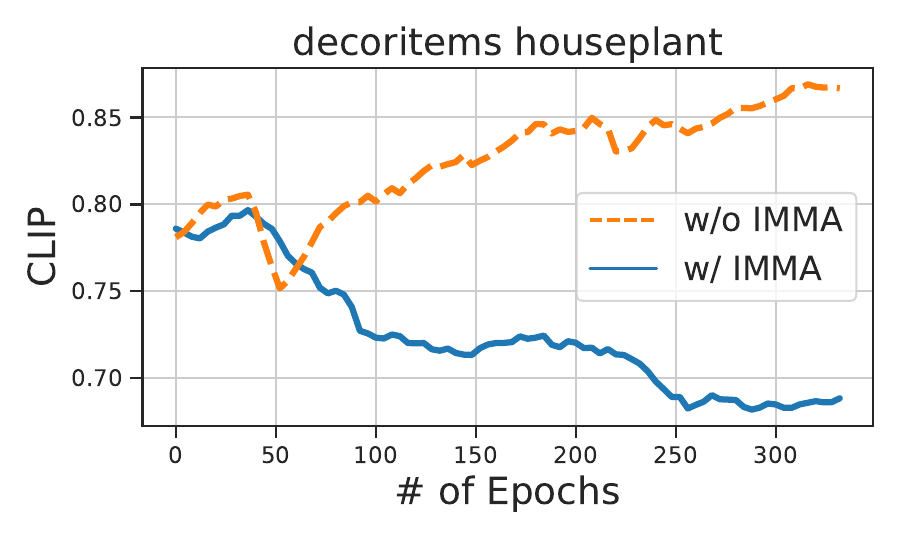} & \includegraphics[height=1.7cm, trim={0.3cm 0.35cm 0.3cm 0.35cm},clip]{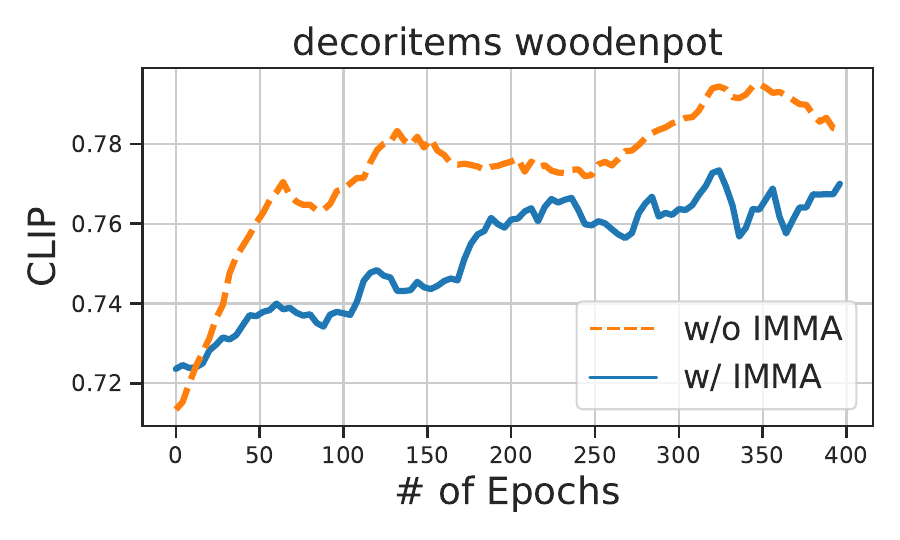} & \includegraphics[height=1.7cm, trim={0.3cm 0.35cm 0.3cm 0.35cm},clip]{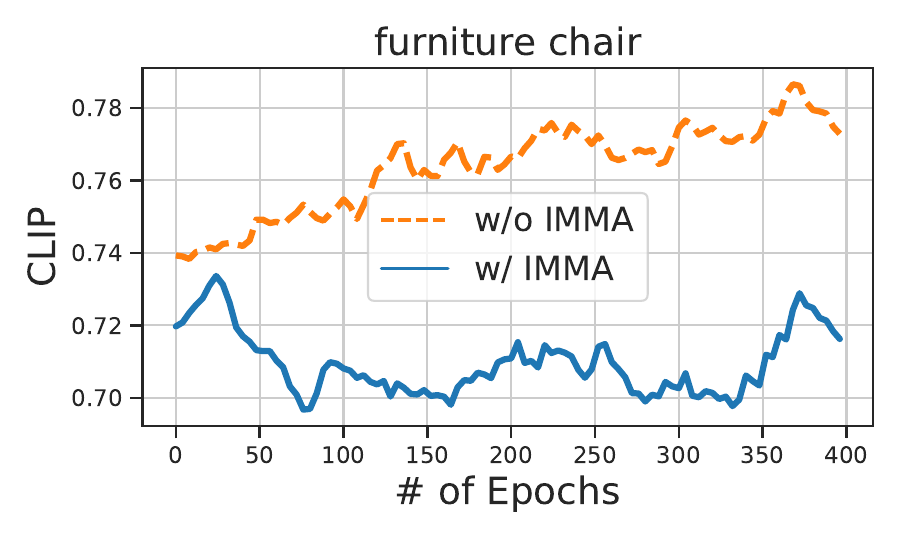} \\ 
     \includegraphics[height=1.7cm, trim={0.3cm 0.35cm 0.3cm 0.35cm},clip]{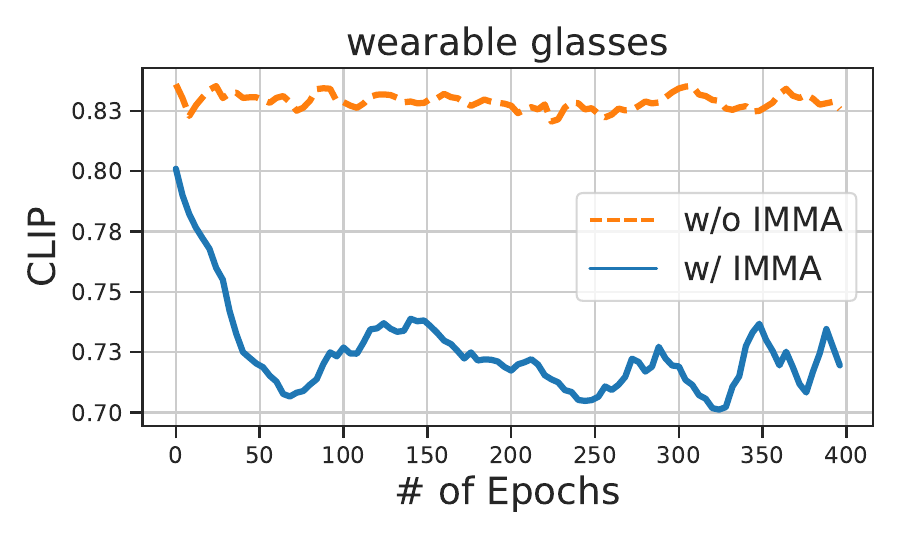} & \includegraphics[height=1.7cm, trim={0.3cm 0.35cm 0.3cm 0.35cm},clip]{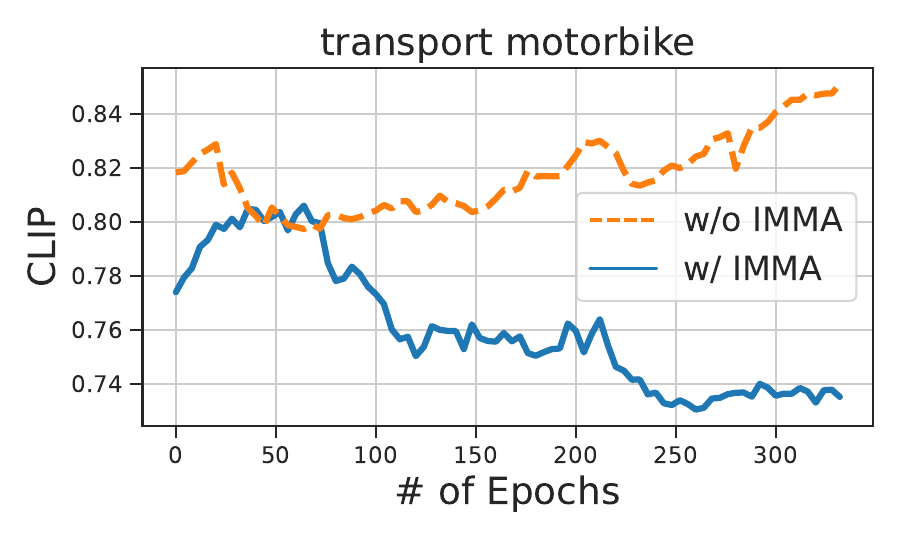} & \includegraphics[height=1.7cm, trim={0.3cm 0.35cm 0.3cm 0.35cm},clip]{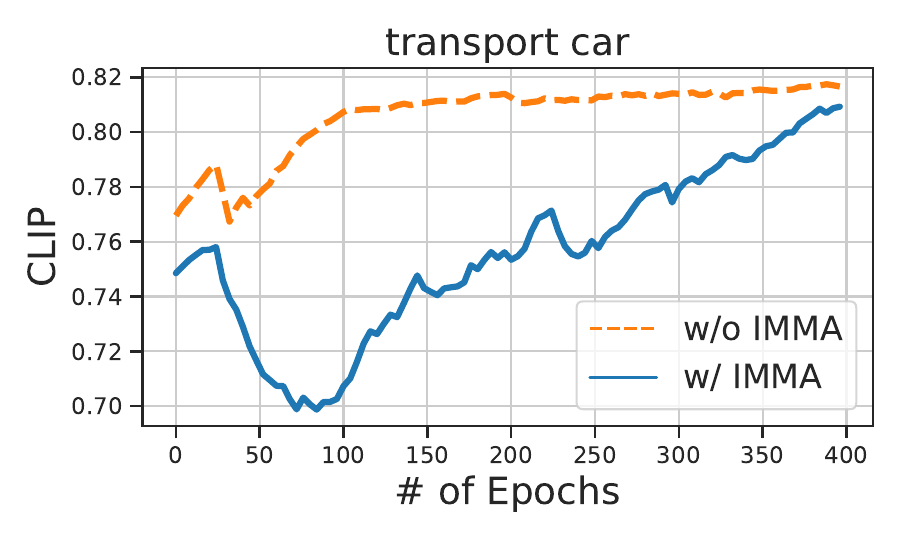} & \includegraphics[height=1.7cm, trim={0.3cm 0.35cm 0.3cm 0.35cm},clip]{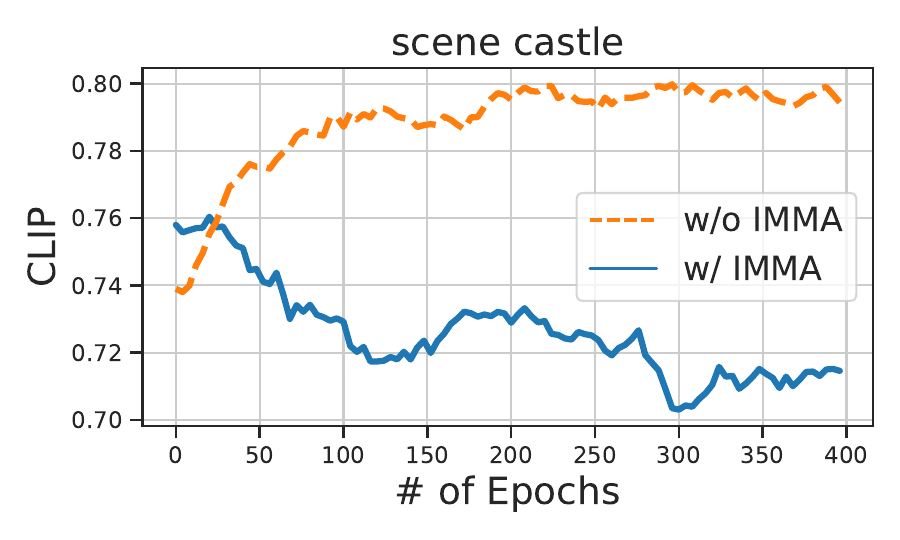} \\
     \includegraphics[height=1.7cm, trim={0.3cm 0.35cm 0.3cm 0.35cm},clip]{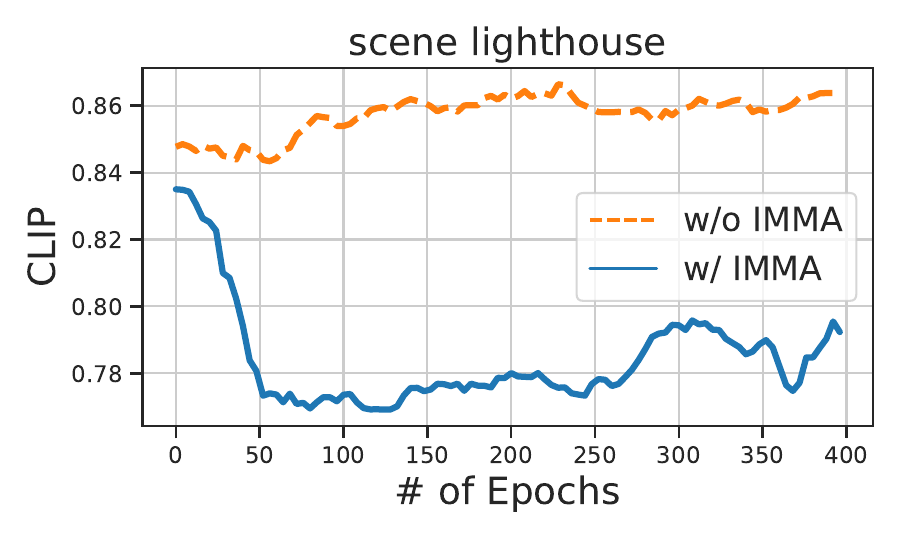}& \includegraphics[height=1.7cm, trim={0.3cm 0.35cm 0.3cm 0.35cm},clip]{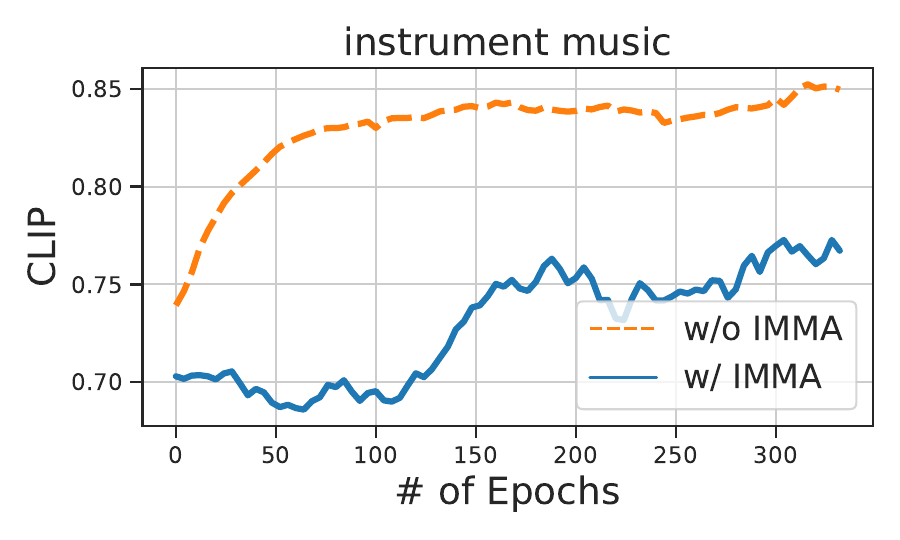} \\
     \includegraphics[height=1.7cm, trim={0.3cm 0.35cm 0.3cm 0.35cm},clip]{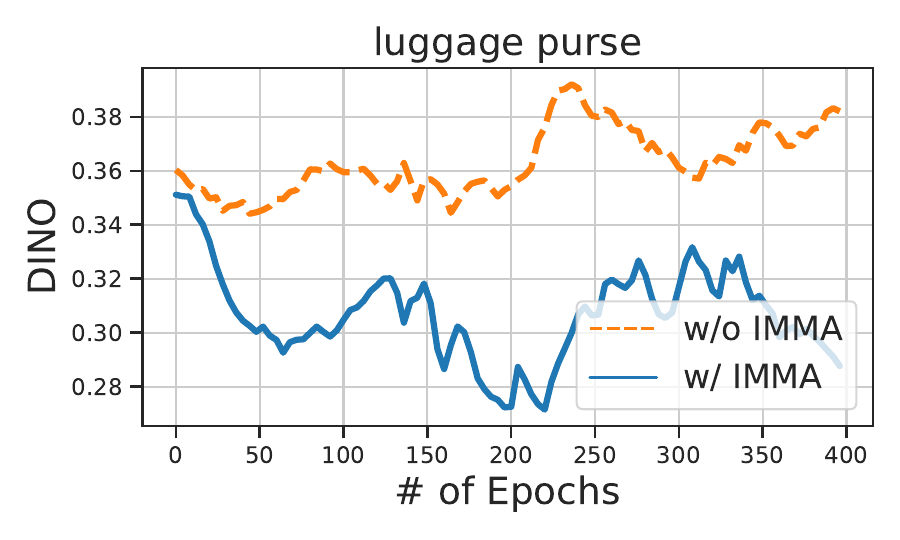} & \includegraphics[height=1.7cm, trim={0.3cm 0.35cm 0.3cm 0.35cm},clip]{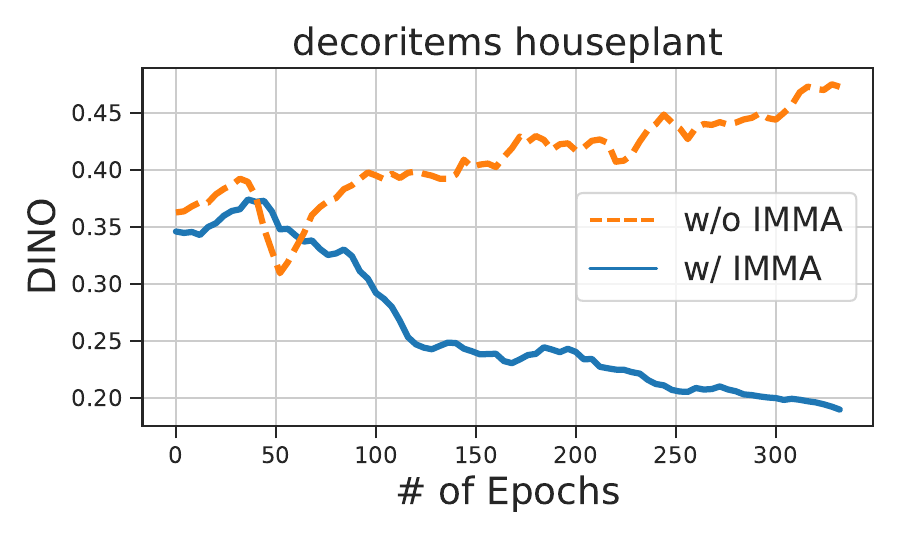} & \includegraphics[height=1.7cm, trim={0.3cm 0.35cm 0.3cm 0.35cm},clip]{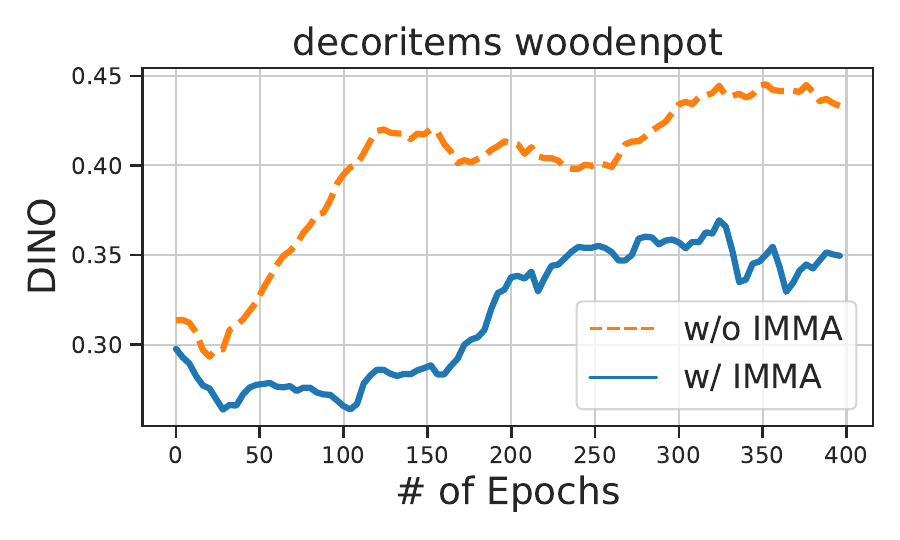} & \includegraphics[height=1.7cm, trim={0.3cm 0.35cm 0.3cm 0.35cm},clip]{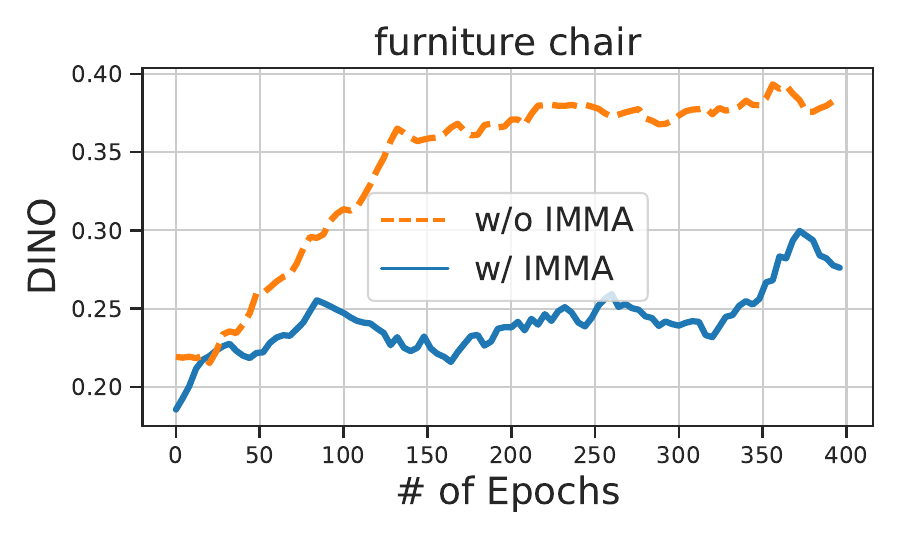}\\
     \includegraphics[height=1.7cm, trim={0.3cm 0.35cm 0.3cm 0.35cm},clip]{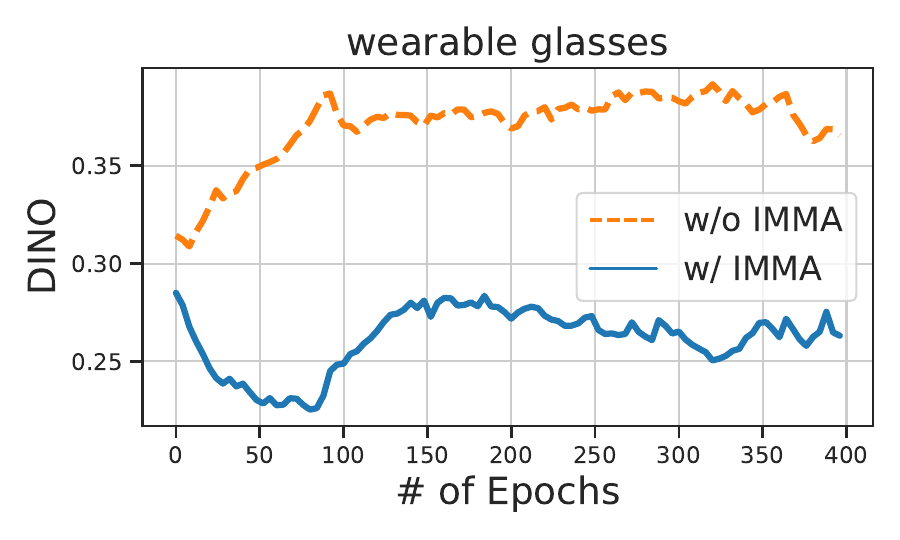} & \includegraphics[height=1.7cm, trim={0.3cm 0.35cm 0.3cm 0.35cm},clip]{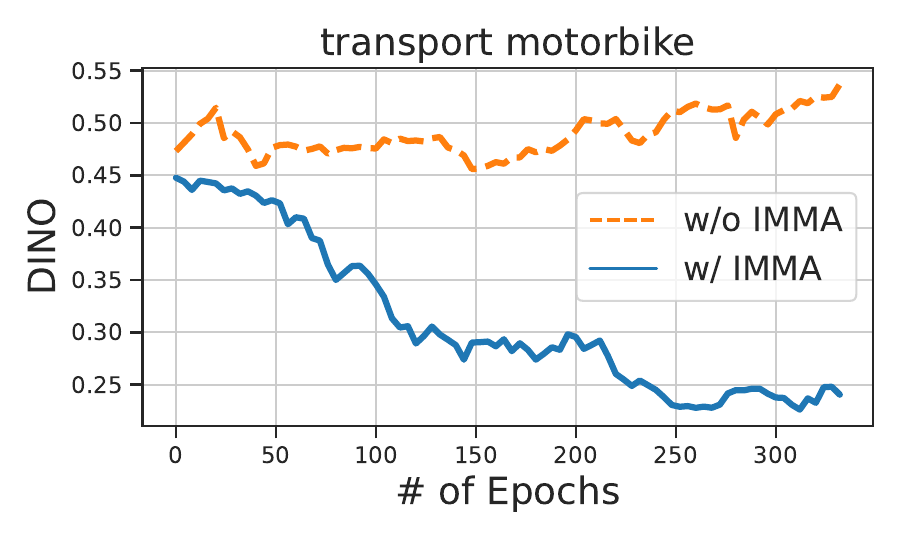} & \includegraphics[height=1.7cm, trim={0.3cm 0.35cm 0.3cm 0.35cm},clip]{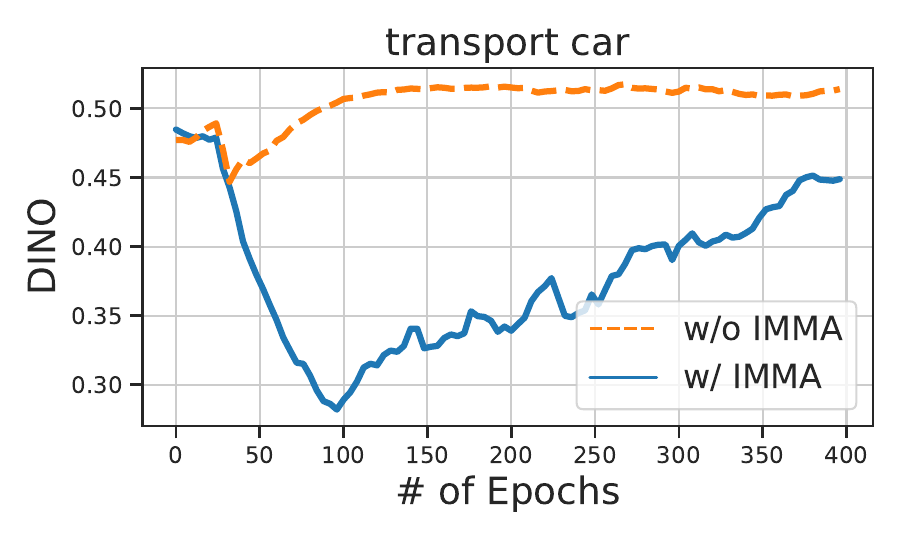} & \includegraphics[height=1.7cm, trim={0.3cm 0.35cm 0.3cm 0.35cm},clip]{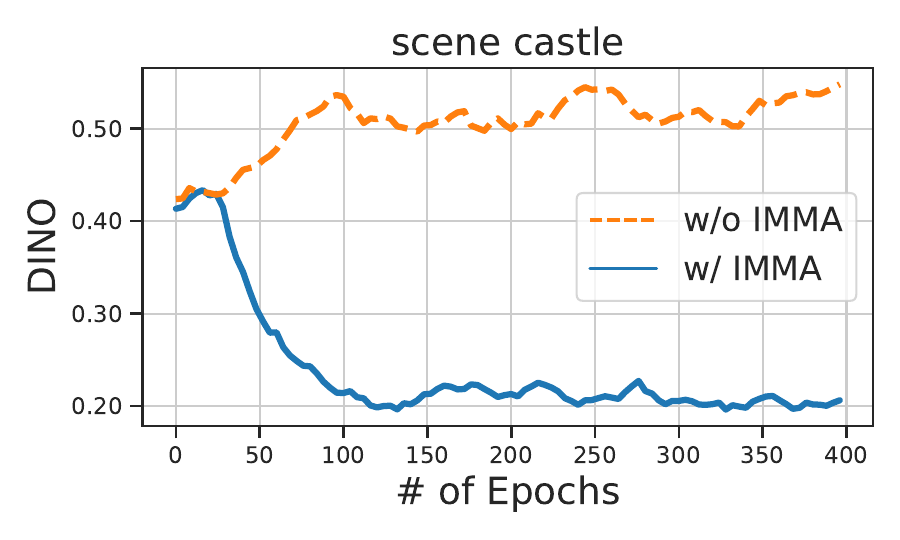} \\
     \includegraphics[height=1.7cm, trim={0.3cm 0.35cm 0.3cm 0.35cm},clip]{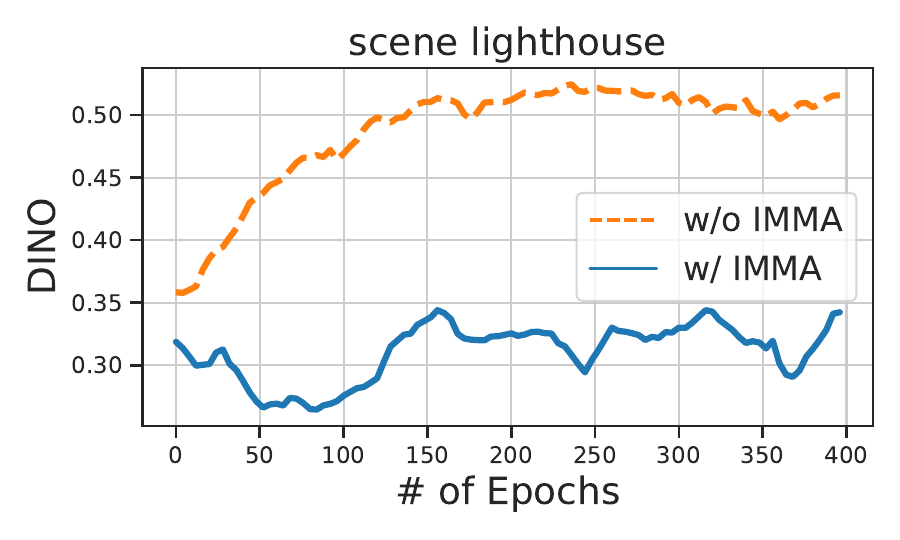}& \includegraphics[height=1.7cm, trim={0.3cm 0.35cm 0.3cm 0.35cm},clip]{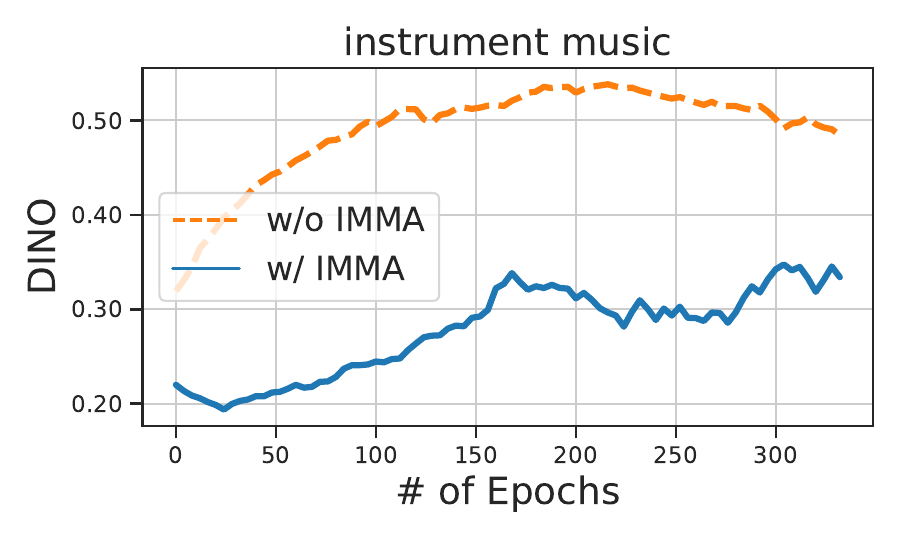} \\
    \end{tabular}
    \caption{\bf CLIP and DINO versus reference images after Textual Inversion Adaptation {\color{noimma}w/o} and {\color{imma} w/} IMMA.}
    \label{fig:ti_ref}
\end{figure*}

\begin{figure*}[t]
    \centering
    \setlength{\tabcolsep}{1pt}
    \begin{tabular}{ccccc}
    \includegraphics[height=1.7cm, trim={0.3cm 0.35cm 0.3cm 0.35cm},clip]{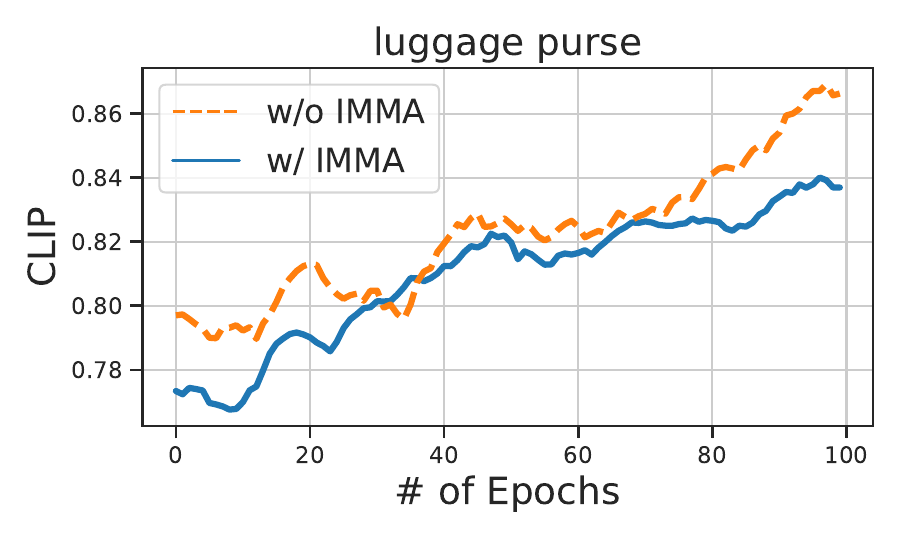} & \includegraphics[height=1.7cm, trim={0.3cm 0.35cm 0.3cm 0.35cm},clip]{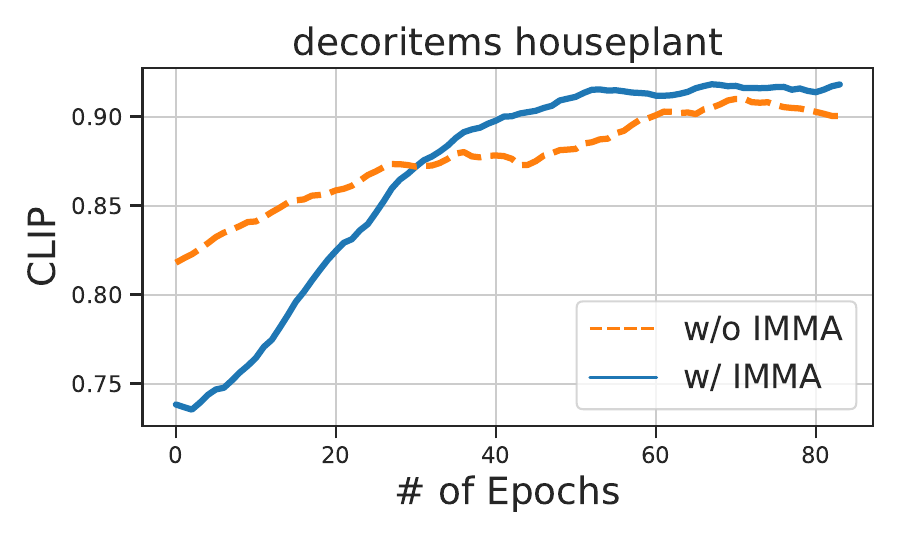} & \includegraphics[height=1.7cm, trim={0.3cm 0.35cm 0.3cm 0.35cm},clip]{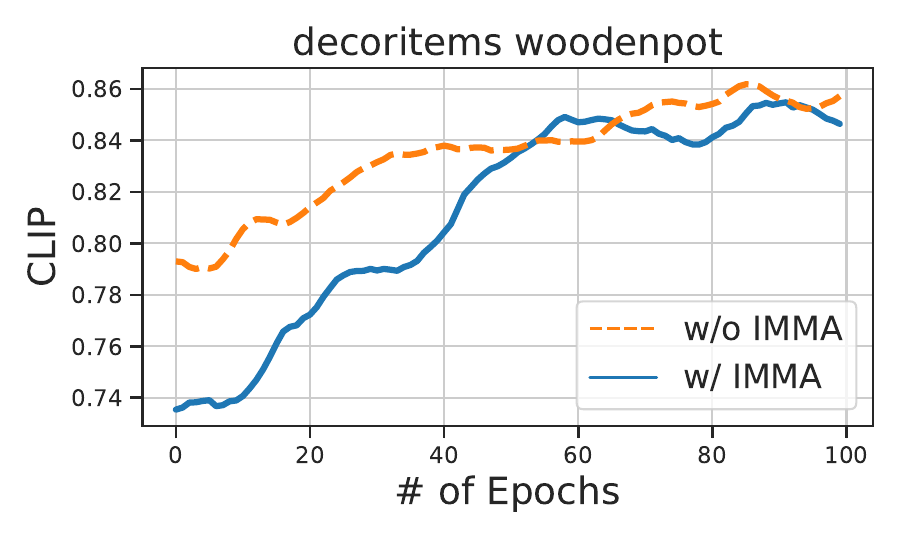} & \includegraphics[height=1.7cm, trim={0.3cm 0.35cm 0.3cm 0.35cm},clip]{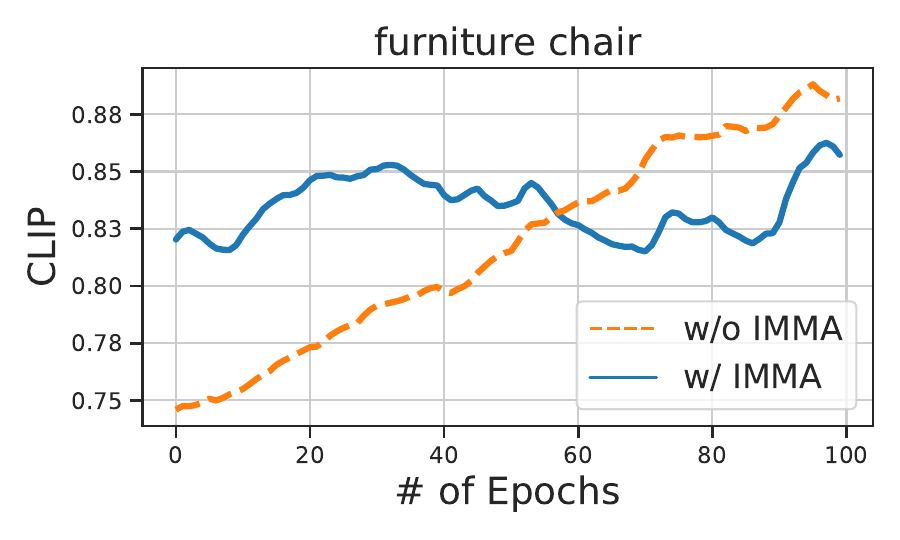} \\
    \includegraphics[height=1.7cm, trim={0.3cm 0.35cm 0.3cm 0.35cm},clip]{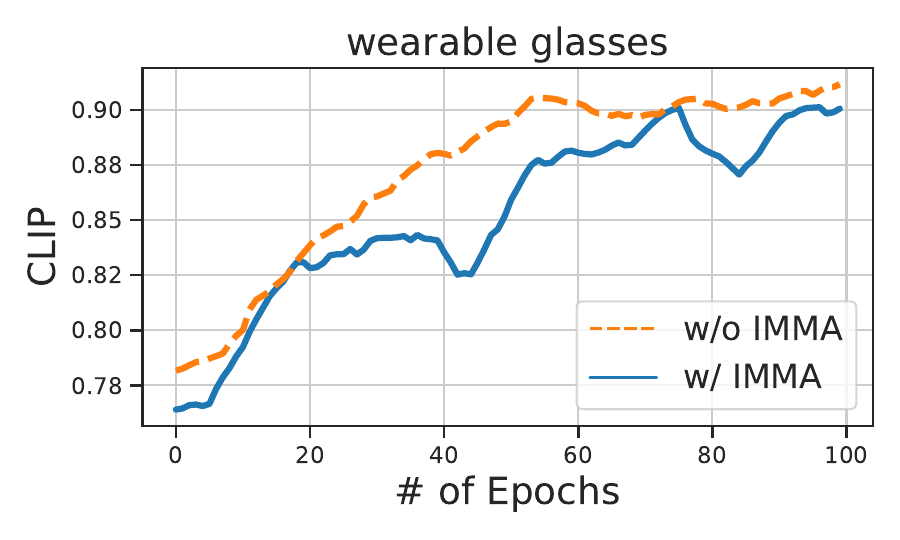} & \includegraphics[height=1.7cm, trim={0.3cm 0.35cm 0.3cm 0.35cm},clip]{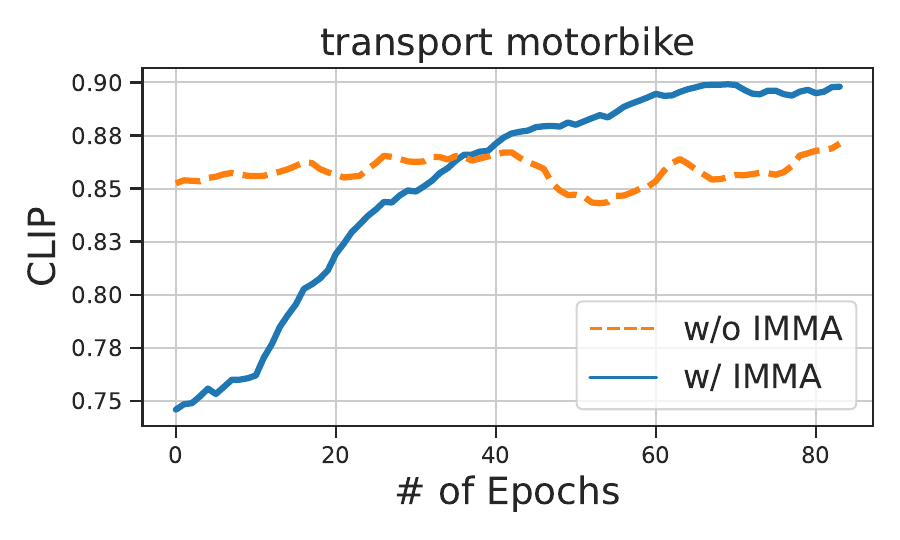} & \includegraphics[height=1.7cm, trim={0.3cm 0.35cm 0.3cm 0.35cm},clip]{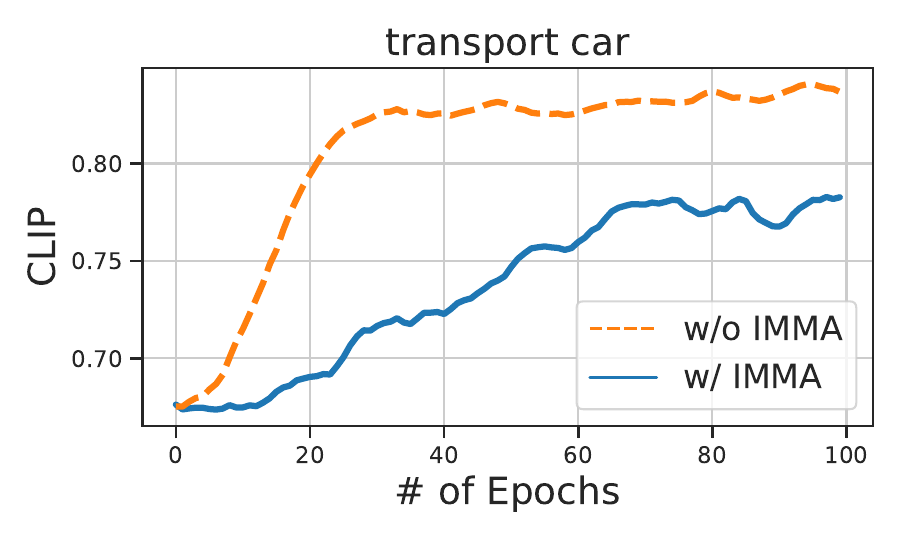} & \includegraphics[height=1.7cm, trim={0.3cm 0.35cm 0.3cm 0.35cm},clip]{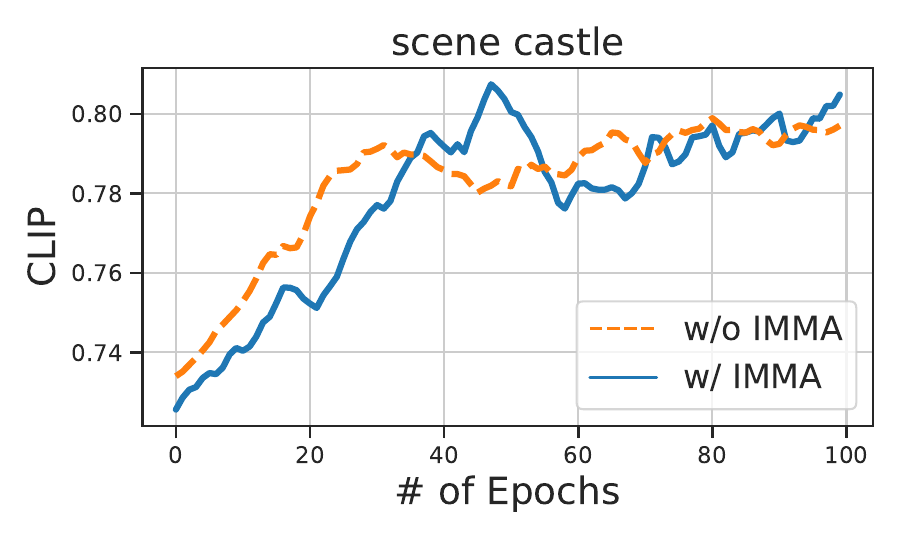} \\
    \includegraphics[height=1.7cm, trim={0.3cm 0.35cm 0.3cm 0.35cm},clip]{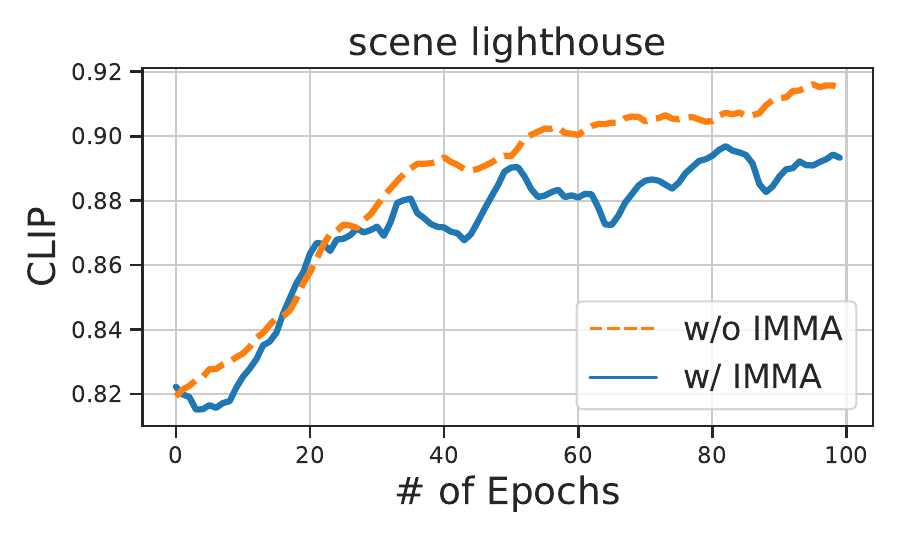}& \includegraphics[height=1.7cm, trim={0.3cm 0.35cm 0.3cm 0.35cm},clip]{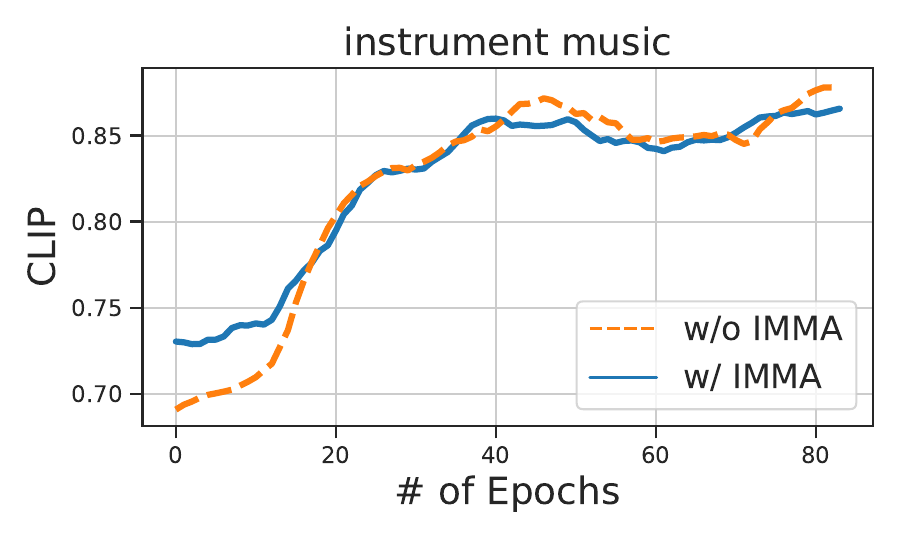} \\
     \includegraphics[height=1.7cm, trim={0.3cm 0.35cm 0.3cm 0.35cm},clip]{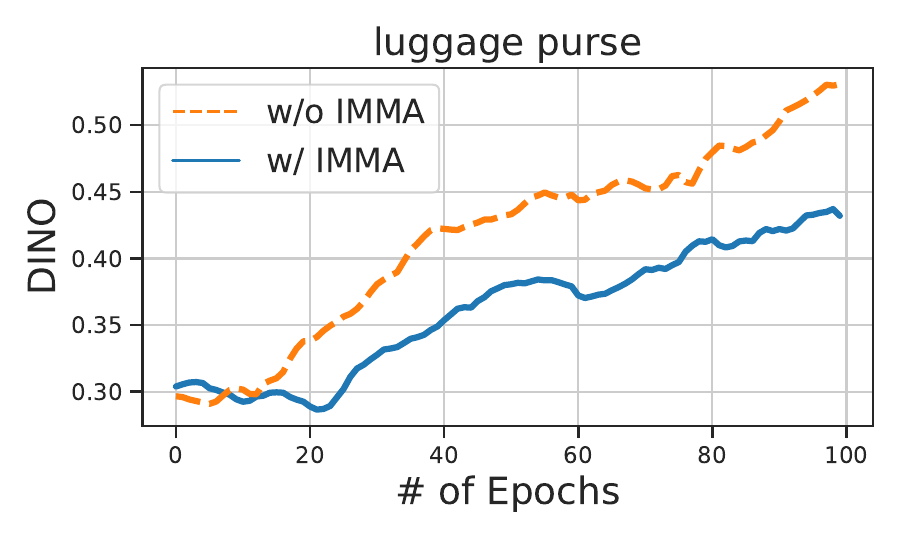} & \includegraphics[height=1.7cm, trim={0.3cm 0.35cm 0.3cm 0.35cm},clip]{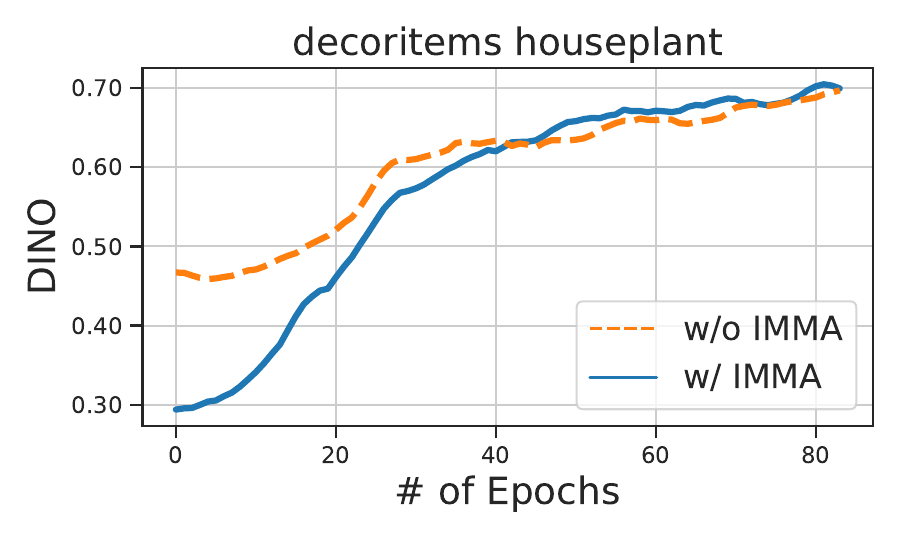} & \includegraphics[height=1.7cm, trim={0.3cm 0.35cm 0.3cm 0.35cm},clip]{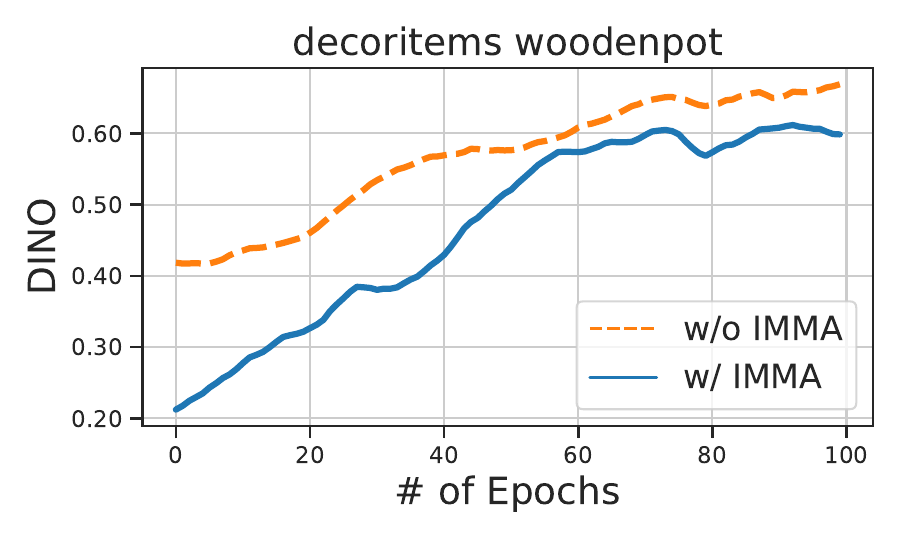} & \includegraphics[height=1.7cm, trim={0.3cm 0.35cm 0.3cm 0.35cm},clip]{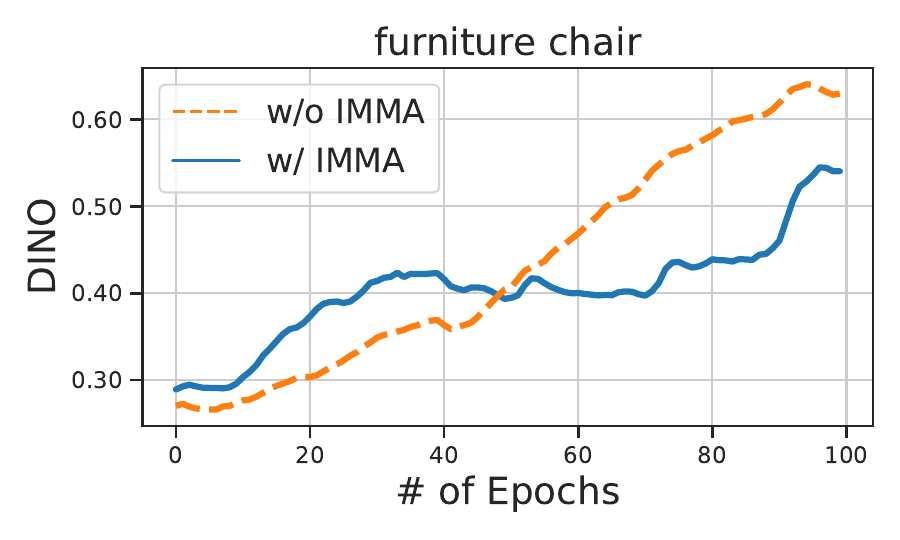} \\
     \includegraphics[height=1.7cm, trim={0.3cm 0.35cm 0.3cm 0.35cm},clip]{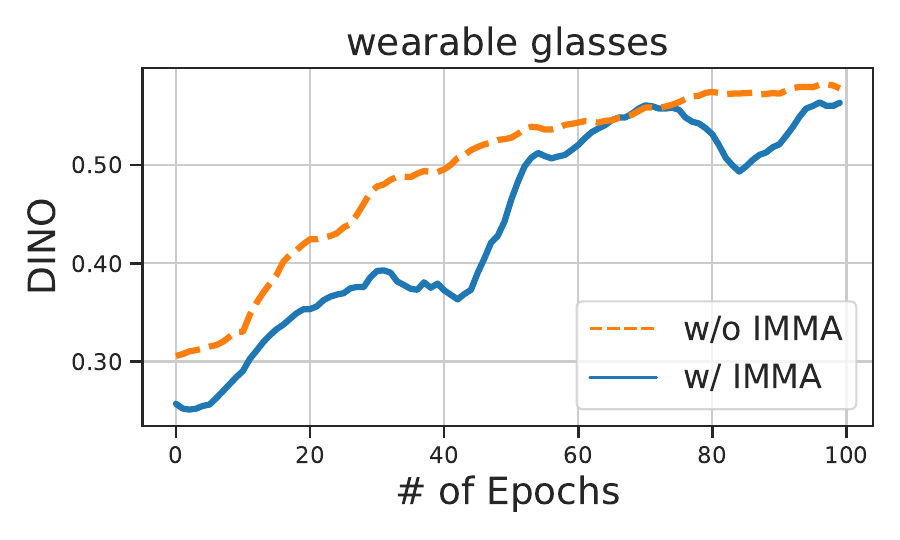} & \includegraphics[height=1.7cm, trim={0.3cm 0.35cm 0.3cm 0.35cm},clip]{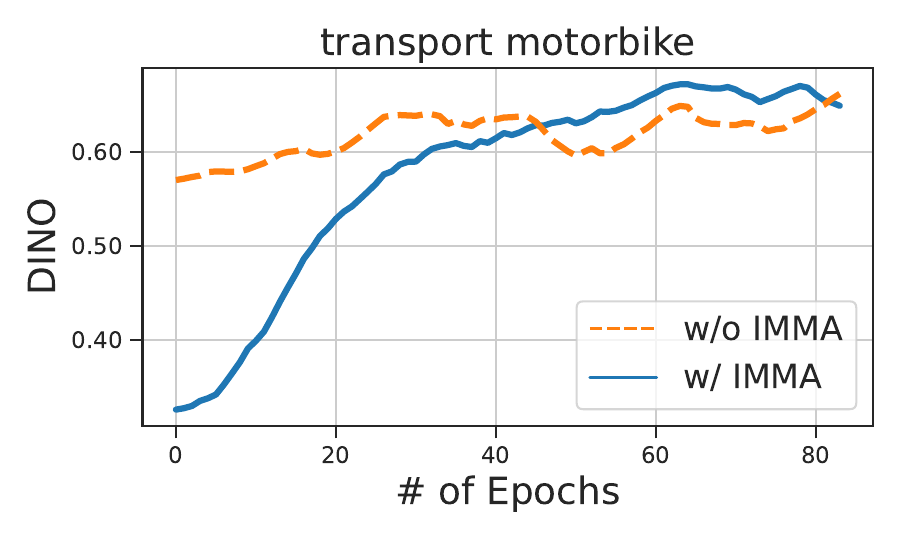} & \includegraphics[height=1.7cm, trim={0.3cm 0.35cm 0.3cm 0.35cm},clip]{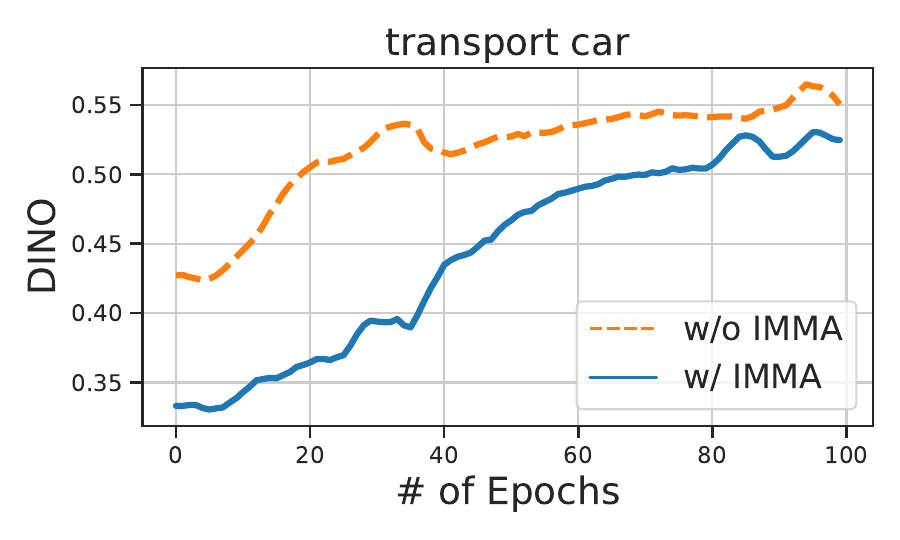} & \includegraphics[height=1.7cm, trim={0.3cm 0.35cm 0.3cm 0.35cm},clip]{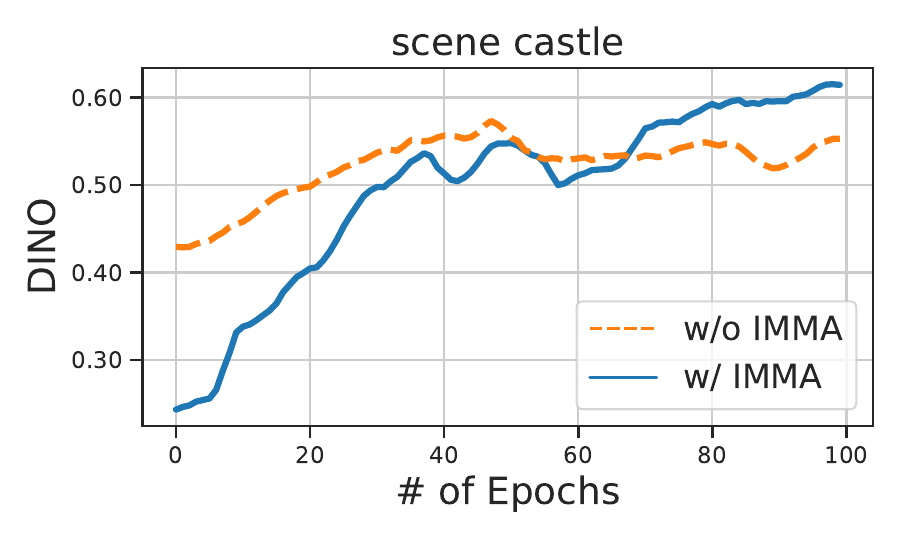} \\
     \includegraphics[height=1.7cm, trim={0.3cm 0.35cm 0.3cm 0.35cm},clip]{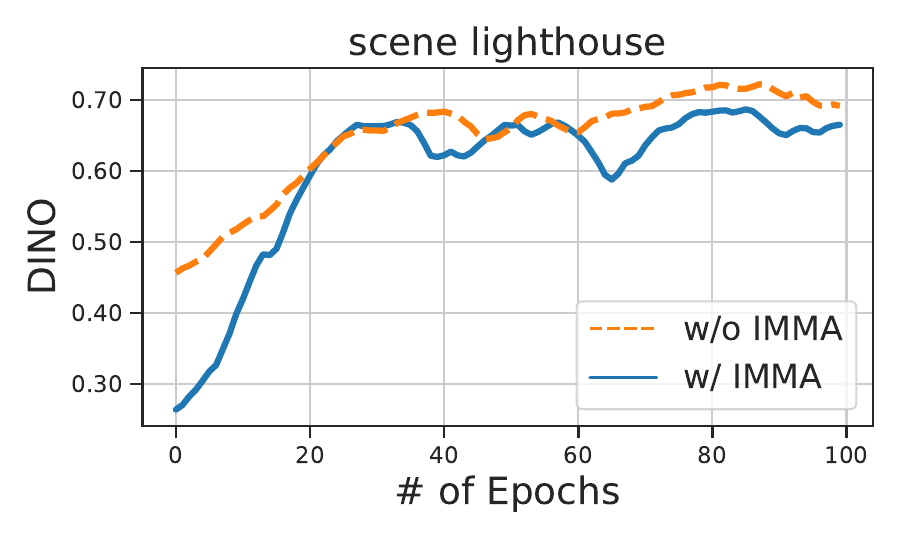}& \includegraphics[height=1.7cm, trim={0.3cm 0.35cm 0.3cm 0.35cm},clip]{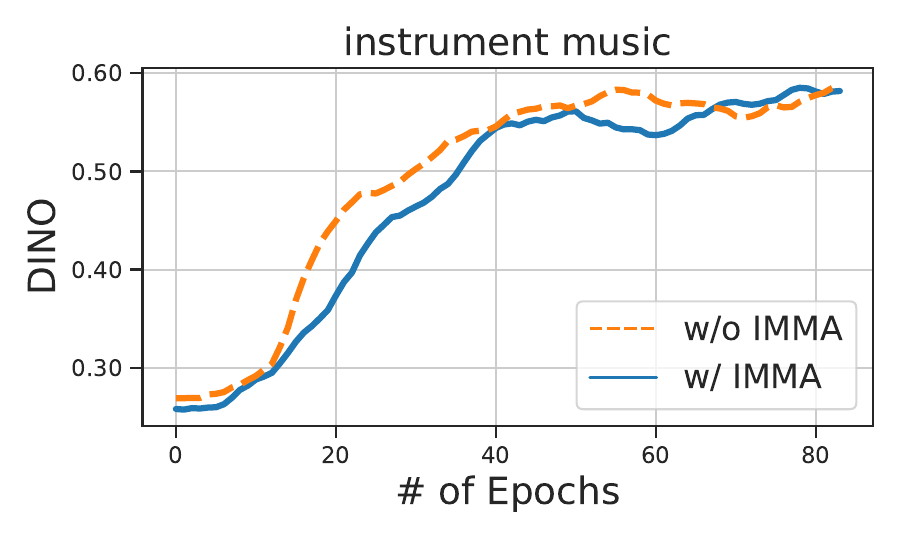} \\
    \end{tabular}
    \caption{\bf CLIP and DINO versus reference images after Dreambooth Adaptation {\color{noimma}w/o} and {\color{imma} w/} IMMA.}
    \label{fig:db_ref}
\end{figure*}

\begin{figure*}[t]
    \centering
    \setlength{\tabcolsep}{1pt}
    \begin{tabular}{ccccc}
     \includegraphics[height=1.7cm, trim={0.3cm 0.35cm 0.3cm 0.35cm},clip]{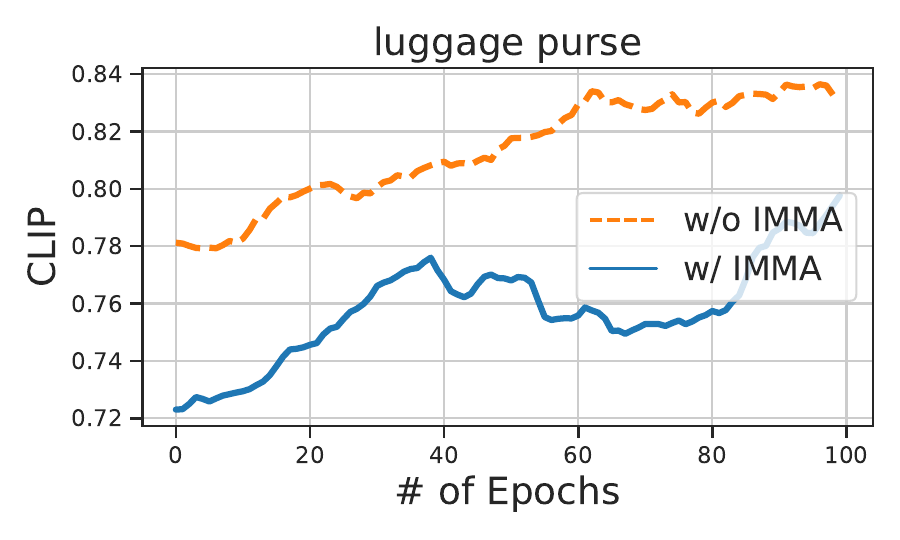} & \includegraphics[height=1.7cm, trim={0.3cm 0.35cm 0.3cm 0.35cm},clip]{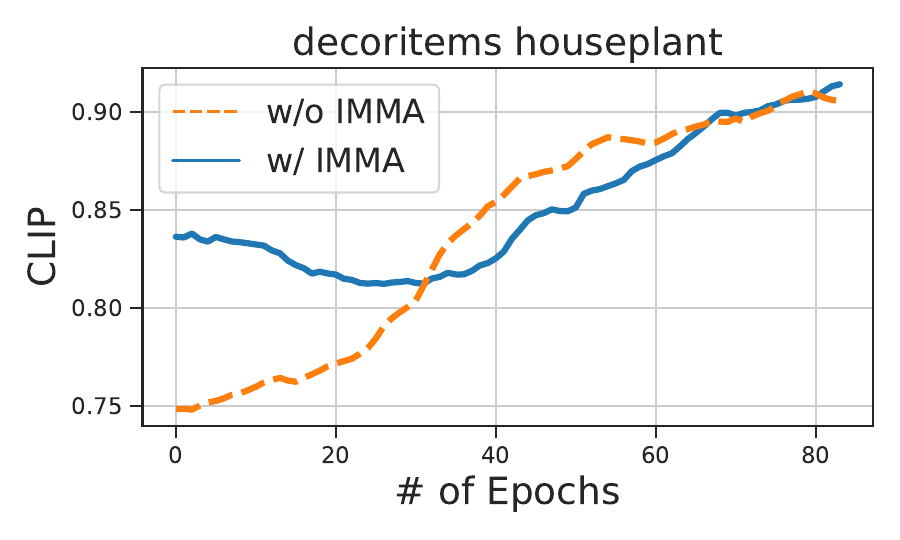} & \includegraphics[height=1.7cm, trim={0.3cm 0.35cm 0.3cm 0.35cm},clip]{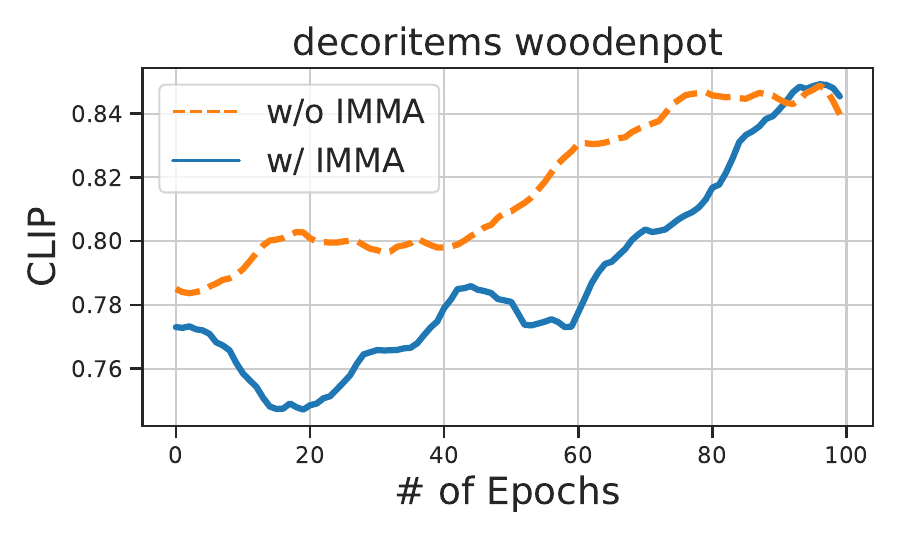} & \includegraphics[height=1.7cm, trim={0.3cm 0.35cm 0.3cm 0.35cm},clip]{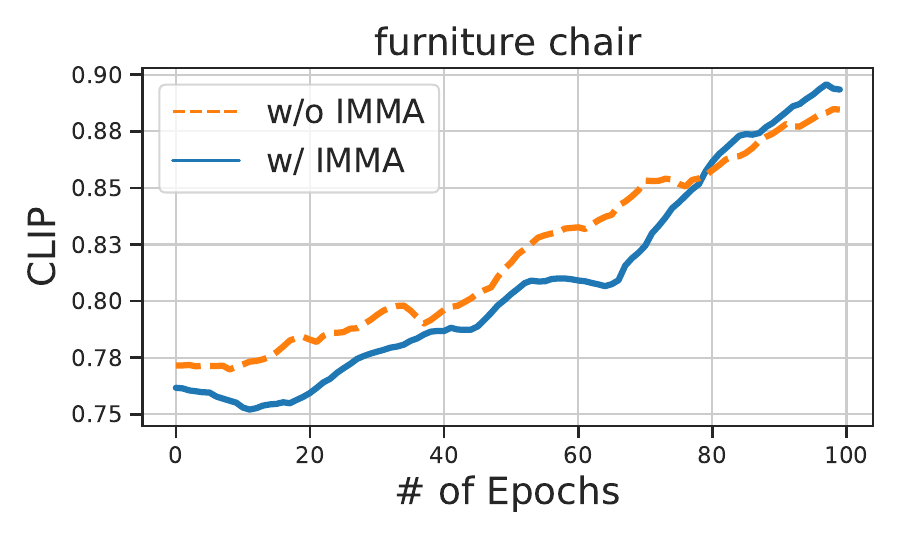}\\
     \includegraphics[height=1.7cm, trim={0.3cm 0.35cm 0.3cm 0.35cm},clip]{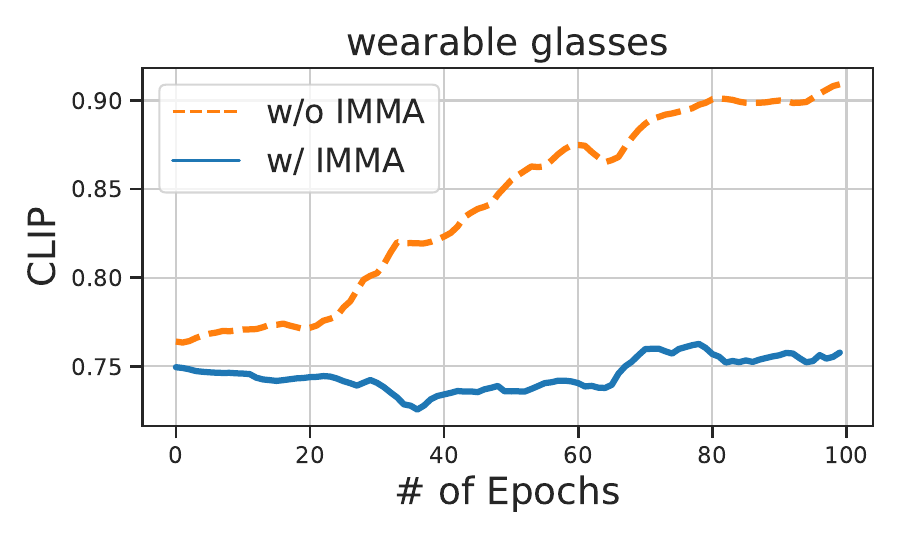} & \includegraphics[height=1.7cm, trim={0.3cm 0.35cm 0.3cm 0.35cm},clip]{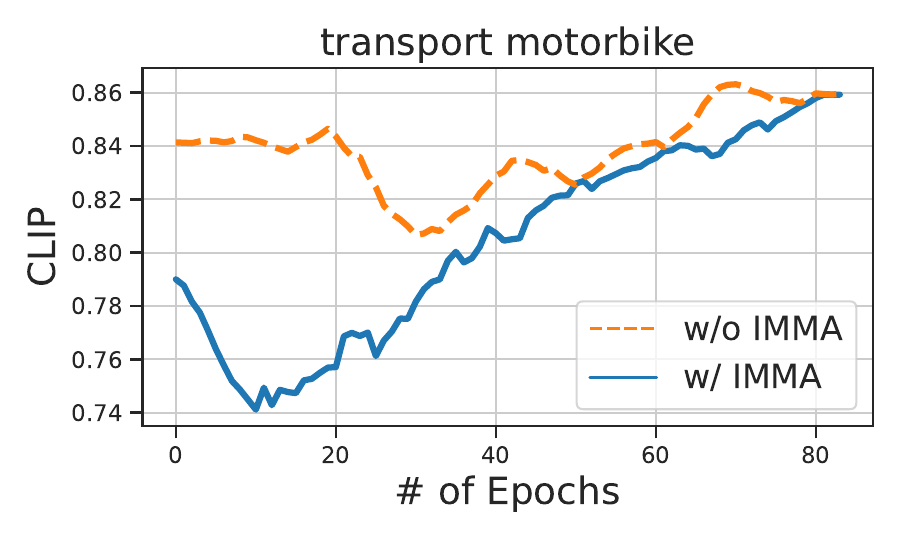} & \includegraphics[height=1.7cm, trim={0.3cm 0.35cm 0.3cm 0.35cm},clip]{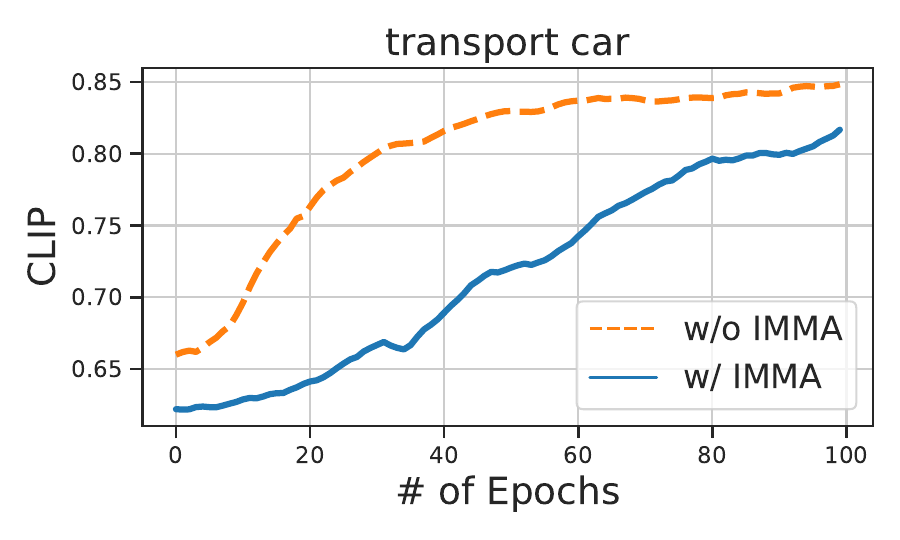} & \includegraphics[height=1.7cm, trim={0.3cm 0.35cm 0.3cm 0.35cm},clip]{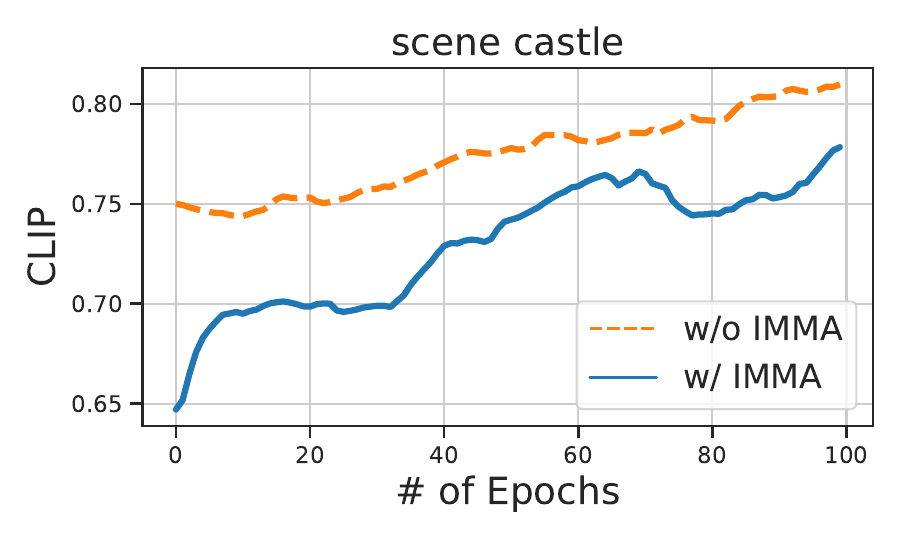} \\
     \includegraphics[height=1.7cm, trim={0.3cm 0.35cm 0.3cm 0.35cm},clip]{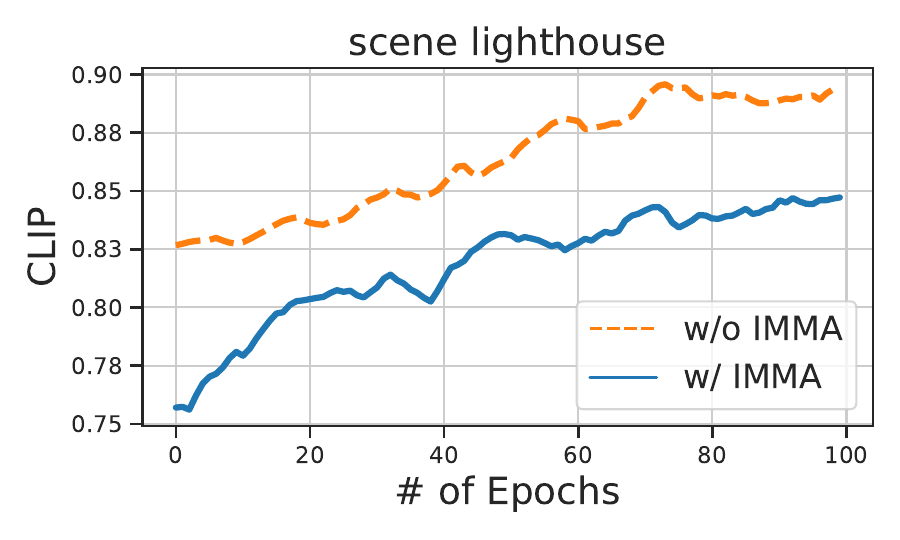}& \includegraphics[height=1.7cm, trim={0.3cm 0.35cm 0.3cm 0.35cm},clip]{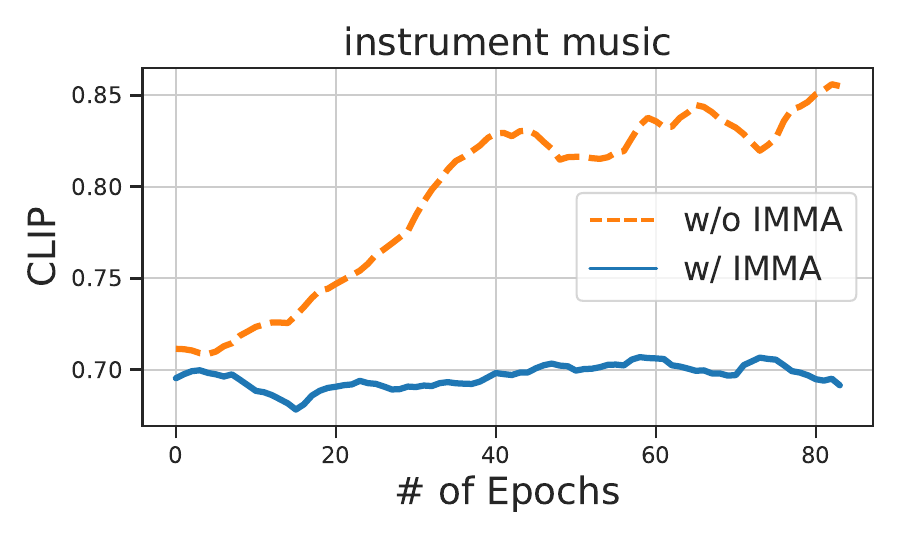} \\
     \includegraphics[height=1.7cm, trim={0.3cm 0.35cm 0.3cm 0.35cm},clip]{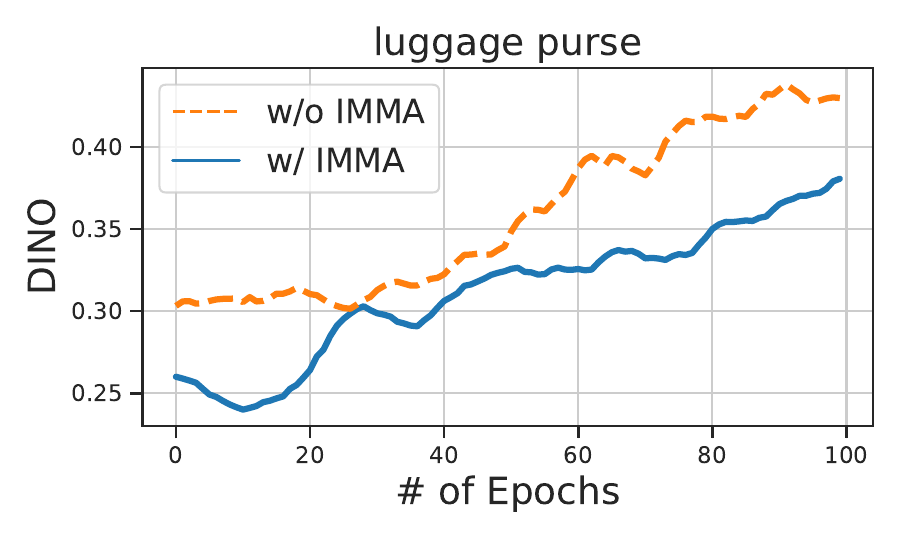} & \includegraphics[height=1.7cm, trim={0.3cm 0.35cm 0.3cm 0.35cm},clip]{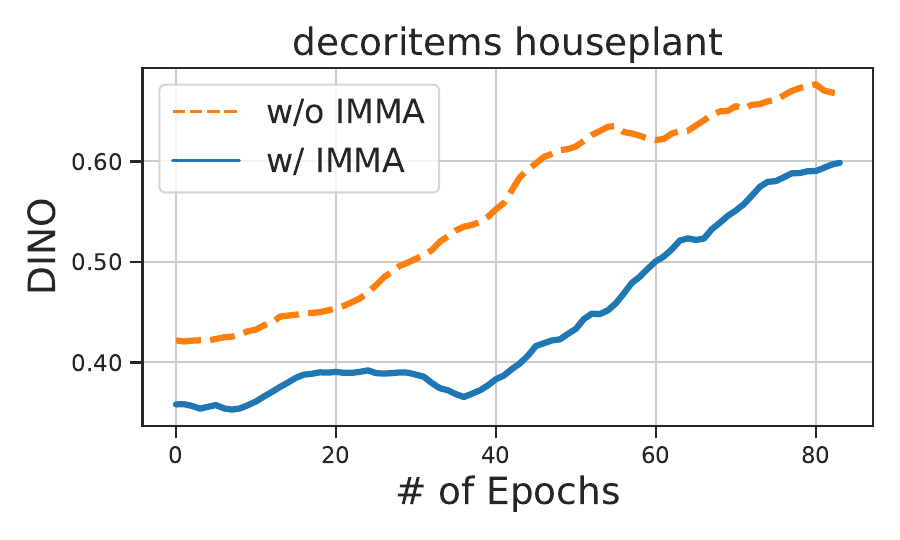} & \includegraphics[height=1.7cm, trim={0.3cm 0.35cm 0.3cm 0.35cm},clip]{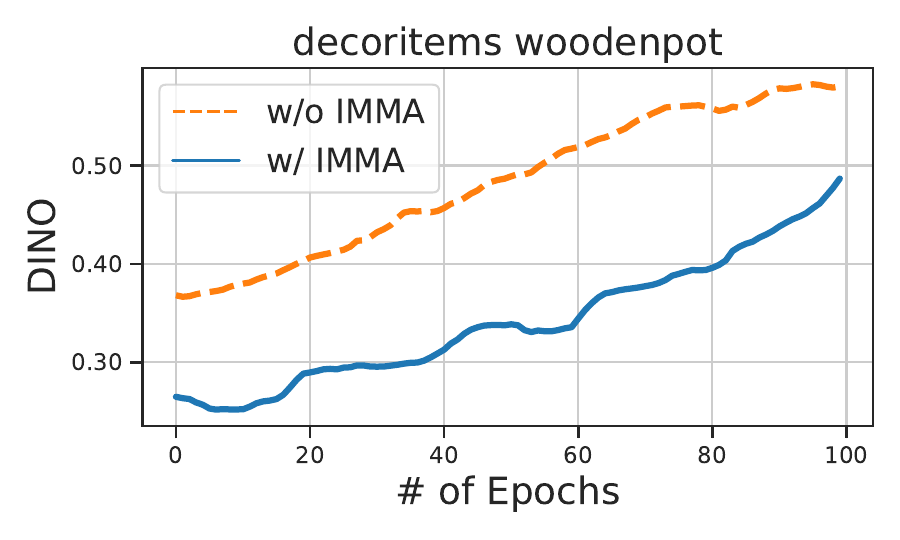} & \includegraphics[height=1.7cm, trim={0.3cm 0.35cm 0.3cm 0.35cm},clip]{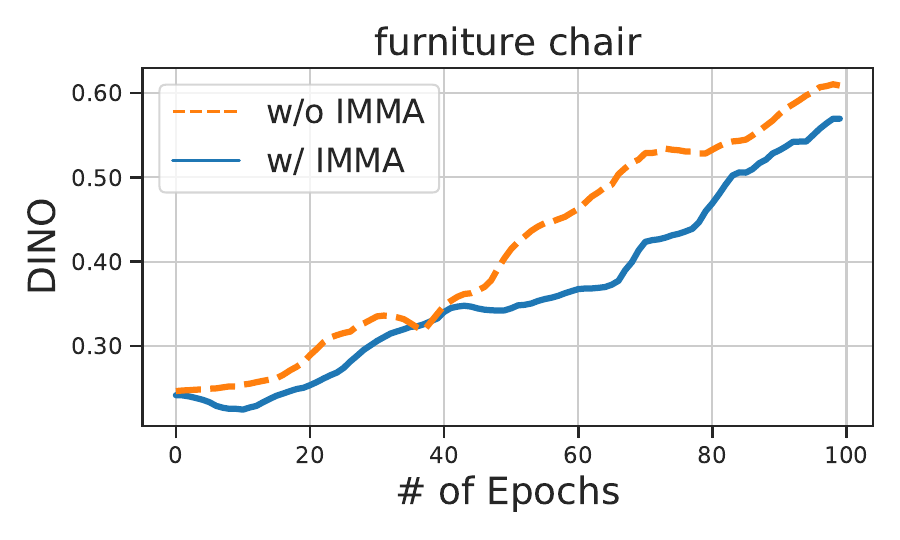}\\
     \includegraphics[height=1.7cm, trim={0.3cm 0.35cm 0.3cm 0.35cm},clip]{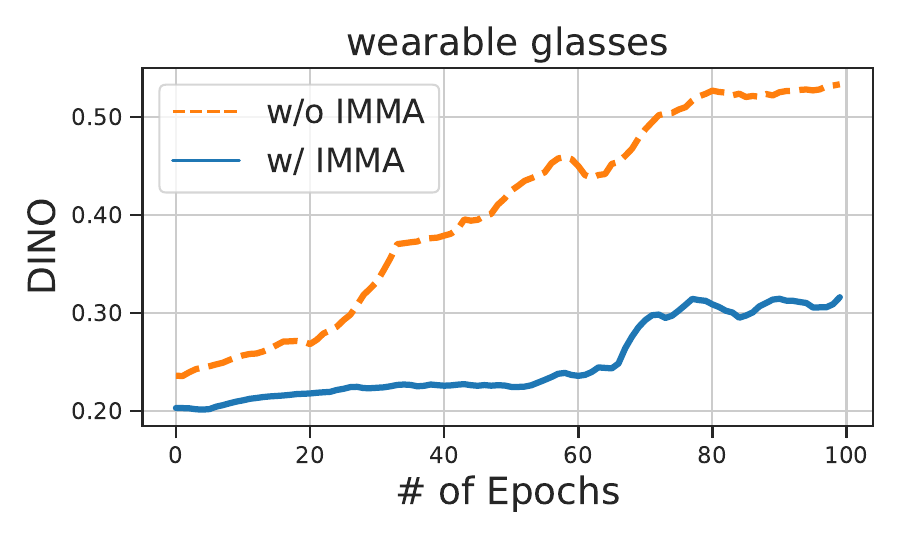} & \includegraphics[height=1.7cm, trim={0.3cm 0.35cm 0.3cm 0.35cm},clip]{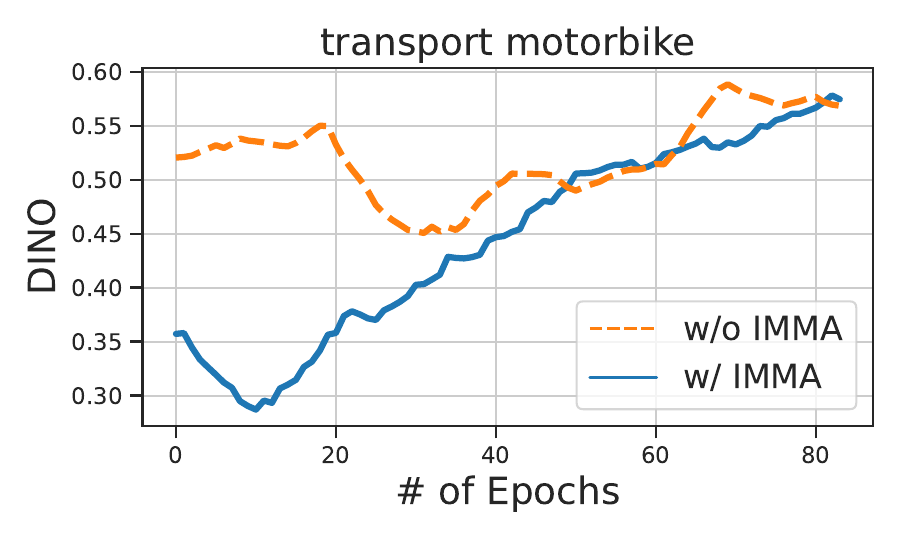} & \includegraphics[height=1.7cm, trim={0.3cm 0.35cm 0.3cm 0.35cm},clip]{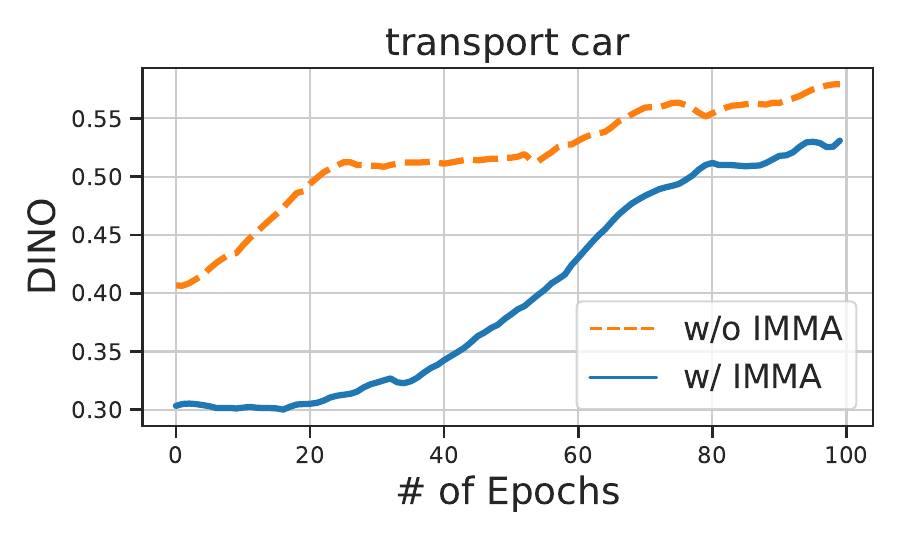} & \includegraphics[height=1.7cm, trim={0.3cm 0.35cm 0.3cm 0.35cm},clip]{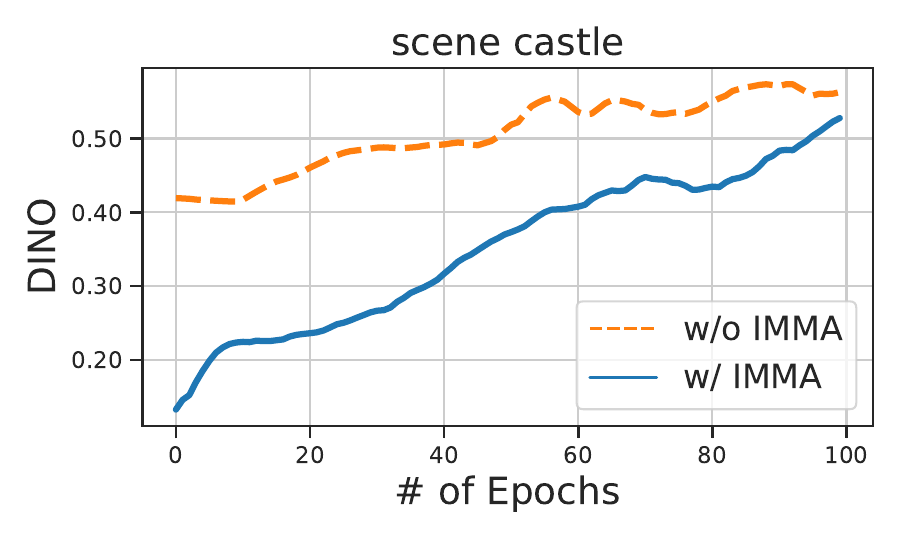} \\
     \includegraphics[height=1.7cm, trim={0.3cm 0.35cm 0.3cm 0.35cm},clip]{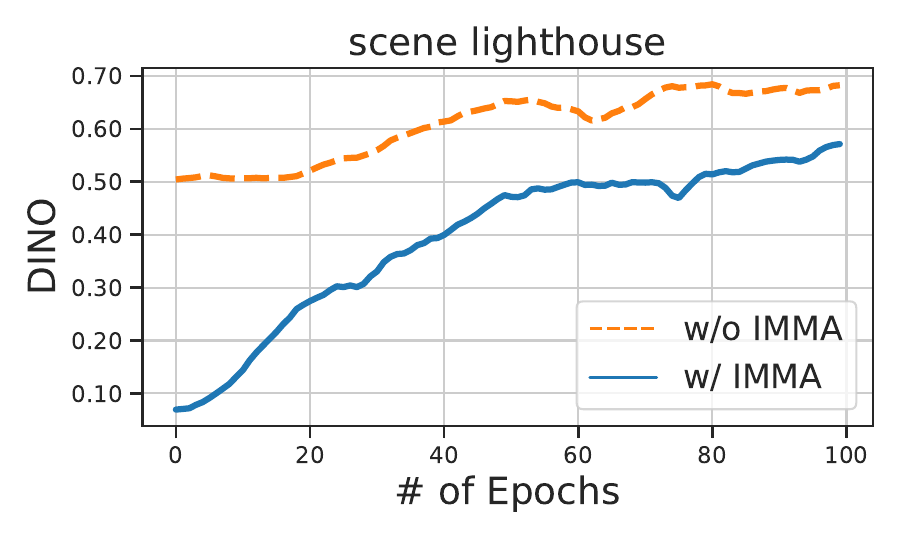}& \includegraphics[height=1.7cm, trim={0.3cm 0.35cm 0.3cm 0.35cm},clip]{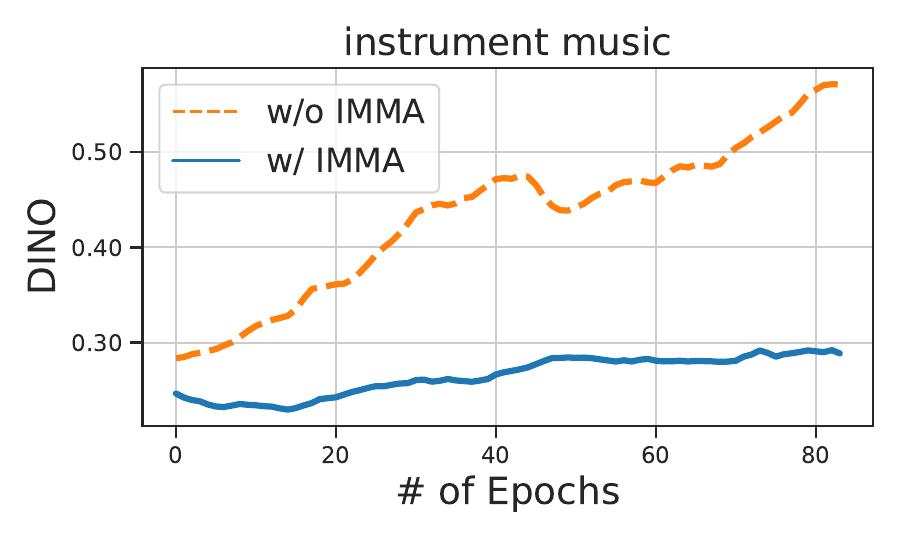} \\
    \end{tabular}
    \caption{\bf CLIP and DINO versus reference images after Dreambooth LoRA Adaptation {\color{noimma}w/o} and {\color{imma} w/} IMMA.}
    \label{fig:dbl_ref}
\end{figure*}

\begin{figure*}[t]
    \centering
    \setlength{\tabcolsep}{1pt}
    \begin{tabular}{ccccc}
     \includegraphics[height=1.7cm, trim={0.3cm 0.35cm 0.3cm 0.35cm},clip]{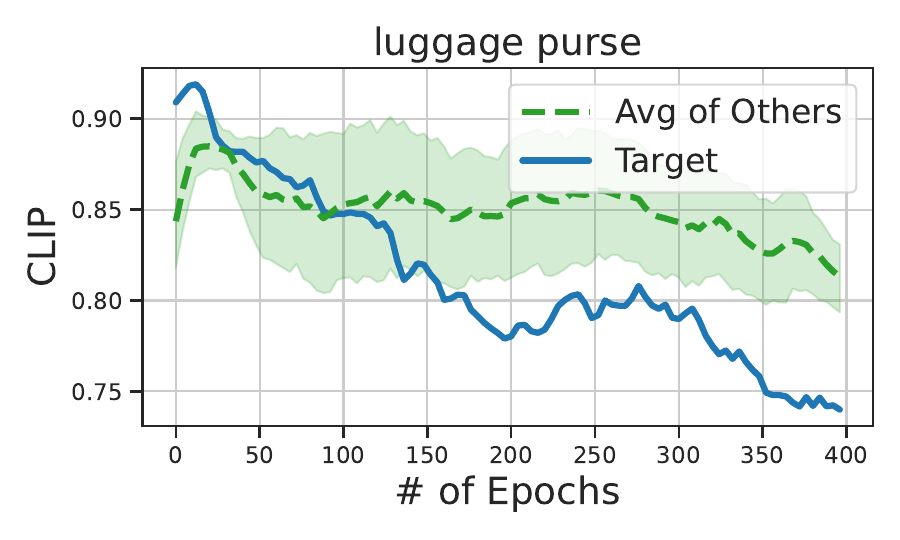} & \includegraphics[height=1.7cm, trim={0.3cm 0.35cm 0.3cm 0.35cm},clip]{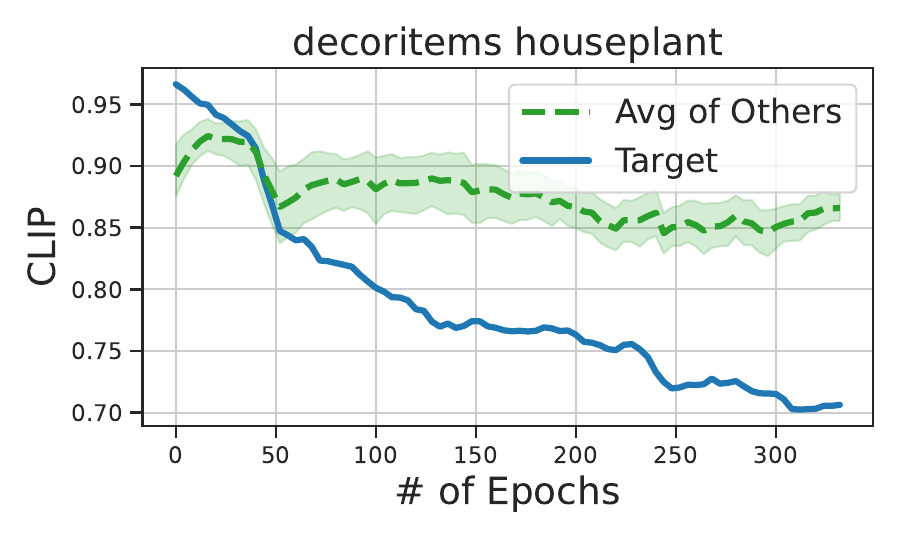} & \includegraphics[height=1.7cm, trim={0.3cm 0.35cm 0.3cm 0.35cm},clip]{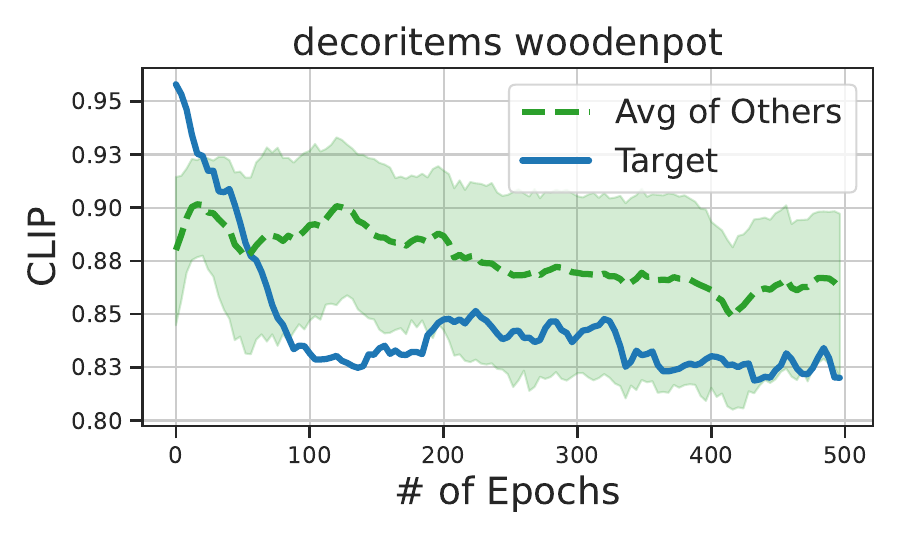} & \includegraphics[height=1.7cm, trim={0.3cm 0.35cm 0.3cm 0.35cm},clip]{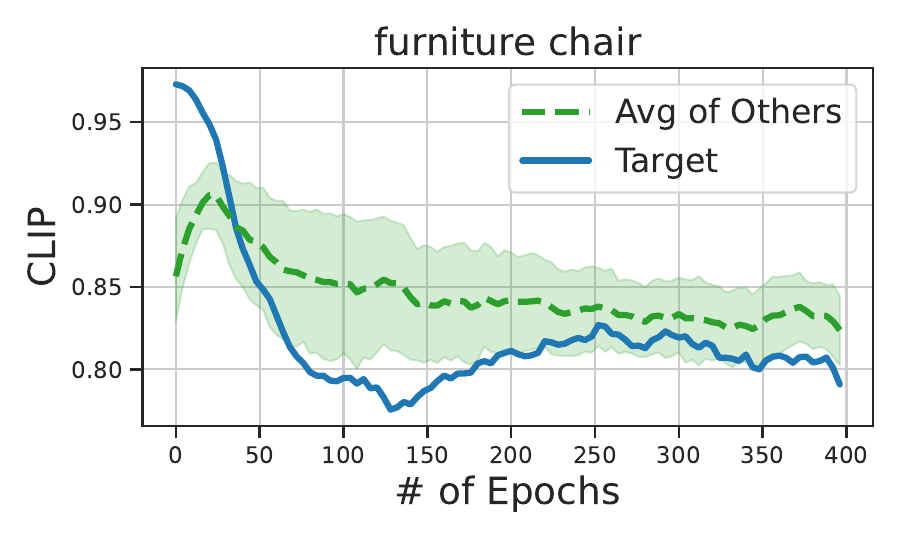}\\
     \includegraphics[height=1.7cm, trim={0.3cm 0.35cm 0.3cm 0.35cm},clip]{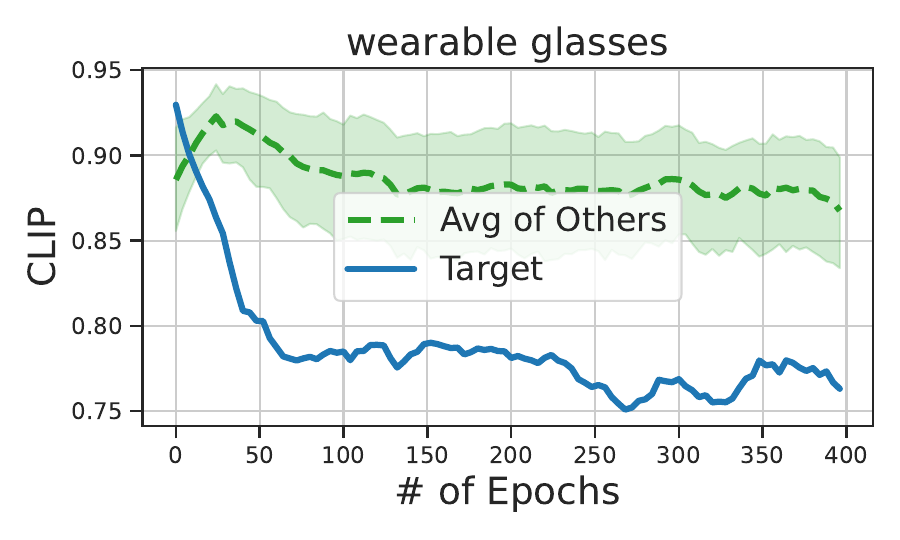} & \includegraphics[height=1.7cm, trim={0.3cm 0.35cm 0.3cm 0.35cm},clip]{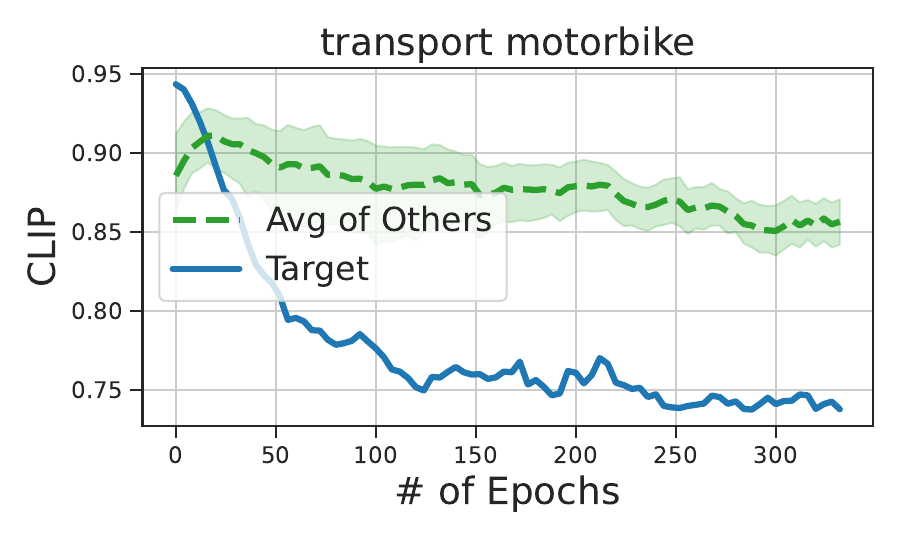} & \includegraphics[height=1.7cm, trim={0.3cm 0.35cm 0.3cm 0.35cm},clip]{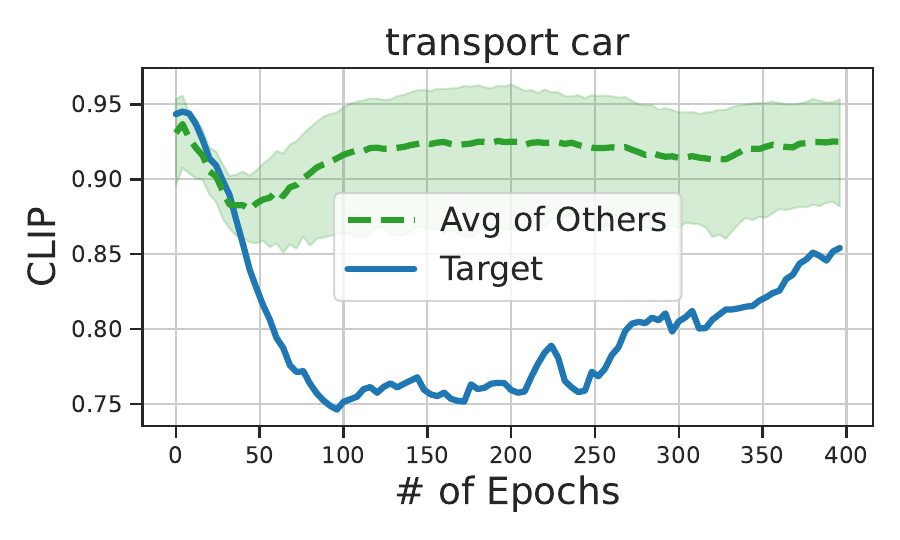} & \includegraphics[height=1.7cm, trim={0.3cm 0.35cm 0.3cm 0.35cm},clip]{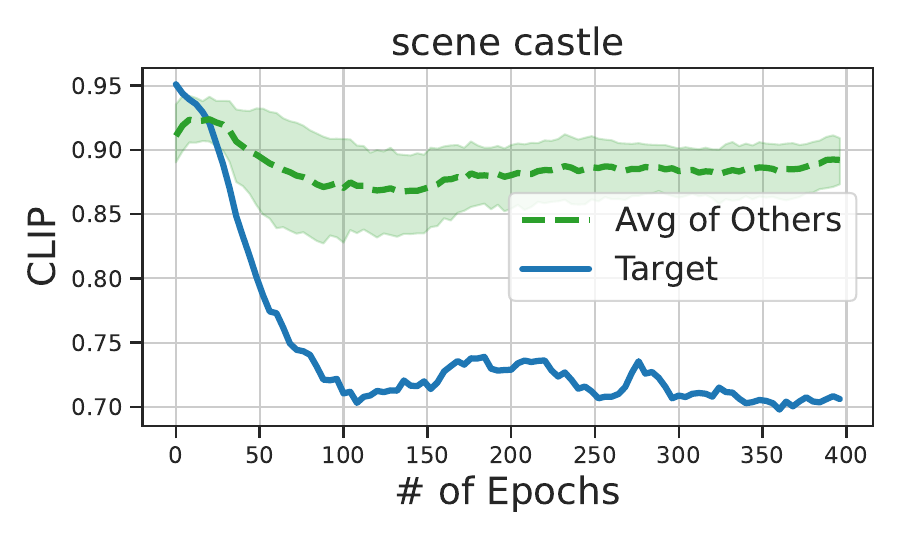} \\
     \includegraphics[height=1.7cm, trim={0.3cm 0.35cm 0.3cm 0.35cm},clip]{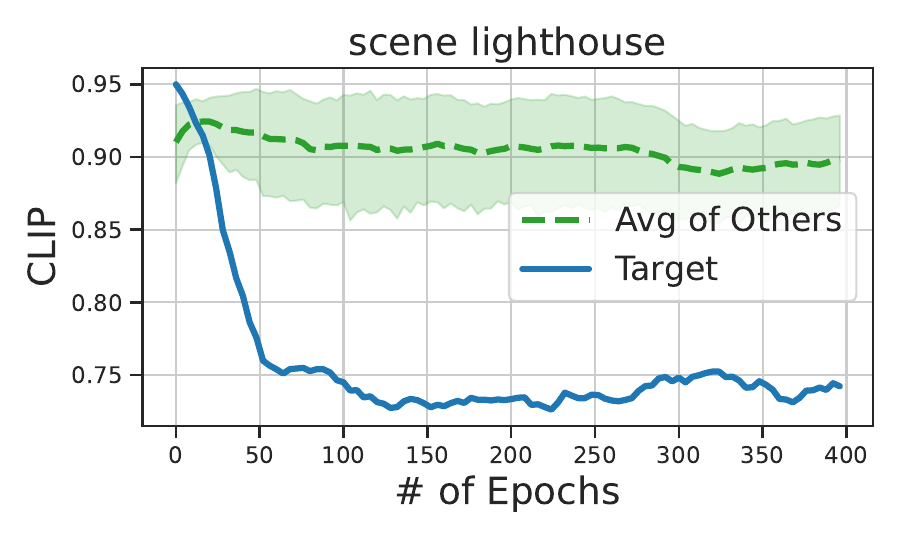}& \includegraphics[height=1.7cm, trim={0.3cm 0.35cm 0.3cm 0.35cm},clip]{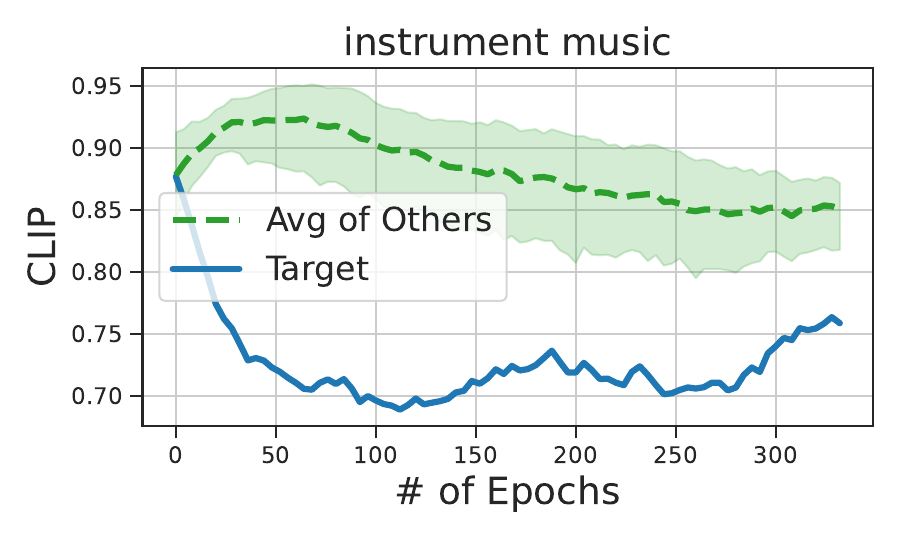} \\
     \includegraphics[height=1.7cm, trim={0.3cm 0.35cm 0.3cm 0.35cm},clip]{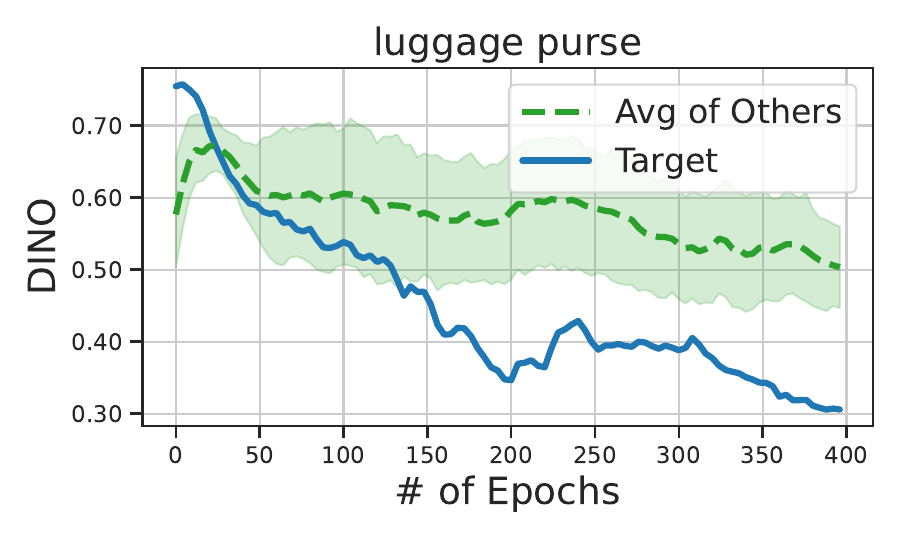} & \includegraphics[height=1.7cm, trim={0.3cm 0.35cm 0.3cm 0.35cm},clip]{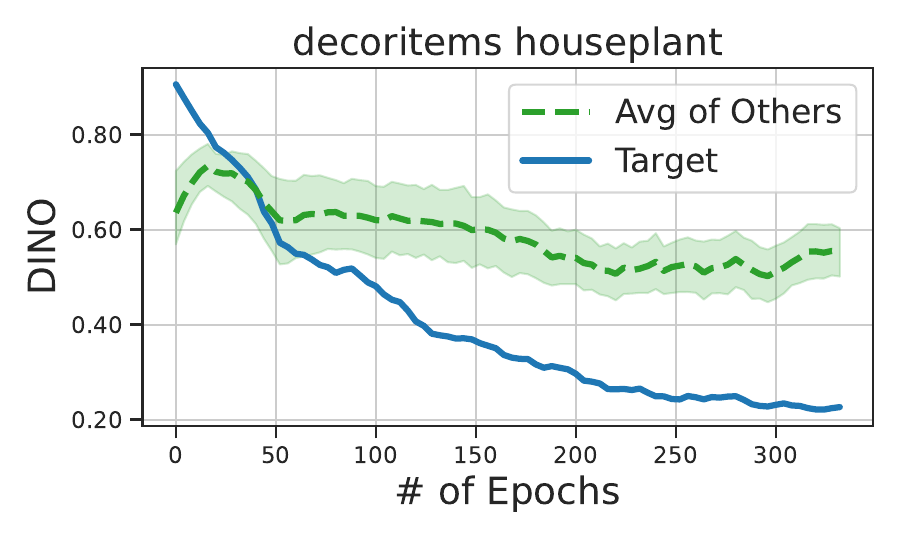} & \includegraphics[height=1.7cm, trim={0.3cm 0.35cm 0.3cm 0.35cm},clip]{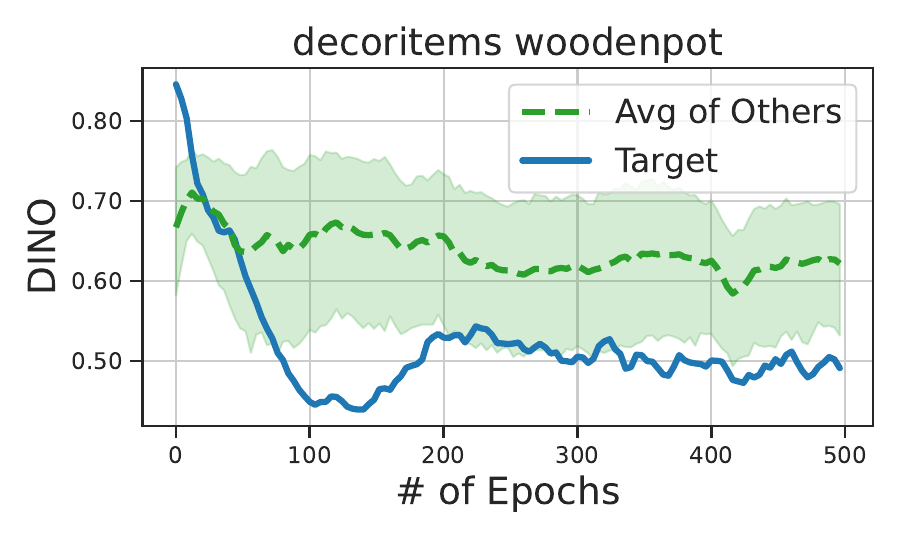} & \includegraphics[height=1.7cm, trim={0.3cm 0.35cm 0.3cm 0.35cm},clip]{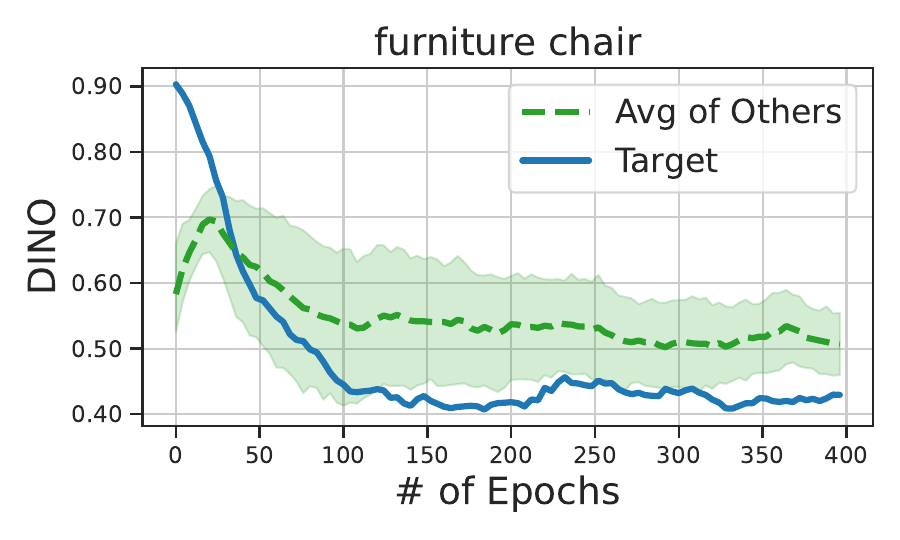}\\ 
     \includegraphics[height=1.7cm, trim={0.3cm 0.35cm 0.3cm 0.35cm},clip]{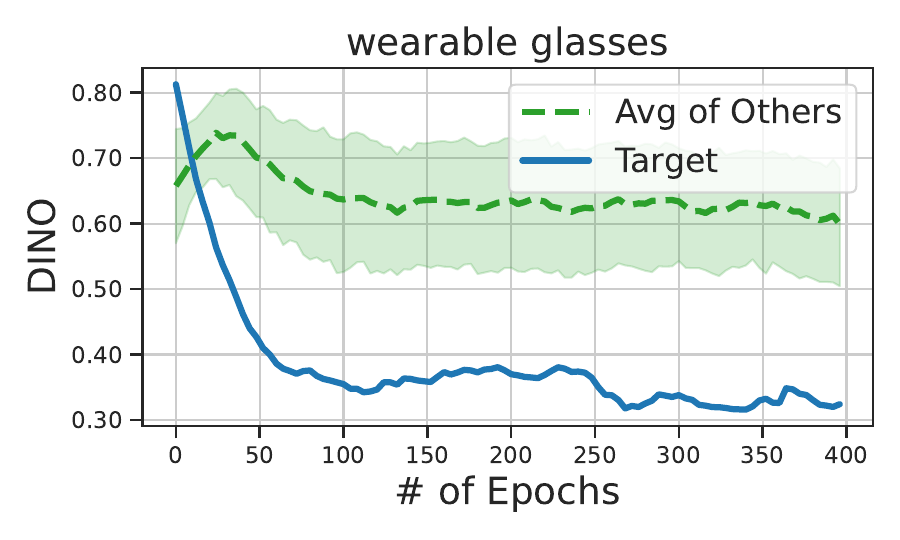} & \includegraphics[height=1.7cm, trim={0.3cm 0.35cm 0.3cm 0.35cm},clip]{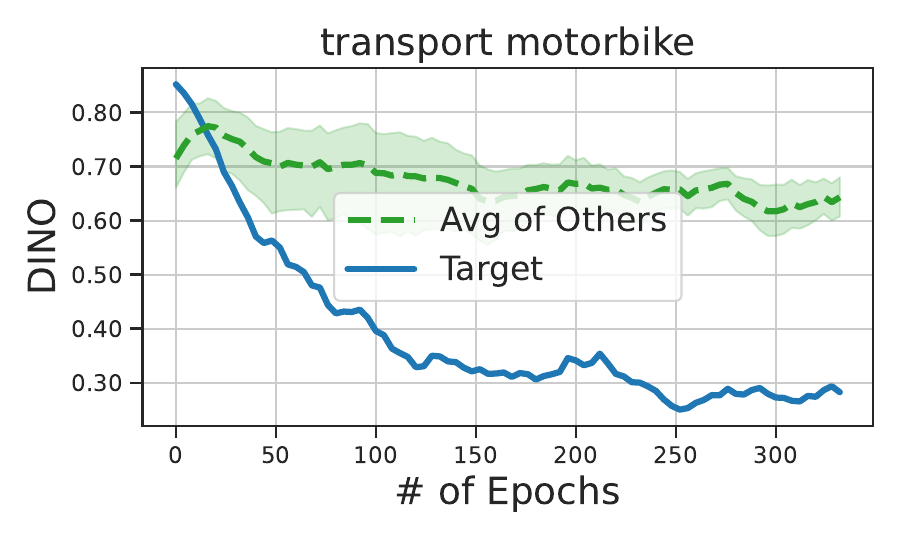} & \includegraphics[height=1.7cm, trim={0.3cm 0.35cm 0.3cm 0.35cm},clip]{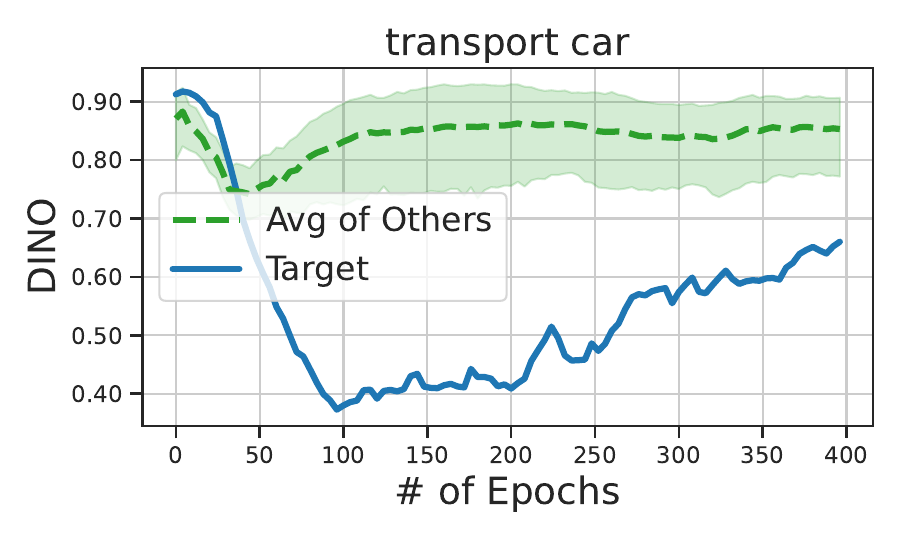} & \includegraphics[height=1.7cm, trim={0.3cm 0.35cm 0.3cm 0.35cm},clip]{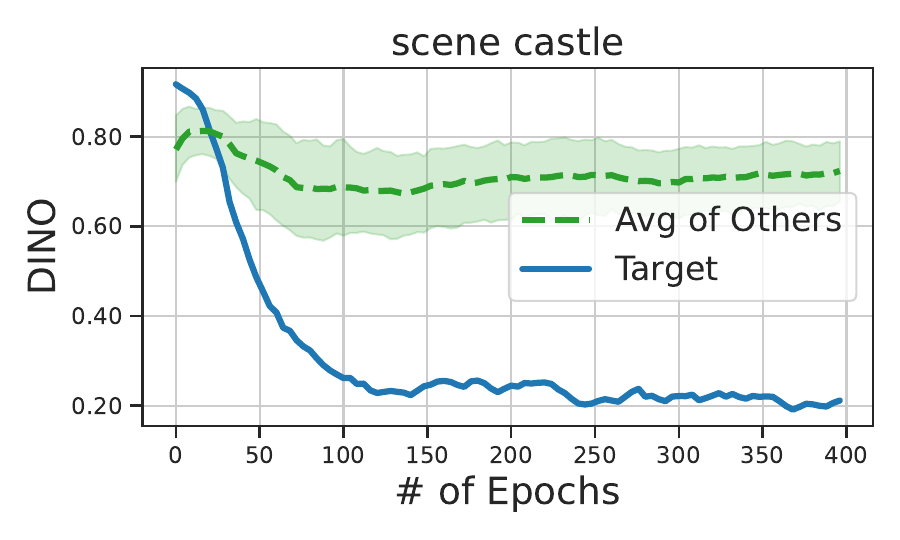} \\
     \includegraphics[height=1.7cm, trim={0.3cm 0.35cm 0.3cm 0.35cm},clip]{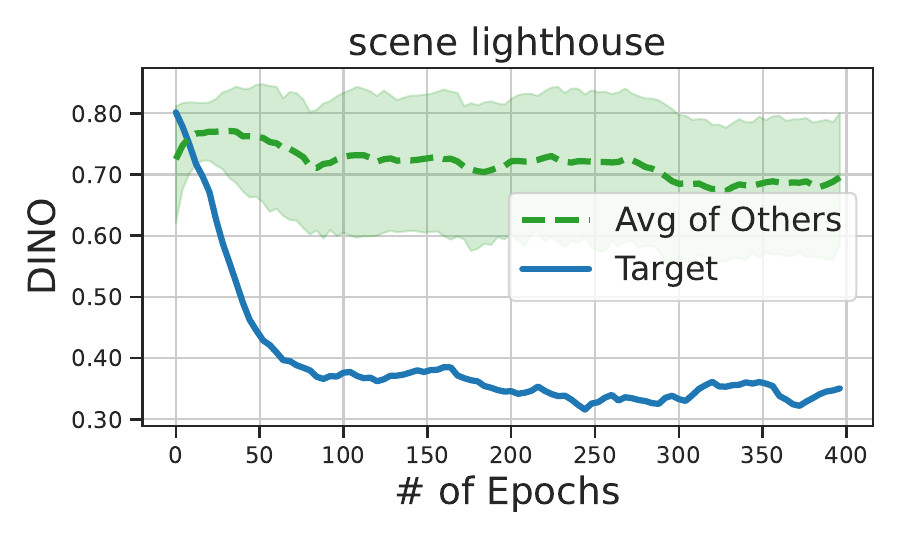}& \includegraphics[height=1.7cm, trim={0.3cm 0.35cm 0.3cm 0.35cm},clip]{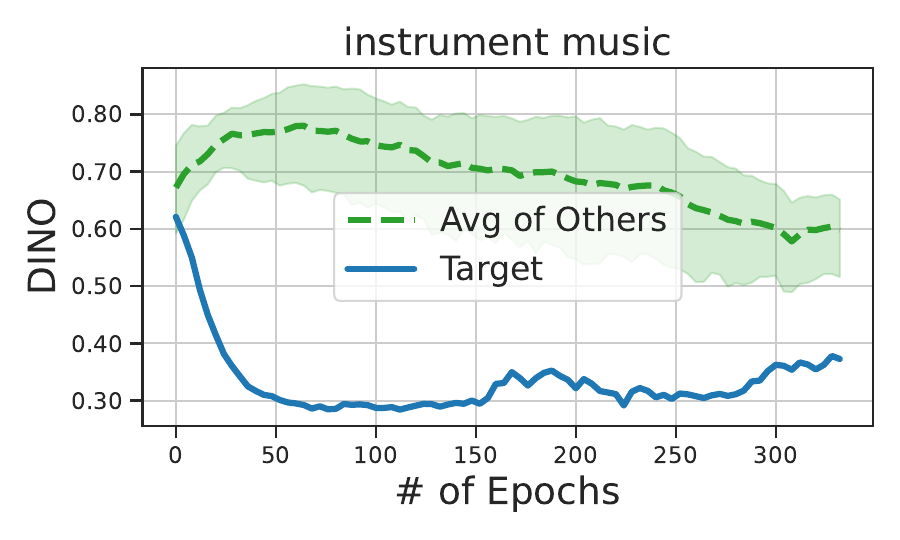} \\
    \end{tabular}
    \caption{\bf CLIP and DINO of Textual Inversion Adaptation {\color{noimma}w/o} and {\color{imma} w/} IMMA.}
    \label{fig:ti_quan}
\end{figure*}

\begin{figure*}[t]
    \centering
    \setlength{\tabcolsep}{1pt}
    \begin{tabular}{ccccc}
     \includegraphics[height=1.7cm, trim={0.3cm 0.35cm 0.3cm 0.35cm},clip]{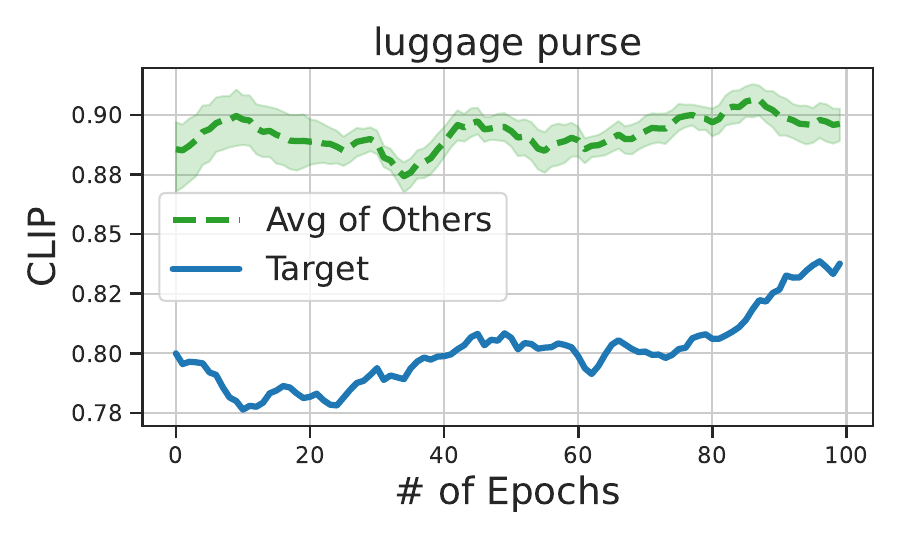} & \includegraphics[height=1.7cm, trim={0.3cm 0.35cm 0.3cm 0.35cm},clip]{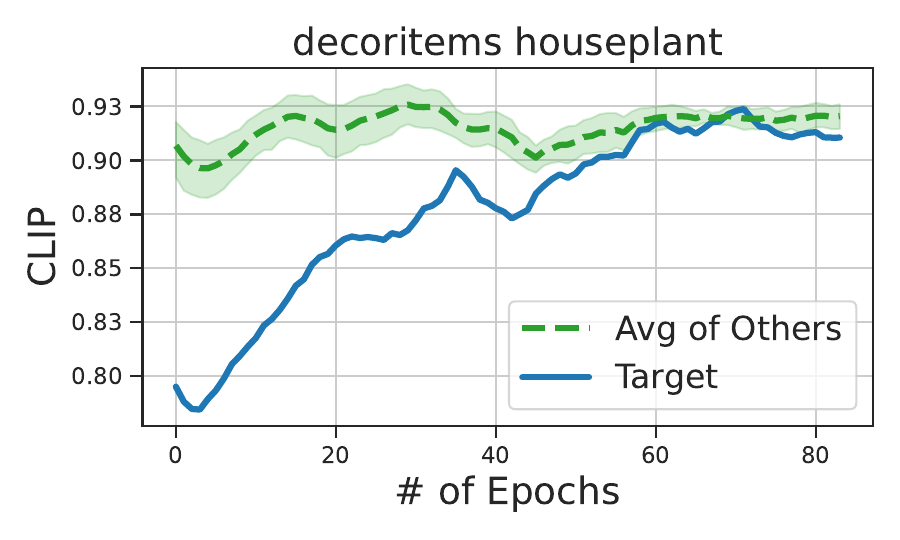} & \includegraphics[height=1.7cm, trim={0.3cm 0.35cm 0.3cm 0.35cm},clip]{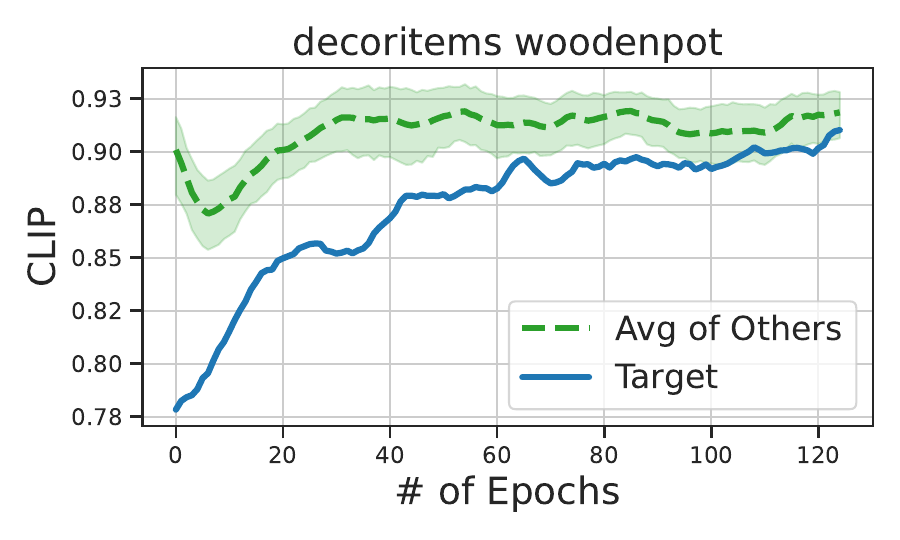} & \includegraphics[height=1.7cm, trim={0.3cm 0.35cm 0.3cm 0.35cm},clip]{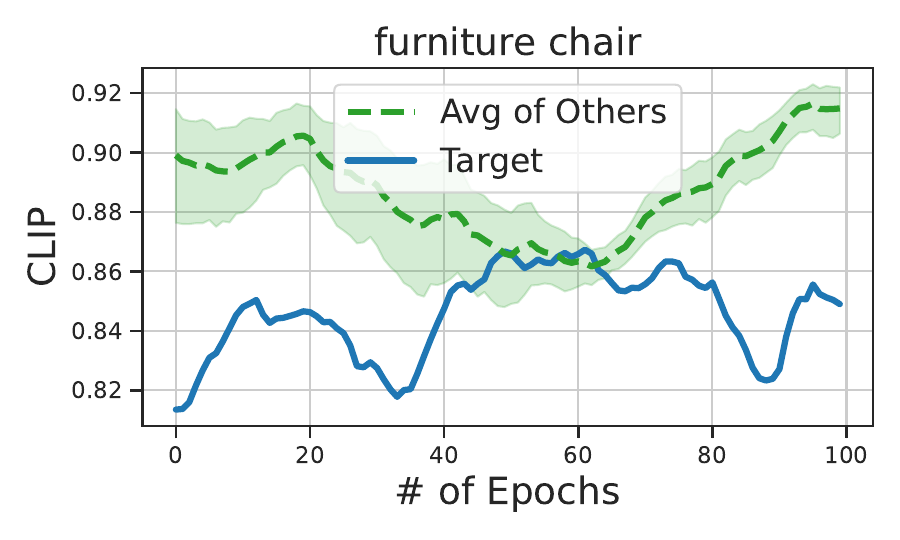}\\ 
     \includegraphics[height=1.7cm, trim={0.3cm 0.35cm 0.3cm 0.35cm},clip]{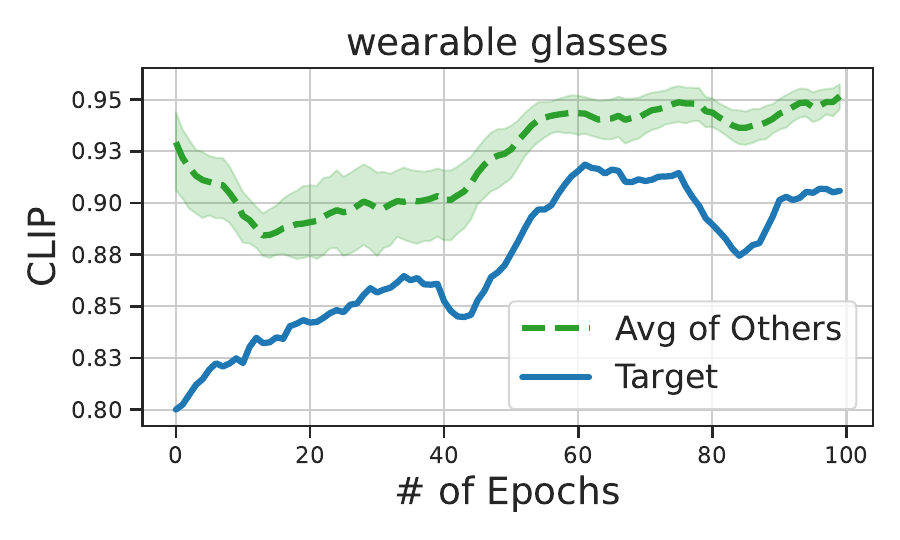} & \includegraphics[height=1.7cm, trim={0.3cm 0.35cm 0.3cm 0.35cm},clip]{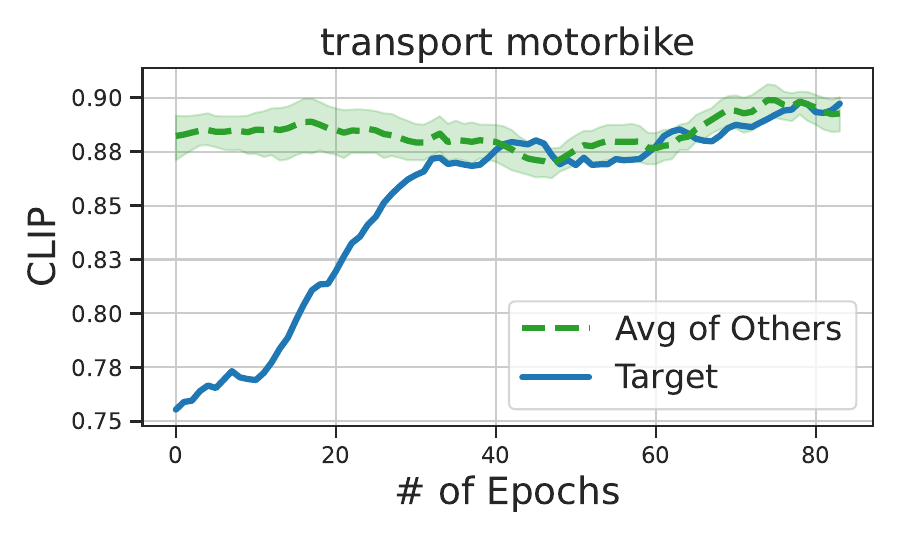} & \includegraphics[height=1.7cm, trim={0.3cm 0.35cm 0.3cm 0.35cm},clip]{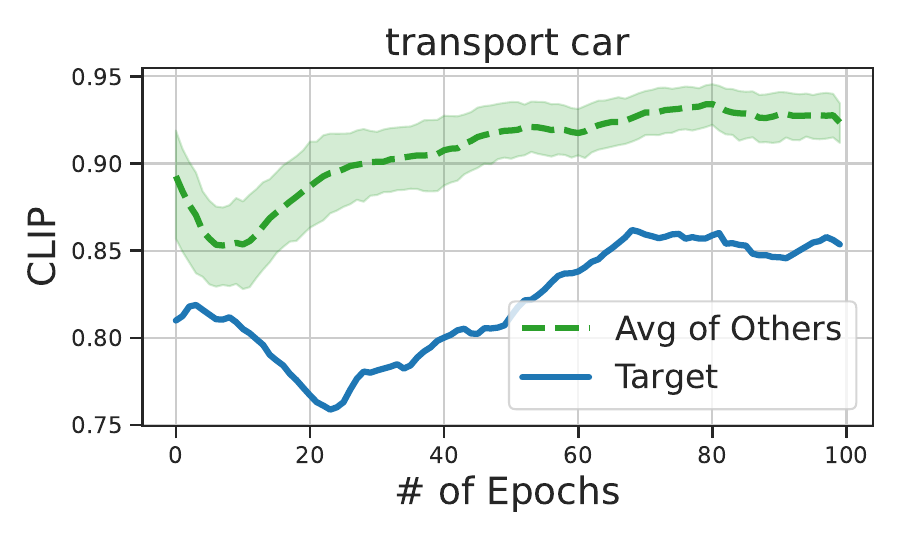} & \includegraphics[height=1.7cm, trim={0.3cm 0.35cm 0.3cm 0.35cm},clip]{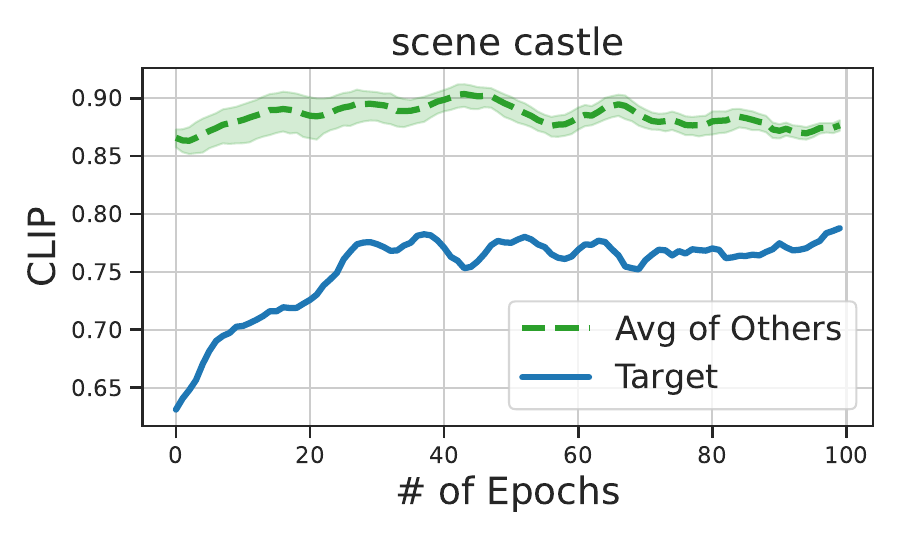} \\
     \includegraphics[height=1.7cm, trim={0.3cm 0.35cm 0.3cm 0.35cm},clip]{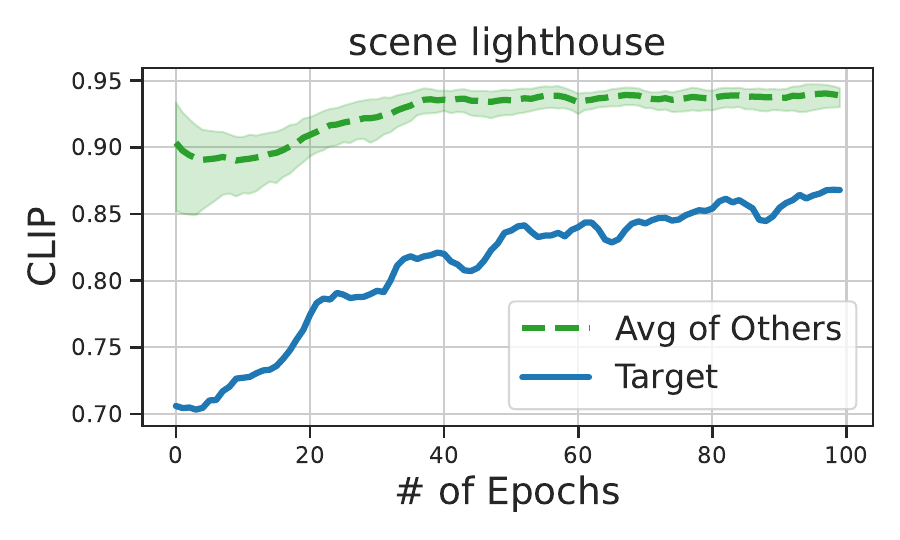}& \includegraphics[height=1.7cm, trim={0.3cm 0.35cm 0.3cm 0.35cm},clip]{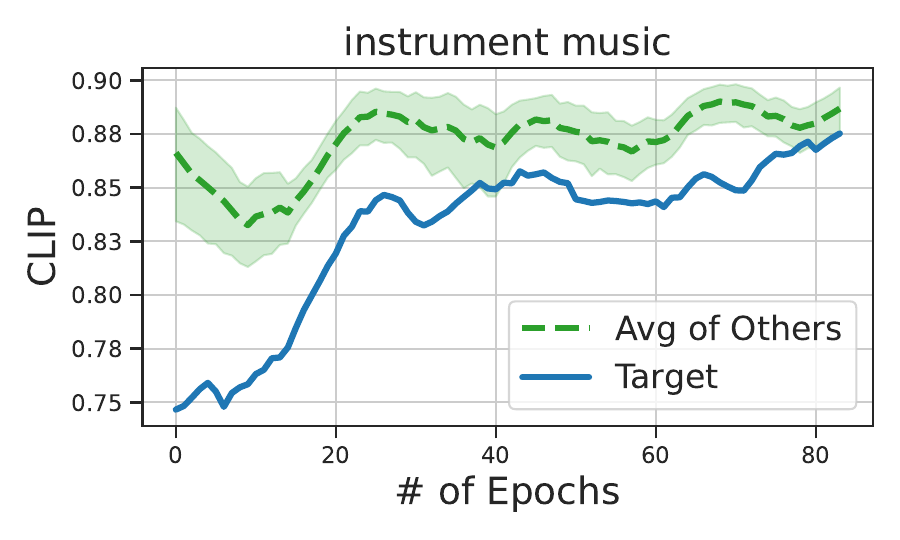} \\
      \includegraphics[height=1.7cm, trim={0.3cm 0.35cm 0.3cm 0.35cm},clip]{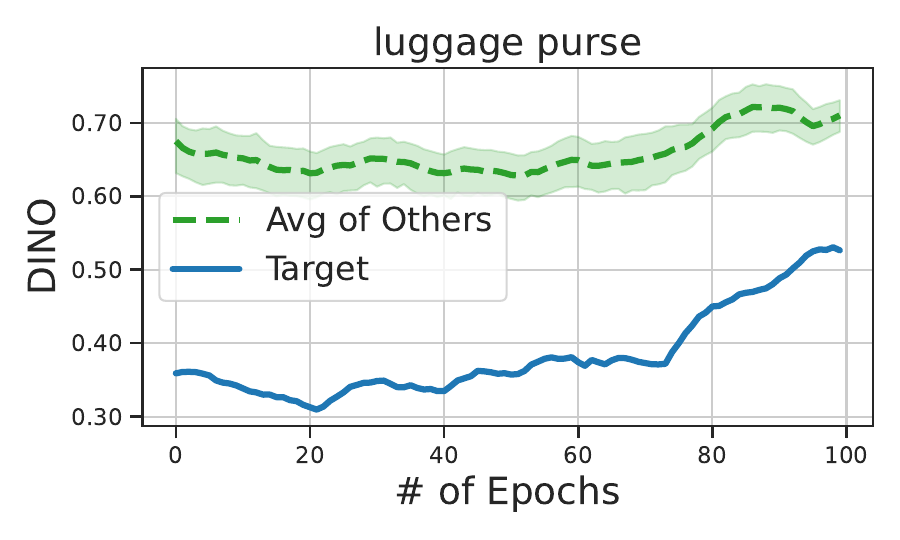} & \includegraphics[height=1.7cm, trim={0.3cm 0.35cm 0.3cm 0.35cm},clip]{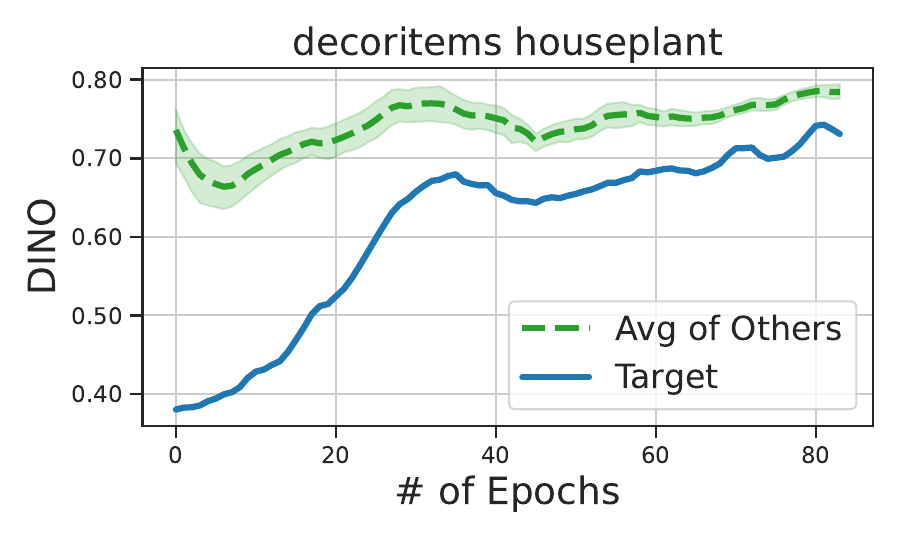} & \includegraphics[height=1.7cm, trim={0.3cm 0.35cm 0.3cm 0.35cm},clip]{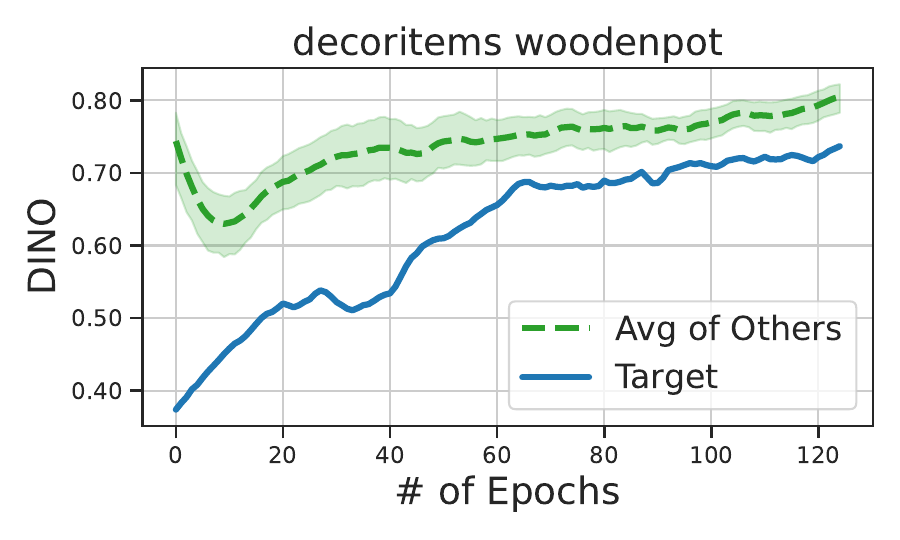} & \includegraphics[height=1.7cm, trim={0.3cm 0.35cm 0.3cm 0.35cm},clip]{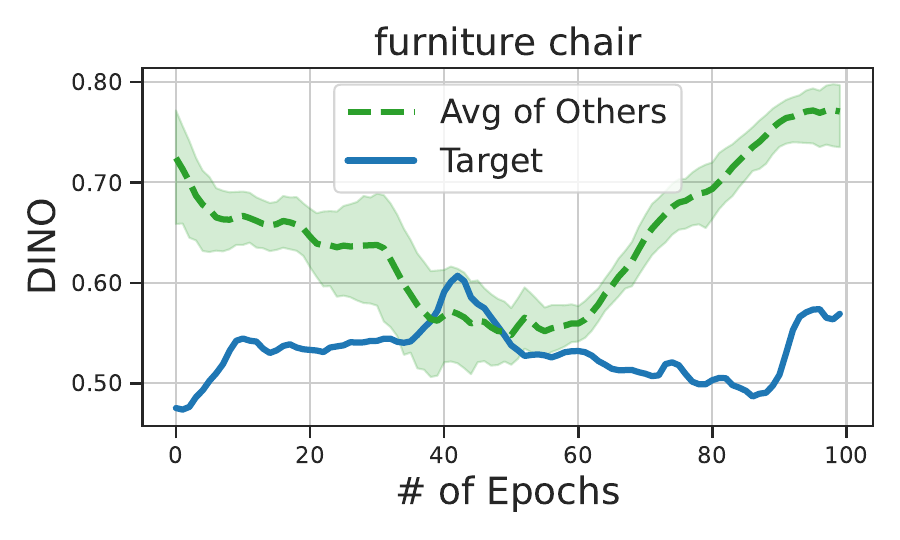}\\ 
      \includegraphics[height=1.7cm, trim={0.3cm 0.35cm 0.3cm 0.35cm},clip]{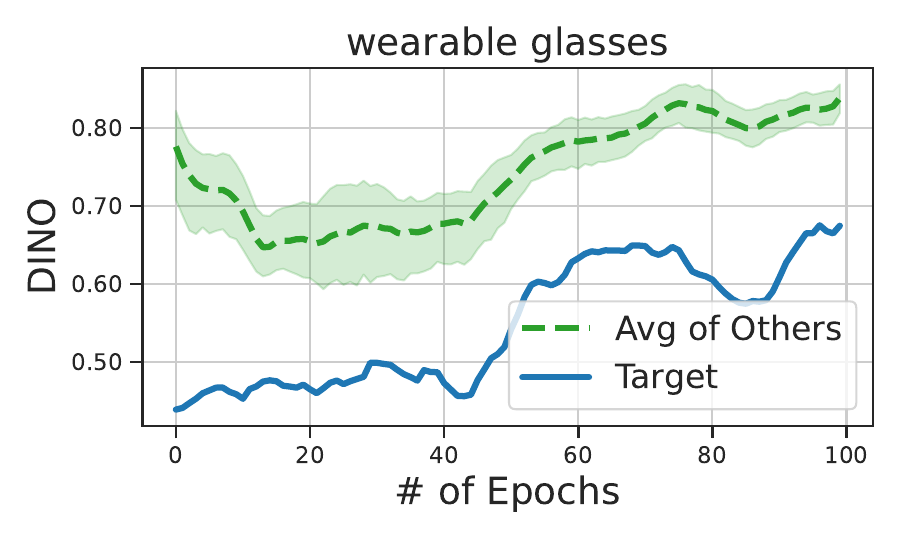} & \includegraphics[height=1.7cm, trim={0.3cm 0.35cm 0.3cm 0.35cm},clip]{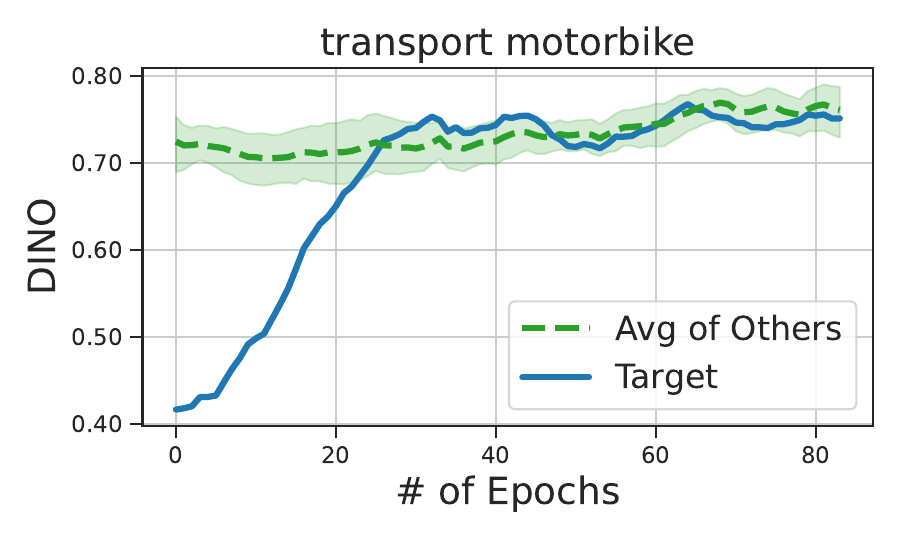} & \includegraphics[height=1.7cm, trim={0.3cm 0.35cm 0.3cm 0.35cm},clip]{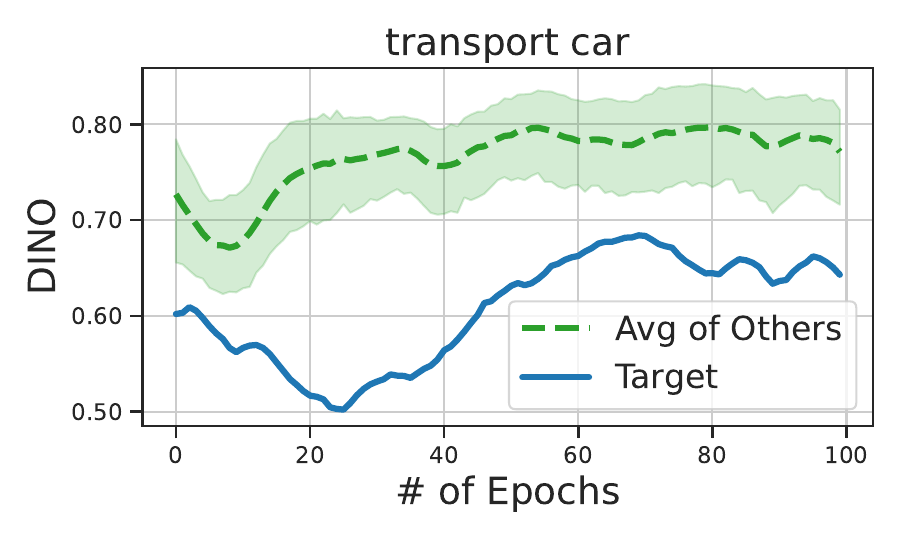} & \includegraphics[height=1.7cm, trim={0.3cm 0.35cm 0.3cm 0.35cm},clip]{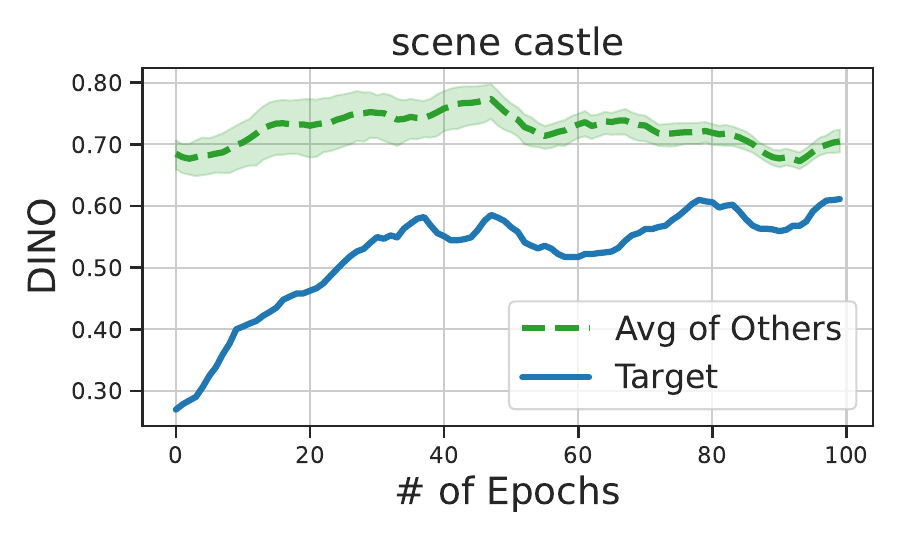} \\
      \includegraphics[height=1.7cm, trim={0.3cm 0.35cm 0.3cm 0.35cm},clip]{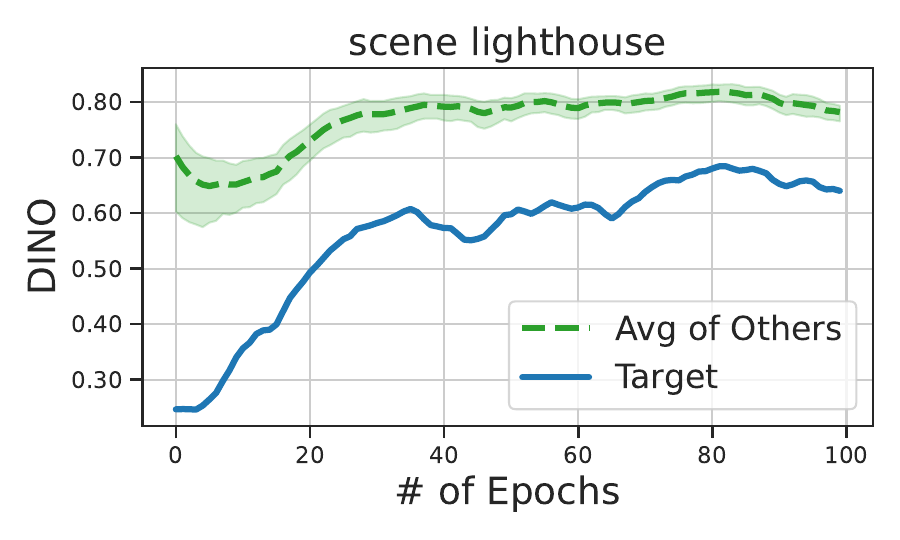}& \includegraphics[height=1.7cm, trim={0.3cm 0.35cm 0.3cm 0.35cm},clip]{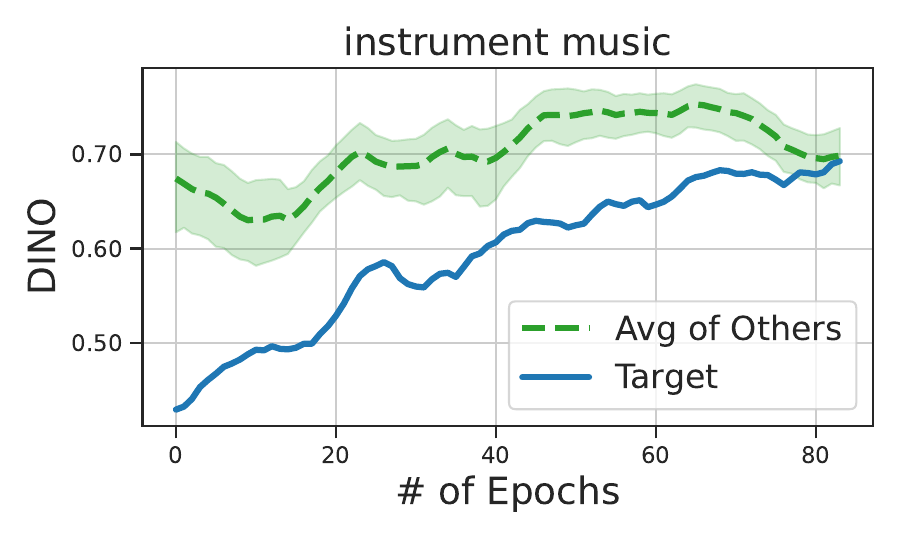} \\
    \end{tabular}
    \caption{\bf CLIP and DINO of Dreambooth Adaptation {\color{noimma}w/o} and {\color{imma} w/} IMMA.}
    \label{fig:db_quan}
\end{figure*}

\begin{figure*}[t]
    \centering
    \setlength{\tabcolsep}{1pt}
    \begin{tabular}{ccccc}
     \includegraphics[height=1.7cm, trim={0.3cm 0.35cm 0.3cm 0.35cm},clip]{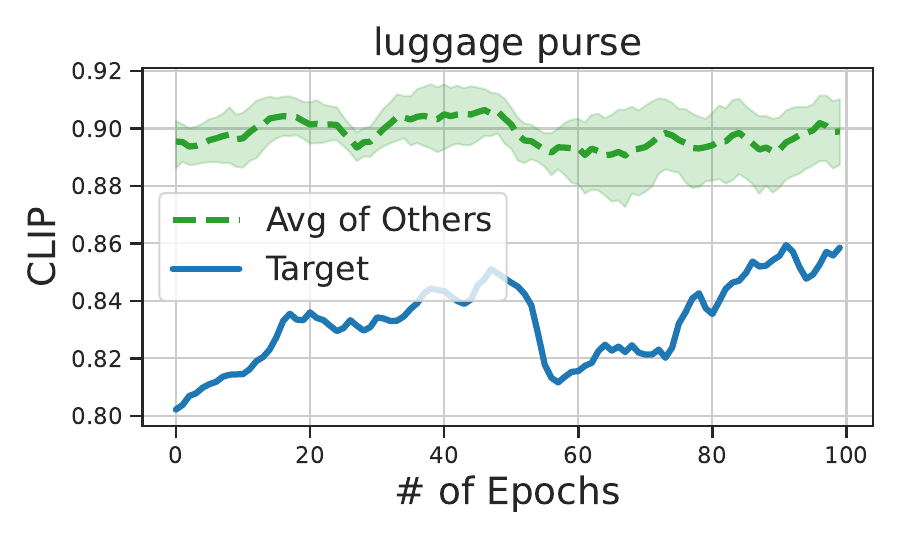} & \includegraphics[height=1.7cm, trim={0.3cm 0.35cm 0.3cm 0.35cm},clip]{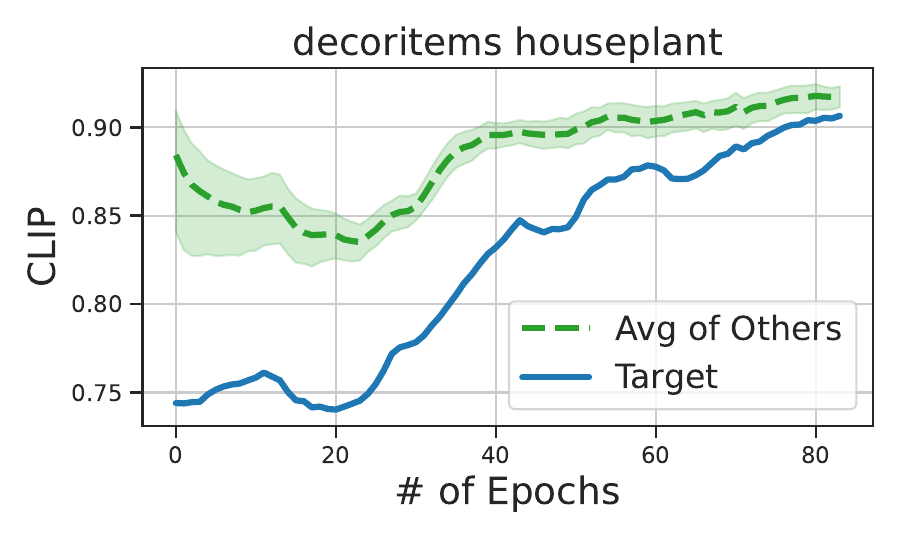} & \includegraphics[height=1.7cm, trim={0.3cm 0.35cm 0.3cm 0.35cm},clip]{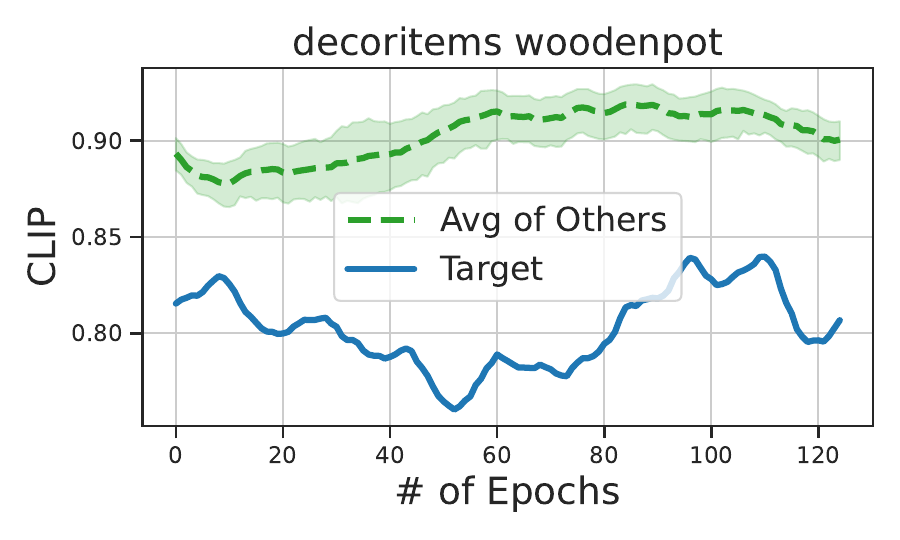} & \includegraphics[height=1.7cm, trim={0.3cm 0.35cm 0.3cm 0.35cm},clip]{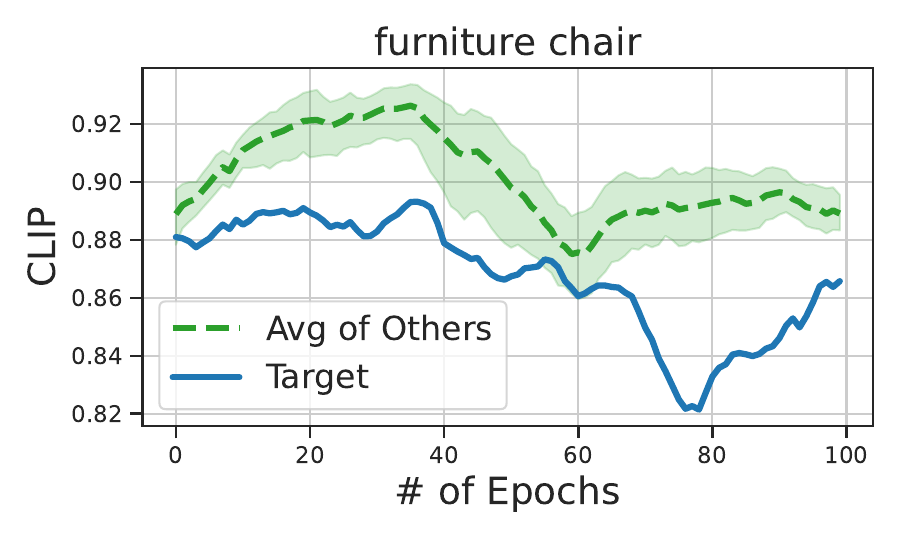} \\ 
     \includegraphics[height=1.7cm, trim={0.3cm 0.35cm 0.3cm 0.35cm},clip]{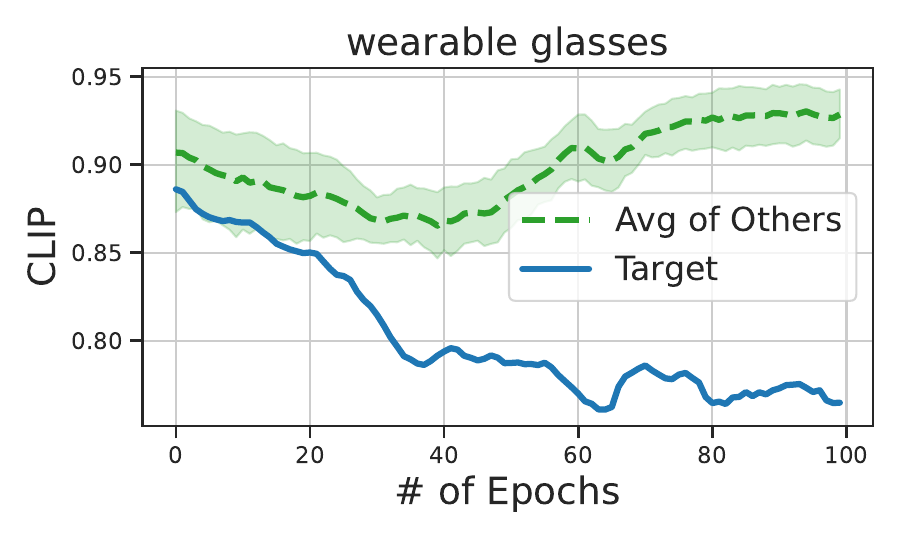} & \includegraphics[height=1.7cm, trim={0.3cm 0.35cm 0.3cm 0.35cm},clip]{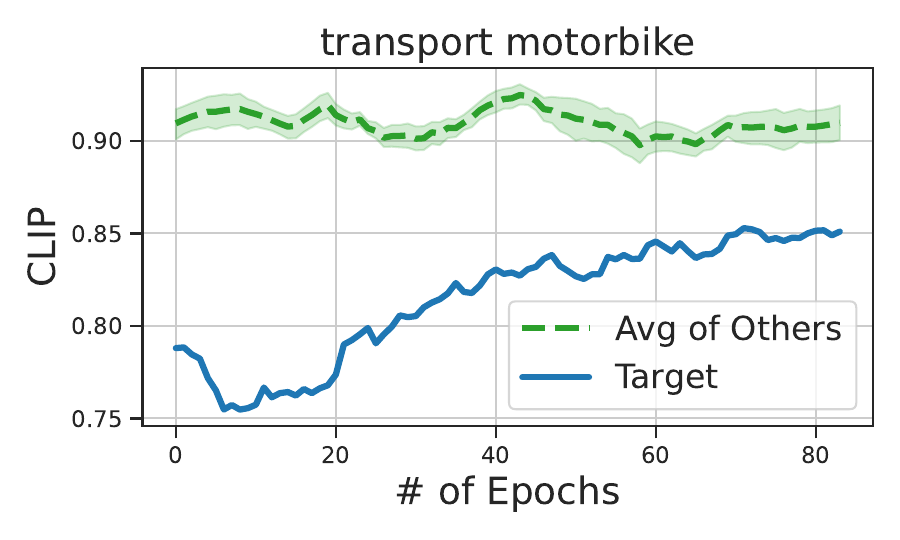} & \includegraphics[height=1.7cm, trim={0.3cm 0.35cm 0.3cm 0.35cm},clip]{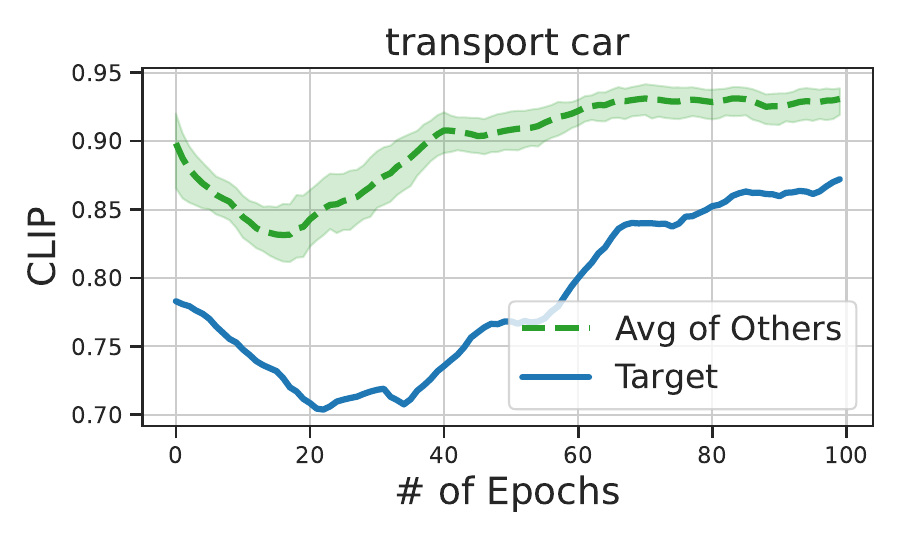} & \includegraphics[height=1.7cm, trim={0.3cm 0.35cm 0.3cm 0.35cm},clip]{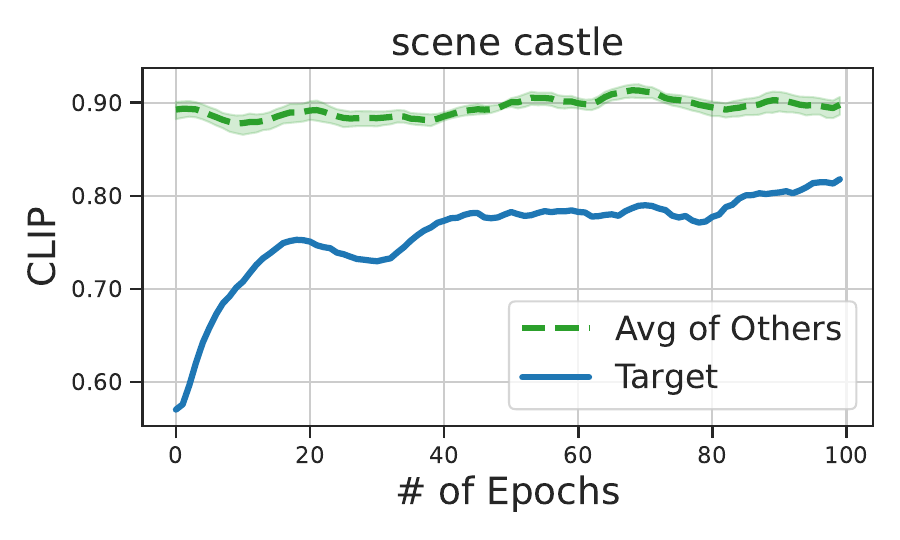} \\
     \includegraphics[height=1.7cm, trim={0.3cm 0.35cm 0.3cm 0.35cm},clip]{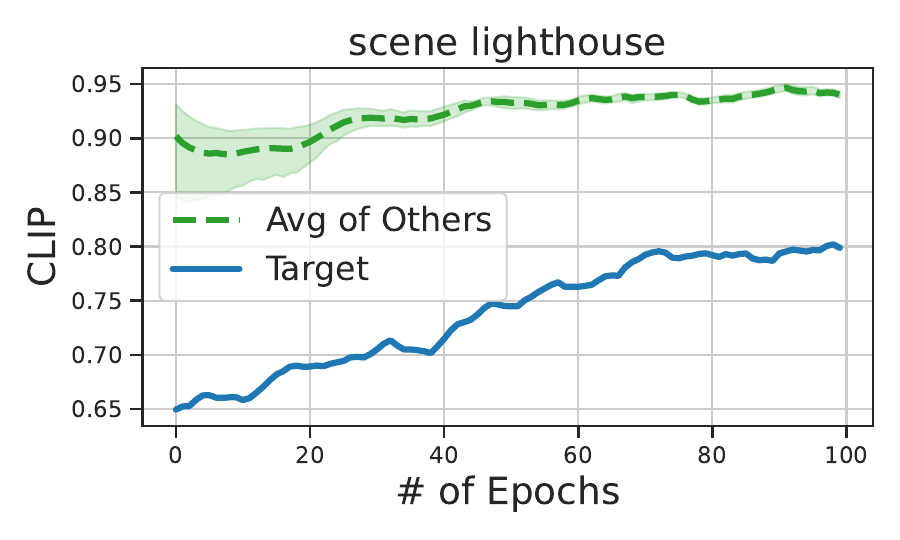}& \includegraphics[height=1.7cm, trim={0.3cm 0.35cm 0.3cm 0.35cm},clip]{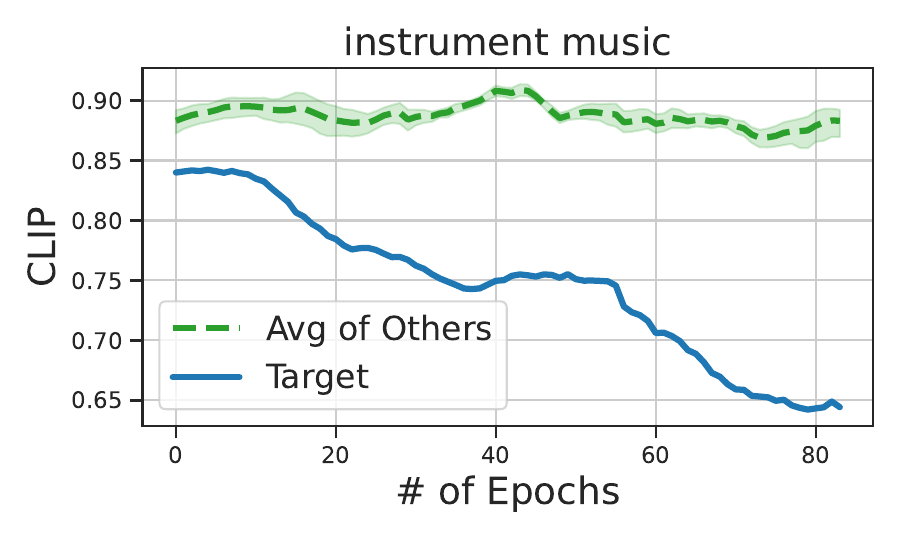} \\
     \includegraphics[height=1.7cm, trim={0.3cm 0.35cm 0.3cm 0.35cm},clip]{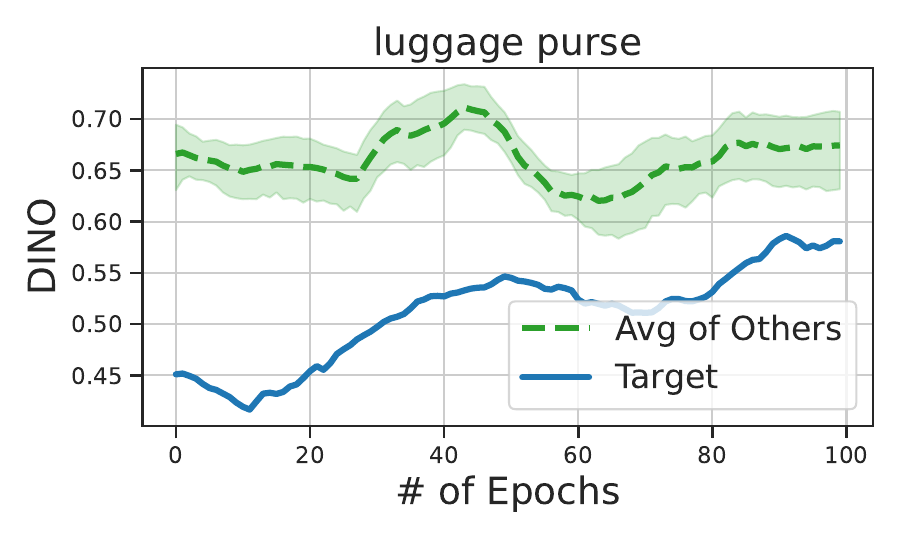} & \includegraphics[height=1.7cm, trim={0.3cm 0.35cm 0.3cm 0.35cm},clip]{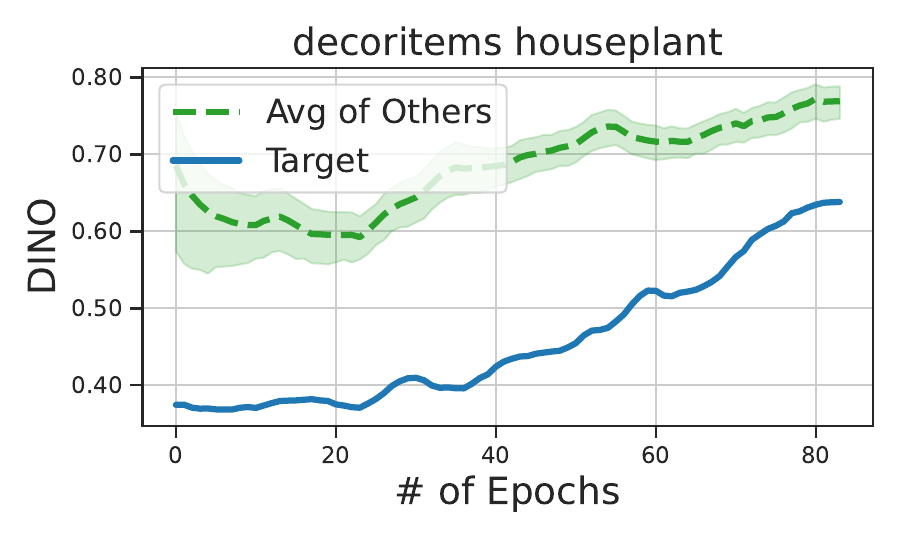} & \includegraphics[height=1.7cm, trim={0.3cm 0.35cm 0.3cm 0.35cm},clip]{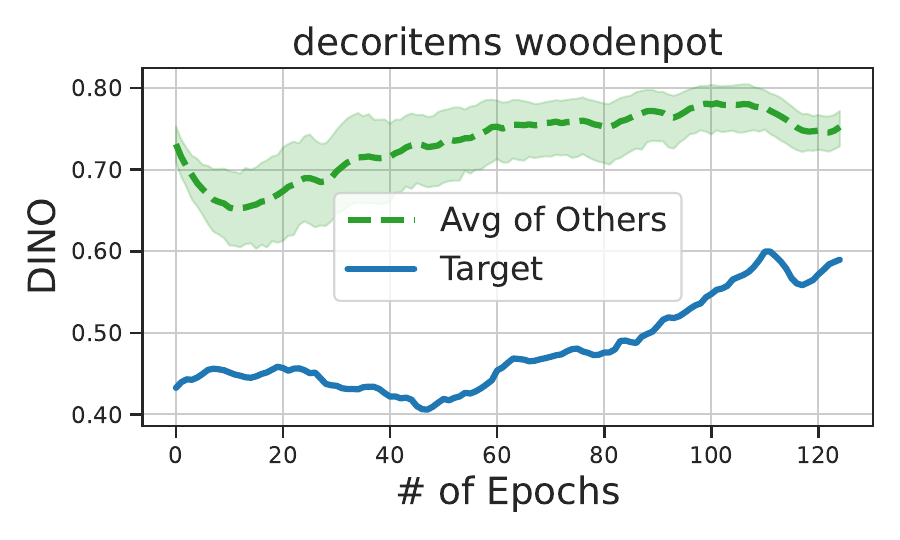} & \includegraphics[height=1.7cm, trim={0.3cm 0.35cm 0.3cm 0.35cm},clip]{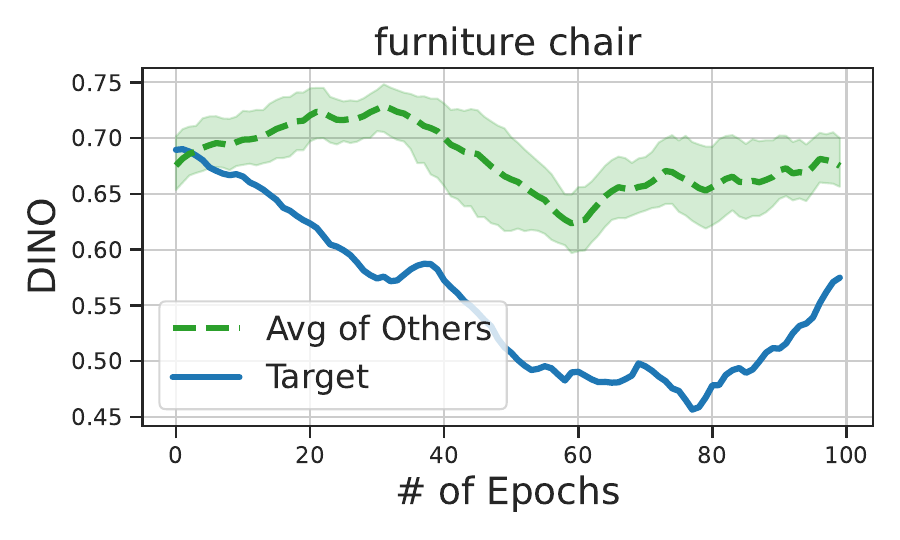}\\
     \includegraphics[height=1.7cm, trim={0.3cm 0.35cm 0.3cm 0.35cm},clip]{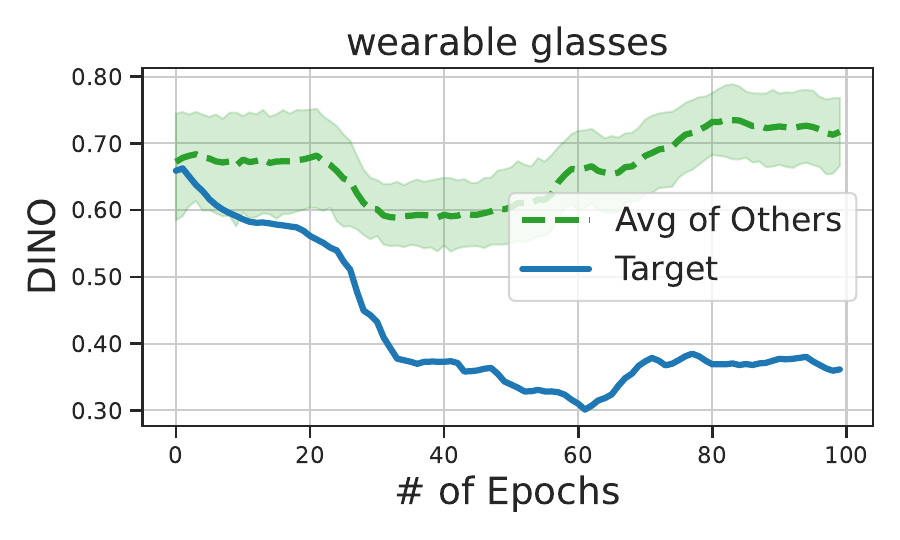} & \includegraphics[height=1.7cm, trim={0.3cm 0.35cm 0.3cm 0.35cm},clip]{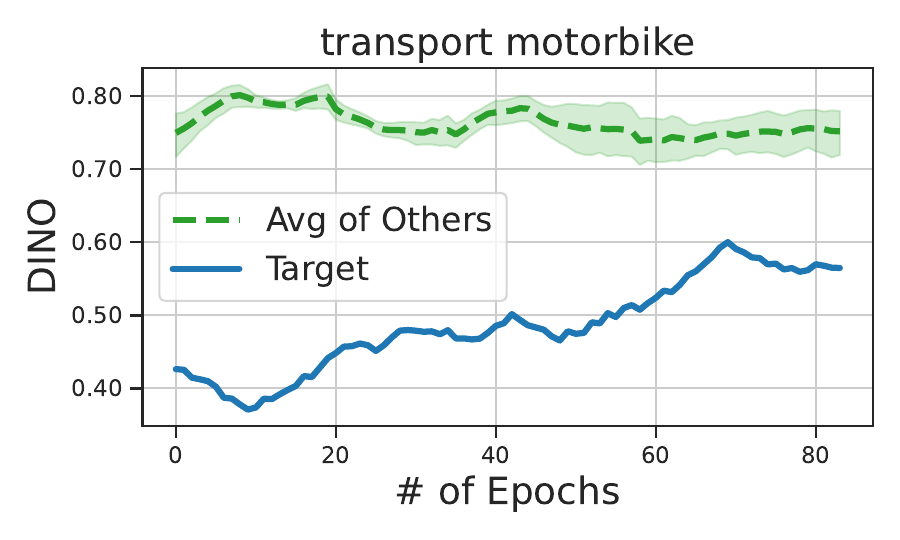} & \includegraphics[height=1.7cm, trim={0.3cm 0.35cm 0.3cm 0.35cm},clip]{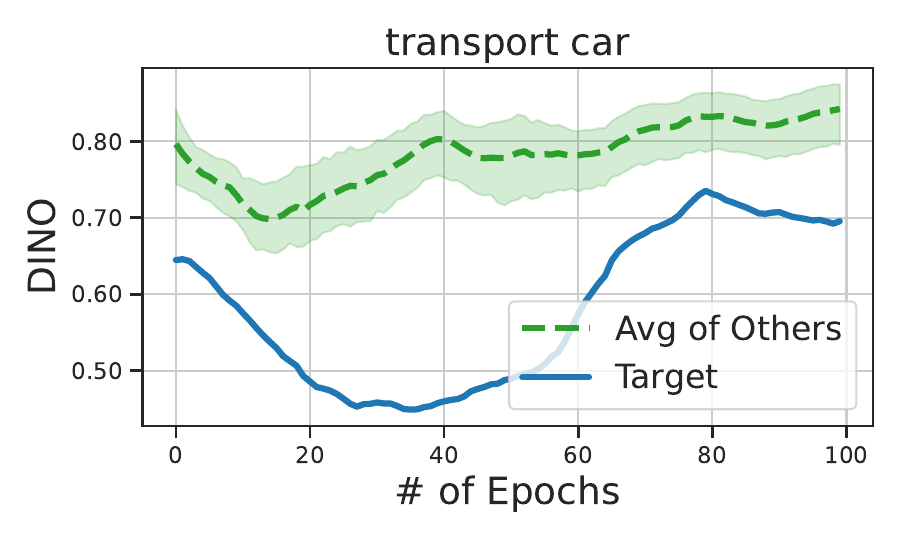} & \includegraphics[height=1.7cm, trim={0.3cm 0.35cm 0.3cm 0.35cm},clip]{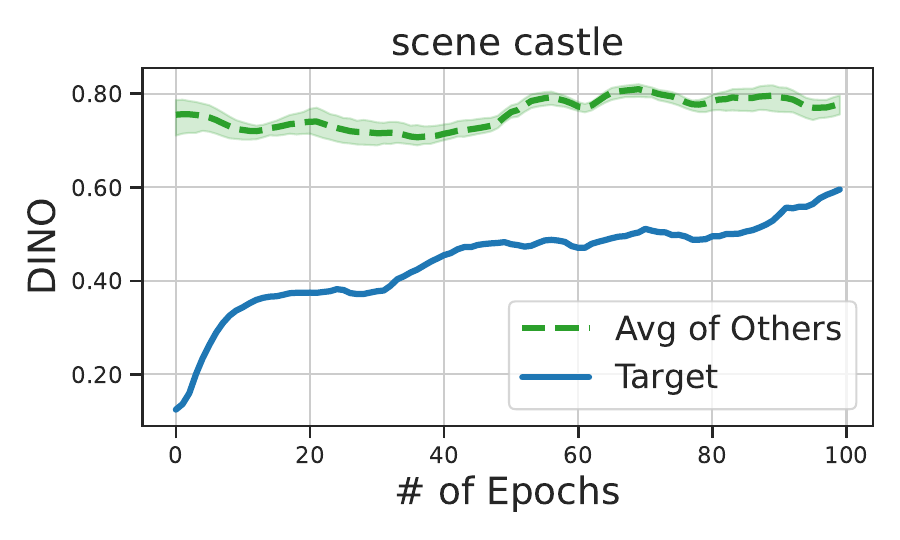} \\
     \includegraphics[height=1.7cm, trim={0.3cm 0.35cm 0.3cm 0.35cm},clip]{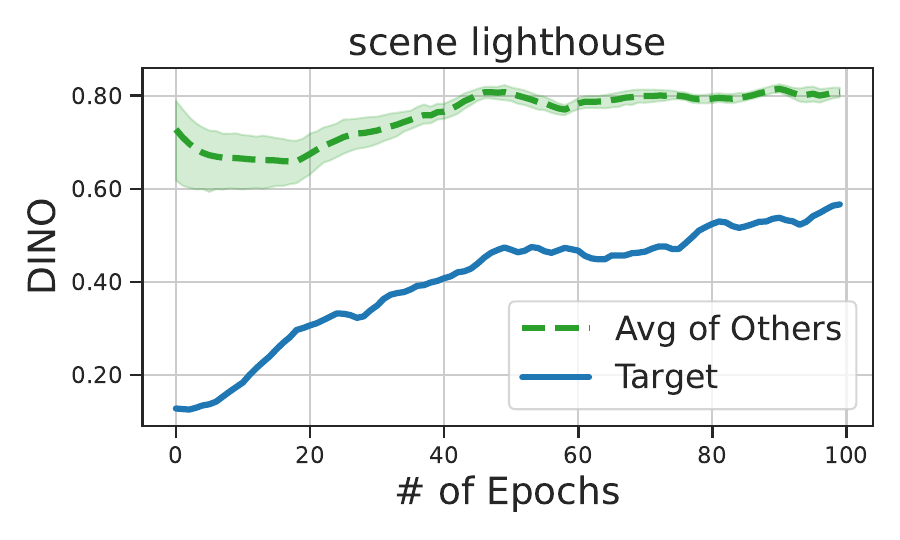}& \includegraphics[height=1.7cm, trim={0.3cm 0.35cm 0.3cm 0.35cm},clip]{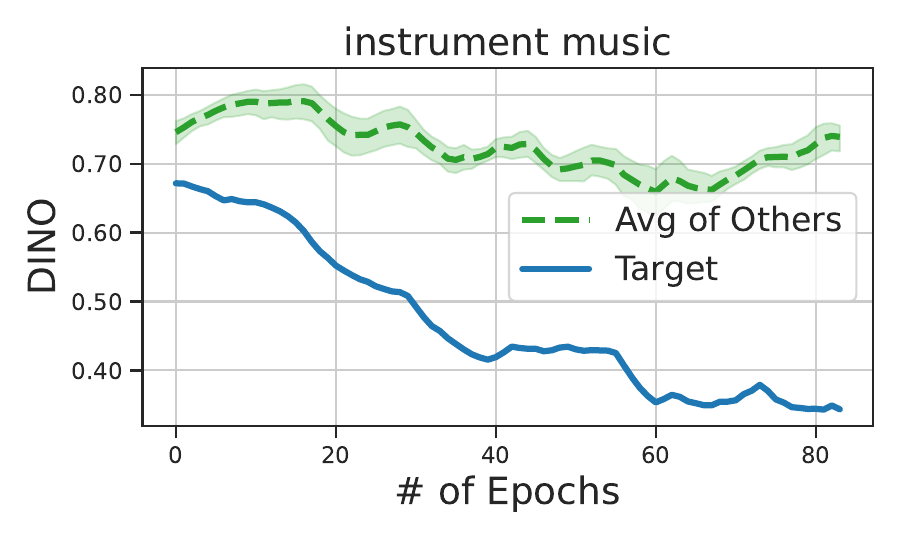} \\
    \end{tabular}
    \caption{\bf CLIP and DINO of Dreambooth LoRA Adaptation {\color{noimma}w/o} and {\color{imma} w/} IMMA.}
    \label{fig:dbl_quan}
\end{figure*}

\myparagraph{Results on immunizing against personalized content.} 
For personalization adaptation, we show more quantitative results on ten datasets from \citet{kumari2022customdiffusion}. The metric values of Textual Inversion, Dreambooth, and Dreambooth LoRA are shown in~\figref{fig:ti_ref},~\figref{fig:db_ref}, and~\figref{fig:dbl_ref}, respectively. We observe there is a consistent gap between the values with and without IMMA, which indicates that IMMA prevents the model from learning the personalized content effectively. We also show the results of target and other concepts in~\figref{fig:ti_quan},~\figref{fig:db_quan}, and~\figref{fig:dbl_quan}. The gap between the two lines shows IMMA immunized the pre-trained model from the target concept while maintaining the ability to be fine-tuned and generate images of other concepts.

\subsection{IMMA for multiple concepts}
\begin{figure*}[t]
    \centering
    \setlength{\tabcolsep}{1pt}
    \begin{tabular}{ccccc}
    Castle & Chair & Guitar \\
     \includegraphics[height=1.6cm, trim={0.3cm 0.35cm 0.3cm 0.35cm},clip]{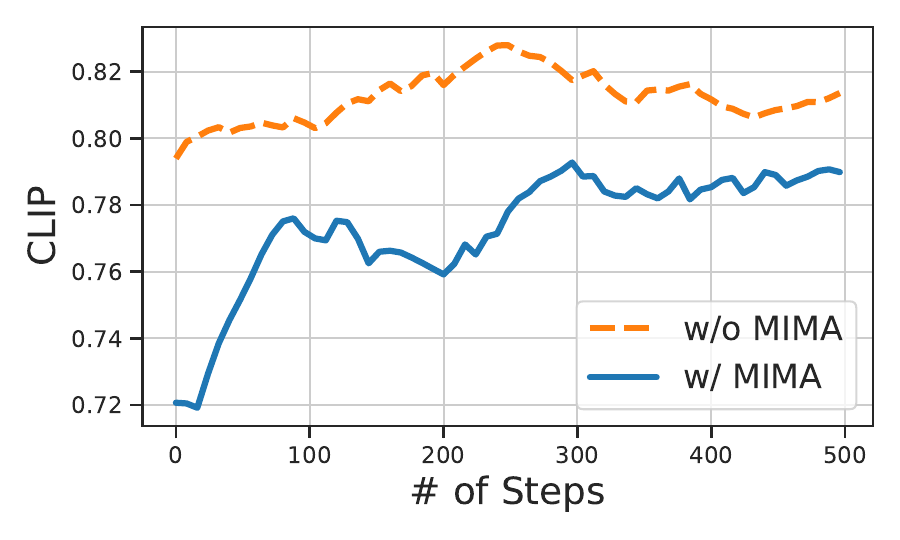} & 
     \includegraphics[height=1.6cm, trim={0.3cm 0.35cm 0.3cm 0.35cm},clip]{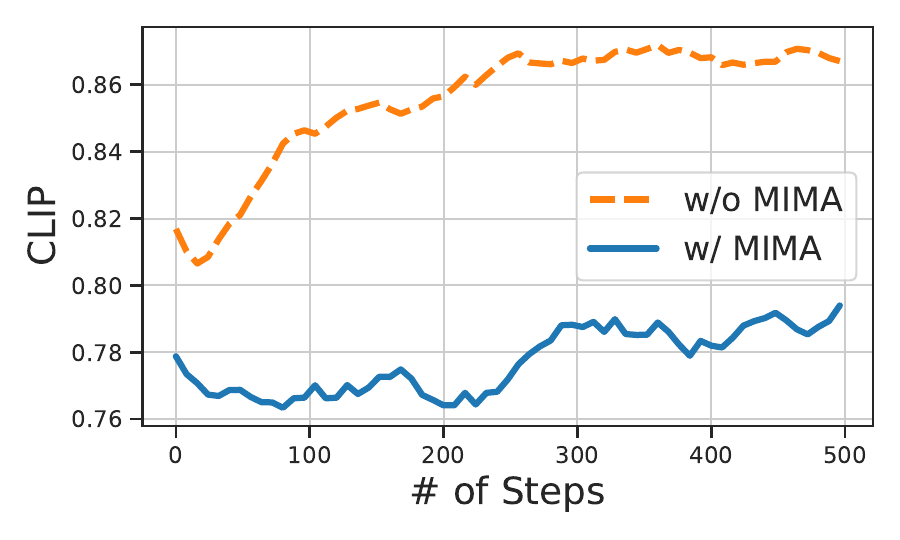} & 
     \includegraphics[height=1.6cm, trim={0.3cm 0.35cm 0.3cm 0.35cm},clip]{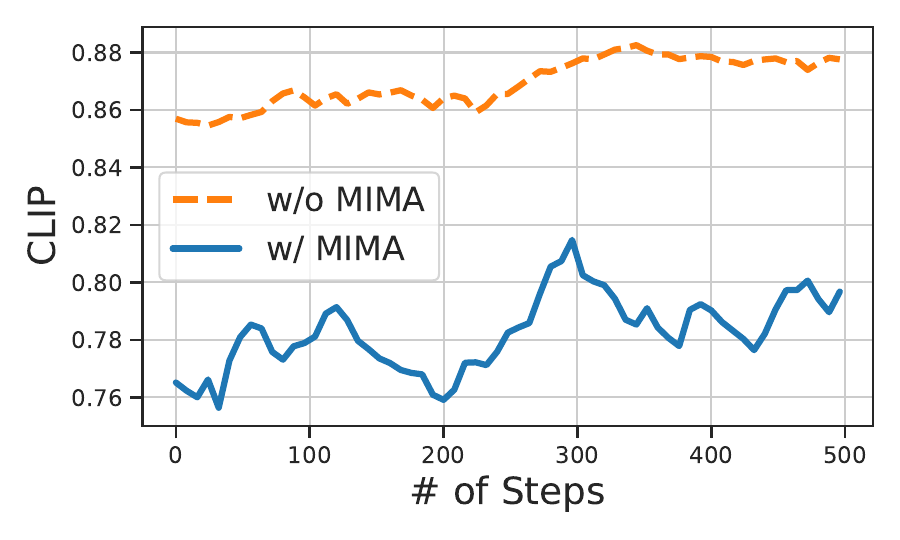}  \\
     \includegraphics[height=1.6cm, trim={0.3cm 0.35cm 0.3cm 0.35cm},clip]{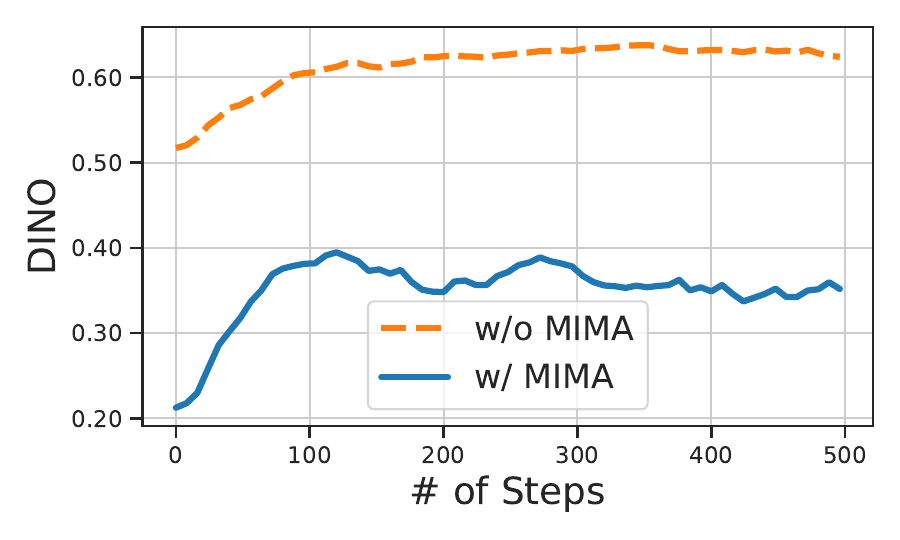} & 
     \includegraphics[height=1.6cm, trim={0.3cm 0.35cm 0.3cm 0.35cm},clip]{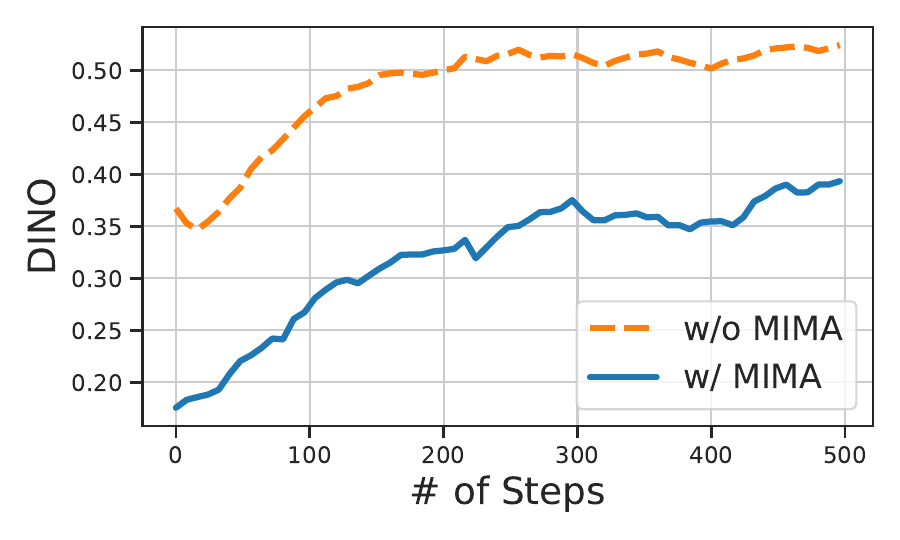} & 
     \includegraphics[height=1.6cm, trim={0.3cm 0.35cm 0.3cm 0.35cm},clip]{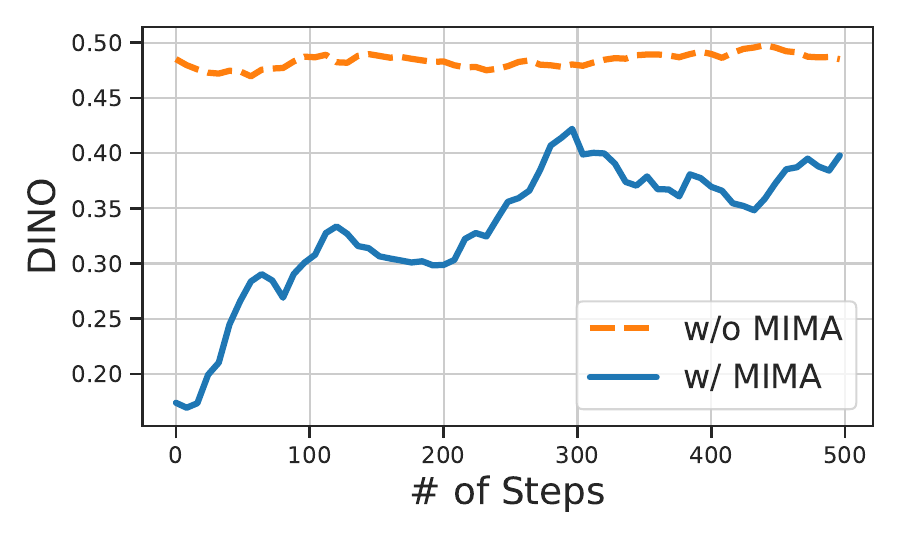}  \\
    \end{tabular}
    \caption{\bf CLIP and DINO of Textual Inversion {\color{noimma}w/o} and {\color{imma} w/} IMMA on multiple concepts.}
    \label{fig:multi_quan_sgr}
\end{figure*}

\begin{figure*}[t]
    \centering
    \setlength{\tabcolsep}{1pt}
    \begin{tabular}{ccccc}
     \includegraphics[height=1.6cm, trim={0.3cm 0.35cm 0.3cm 0.35cm},clip]{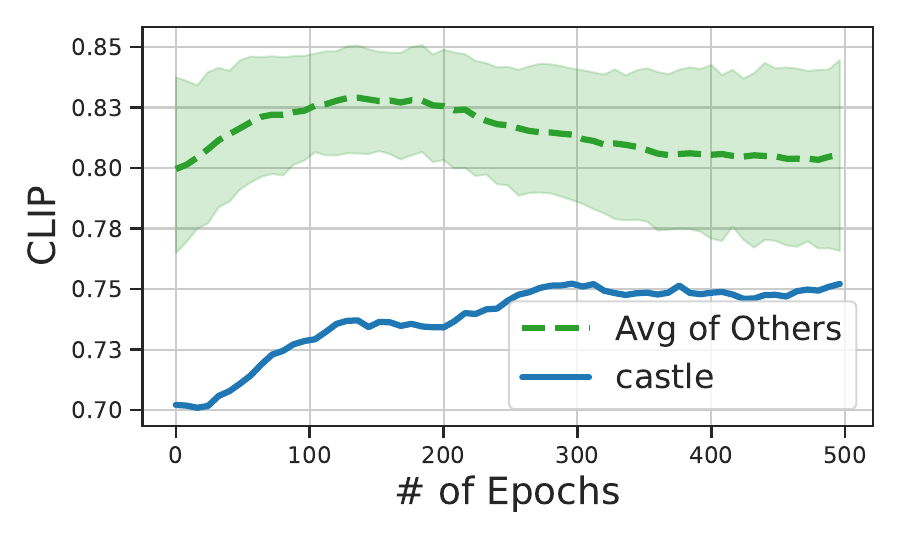} & 
     \includegraphics[height=1.6cm, trim={0.3cm 0.35cm 0.3cm 0.35cm},clip]{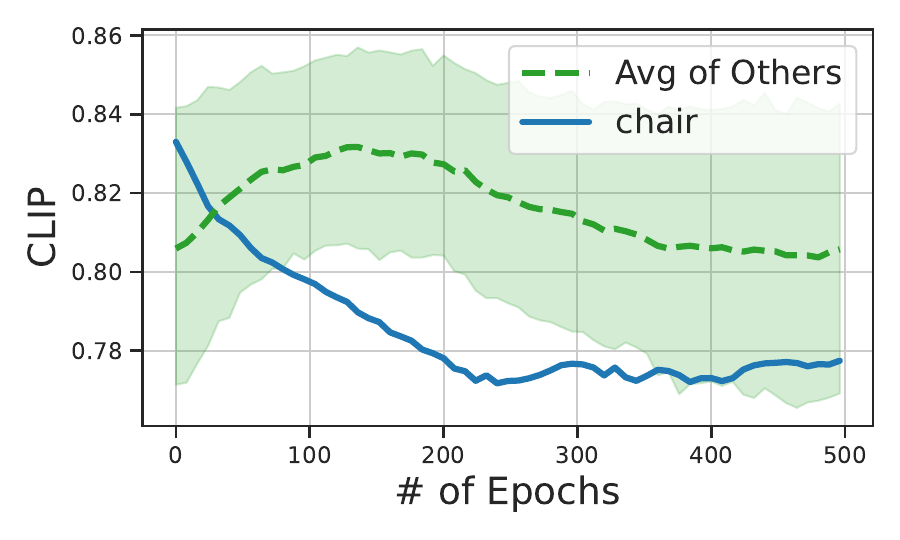} & 
     \includegraphics[height=1.6cm, trim={0.3cm 0.35cm 0.3cm 0.35cm},clip]{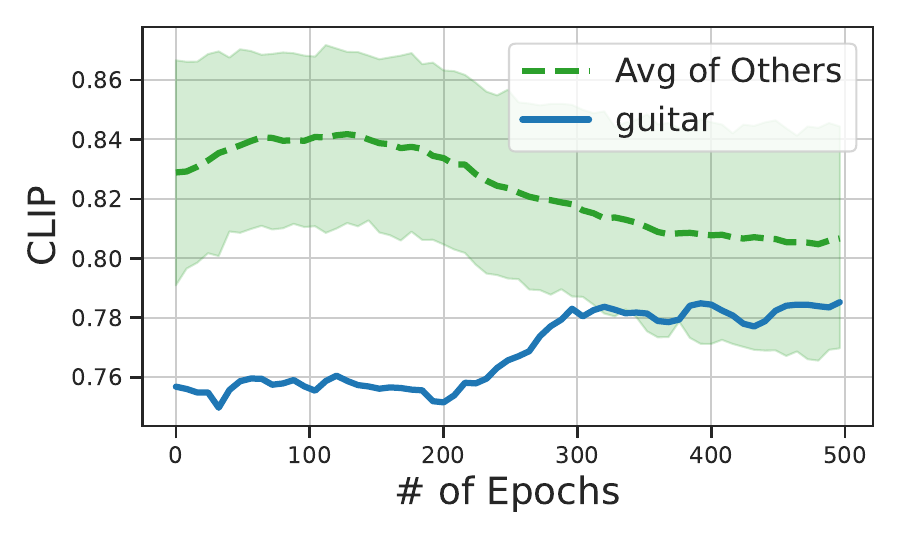}  \\
     \includegraphics[height=1.6cm, trim={0.3cm 0.35cm 0.3cm 0.35cm},clip]{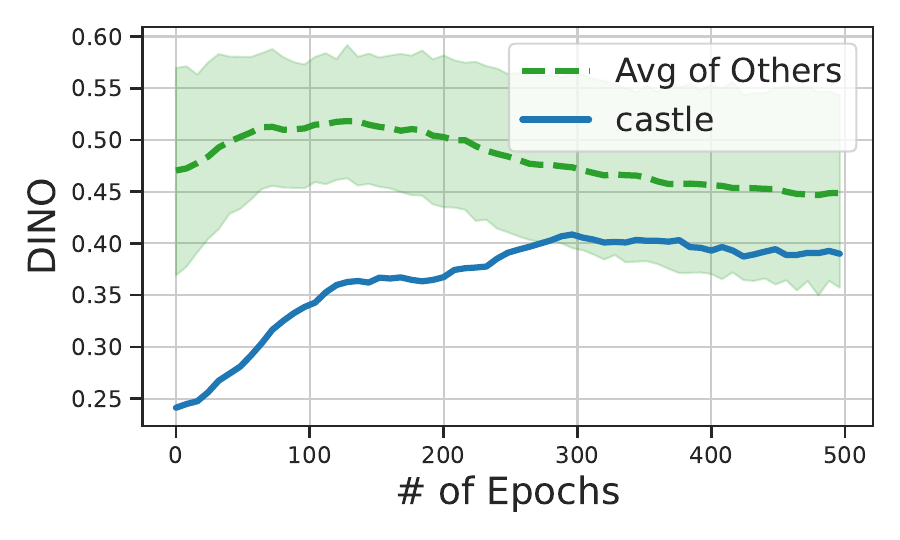} & 
     \includegraphics[height=1.6cm, trim={0.3cm 0.35cm 0.3cm 0.35cm},clip]{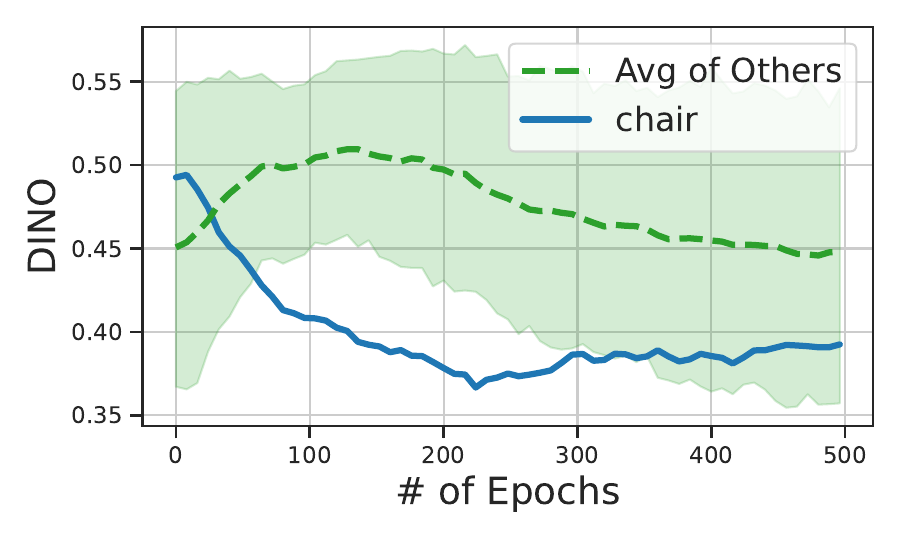} & 
     \includegraphics[height=1.6cm, trim={0.3cm 0.35cm 0.3cm 0.35cm},clip]{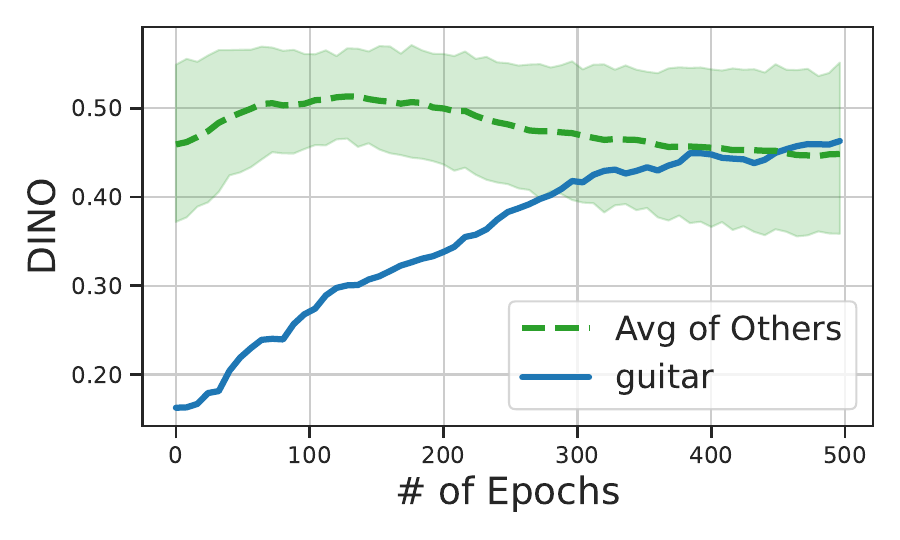}  \\
    \end{tabular}
    \caption{{\bf CLIP and DINO similarity on other concept \vs target concept. } The adaptation method is Textual Inversion. The gap between the two lines shows \rsgr.}
    \label{fig:multi_quan_rsgr}
\end{figure*}

We conducted experiments of immunization on multiple concepts by running IMMA on each target concept sequentially. As shown in~\figref{fig:multi_quan_sgr} and~\figref{fig:multi_quan_rsgr}, after running IMMA three times on three target concepts: castle, chair and guitar, the immunized model can protect the model from using Textual Inversion to learn any of the concepts. Thus, IMMA has the potential to be extended to multiple-concept scenarios. However, the similarities of other concepts may drop more than the single-concept case, which is worth further research.

\section{Additional qualitative results}
\label{sec:supp_qual}

\subsection{Comparison with MIST}
\begin{figure}[h]
\setlength{\tabcolsep}{2pt}
\centering
\footnotesize
\begin{tabular}{cc@{\hskip 4pt}|cc|c}
Reference & MISTed Reference & {TI} & {TI (JPEG)} & DB \\
\includegraphics[width=0.16\linewidth]{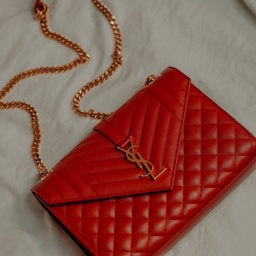} & 
\includegraphics[width=0.16\linewidth]{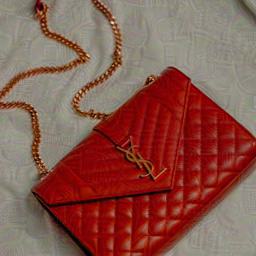} & 
\includegraphics[width=0.16\linewidth]{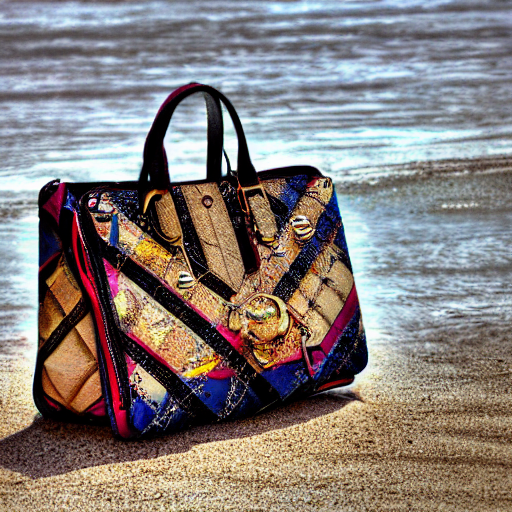} &
\includegraphics[width=0.16\linewidth]{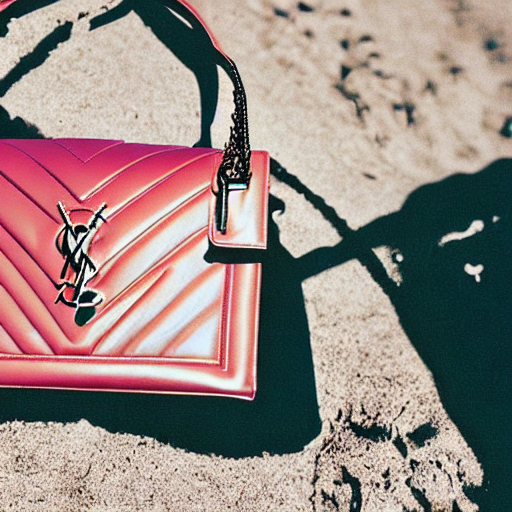} &
\includegraphics[width=0.16\linewidth]{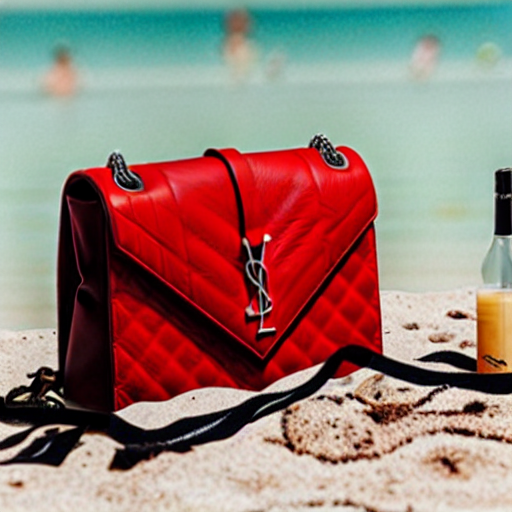}\\
\includegraphics[width=0.16\linewidth]{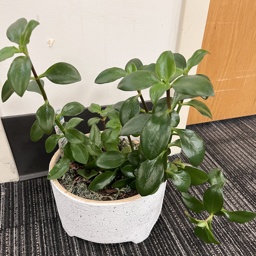} & 
\includegraphics[width=0.16\linewidth]{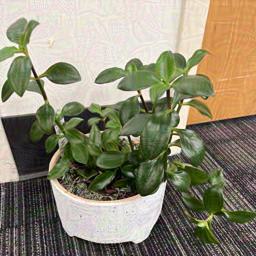} & 
\includegraphics[width=0.16\linewidth]{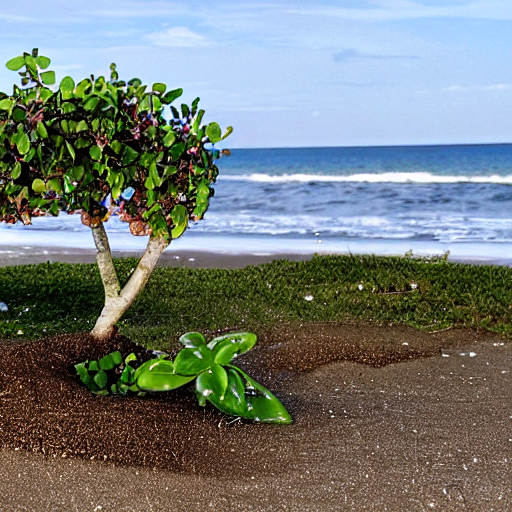} &
\includegraphics[width=0.16\linewidth]{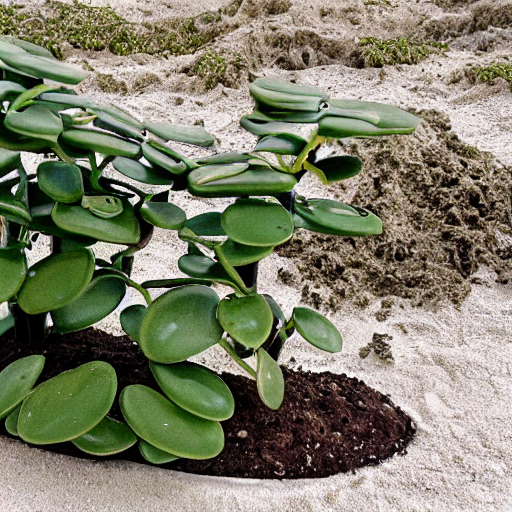} &
\includegraphics[width=0.16\linewidth]{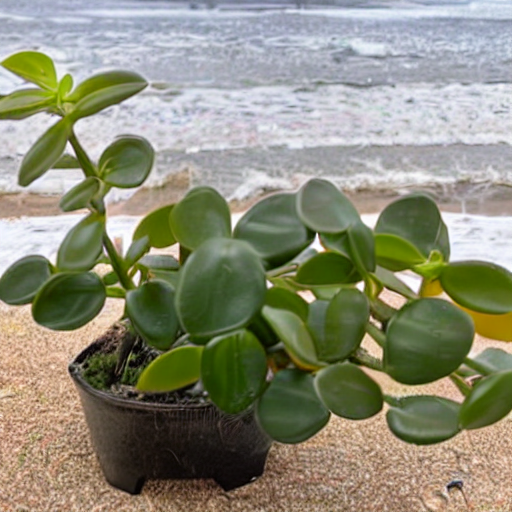}\\
\includegraphics[width=0.16\linewidth]{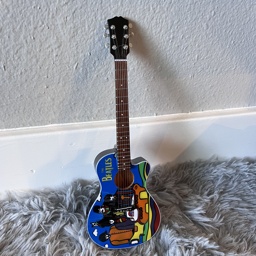} & 
\includegraphics[width=0.16\linewidth]{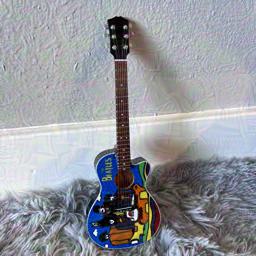} & 
\includegraphics[width=0.16\linewidth]{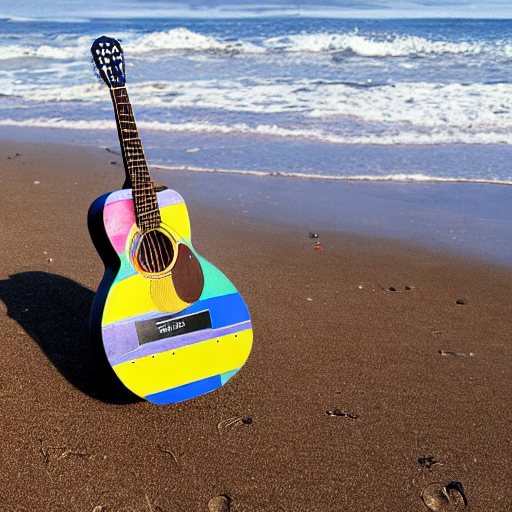} &
\includegraphics[width=0.16\linewidth]{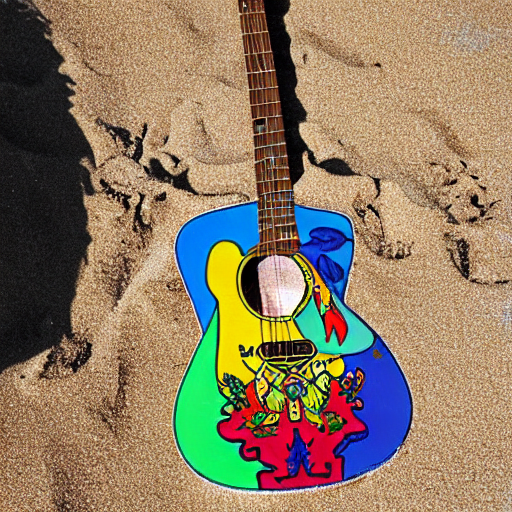}&
\includegraphics[width=0.16\linewidth]{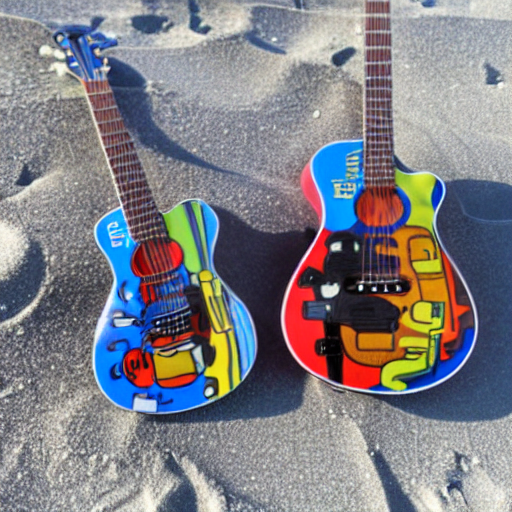}\\
\end{tabular}
\caption{\textbf{Additional results on MIST.} {We observe that MIST successfully prevented personalization against Textual Inversion. However, MIST is unsuccessful when JPEG is applied to the image, as reported in their paper, or when DreamBooth is used for adaptation.}}
\label{fig:mist_supp}
\end{figure}

In~\figref{fig:mist_supp}, we show additional results of MIST~\cite{liang2023adversarial}. Textual Inversion using images noised by MIST fails to learn the concept ($3^{\tt rd}$ column). However, after compressing the MISTed images with JPEG, such protection disappears ($4^{\tt th}$ column). Finally, we show that MIST fails to protect personalization items against the adaptation of DreamBooth ($5^{\tt th}$ column). We followed the default parameters of MIST using the strength of the adversarial attack being 16 and the iterations of the attack being 100.

\subsection{IMMA on preventing cloning of face images}
\begin{figure}[t]
    \centering
    \setlength{\tabcolsep}{1.5pt}
    \renewcommand{\arraystretch}{1.2}
    \begin{tabular}{c@{\hskip 8pt}cc}
    Reference & \color{noimma} w/o IMMA &  \color{imma} w/ IMMA\\
    \includegraphics[height=2.5cm]{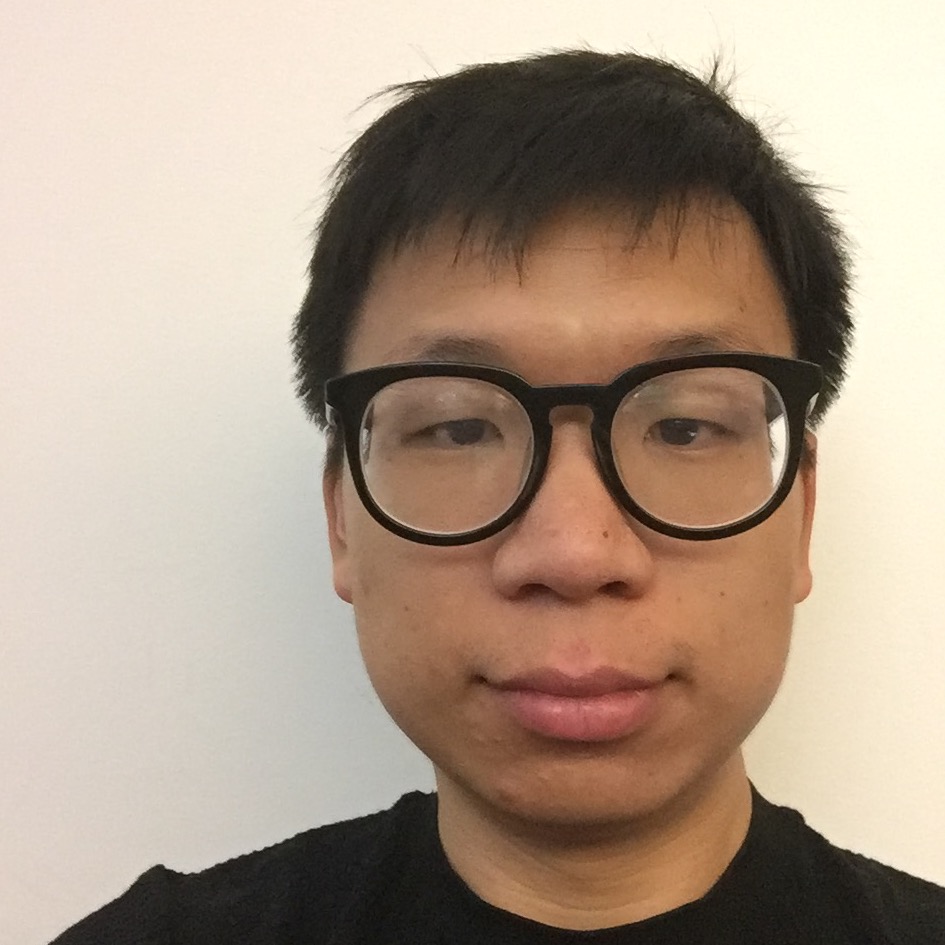} & 
    \includegraphics[height=2.5cm]{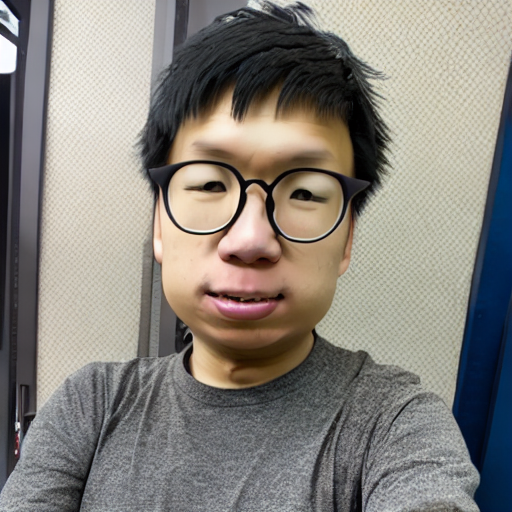} & 
    \includegraphics[height=2.5cm]{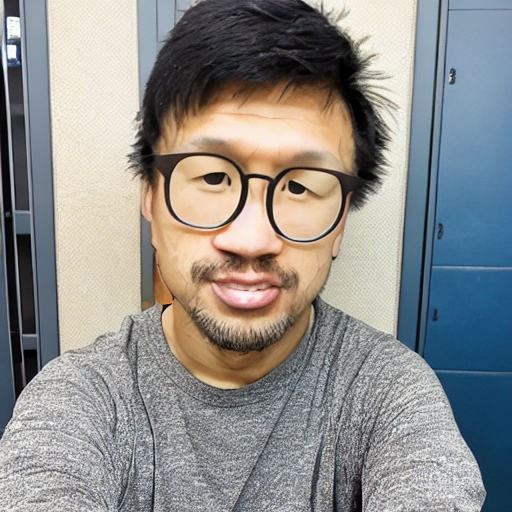}  \\
    \includegraphics[height=2.5cm]{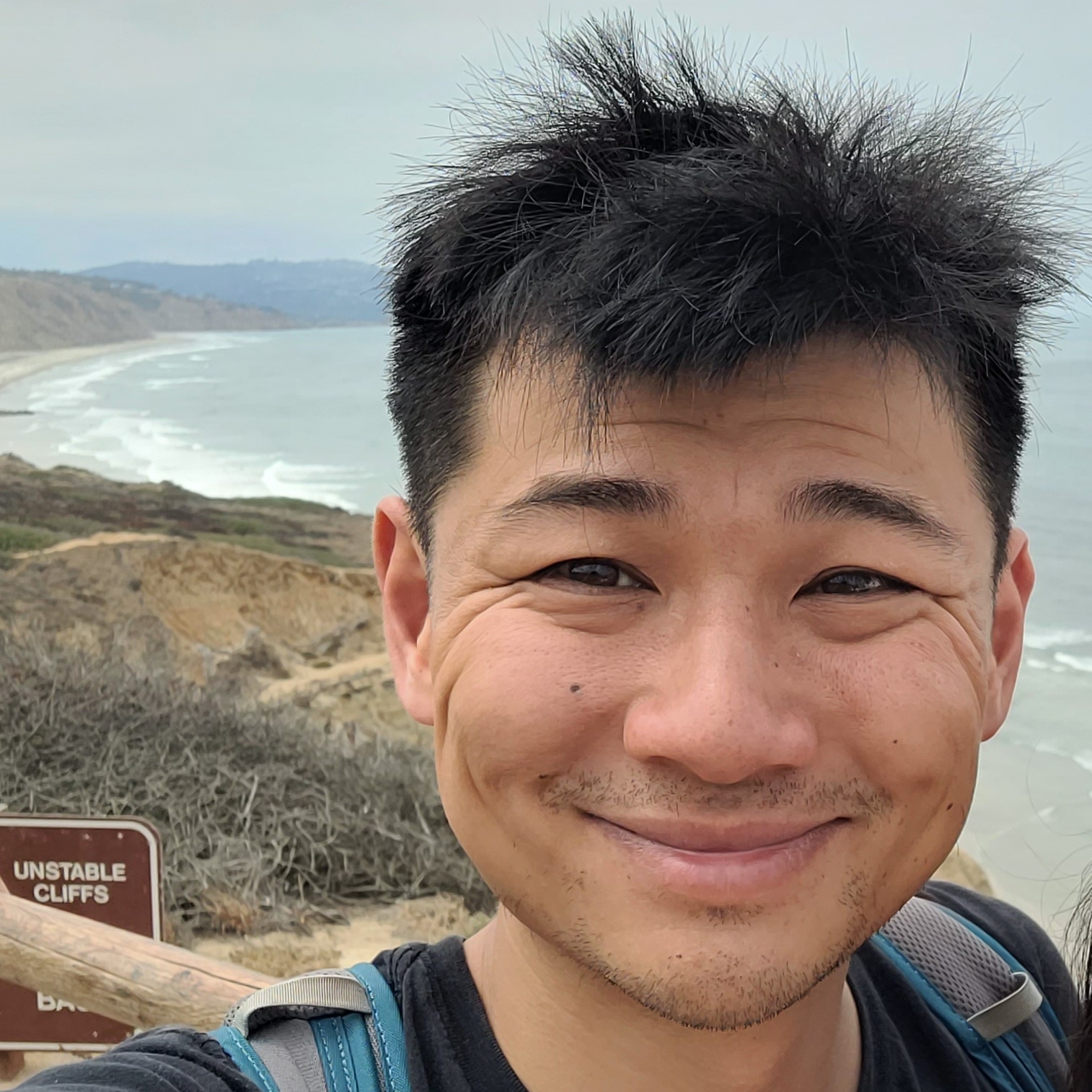} & 
    \includegraphics[height=2.5cm]{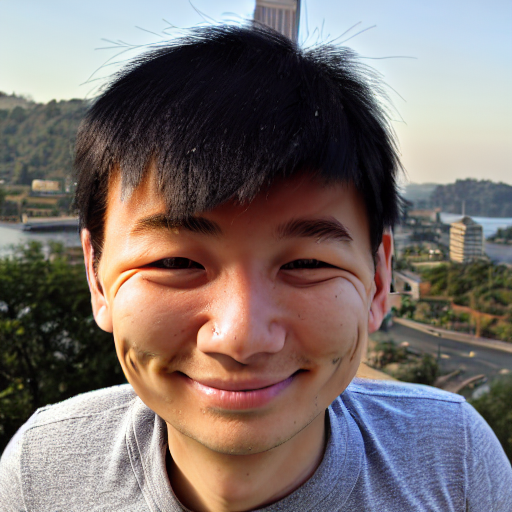} & 
    \includegraphics[height=2.5cm]{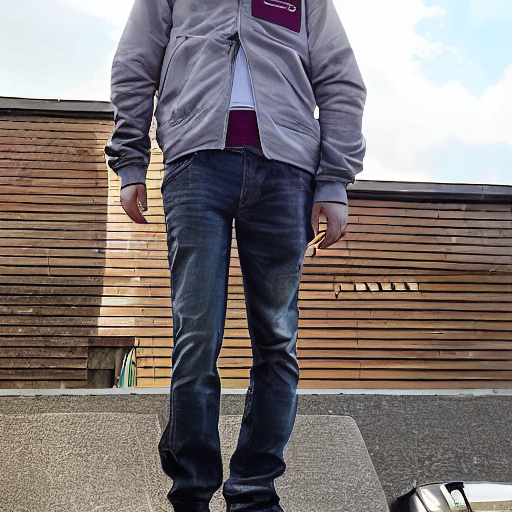}  \\
    \includegraphics[height=2.5cm]{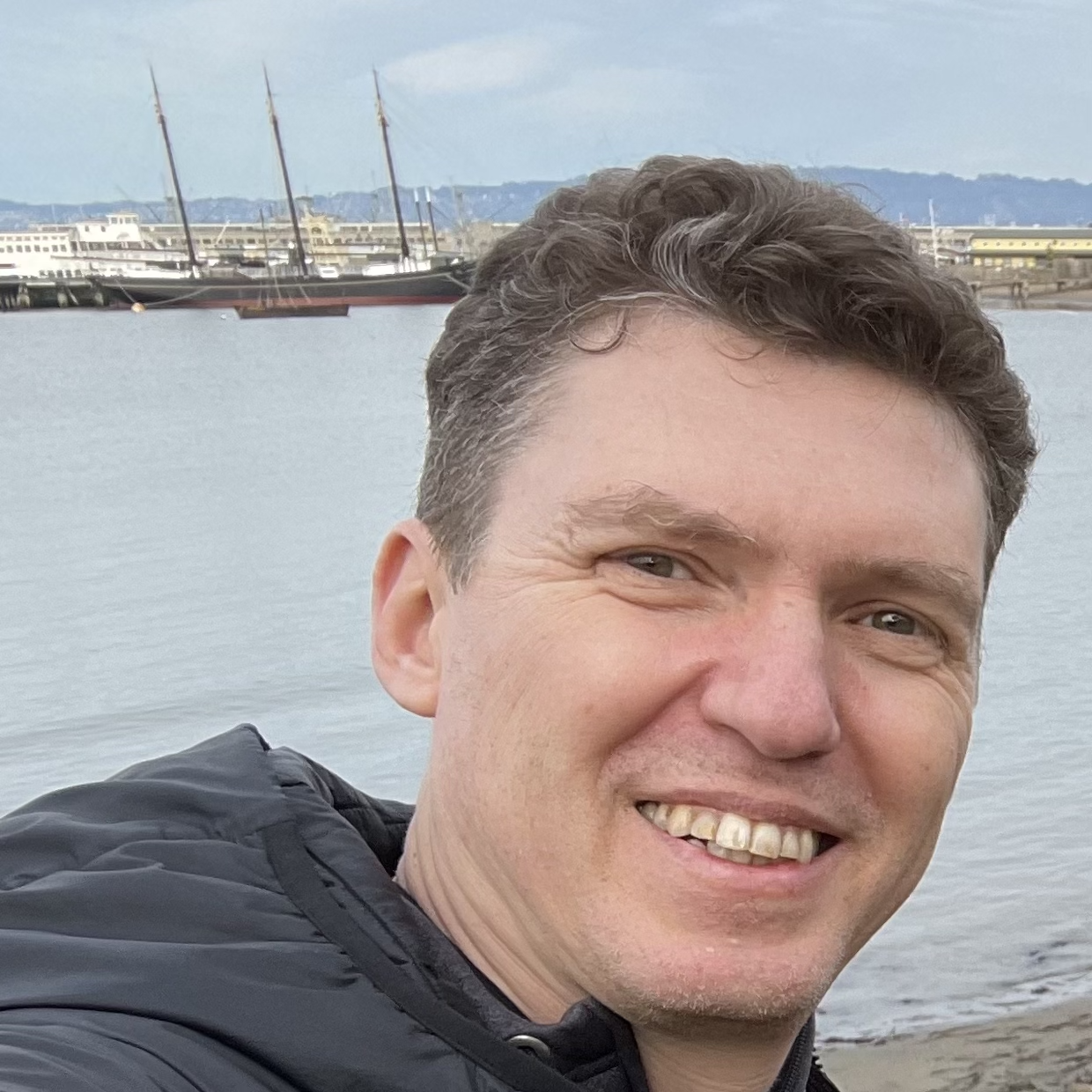} & 
    \includegraphics[height=2.5cm]{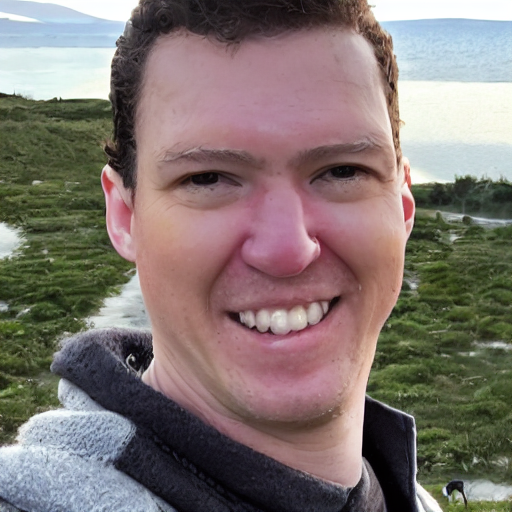} & 
    \includegraphics[height=2.5cm]{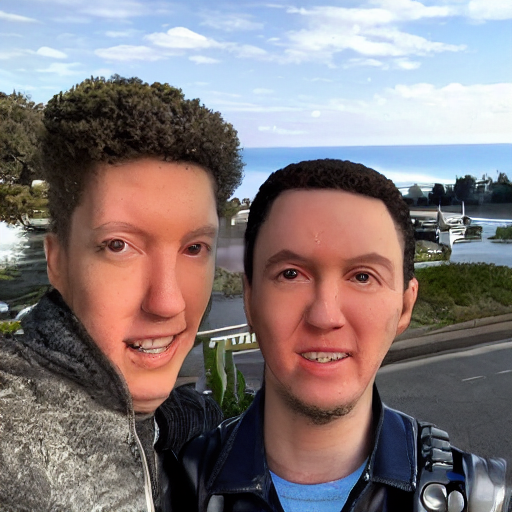}  \\
    \end{tabular}
    \vspace{-0.3cm}
    \caption{\textbf{IMMA on celebrities with adaptation of DreamBooth LoRA.}}
    \vspace{-0.3cm}
    \label{fig:celebrity}
\end{figure}

We now show that IMMA can effectively restrict the model from duplicating face images of a particular person. We conduct experiments using the datasets from~\citet {kumari2022customdiffusion} which contain face images. We use \textit{``a photo of [V] person''} as the prompt. From~\figref{fig:celebrity}, we observe that after implementing IMMA on the target person, the model loses its capacity to generate images of that identity using Dreambooth LoRA.

\subsection{IMMA on datasets from Dreambooth~\cite{ruiz2022dreambooth} and Textual Inversion~\cite{gal2022textual} directly reported in their paper}

\begin{figure}[t]
    \centering
    \setlength{\tabcolsep}{1.5pt}
    \renewcommand{\arraystretch}{1.4}
    \begin{tabular}{c@{\hskip 10pt}cc@{\hskip 8pt}cc}
      Reference & \multicolumn{2}{c}{\color{noimma} w/o IMMA} & \multicolumn{2}{c}{\color{imma} w/ IMMA} 
     \\
    \includegraphics[height=2.2cm]{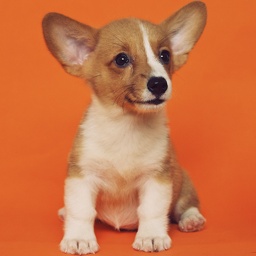}
    & \includegraphics[height=2.2cm]{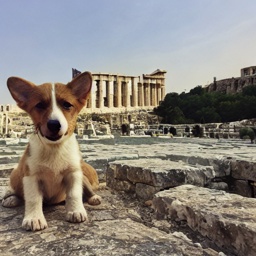} & \includegraphics[height=2.2cm]{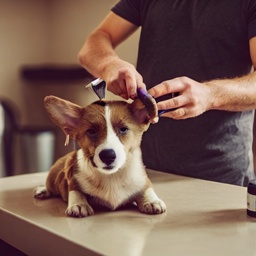} 
    &\includegraphics[height=2.2cm]{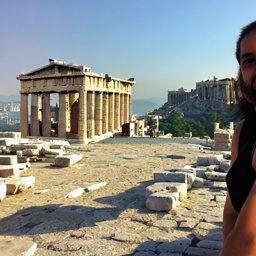} & \includegraphics[height=2.2cm]{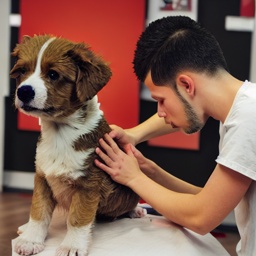} \\
    \includegraphics[height=2.2cm]{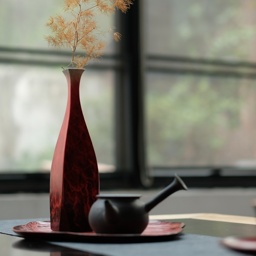}
    & \includegraphics[height=2.2cm]{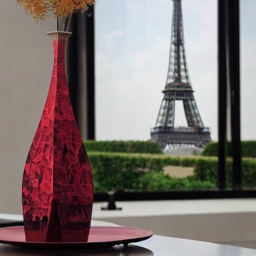} & \includegraphics[height=2.2cm]{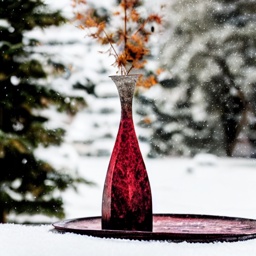} 
    &\includegraphics[height=2.2cm]{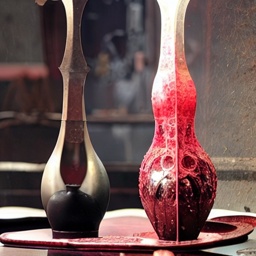} & \includegraphics[height=2.2cm]{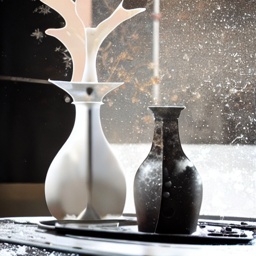} \\ 
    \hdashline
    \vspace{-0.45cm}
    \\
    \includegraphics[height=2.2cm]{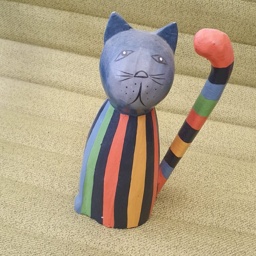}
    & \includegraphics[height=2.2cm]{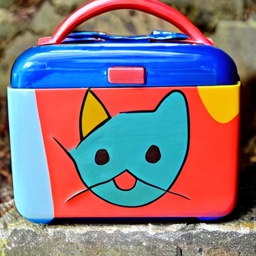} & \includegraphics[height=2.2cm]{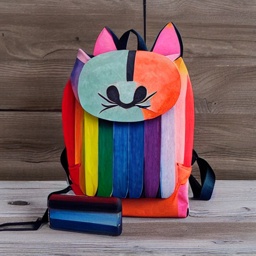} 
    &\includegraphics[height=2.2cm]{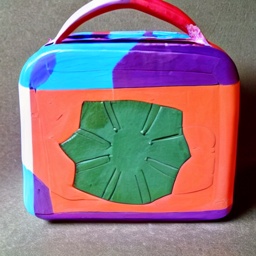} & \includegraphics[height=2.2cm]{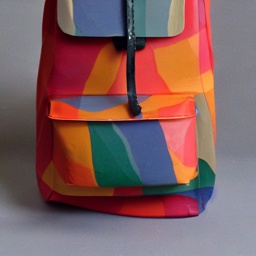} \\
    \includegraphics[height=2.2cm]{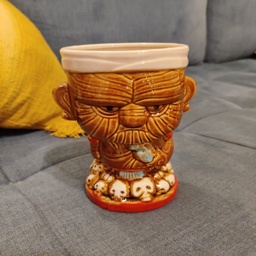} 
    & \includegraphics[height=2.2cm]{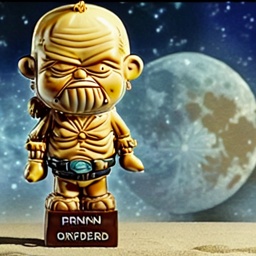} & \includegraphics[height=2.2cm]{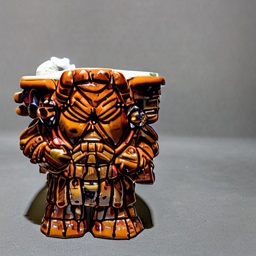} 
    &\includegraphics[height=2.2cm]{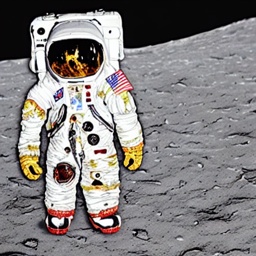} & \includegraphics[height=2.2cm]{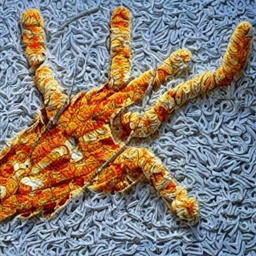}\\
    \end{tabular}
    \caption{Results of IMMA on Dreambooth (upper block) and Textual Inversion (lower block) datasets.}
    \label{fig:finetune_paper_dataset}
\end{figure}

For the four sets of images shown in~\figref{fig:finetune_paper_dataset}, we follow the prompts provided in the corresponding papers. The upper block shows the results of Dreambooth, and the lower block shows the results of Textual Inversion. As we can see, the generation with IMMA successfully prevents the model from generating content of target concepts.

\subsection{Visualization of generation with negative metrics 
 in~\tabref{tab:personalization_sgr}}

\begin{figure}[t]
    \centering
    \setlength{\tabcolsep}{1.5pt}
    \renewcommand{\arraystretch}{1.2}
    \begin{tabular}{lcc@{\hskip 8pt}cc}
    &Reference & \color{noimma} w/o IMMA &  \color{imma} w/ IMMA\\
    \rotatebox[origin=lc]{90}{\hspace{0.18cm}furniture chair} &
    \includegraphics[height=2.5cm]{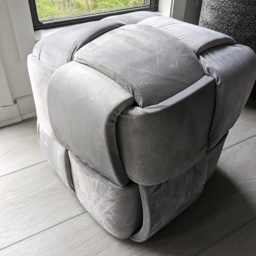} & 
    \includegraphics[height=2.5cm]{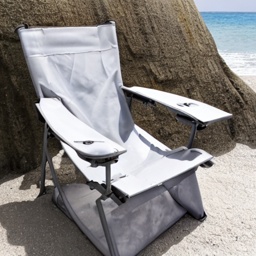}  &
    \includegraphics[height=2.5cm]{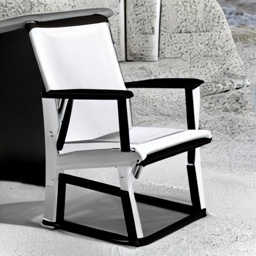}  \\
    \rotatebox[origin=lc]{90}{scene lighthouse} & 
    \includegraphics[height=2.5cm]{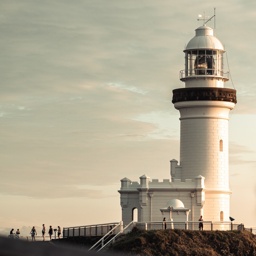} & 
    \includegraphics[height=2.5cm]{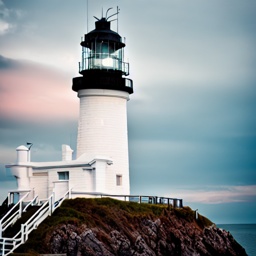}  &
    \includegraphics[height=2.5cm]{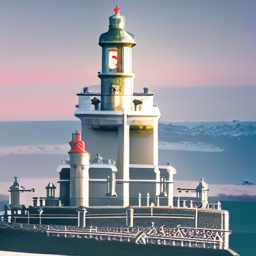}  \\
    \end{tabular}
    \vspace{-0.3cm}
    \caption{\textbf{Generation of datasets with negative metric values in~\tabref{tab:personalization_sgr}.} We observe that the base adaptation of DreamBooth's personalization adaptation failed even without IMMA.
    }
    \vspace{-0.3cm}
    \label{fig:failure_case}
\end{figure}

In~\tabref{tab:personalization_sgr}, there are two datasets shown with negative evaluation metric values. 
To study this, we provide the qualitative results for those corresponding datasets in~\figref{fig:failure_case}. %
In both cases, we observe that DreamBooth \textit{failed to learn the target concept} even without using IMMA.

\section{User Study}
\label{sec:supp_user_study}

The user study is designed to evaluate the generation quality and similarity to the reference images after adaptation with and without IMMA. The question includes both relearning erased styles and personalization adaptation.

\begin{figure*}[h]
\centering
\begin{tabular}{cc}
\includegraphics[height=6.5cm]{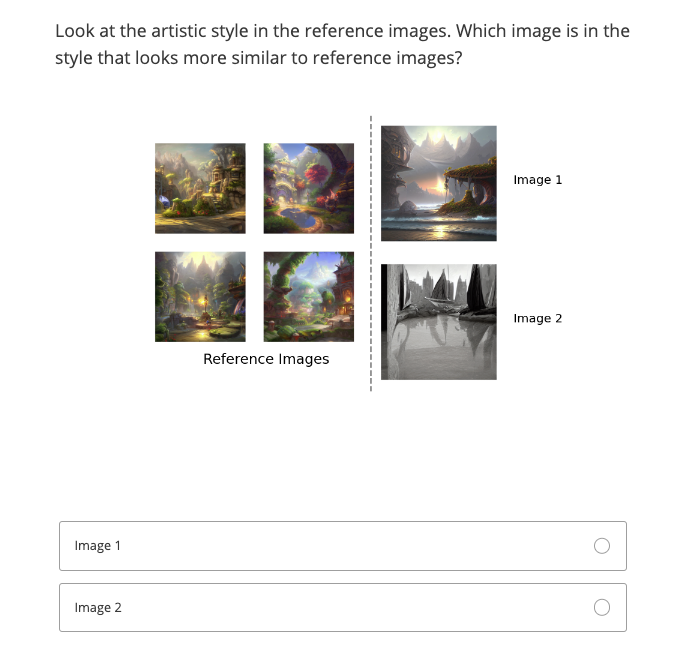} &
\includegraphics[height=6.5cm]{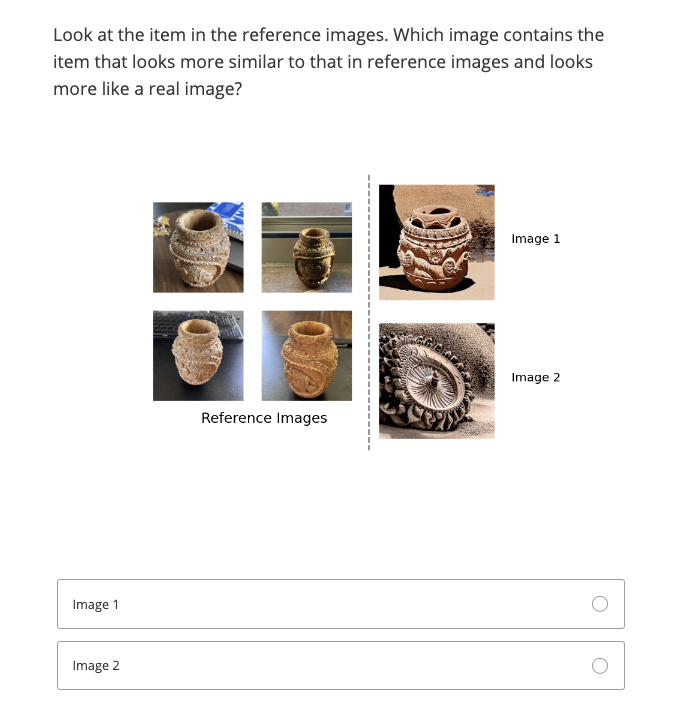}
\end{tabular}
\vspace{-0.3cm}
\caption{\textbf{Illustration of our user study survey.} We show four reference images to the user and ask them to select images generated by different methods for comparison.
}
\label{fig:user_study}
\end{figure*}

\myparagraph{Re-learning artistic styles.} We evaluate IMMA on preventing style relearning with erased models on eight artistic styles. For each style, the participants are shown four reference images randomly selected from the training images of that artist, \ie, the images generated by SD V1-4 conditioned on \textit{``an artwork of \{artist\}''}. We provide two images per question for participants to choose from, generation with and without IMMA, respectively. The judgment criteria are image quality (reality) and similarity to reference images. We provide the interface of the user study in~\figref{fig:user_study}.

\myparagraph{Personalization adaptation.} We also evaluate IMMA on personalization adaptation. For each personal item, the participants are shown four reference images that serve as training images for adaptation. We provide two images per question for participants to choose from, generation with and without IMMA, respectively. The judgment criteria are image quality (reality) and similarity to reference images.

\myparagraph{User study for MIST.} To evaluate the effect of MIST, we conducted a user study for MIST on personalization adaptation. The setting is identical to that of IMMA except that one of the images to choose from is the generation with MIST instead of IMMA.

\section{Additional Experimental Details}
\label{sec:exp_details}

We build our codes on the example code from Diffusers (\url{https://github.com/huggingface/diffusers/tree/main/examples}). The pre-trained diffusion model is downloaded from the checkpoint of Stable Diffusion V1-4(\url{https://huggingface.co/CompVis/stable-diffusion-v1-4}). Please refer to README.md of our attached code for hyperparameters and experimentation instructions. Note that we use the same set of hyperparameters for each adaptation method across all datasets.

\myparagraph{Backbone model for evaluation.} We use `ViT-B/32' for CLIP, `ViT-S/16' for DINO and AlexNet for LPIPS.

\myparagraph{Datasets.} We collected our datasets from the following sources: (i) ImgaeNet~\cite{deng2009imagenet} as in ESD~\cite{gandikota2023erasing} for object relearning. (ii) Eight artistic styles as in ESD~\cite{gandikota2023erasing} for style relearning. (iii) CustomConcept101~\cite{kumari2022customdiffusion} for personalization adaptation.

\myparagraph{Run time and memory consumption.} The running time and memory consumption for IMMA on a specific fine-tuning algorithm $\gA$ are comparable with adapting $\gA$ on the pre-trained models, \eg, training IMMA against concept relearning with LoRA takes 6 minutes and 15GB GPU memory usage in one Nvidia A30 GPU, where the training step is 1000  with a batch of one.

\end{document}